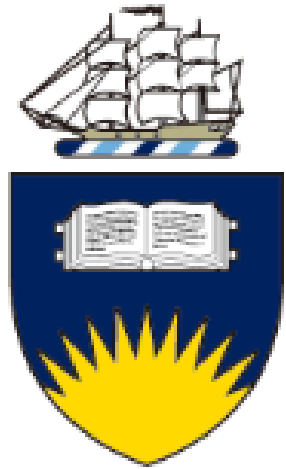

# Autonomous Reactive Mission Scheduling and Task-Path Planning Architecture for Autonomous Underwater Vehicle

by

**Somaiyeh Mahmoud.Zadeh**
*B.Eng.* (*Software Engineering.*), *M.Sc.* (*Computer Science*)
School of Computer Science, Engineering and Mathematics,
Faculty of Science and Engineering

Dec 2016

The thesis presented to the
Flinders University of South Australia
in total fulfilment of the requirements for the degree of
Doctor of Philosophy

# Contents

*CONTENTS*

# List of Algorithms

# List of Figures

## LIST OF FIGURES

*LIST OF FIGURES*

# List of Tables

# Abstract

An Autonomous Underwater Vehicle (AUV) should carry out complex tasks in a limited time interval. Since existing AUVs have limited battery capacity and restricted endurance, they should autonomously manage mission time and the resources to perform effective persistent deployment in longer missions. Task assignment requires making decisions subject to resource constraints, while tasks are assigned with costs and/or values that are budgeted in advance. Tasks are distributed in a particular operation zone and mapped by a waypoint covered network. Thus, design an efficient routing-task priority assign framework considering vehicle's availabilities and properties is essential for increasing mission productivity and on-time mission completion. This depends strongly on the order and priority of the tasks that are located between node-like waypoints in an operation network. On the other hand, autonomous operation of AUVs in an unfamiliar dynamic underwater and performing quick response to sudden environmental changes is a complicated process. Water current instabilities can deflect the vehicle to an undesired direction and perturb AUVs safety. The vehicle's robustness to strong environmental variations is extremely crucial for its safe and optimum operations in an uncertain and dynamic environment. To this end, the AUV needs to have a general overview of the environment in top level to perform an autonomous action selection (task selection) and a lower level local motion planner to operate successfully in dealing with continuously changing situations.

This research deals with developing a novel reactive control architecture to provide a higher level of decision autonomy for the AUV operation that enables a single vehicle to accomplish multiple tasks in a single mission in the face of periodic disturbances in a turbulent and highly uncertain environment. The system incorporates two different

*ABSTRACT*

execution layers, deliberative and reactive, to satisfy the AUV's autonomy requirements in high-level task-assign/routing and low-level path planning.

The proposed architecture encompasses a Task-Assign/Routing (TAR) approach as a high-level decision maker and a local Online Real-time Path Planner (ORPP) as a low-level action generator. This model can simultaneously plan the complete mission by accurate organizing the tasks to ensure a productive mission; while providing dynamic manoeuvre in dealing with uncertain variable environment. The ORPP operates concurrently, back feed the environmental condition to higher level TAR, and system makes decisions according to the raised situation. Static-dynamic current map data, uncertain dynamic-static obstacles, vehicles Kino-dynamic constraints are taken into account. The system has a cooperative and consistent manner in which the higher level mission planner handles mission scenario and gives a general overview of the terrain by cutting off the operating area to smaller beneficial zones for vehicles deployment in the feature of a global route (sequence of tasks).

On the other hand, path planning is an excellent strategy for guiding vehicle between two points, which accurately handles environmental changes when operating in a small scale area. Splitting the large area to smaller sections (carried out by higher level) solves the weakness of path planning methods associated with large scale operations. This leads to problem space reduction and significant reducing of the computational burden for the path planning; thus, re-planning a new trajectory requires rendering and re-computing less information.

Numerical simulations for analysis of different situations of the real-world environment is accomplished separately for each layer and also for the entire ARMSP model at the end. Performance and stability of the model is investigated through employing meta-heuristic algorithms toward furnishing the stated mission goals. It is demonstrated by analysing the simulation results that the proposed model provides a perfect mission timing and task managing while guarantying a secure deployment during the mission.

# Nomenclature

| | |
|---|---|
| $\Gamma_{3D}$ | Symbol of the 3-D volume |
| $\aleph_i$ | Task index |
| $\rho_i$ | Priority of task $i$ |
| $\xi_i$ | Risk percentage associated with task $i$ |
| $\delta_i$ | Absolute time required for completion of task $i$ |
| $P$ | Vertices of the network that corresponds to waypoints |
| $E$ | Edges of the network |
| $m$ | Number of waypoints in the network |
| $k$ | Number of edges in the network |
| $p^i_{x,y,z}$ | Position of arbitrary waypoint $i$ in 3-D space |
| $e^{ij}$ | An arbitrary edge that connects $p^i_{x,y,z}$ to $p^j_{x,y,z}$ |
| $w_{ij}$ | The weight assigned to $e^{ij}$ |
| $d_{ij}$ | Distance between position of $p^i_{x,y,z}$ and $p^j_{x,y,z}$ |
| $t_{ij}$ | Time required for traversing edge $e^{ij}$ |
| $\Re$ | An arbitrary route including sequences of tasks and waypoints |
| $T_\Re$ | Estimated mission time (route time) |
| $T_\nabla$ | The total available time for the mission based on battery capacity |
| $C_{TAR}$ | The task-assign/routing cost |
| $Se_{ij}$ | Selection variable {0: unselected, 1: selected} |
| $p^S$ | Position of the start waypoint |
| $p^D$ | Position of the destination waypoint |
| $C_i$ | Cluster centre |
| $\Theta$ | Obstacle |
| $\Theta_p$ | Obstacle's position |
| $\Theta_r$ | Obstacle's radius |
| $\Theta_{Ur}$ | Obstacle's uncertainty rate |
| $V_C$ | The current velocity vector |
| $u_c, v_c, w_c$ | $x,y,z$ components of water current vector |
| $\lambda_w$ | Covariance matrix for radius of vortex |
| $U_R^C(t)$ | Current update rate at time $t$ |
| $\{n\}$ | North-East-Depth (NED) frame |
| $\{b\}$ | AUVs Body-Fixed frame |

# NOMENCLATURE

| | |
|---|---|
| $[^n_b R]$ | The rotation matrix that transforms the body frame $\{b\}$ into $\{n\}$ frame |
| $S$ | Two-dimensional x-y space |
| $S^o$ | The centre of the vortex in the current map |
| $\ell$ | The radius of the vortex in the current map |
| $\mathfrak{I}$ | The strength of the vortex in the current map |
| $\eta$ | The AUV state in the $\{n\}$ frame |
| $\upsilon$ | Vehicle's velocity vector in the $\{b\}$ frame |
| $u,v,w$ | Linear components of $\upsilon$: surge, sway, heave |
| $X,Y,Z$ | Position of the AUV along the path $\wp$ in $\{n\}$ frame |
| $\phi,\theta,\psi$ | AUV's Euler angles of roll, pitch, yaw in $\{n\}$ frame |
| $p,q,r$ | AUV's angular velocities in the *x-y-z* axis |
| $\wp$ | The potential path produced by LOP-P |
| $\vartheta$ | Control point for B-Spline curve |
| $n$ | Number of control points along an arbitrary path $\wp$ |
| $L_\wp$ | Length of the candidate path $\wp$ |
| $T_\wp$ | The local path flight time |
| $T_\varepsilon$ | Time expected for traversing an edge $\approx t_{ij}$ |
| $L_\vartheta, U_\vartheta$ | lower and upper bounds for $\vartheta$ and, respectively |
| $B_{i,K}$ | Blending functions used to generate B-Spline curve |
| $C_\wp$ | The cost of local path generated by path planner |
| $\nabla_\wp$ | Path Violation of Kino-dynamic constraints |
| $\nabla_{\sum M,\Theta}$ | Path Violation of Collision violation |
| $\wp_{CPU}$ | computational time for generating a local path |
| $TAR_{CPU}$ | The task-assign/routing CPU time |
| $T_{compute}$ | Computation time for checking mission re-planning criterion |
| $Cost_{Total}$ | The total cost of the mission including $C_\wp$ and $C_{TAR}$ |
| $t_{max}$ | Maximum number of iterations |
| $i_{max}$ | Maximum number of population |
| $\chi^{G\text{-}best}$ | ACO: The global best solution in the population |
| $\chi_t^{best}$ | ACO: The best solution |
| $\chi_t^{worst}$ | ACO: The worst solution |
| $\tau^k_{ij}$ | ACO: Pheromone concentration on edge for ant *k* from node *i* to *j* |
| $\alpha$ | ACO: The pheromone factor |
| $\beta$ | ACO: The heuristic factor |
| $N_i^{(k)}$ | ACO: The set of neighbour of ant *k* when located at node *i* |
| $\eta_i$ | ACO: The heuristic information in $i^{th}$ node |
| $Q$ | ACO: The released pheromone on nodes taken by the best ant(s) |
| $\varpi$ | ACO: The coefficient of pheromone evaporation |
| $r_e$ | ACO: The evaporation rate |
| $\mu$ | BBO: The emigration rate |

*NOMENCLATURE*

| | |
|---|---|
| $\lambda$ | BBO: The immigration rate |
| $h_i$ | BBO: The habitat as a solution |
| $S_p$ | BBO: Number of species in a habitat |
| $m_r$ | BBO: The mutation rate |
| $m_{r,max}$ | BBO: The maximum mutation rate |
| $S_{p,max}$ | BBO: The habitat with maximum number of species |
| $P_{Sp,max}$ | BBO: Probability the habitat with maximum number of species $S_{p,max}$ |
| $I_r$ | BBO: The maximum immigration rate |
| $E_r$ | BBO: The maximum emigration rate. |
| $P_{Sp}(t)$ | BBO: The probability of existence of the $S_p$ species at time $t$ |
| $\chi_i, \upsilon_i$ | PSO: The position and velocity of a particle, respectively |
| $\chi^{P\text{-}best}$ | PSO: The personal positions of a particle |
| $\chi^{G\text{-}best}$ | PSO: The global best positions of a particle |
| $c_1, c_2$ | PSO: The acceleration parameters |
| $r_1, r_2$ | PSO: The randomness factors |
| $\omega$ | PSO: The weight for balancing the local-global search |
| $\chi_i$ | FA: The firefly in the population. |
| $L$ | FA: The distance between fireflies |
| $\beta$ | FA: The received brightness from the adjacent fireflies |
| $\beta_0$ | FA: The initial attraction value of firefly $\chi_i$ |
| $\alpha_t$ | FA: The randomization factor |
| $\alpha_0$ | FA: The initial randomness scaling value |
| $\gamma$ | FA: The light absorption coefficient |
| $\delta$ | FA: The damping factor |
| $\chi_i$ | DE: The solution vector in the population |
| $\chi_{ri,t}$ | DE: The randomly picked solution vector in iteration $t$ |
| $\dot{\chi}_{i,t}$ | DE: The mutant solution vector |
| $\ddot{\chi}_{i,t}$ | DE: The produced offspring solution vector by crossover operation |
| $S_f$ | DE: The scaling factor |
| $r_C$ | DE: The crossover rate |

# List of Abbreviations

| | |
|---:|---|
| ACO | Ant Colony Optimisation |
| ARMSP | Autonomous Reactive Mission Scheduling Path-planning |
| AUVs | Autonomous Underwater Vehicles |
| APF | Artificial Potential Field |
| BBO | Biogeography-Based Optimisation |
| BKP | Bounded Knapsack Problem |
| CBR | Case Based Reasoning |
| DE | Differential Evolution |
| DEMiR-CF | Distributed and Efficient Multi Robot Cooperation Framework |
| DTSPMS | Double Traveling Salesman Problem with Multiple Stacks |
| FM | Fast Marching |
| FA | Firefly Algorithm |
| GA | Genetic Algorithm |
| HSI | Habitat Suitability Index |
| HADCP | Horizontal Acoustic Doppler Current Profiler |
| ICA | Imperialist Competitive Algorithm |
| INGA | Improved Niche Genetic Algorithm |
| MARL | Multi-Agent Reinforcement Learning |
| MILP | Mixed Integer Linear Programming |
| NSM | Nearest Sequence Matching |
| NED | North-East-Depth |
| NP | Non-deterministic Polynomial-time |
| NSGA-II | Non-dominated Sorting Genetic Algorithm |
| ORPP | Online Real-Time Path Planning |

*LIST OF ABBREVIATIONS*

| | |
|---|---|
| PSO | Particle Swarm Optimization |
| QVDP | Q-Value Based Dynamic Programming |
| TAR | Task-Assign/Routing |
| RFETA | Robust Filter Embedded Task Assignment |
| SOM | Self-Organizing Map |
| SA | Situation Awareness |
| SIVs | Suitability Index Variables |
| TS | Tabu Search |
| TA | Task Assignment |
| TSP | Traveller Salesman Problem |
| UAVs | Unmanned Aerial Vehicle |
| VRP | Vehicle Routing Problem |

# Declaration

I certify that this thesis does not incorporate without acknowledgment any material previously submitted for a degree or diploma in any university; and that to the best of my knowledge and belief it does not contain any material previously published or written by another person except where due reference is made in the text.

Somaiyeh MahmoudZadeh

Adelaide, 24 May 2017

# Acknowledgements

Firstly, I would like to express my sincere gratitude to my advisor Prof. *David Powers* for the continuous support of my PhD study and related research, for his patience, motivation, enthusiasm, and immense knowledge. His guidance helped me in all the time of research and writing of this thesis. His pursuit of excellence in research and his constant desire to push the limits were very inspiring.

My appreciation also extends to my co-supervisor A/Prof. *Karl Sammut*, who provided me an opportunity to join his team, and for his advice, motivation and inspiration. I would like to thank my other co-supervisors Dr *Adham Atyabi* and Dr *Trent Lewis* for their guidance and support over the years of PhD study. My special appreciation and thanks goes to *Adham* for his technical support and guidance in writing of this thesis.

I would like to thank the School of Computer science, Engineering and Mathematics at the Flinders University of South Australia for their assistance and support. I would like to thank my colleagues and lab mates from Centre for Maritime Engineering, Control and Imaging of Flinders University for their wonderful collaboration. My sincere thanks also goes to Dr. *Andrew Lammas* for his excellent guidelines and encouragement. I thank my dear colleague *Amirmehdi Yazdani* in for the endless support, for the sleepless nights we were working before deadlines, and for all the fun we have had in the past years. *Amir*, without your precious support it would not be possible to conduct this research. In addition, I would like to thank my dear friend *Zheng Zeng* for his accompany and compassionate support. You made my PhD years very memorable, and I will always cherish our friendship.

Last but not the least; I would like to thank my family: my parents, my dear fiancé *Reza*, my brother and sisters and *Asal Jigulie* for their wise counsel and sympathetic ear and for supporting me spiritually throughout my life. *Ahad dadash*, and *Akram* you were the first to enlighten me with my first glimpse of research. I would particularly like to single out my fiancé *Reza* for all the love he has gave me. You are always there for me.

*To those stories that happened*

# Chapter 1

# Introduction

## 1.1 Research Background

Autonomous Underwater Vehicles (AUVs) have proven their capabilities in cost effective underwater explorations up to thousands of meters, far beyond what humans are capable of reaching, and today are the first choice to navigate autonomously and undertake various underwater missions. AUVs are largely employed for purposes such as inspection and survey [1], scientific underwater explorations [2,3], sampling and monitoring coastal areas [4], benthic habitat mapping [5], to measure turbulence and obtain related data [6,7], offshore installations and mining industries [8], and military missions [9]. AUVs generally are capable of spending long periods of time carrying out underwater missions at lower costs comparing manned vessels [10].

Autonomous operation of AUV in a vast, unfamiliar, and dynamic underwater environment is a complicated process, specifically when the AUV is obligated to react promptly to environmental changes. An AUV thus needs to ensure safe deployment at all stages of a mission, as failure is not acceptable due to the prohibitive recovery and repair costs. On the other hand, diversity of underwater scenarios and missions necessitates robust decision-making based on proper awareness of the situation. A proper awareness of the situation is a key element in the process of decision-making. Hence, an overview of the possible environmental situations has a great impact in enhancing vehicle's overall autonomy. However, autonomous expectation of AUVs in performing various tasks in dynamic and continuously changing environment is not completely fulfilled yet, where it is strongly necessary for the operators to remain in the loop of considering and making decisions [11]. Hence, an advanced intelligence (at the same level as human operator) is an essential prerequisite to compromise the tasks importance and the problem constraints while dealing with the environmental changes. Increasing the levels of autonomy leads to minimizing the reliance on the human supervisor. Development of such self-controlled systems tightly depends on the design of robust mission-motion planning and task allocation scheme. This research aims to increase AUV's degree of intelligence in both high and low level task-assign/mission and motion planning.

**1.1.1 Mission Planning, Task Assigning and Routing in the Current Research**

An AUV should carry out complex mission in a confined time. However, it has limited battery capacity and restricted endurance, which these limitations arose the necessity of managing mission time. Obviously, a single AUV is not able to meet all specified tasks in a single mission with limited time and energy. The vehicle has to manage its resources effectively to perform effective and persistent deployment in longer missions without human interaction.

The tasks can have different characteristics such as priority value, risk percentage, and absolute completion time, which are known in advance by AUV. Assuming that different tasks are distributed over a specific operation area, they can be mapped by a network in which their start and ending point is presented as waypoints (a sample map is illustrated in Figure 1.1). In this way, the complication of vehicle routing in a graph-like terrain can be addressed straightforwardly.

Task assignment is a subprocess of routing. Mission planning in this research is presented as a combination of the Vehicle Routing Problem (VRP) and task assignment-allocation, presented by Figure 1.2. The VRP is analogous to Traveller Salesman Problem (TSP), which is contiguous sequencing problem in which order of solutions is important. On the other hand, the tasks assignment with time restriction is a discrete (synthetic) natured problem analogous to Bounded Knapsack Problem (BKP), in which combination of the solutions is important. In this respect, mission productivity and time management is a fundamental requirement towards mission success that tightly depends on the optimality (order and priority) of the selected tasks between start and destination point in a graph-like operation terrain.

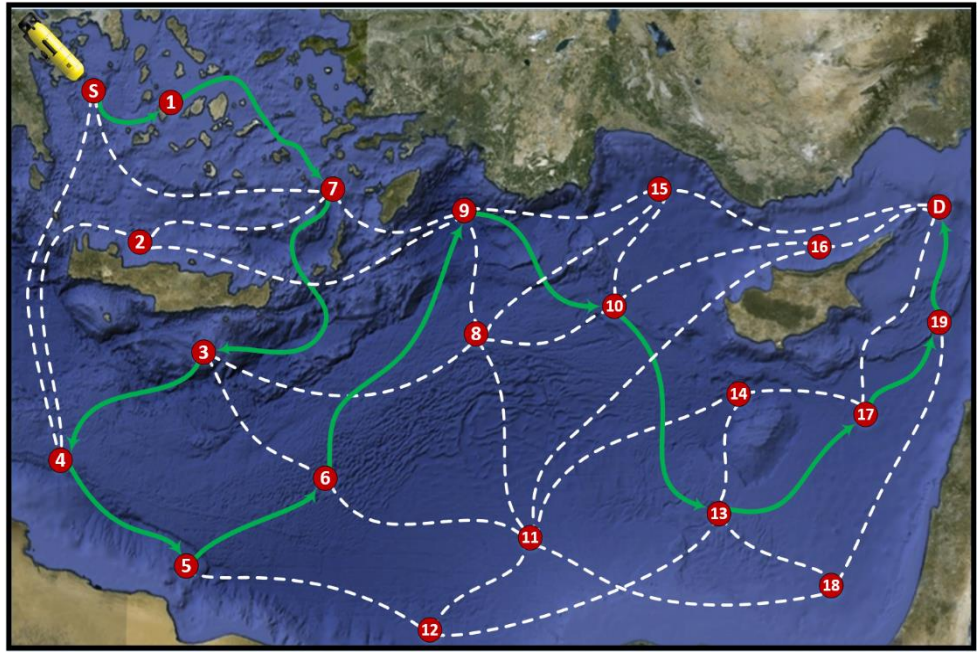

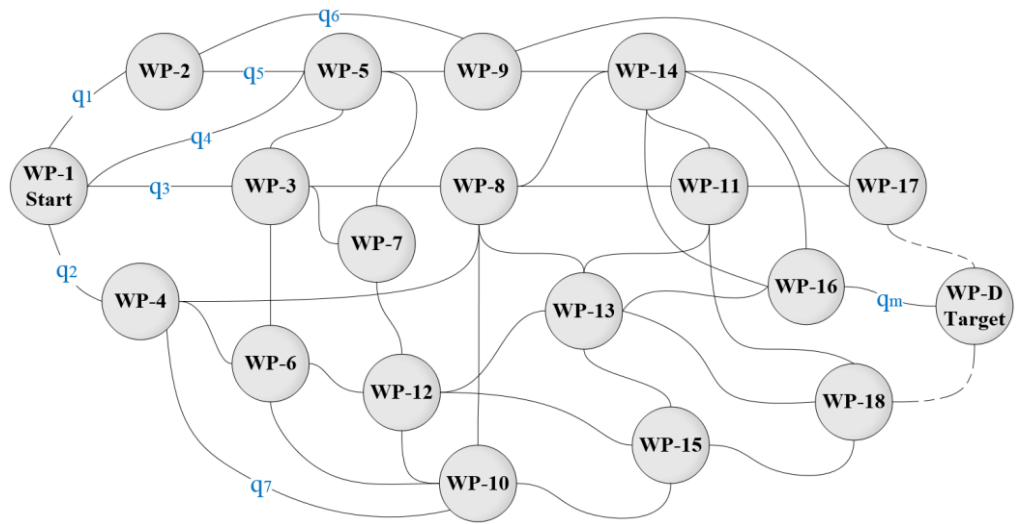

**Figure 1.1. (a)** Sample of graph representation of operating area covered by waypoints, in which AUV starts its operation from a specific waypoint and it is expected to reach to the destination waypoint passing adequate number of edges; **(b)** The graph representation of the operation area in which any edge ($q_i$) is assigned with a specific task and it is weighted with the characteristics of the corresponding task.

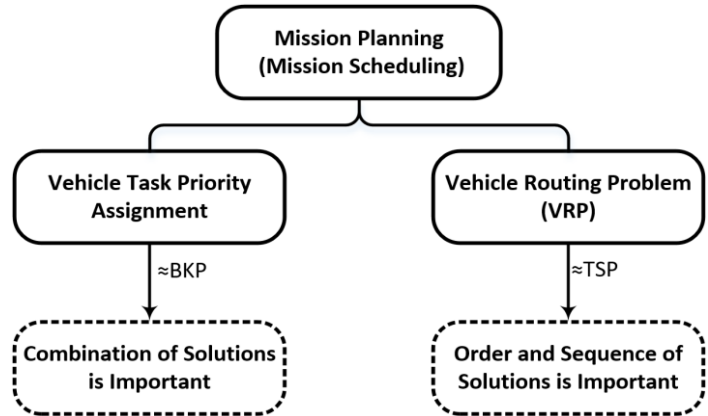

**Figure 1.2.** Components of the mission planning.

Based on given explanations, designing of an efficient task-assign/routing framework considering the vehicle's availabilities and capabilities is an essential requirement for maximizing mission productivity and on-time termination of the mission.

### 1.1.2 Path Planning in the Current Research

AUVs autonomous operation in an unfamiliar and dynamic underwater environment is a complicated process, especially when it is required to respond to environmental changes, where usually *a priori* information is not available (local path planning). A vehicle interacts with the environment according to its observation and comprehension from surrounding area to have a safe deployment. The underwater environment can be considered from three different perspectives. First is the known environment. In this category the information about undersea environment and probable possibilities are provided by an offline map and is fed to the AUV beforehand. Second is the partially known environment in which, the vehicle has a general overview of the environment, but the details and unforeseen events remain as a hazy part of the vehicle's observation. The third category is the unknown environment. In such conditions, the vehicle has no or little information about the surrounding environment and it is expected to observe, find out, categorise and understand the events in order to appropriately interact with the unknown phenomena. To handle the first category (known environment), several artificial intelligence and machine learning algorithms such as, Case Based Reasoning (CBR) and Nearest Sequence Matching (NSM) [12-14], have been employed previously. These strategies can help vehicle to use embedded knowledge and taking an action facing similar situation based on previous experiences. This mechanism is helpful for the vehicle's interaction with environment and helps it to follow the defined tasks and goals. However, some events do not exist in the vehicle's experience library especially when the vehicle deals with the unknown or partially known environment. In order to overcome these difficulties, the vehicle should be able to perform motion planning in real-time.

Recent advancements in sensor technology and embedded computer systems has opened new possibilities in underwater path planning and made AUVs more capable of handling long-range operations. However, there still exist major challenges for this class of the vehicle, where the surrounding environment has a complex spatiotemporal variability and uncertainty. Water currents instabilities can strongly affect its motion and probably push it to an undesired direction. This variability can also perturb AUVs safety conditions [15]. Robustness of AUVs to this strong environmental variability is a key element to carry out safe and optimum operations. Figure 1.3 provides a general understanding of a motion planning system.

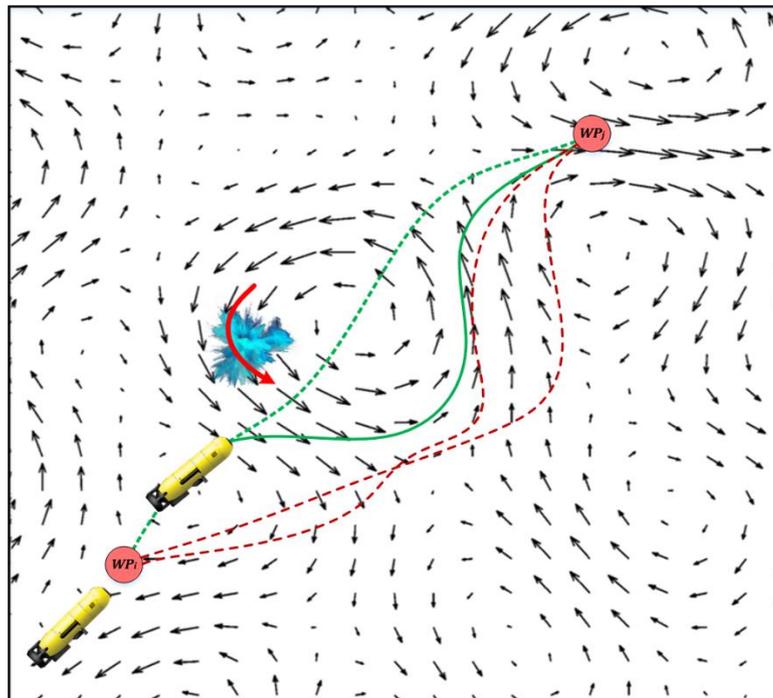

**Figure 1.3.** A sample representation of an adaptive path planning in presence of water current (presented by black arrows) moving obstacles affected by current flow (presented by the blue object) in which for coping the environmental variations the generated path (presented by dashed lines) is refined to the new path (presented by thicker green line) from the vehicles existing position to the target point ($WP_j$).

## 1.2 Problem Statements and Research Motivation

Considerable effort has been devoted in recent years to enhancing AUV's capability in robust motion planning, vehicle routing and efficient task assignment [1]. Although some improvements have been achieved in other autonomous systems, there are still many challenges to achieve a satisfactory level of intelligence and robustness for AUV in this regard.

### 1.2.1 Challenges in the Scope of Vehicle Routing and Task Assigning

Effective routing and task priority assigning have a great impact on a vehicle's time management as well as mission performance by appropriate selection and arrangement of the tasks sequence. The AUV should handle the mission in a bounded time and thus needs to manage its resources effectively to perform resilient operation without human interaction and to succeed in longer missions. This issue tightly depends on optimality of the selected route between start and destination point. On the other hand, quality of vehicle's deployment is another important consideration to ensure its safety in an uncertain and dynamic environment. A mission planning strategy is capable of organizing tasks in an accurate manner to manage the available time and ensure a productive mission; however, this strategy is not designed for handling prompt changes of the environment, which is a critical concern in underwater autonomous operations.

### 1.2.2 Challenges in the Scope of the AUVs' Path Planning

The path planning techniques are specifically designed to deal with quality of vehicle's motion encountering environmental characteristics and variations. As mentioned earlier, robustness of the trajectory planning to current variability and terrain uncertainties is essential to mission success and AUV's safe deployment. Existence of a priori knowledge about the variability of the current and the environment facilitates the AUV to minimize the undesirable effects of the environment on its operation. However, present technology is only capable of forecasting limited components of the ocean variability. This limited knowledge about later conditions of the environment reduces AUVs autonomy, safety, and its robustness. The shortcomings with the path planner specifically appears when the AUV is required to operate in a large scale terrain, as it should compute a large amount of data repeatedly and estimate dynamicity of the underwater adaptively. Proper estimation of the underwater behaviour beyond the sensor coverage is impractical and inaccurate. Moreover, path-planning strategies are designed to deal with vehicles guidance from one point to another and do not deal with mission scenario or task assignment considerations.

---

1. Literature review on the addressed scopes is provided by Chapter 2.

### 1.2.3 Research Motivation

Existing approaches mainly are able to cover only small parts of this problem, that is, either task assignment together with time management or path planning along with obstacle avoidance as safety consideration. To fill the gaps of each strategy, the AUV needs to have a general overview of the environment in top level to perform an autonomous action selection (task selection) and a lower level local motion planner to operate successfully in dealing with continuously changing situations. This research aims to address the AUV's requirements in both high and low level autonomy toward making AUV systems more intelligent and robust in managing its availabilities and adapting to the dynamic environment (pointed out in Section 1.3).

## 1.3 Research Objectives

It is important for an autonomous vehicle to operate successfully in dealing with continuously changing situations. The main goal of AUV operation is to complete mission objectives while ensuring the vehicle's safety at all times. Appropriate mission planning and path planning strategy along with efficient synchronization between two planners maximizes the achievements of a mission. To this end, following objectives are introduced to furnish the mentioned above expectations:

**1)** To develop an appropriate Task-Assign/Routing system:
    **a.** To provide an efficient routing for AUV to take a maximum use of time that battery capacity allows;

      **b.** To carry out an accurate task assignment by selecting the highest priority tasks with minimum risk percentage fitted to the available time;

      **c.** To guarantee on-time termination of the mission;

**2)** To develop a reliable on-line path planning strategy to provide a safe trajectory for vehicles deployment:

      **a.** To avoid colliding uncertain static and moving obstacles and avoid entering no-flying zones;

      **b.** To cope with adverse water current;

      **c.** To use desirable water current for saving energy;

      **d.** To carry out on-line re-planning at any time that a prompt change in the environment is appeared;

**3)** To develop a synchronous control architecture:

      **a.** To provide a concurrent operation and synchronization among mission planner and path planner;

      **b.** To manage mission time by reorganizing the tasks and carrying out mission re-scheduling;

      **c.** To split the operating area to smaller and more useful zones for path planner;

      **d.** To compensate any lost time during path re-planning process;

## 1.4 Research Assumptions and Scope

The AUV is required to complete the maximum possible number of tasks of the highest priority and reach the destination on time, passing through the appropriate sequence of waypoints while considering three-dimensional underwater variations (water current or/and obstacles, etc.), problem constraints (time/battery restriction, collision avoidance constraints, etc.), and restrictions of vehicle's kinematics. Selecting the sequence of waypoints is a very important issue in saving time and energy and satisfying the task assignment purpose of the mission. The construction of this study is defined in threefold. First attempt will be developing a task-assign/routing framework to find a time efficient route in a graph-like terrain and appropriate arranging the tasks to ensure the AUV has a plenteous journey and efficient timing. In this framework, the operation network is generated in advance and different tasks are distributed to different parts of the generated map.

The second step will be utilizing an accurate and reliable path planning system to provide an overall situational awareness for the vehicle in dealing with unknown and uncertain events. In this framework, the path planner deals with quality and safety of deployment along the split area bounded to distance between waypoints, where persistent variation of the sub-area in proximity of the vehicle is considered simultaneously. To the purpose of this research, static and time-varying current map, kinematic of the AUV, different uncertain obstacles along with real map data are taken into account for modelling a realistic underwater environment and for improving the proposed path planner in handling real-world underwater situations.

The third and last endeavour of this study will be developing a modular control architecture with reactive re-planning capability to carry out the underwater missions in a large scale environment in the presence of severe environmental disturbances. The system incorporates two different execution layers, deliberative and reactive, to satisfy the AUV's autonomy requirements in high-level mission planning and low-level path planning. In this autonomous architecture, the mission-planning framework operates at the top level as a deliberative layer, and the path planning system operates at the lower level as a reactive layer. The path planner operates concurrently and back feeds the environmental condition to higher-level mission planner and system makes decisions according to the raised situation. The operation of each module may take longer than expectation due to unexpected dynamic changes in the environment. In order to reclaim the missed time, an efficacious synchronization scheme (named "Synchron") is added to the architecture to keep pace of modules in different layers of the system.

## 1.5 Statements of Contributions

- ▪ The proposed approach in this research is advantageous as it joins two disparate aspects of vehicle autonomy, combining high-level task organization and self/environment awareness by low-level motion planning.

- The system has a cooperative and consistent manner in which the higher-level mission planner handles mission scenario by tasks priority assignment and gives a general overview of the terrain by cutting off the operating area to smaller beneficial zones for vehicles deployment in the feature of a global route including a sequence of tasks. On the other hand, path planning is an excellent strategy for guiding vehicle between two points, which accurately handles environmental changes when operating in a small-scale area. Splitting the large area to smaller sections (carried out by higher-level mission planner) solves the weakness of path planning methods associated with large-scale operations. This leads to problem space reduction and significantly reducing the computational burden for the path planning; thus, re-planning a new trajectory requires rendering and re-computing less information.

- Parallel execution of the higher and lower level systems speeds up the computation process, which is another reason for fast operation of the proposed approach, in which total operation time is in the range of seconds that is a remarkable achievement for such a real-time system.

- The performance of the proposed architecture is mainly independent of the employed algorithms by modules, where the use of diverse algorithms (subjected to real-time performance of the algorithm) does not negatively affect the real-time performance of the system.

- The main reason for remarkable performance of this system is the fashion of mixing and matching two strategies from two different perspectives and constructing an accurate synchronization between them. A significant benefit of such modular construction is that the modules can employ different methods or their functionality can be upgraded without manipulating the system's structure. This advantage specifically increases the reusability and versatility of the control architecture and eases updating/upgrading AUV's functionalities to be compatible with other applications, and in particular implement it for other autonomous systems such as unmanned aerial, ground or surface vehicles.

## 1.6 Publications and Thesis Outline

This thesis is the outcome of research carried out whilst studying at the Flinders University of South Australia. The architecture introduced in this thesis has been developed gradually, and different components of the work have been completed, reviewed and presented in several international conferences or journals:

**[P.1]** Optimal Route Planning with Prioritized Task Scheduling for AUV Missions, *IEEE International Symposium on Robotics and Intelligent Sensors* **(ISIR)**. pp 7-15, 2015.

**[P.2]** A Novel Efficient Task-Assign Route Planning Method for AUV Guidance in a Dynamic Cluttered Environment. *IEEE Congress on Evolutionary Computation* **(CEC)**. Vancouver, Canada. July 24-29, 2016. **(Published)**. arXiv:1604.02524

**[P.3]** Differential Evolution for Efficient AUV Path Planning in Time Variant Uncertain Underwater Environment. *Journal of Marine Science and Application* **(JMSA)**. 2017. **(Submitted)**. arXiv:1604.02523

**[P.4]** AUV Rendezvous Online Path Planning in a Highly Cluttered Undersea Environment Using Evolutionary Algorithms. *Journal of Applied Soft Computing* **(ASOC)**. 2016. **(Accepted)**. arXiv:1604.07002

**[P.5]** Toward Efficient Task Assignment and Motion Planning for Large Scale Underwater Mission. *Journal of Advanced Robotic Systems* **(JARS-SAGE)**. October 2016. Volume 13: pp. 1-13. DOI: 10.1177/1729881416657974

**[P.6]** Biogeography-Based Combinatorial Strategy for Efficient AUV Motion Planning and Task-Time Management. *Journal of Marine Science and Application* **(JMSA)**. December 2016, Volume 15, Issue 4, pp 463–477. DOI: 10.1007/s11804-016-1382-6

**[P.7]** A Novel Versatile Architecture for Autonomous Underwater Vehicle's Motion Planning and Task Assignment. *Journal of Soft Computing* **(SOCO)**. November 2016. Volume 20 (188): pp.1-24. DOI 10.1007/s00500-016-2433-2

**[P.8]** An Efficient Hybrid Route-Path Planning Model for Dynamic Task Allocation and Safe Maneuvering of an Underwater Vehicle in a Realistic Environment. *Journal of Autonomous Robots* **(JARO)**. 2016. **(Submitted)**. arXiv:1604.07545

**[P.9]** A Hierarchal Planning Framework for AUV Mission Management in a Spatio-Temporal Varying Ocean. *Journal of Computers and Electrical Engineering* **(JCEE)**. 2016. **(Submitted)**. arXiv: 1604.07898

**[P.10]** An Autonomous Dynamic Motion-Planning Architecture for Efficient AUV Mission Time Management in Realistic Sever Ocean Environment. *Journal of Robotics and Autonomous Systems* **(JRAS)**. January 2017. Volume 87: pp. 81–103. http://dx.doi.org/10.1016/j.robot.2016.09.007

**[P.11]** Persistent AUV Operations Using a Robust Reactive Mission and Path Planning (R$^{II}$MP$^{II}$). ***Journal of Soft Computing* (SOCO)**. 2016. **(Submitted)**. arXiv: 1605.01824

The papers are attached to thesis by Appendices A to K. The flow of the completion process, in both assumptions and algorithms, is demonstrated by the following diagram in Figure 1.4 and Table.1.1.

**Table 1.1.** The provided publications toward completion of the thesis chapters.

| | Publications | Ref | Appendix # | Algorithms | Assumptions and Improvements |
|---|---|---|---|---|---|
| **Chapter-3 Mission Planner** | Conference (IRIS) Published | [P.1] | A | GA, ICA | A simple static mission planner is developed for task ordering implying target points/trajectories for later path planning |
| | Conference (WCCI-CEC) Published | [P.2] | B | PSO, BBO | The mission planner in [P.1] is improved and implemented on a dynamic network and a rerouting capability is added. |
| **Chapter-4 Path Planner** | Journal (JMSA) Accepted | [P.3] | C | DE | A small-scale offline path planner is developed considering static water current and static environment without re-planning. |
| | Journal (ASOC) Published | [P.4] | D | DE-FA BBO-PSO | An online path planner-replanner is developed taking current, uncertainty and variability of the environment into account. |
| **Chapter-5 Hybrid Planners & Modular Architecture** | Journal (IJARS) Published | [P.5] | E | Layer-1: **PSO** Layer-2: **PSO** | A hierarchal model of static mission planner and simple path planner is developed. + Synchronization process is added. |
| | Journal (JMSA) Published | [P.6] | F | Layer-1: **BBO** Layer-2: **BBO** | + Offline path planning is implemented - Re-planning capability is *not* provided. + Static and floating/suspended obstacles are handled. - Dynamic obstacles are *not* considered - Currents are *not* considered. - Vehicle Kino-dynamics are *not* considered. - *No* map data is considered. |
| | Journal (SOCO) Accepted | [P.7] | G | Route planning module: **GA** Path planning module: **PSO** | The hierarchal model presented in the last two papers [P.5, P.6] is upgraded to a modular architecture. + Dynamic uncertain obstacles are now taken into account. |

| | Journal (JARO) Submitted | [P.8] | H | Layer-1: **DE** <br><br> Layer-2: **FA** | The hierarchal model introduced in [P.5, P.6, P.7] is upgraded with a rerouting capability. <br> + Static water current map is taken into account. <br> + Environmental obstacles assumed to be uncertain and static. <br> + Real map data is added into account. <br> Path planning is implemented considering vehicle's Kino-dynamic and water current effect on vehicle's motion. <br> Task risk and task completion time are added into account. <br> The path planner in this paper implemented as an offline planner. |
|---|---|---|---|---|---|
| **Chapter-5** <br> **Hybrid Planners & Modular Architecture** | Journal (JCEE) Submitted | [P.9] | I | Layer-1: **BBO** <br><br> Layer-2: **BBO** | The hybrid model of [P.8] is further developed as follows: <br> Online path planning and reactive mission planning capability are integrated. <br> The environment is modelled as dynamic and uncertain covering all types of static-floating-suspended-moving obstacles. <br> The dynamic time-varying water current is taken into account in online path planning. |
| | Journal (JRAS) Published | [P.10] | J | Task-assign/routing module: **ACO** <br><br> Online path planning module: **FA** | This paper is an extension of [P.8,P.9] in which the construction of the higher and lower level mission-motion planners are enhanced, moving to a consistent architecture in which the performance of the proposed architecture is mainly independent of the employed algorithms by modules. A synchro module is proposed and added to the architecture. |
| | Journal (SOCO) Submitted | [P.11] | K | TAR Module: <br><br> **ACO-BBO-PSO-GA** <br><br> ORPP Module: <br><br> **DE-FA-BBO-PSO** | This paper presents an integrated mode of all the previous studies with all variables considered [P.1 to P.10]. This research thus contributes a novel autonomous ARMSP architecture covering shortcomings associated with previous work. This architecture contains both high-level decision making and low-level action generator, and it can simultaneously organize the mission completion and the manoeuvrability of the vehicle increasing the autonomy of the vehicle for operation in a cluttered and uncertain environment and carrying out a successful mission. The proposed architecture offers a degree of flexibility for employing diverse algorithms and can effectively synchronize them to have a hitherto unachieved level of performance. |

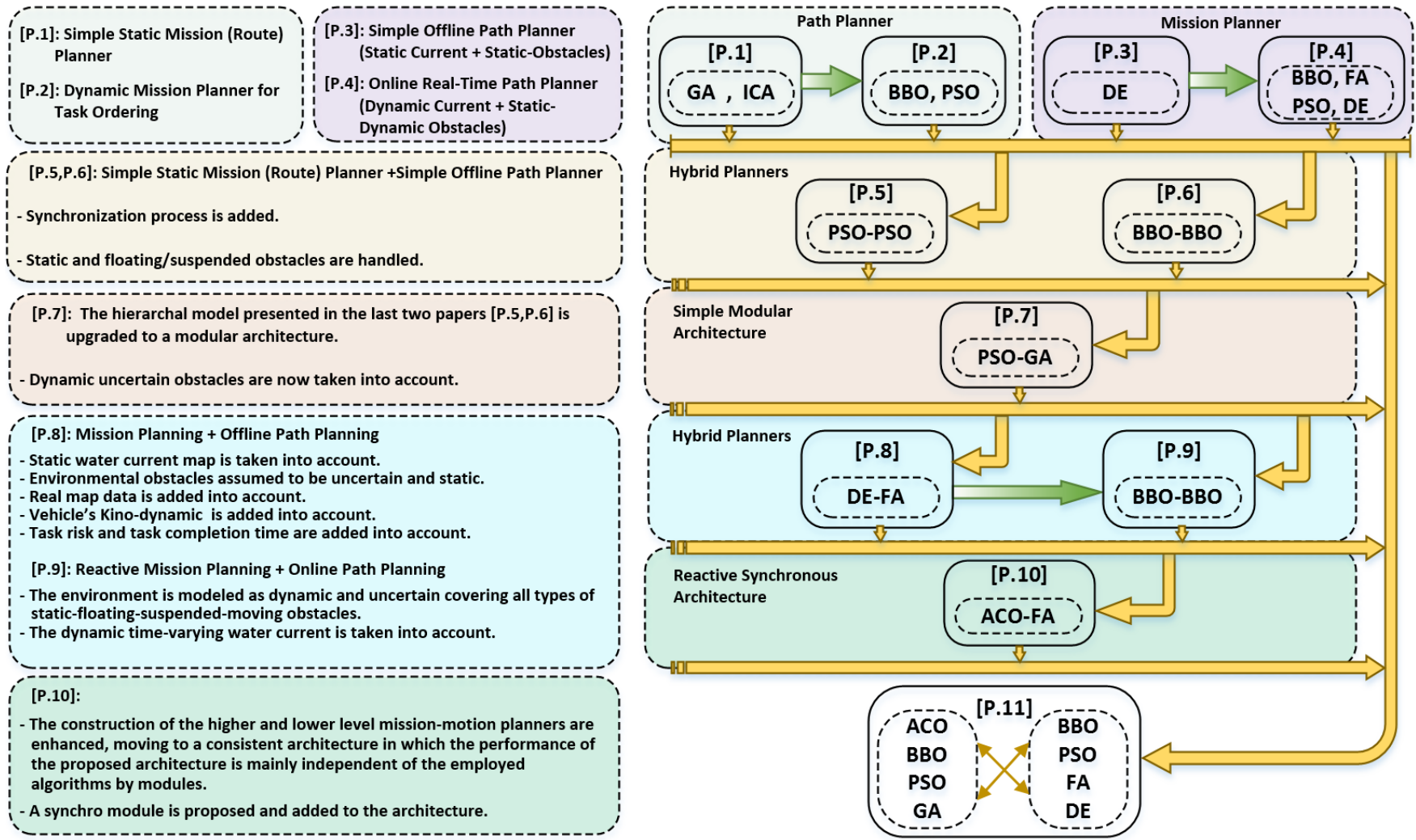

**Figure 1.4.** The flow diagram of the published papers in thesis completion process (both assumptions and algorithms).

*CHAPTER 1. INTRODUCTION*

The thesis is organized as follows.

Chapter 1 provides a background relating to the challenges of autonomous operation of AUV. Detailed definitions of task-assign/routing and path planning, research contribution and structure of thesis is provided in this chapter.

Chapter 2 provides a basic definition of autonomy and its dependencies to accurate decision-making and efficient situation awareness. The relation of mission planning as a higher level decision maker and the path planning based on vehicle's awareness from different situations, will be explained by Sections 2.3 and 2.4, respectively. A principle review of the existing methods for autonomous vehicles' routing-task assignment problem and state of the art in autonomous vehicles motion (path-route-trajectory) planning will be provided and discussed in this chapter.

The mission-planning problem will be discussed and mathematically formulated by Chapter 3. A Task-Assign/Routing (TAR) system will be developed and simulated through a set of given constraints. The TAR system will be tested and evaluated on different graphs of varying complexity applying Genetic Algorithm (GA), Ant Colony Optimisation (ACO), Biogeography-Based Optimisation (BBO), and Particle Swarm Optimization (PSO) algorithm.

Chapter 4 provides a definition of the path-planning problem. This chapter mathematically models the underwater environment, optimization criterion and constraints, and the corresponding path-planning problem. Online Real-Time Path Planning (ORPP) system and re-planning based on changing environment will be designed and implemented in this chapter. The performance of the ORPP system in vehicle's accurate and safe deployment will be examined and validated using Firefly Algorithm (FA), PSO, BBO, and Differential Evolution (DE) algorithms.

Subsequently, Chapter 5 addresses problems and shortcomings associated with each planner by developing a novel Autonomous Reactive Mission Scheduling Path-planning (ARMSP) architecture with reactive re-planning capability. Mechanism and structure of this modular control architecture (including deliberative and reactive execution layers) will be explained in this chapter. After all, performance and stability of the ARMSP in satisfying the AUV's high and low level autonomy requirements in motion planning and mission management in vast dynamic environment, will be tested and evaluated using different combination of aforementioned meta-heuristic algorithms.

After all, Chapter 6 provides a summary of each section of the thesis and concludes the final outcome and achievements of this research. This chapter will also provide recommendations for further research. The operation diagram of the developed synchronous ARMSP architecture is illustrated by Figure 1.5.

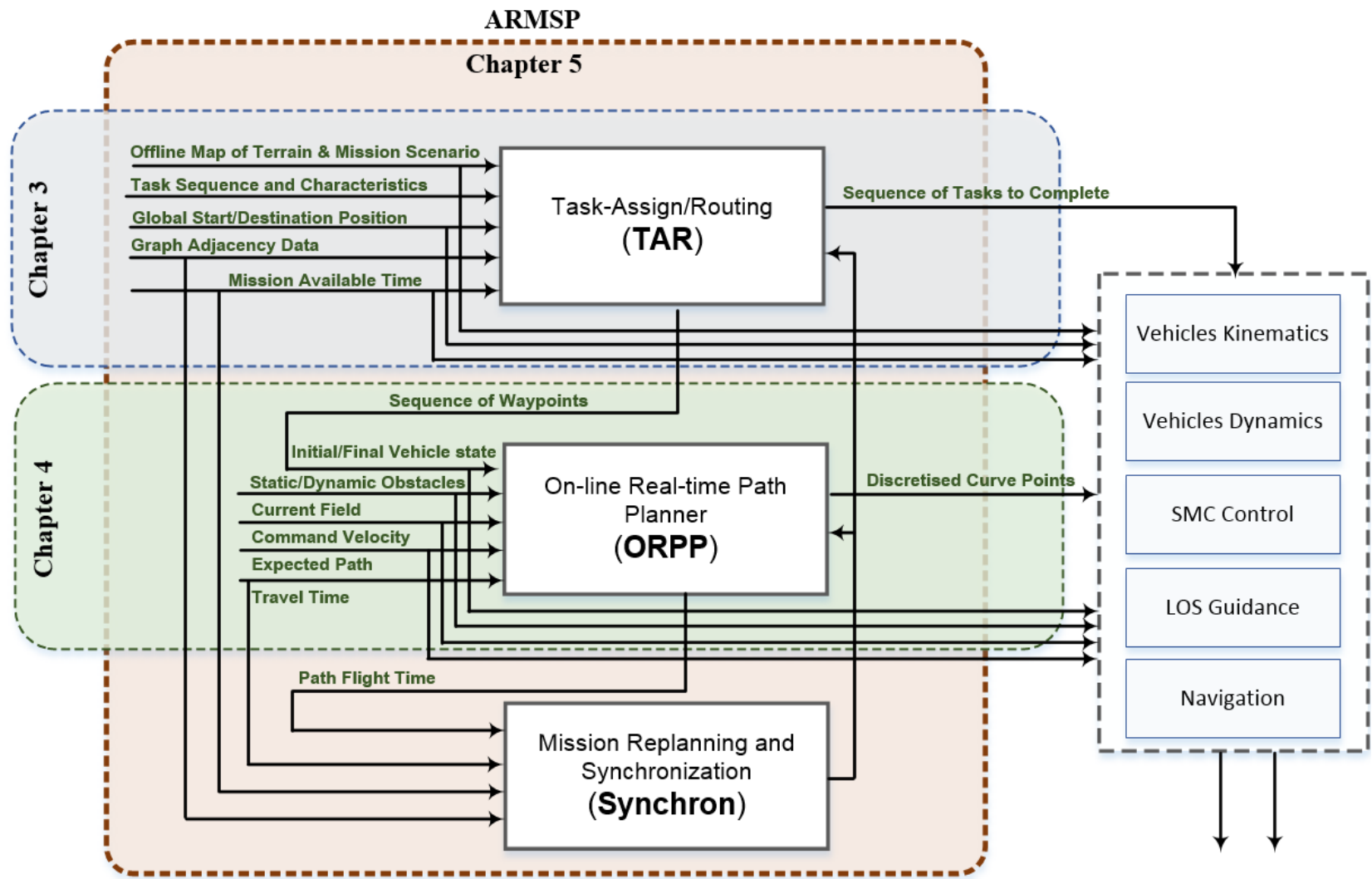

**Figure 1.5.** The operation diagram of the developed synchronous ARMSP architecture and its completion process by each chapter.

# Chapter 2

# Literature Review

## 2.1 Introduction

AUVs are largely employed for various purposes such as scientific underwater explorations [1], inspection and surveys [2], sampling and monitoring coastal areas [3,4], marine biodiversity mapping [16], offshore installations and mining [9], etc. However, majority of the available AUVs operate with a pre-programmed mission scenario while all parameters for the entire mission should be defined in advance and operator's interaction is necessary at all stages of a mission. For any of scientific, surveillance, mine, or military applications of AUV, a sequence of tasks is predefined and fed to vehicle through the series of commands (mission scenario) that limits the mission to executing a list of pre-programmed instructions and completing a predefined sequence of tasks. Accordingly, an advanced level of autonomy is an essential prerequisite to trade-off within importance of tasks and problem restrictions. An appropriate higher-level decision maker furnishes the expectation of such autonomy level.

On the other hand, safe and reliable deployment and handling uncertainties in constantly changing and complex undersea environments is another aspect of autonomy that needs to be addressed. Unforeseen and unpredictable conditions can enforce the mission to abort. This issue sometimes might even cause the loss of the vehicle, as occurred for Autosub2 that was lost under the Fimbul ice shell in Antarctic [17]. Failure of an AUV is not acceptable due to exorbitant expensive maintenance.

*CHAPTER 2. LITERATURE REVIEW*

Unexpected events may happen when operating within an unknown (partly known) environment and a vehicle must deal with these appropriately. Having a clear picture of the present situation and accurate projection of possibilities are essential requirements for safe operation and effective decision making in such an environment. For this purpose, a high level of Situation Awareness (SA) is required. SA is the process of sensing, comprehending and dealing with extremely dynamic and intricate environments. It is one of the most important factors required for facilitating the next generation of AUVs. The vehicle, thus, collects information from surrounding environment, make decisions based on such information and predefined knowledge. Afterward, it takes actions according to its decisions and then obtain feedback from the environment in response to the specific action, and update its knowledge to produce better decisions in the next step.

Current research aims to develop a comprehensive structure for providing situation awareness in order to promote the autonomous decision making for AUV mission management. To this end, an accurate mission planner and task allocator is developed to carry out the higher-level decision making procedure that increases the productivity of the AUV's operation and probability of success. On the other hand, for adapting the terrain changes during the operation, a robust motion planning approach is proposed to satisfy the situation awareness requirements of lower level operations. The following definitions are proposed in this study for clarification and to prevent confusion:

**Definition-1:** The mission planning in this study is a joint problem of discrete natured task priority assignment and contiguous natured vehicle routing toward the destination in graph-like terrain.

**Definition-2:** The path (motion) planning is a problem of guiding the vehicle between two specific points while dealing with dynamicity of the underwater environment.

In the rest of this chapter the importance of decision making based on situational awareness and its impact on improving vehicles autonomy is going to be discussed in Section 2.2. The relation of decision-making and autonomous mission-motion planning also will be clarified in this section. We are going to talk about what have been done for solving these problems (Definitions 1 and 2) in literature. The state of the art in mission planning including autonomous vehicle routing problem, task assignment and also challenges and shortcoming associated with these approaches is outlined in Section 2.3. Section 2.4 outlines the previous attempts in the scope of autonomous vehicles motion planning including path-route-trajectory planning and claims the existing challenges and open problems in this scope. There are several similarities between Unmanned Aerial Vehicles (UAVs) [18], AUVs and Unmanned Ground Vehicles (UGVs) in the context of the mission planning problem (Definition-1), and the path planning problem (Definition-2) so we are also going to discuss the way that these problems are addressed in UAV and UGV platforms as well. Section 2.5 presents the chapter's summary.

## 2.2 Autonomy, Decision Making and Situational Awareness

There is a big gap between concept of automatic and autonomous operations. In automatic systems, a machine follows programmed commands and it is not able to make a decision. In contrast, autonomous systems have the capability of recognizing different situations and making decision accordingly, without human interaction. Hence, developing more intelligent autonomous systems, which are capable of reconfiguring based on varying situation, and carrying out an efficient mission planning according to new situation, is one of the major fields of interest for many system designers [19].

Autonomous robotic platforms arouse more interest and broadly used in maritime approaches to achieve routine and permanent access to the underwater environment. However, their operations are still restricted to particular tasks with low-level autonomy. AUVs' communication is restricted by high latency and excessively low bandwidth. Therefore, AUV can communicate to an operator and exchange data only before or after the mission. Hence, the vehicle must be highly intelligent and trustable to carrying out the mission accurately [20].

A major concern about an AUV is its survivability by protecting itself and recovering from faults (if applicable) as the environment changes, which helps to maintain its operations at a near optimal performance. Highly complex approaches with high levels of autonomy still rely on expert's observation and control, so the platform always requires the involvement of an expert's knowledge and suggestions. Therefore, to make AUVs truly reliable and autonomous, more advanced situational awareness is required [21]. For a successful mission, the autonomous system should continuously monitor its resources and reorder the tasks to maximize mission performance during the operation. Thus, the system should dynamically track the available time, the lost time, environmental situations; then it should be able to manage its resources by adjusting its parameters according to the new situation. Consequently, it should reorder the tasks so that to be fitted to total available time for manoeuvre. Most of today AUV applications are supervised from a support vessel, in which an operator provides higher-level decisions in critical situation. This procedure costs highly expensive during a mission (e.g. communication cost, or human precision faults) [22]. Complicated missions that cannot be defined accurately in advance will need to be resolved through intermittent communication with a human supervisor. Undoubtedly, this will restrict the applicability and accuracy of such vehicles. Higher autonomy can improve AUVs' operation in the following areas:

− **Energy consumption**: optimize the power sources and vehicle's power consumption to improve its capabilities in long endurance missions.
− **Risk free and optimal navigation**: navigating and positioning properly with minimum position calibration error.
− **Robust decision making**: autonomy in this discipline can be defined as the capacity of sensing, interpreting and acting upon unexpected or unknown changes of environment. An autonomous vehicle should be capable of making efficient decisions when facing different challenging situations.

Autonomy in decision-making and navigation are connected in various standpoints. An AUV's capabilities in handling mission objectives are directly influenced by navigation system performance [23]. A fully autonomous vehicle should have capability to consider its own position as well as, its environment, to react properly to unexpected or dynamic circumstances.

*CHAPTER 2. LITERATURE REVIEW*

A growing attention has been devoted in recent years on increasing the ranges of missions, vehicles endurance, extending vehicles applicability, promoting vehicles autonomy to handle longer missions without supervision, and reducing operation costs [8]. Autonomous application of AUVs in performing different tasks in dynamic and continuously changing environment has not yet been completely fulfilled. In its absence, it is imperative for the operators to remain in the loop of considering and making decisions [17]. In addition, many efforts have been devoted in recent years for improving AUVs' autonomy in mission task assignment, vehicle routing, and path/trajectory planning; however, there are still many challenges on achieving a satisfactory level of autonomy for AUVs in this regard. Accordingly, various methodologies have been discussed and worked out in the literature is divided into the following two disciplines.

## 2.3 State-of-the-art in Autonomous Vehicle Routing Problem (VRP) and Task Assignment Approach (Definition-1)

Effective mission planning (routing-task assignment) refers to appropriate selection and arrangement of the tasks sequence considering time and other restrictions, which has a great impact on vehicle's time management as well as mission performance. The task assignment (allocation) involves a constraint decision-making procedures for one or multiple autonomous vehicles and is usually considered as a combinatorial optimization problem [24].

Generally, the task assignment can be a sub process in routing or sometimes it can be defined as an individual problem. For example, in a waypoint navigation problem routing procedure's job is to identify waypoints to derive the vehicle from its current location to destination while the task assignment can be some sub-routines that make sure point-to-point navigation is done appropriately and tasks that are mapped to different geographical areas are completed correctly. The vehicle routing and task assignment problem have been comprehensively investigated on various frameworks and different environments over the last two decades.

*CHAPTER 2. LITERATURE REVIEW*

Several strategies have been suggested for single or multiple vehicles routing-task assignment, such as graph matching algorithm [25], Tabu Search (TS) algorithm [26], partitioning method [27], and branch and cut algorithm [28]. Kwok et al., [25] used a graph based approach to maximize the utilization of a group of vehicles with a restricted sensor-coverage and restricted time, in which collaboration of the robots covers the limitation of their sensor-coverage. Higgins [26] applied TS for generalized assignment of 50000 tasks to 40 agents in a large-scale environment, in which the main focus is on controlling the CPU time for the whole process. Liu and Shell [27] also handled the similar challenge introduced in [26] for dynamic task assignment to multiple robots in which the change of location of tasks is managed applying a top-down partitioning strategy; however, solution quality is not perfectly satisfied in the simulation results.

Lysgaard et al., [29] employed branch-and-cut method to manage the cutting planes in a capacitated vehicle routing problem using different capacity criterion. Martinhon et al., [28] proposed a stronger Lagrangian-based K-tree approach for the vehicle routing problem in an undirected graph-like operation field and then applied a variable approximation based on linear programming to reduce the costs; but their proposed method does not appear competitive in computational time and cost evaluation when comparing with the recent branch and cut method in [29]. Zhu and Yang [30] developed an improved Self-Organizing Map (SOM)-based approach for multi-robots dynamic task assignment in which the vehicle route planner has been integrated with capability of task assignment. Sanem and Balch [31] presented a generic framework called Distributed and Efficient Multi Robot Cooperation Framework (DEMiR-CF) for distributed task allocation and to cover a complex mission for multi-robots involving simultaneous execution of number of tasks. A special model of Multi-Agent Reinforcement Learning (MARL) algorithms is proposed by [32] for a road network route planning system taking advantages of Q-value based dynamic programming (QVDP) to solve vehicle delay problems, in which the applied *m* method provides ability of learning, planning, recognizing different situations, and action selection through the trial and error mechanism; however, consistent information about the correct action is not always available.

Sadeghi and Kim [33] modelled vehicle route planning with a multi-criteria decision making mechanism and applied a generic ontology-based architecture taking the advantage of an analytic hierarchical process to determine the choice of criteria for applying an impedance function in the route finding algorithm. In the proposed strategy by [33], the domain specific ontology provides foundation for maintaining or extending the domain knowledge. An autonomous goal-based planning approach is presented by Patron and Ruiz [34] using a knowledge based framework, and service oriented architectures. The authors have shown that this methodology enables domain independent planning and capability of discovery for complex missions without experts' involvement in which the system can handle poor and intermittent communications by predicting the intent of the other platforms to accomplish goal selection [34]. The problem associated with knowledge based (ontology-based) systems is that consistent information about the correct action and environmental situations is not always available.

Dominik et al., [35] implemented a route-plan-decision-maker for multi-agent transportation planning, where routing problem is considered with a specific time windows, cluster-dependent tour starts, and the agent decides between distributing orders to customers, traversing edges, competing vendors, increasing production, etc. The outlined studies [25-35] focused on dividing jobs between multiple robots while they are moving towards a destination point and managing the loose (break down) of any of robots in the group by reassignment of its tasks in to the closest robots; however, in these studies, only it is assumed that the environment is ideal, while there are many other uncertainties and considerations in real world. Certainly, many problems will be revealed when it comes to practical applications especially in the case of autonomous unmanned robots.

There have been a number of relevant investigations in the scope of Unmanned Aerial Vehicle (UAVs) task-target assignment so far. Kingston [36] applied a decentralized algorithm to address task assignment path planning problem for a team of UAVs with moving targets, in which a velocity synthesis algorithm is employed for the purpose path planning. Alighanbari and How [37] presented a combinational approach of Robust Filter Embedded Task Assignment (RFETA) for robust path planning and dynamic task assignment of UAVs operating in an uncertain variable environment where two separate approach is proposed for vehicles' SA-based task assignment and the path planning-replanning. Kamil Tulum et al., [38] presented a situation aware agent based approach for proactive route planning problem by finding the best order of waypoints for an UAV using A* algorithm for optimizing the distance and battery cost over the large scale operation network. This approach highlighted the relation between route planning and situation awareness; however, in this research only a restricted tactical situations are considered that is not sufficient to furnish all real world possibilities. The technique needs a pre-computation to analyse the situation due to high computational cost of the applied A*, which is a substantial deficiency for a real-time platform.

Theopisti and Richards [39] used a modified Ant Colony Optimization (ACO) for handling both task allocation, task loitering, and route planning for multiple UAVs. A multiple UAVs' cooperative task assignment problem is efficiently addressed by Tao et al., [40] applying an improved ACO, where a task tree is applied to carry out the decision process of choosing the tasks. Kulatunga et al., [41] also introduced a solution for simultaneous task assignment and route planning of multi automated guided vehicles applying an ACO based meta-heuristic algorithm, in which ACO based approach compared with exhaustive search mechanism for small size problem to evaluate performance of the ACO in this research. Acknowledging the relevant studies in the scope of ground and aerial vehicles, the mission management becomes even further challenging considering the severity and uncertainties of underwater environment.

The underwater environment is generally a vast 3D area, which is rarely populated with obstacles and sparsely known in advance [23]. The challenges associated with such an uncertain environment becomes even greater in long range missions with enlargement of the operation area. Robustness of AUV to strong environmental variability is a key element to mission performance and safe deployment. Restrictions of priori knowledge about later conditions of the environment reduces AUVs autonomy and robustness.

*CHAPTER 2. LITERATURE REVIEW*

In the scope of underwater operations, Karimanzira et al., [42] presented a behaviour-based controller coupled with waypoint tracking scheme for an AUV guidance in large-scale underwater environment. Yan et al., [43] developed an integrated mission planning strategy to serve the AUVs' routing problem, while minimising total energy cost in the presence of ocean currents. Later on the proposed method in [42] is generalized by Iori and Ledesma [44] for double vehicle routing problem with multiple stacks and the AUV routing problem is modelled with a Double Traveling Salesman Problem with Multiple Stacks (DTSPMS) for a single-vehicle pickup-and-delivery problem. They studied the performance of three branch-and-cut, branch-and-price, and branch-and-price-and-cut algorithms on a wide family of benchmark test samples with different graph complexities. Chow [45] proposed an approximated algorithm based on *k*-means method to address task assignment problem that minimizes the travel time for multiple AUVs considering presence of constant ocean current. The approach combined the AUV dynamic model with adjusted Dubins model including ocean currents for route planning and to control the AUVs to move along the route with a constant velocity with respect to the kinematics constraint of AUVs. However, the problem is considered in a 2-D environment that would be problematic when the method is applied to accomplish the similar task in 3-D environment. Moreover, they considered a static ocean current rather than the variable current, which is not prevalent in real world application.

Later on, Zhu et al., [46] integrated improved self-organizing map (SOM) neural network with a velocity synthesis approach for solving multiple AUVs' dynamic task assignment and path planning problem. They considered multiple target locations in a 3-D environment in presence of dynamic current. Although the approach in [46] is capable of keeping an AUV on the desired track during the mission, but the environment is assumed ideal, while there are many other uncertainties and considerations in real world, such as static or moving obstacles and so on. Williams [47] developed an adaptive track-spacing algorithm for an AUV optimal route planning approach used in mine countermeasures mission in which the operation is framed in terms of maximizing underwater mines detection. No on-board processing capability is incorporated in the route planning resulting in non-adaptiveness of the approach to dynamic changes.

The AUV's routing-task assignment problem has a combinatorial nature analogous to both TSP-BKP, and categorized as a Non-deterministic Polynomial-time (NP) hard problem. Many knowledge based strategies, deterministic algorithms and graph search methods have been introduced for solving the vehicle routing or task assignment problems, as mentioned above. The deterministic methods produce better quality solutions; however, these algorithms are computationally complex [48]. Therefore, they are not appropriate approaches for real-time routing applications, especially when the search space is large or the operating network is topologically complex [48]. In contrast, the meta-heuristic methods take less computation time and obtains optimal or near optimal solutions quickly.

Several studies also investigated the advantages of meta-heuristics methods on vehicle routing and task assignment. A comparative Dijkstra-GA approach is used by Sharma et al., [49] to address the vehicle routing problem in a graph like operation network. The evaluation results affirmed that both algorithms provided same solution but with different execution time as Dijkstra's is time-consuming comparing to GA.

A large scale route planning and task assignment joint problem related to the AUV activity has been investigated by M.Zadeh et al., [50] transforming the problem space into a NP-hard graph context. The approach used heuristic search nature of Imperialist Competitive Algorithm (ICA) and GA to find the best waypoints for AUV task assignment and risk management joint problem in a large-scale static terrain. Later on, the approach was extended by M.Zadeh et al., [51] to be implemented on semi dynamic network by taking the use of Biogeography-Based Optimization (BBO) and Particle Swarm Optimization (PSO) algorithms. The merit of the proposed solution by [50,51] is that being independent of the graph size and complexity.

As a modern heuristic algorithm, ACO is also widely used to solve combinatorial task assignment and routing problems. The real-time performance of the application is usually overshadowed by growth of the graph complexity or problem search space. Growth of the search space increases the computational burden that is often a problematic issue with deterministic methods such as Mixed Integer Linear Programming (MILP) proposed by Yilmaz et al., [52] for governing multiple AUVs. The meta-heuristic algorithms offer near-optimal solutions in faster time and with better scalability with number of tasks comparing to deterministic method such as MILP. The MILP is inaccurate in large sized problems as its computational time increases exponentially with the problem size. Table 2.1 and Table 2.2 survey different methods suggested for various autonomous aerial, ground, surface and underwater vehicles.

**Table 2.1**. Autonomous Vehicle Routing Problem (VRP) with/without Task Assignment (TA): beyond AUVs

| Techniques | | Ref | Scope | Comments |
|---|---|---|---|---|
| **Graph Matching algorithm** | | 22 | To cover restricted sensor-coverage and time management of multi ground vehicles (VRP+TA) | Solution quality is not perfectly satisfied in the simulation results. Computationally inaccurate in large size problems. |
| **Tabu Search (TS) algorithm** | | 23 | To control CPU time for process of task assignment to multiple agents (VRP+TA) | |
| **Partitioning Method** | | 24 | Dynamic task assignment to multiple robots (TA) | |
| **Branch and Cut algorithm** | | 25 | To manage the cutting planes in a capacitated (VRP) | Does not appear competitive in computational time and cost evaluation. |
| **Self-Organizing Map (SOM)-based approach** | | 27 | Multi-robots dynamic task assignment and routing (VRP+TA) | The problem associated with knowledge (ontology-based) based systems is that consistent information about the correct action and environmental situations is not always available. It is assumed that the environment is ideal, while there are many other uncertainties and considerations in real world, especially in the case of autonomous unmanned robots. |
| **Generic framework of DEMiR-CF** | | 28 | | |
| **Multi-Agent Reinforcement Learning (MARL) algorithms** | | 29 | Road network route planning system to solve vehicle delay problems (VRP) | |
| **Ontology-based multi-criteria decision making mechanism** | | 30 | The VRP approach in a graph-like operation network | |
| **Knowledge based planning framework** | | 31 | For multi-agent transportation planning with a specific time windows (VRP) | |
| **Knowledge based route-plan-decision-maker** | | 32 | | |
| **Decentralized algorithm** | | 33 | Task assignment path planning problem for a team of UAVs' with moving targets | Appropriate for aerial operations but not for underwater manoeuvre due to environmental and communication restrictions. |
| **Robust Filter Embedded Task Assignment (RFETA) approach** | | 34 | For multi UAVs' SA-based task assignment and the path planning-replanning (VRP+TA) | |
| **A\* Situation Aware Agent Based approach** | | 35 | Proactive route planning for an UAV and optimizing the distance and battery cost. Highlighted the relation between VRP and situation awareness. | Only a restricted tactical situation are considered that is not sufficient to furnish all real world possibilities. Needs a pre-computation to analysis the situation and not appropriate for a real-time platform. |
| **Evolutionary Approaches** | **ACO** | 36 | Task allocation, task loitering, and route planning for multiple UAVs | Offers near-optimal solutions in faster time with better scalability in number of tasks. Appropriate for high-dimensional problems, but may converge to local optima in a finite time. |
| | | 37 | A multiple UAVs' cooperative task allocation problem | |
| | | 38 | Simultaneous task allocation and route planning of multi automated guided vehicles | |

**Table 2.2**. Autonomous Vehicle Routing Problem (VRP) with/without Task Assignment (TA): AUVs

| Techniques | | Ref | Scope | Comments |
|---|---|---|---|---|
| **Behaviour Based Approach** | | 39 | AUV guidance and routing in large-scale graph-like terrain. | Consistent information about the correct action and environmental situations is not always available. |
| | | 41 | Single AUV pickup-and-delivery problem by minimizing the total routing cost in which the performance with different graph complexities. | |
| **Branch and Bound with Velocity Synthesis Approach** | | 40 | To serve the AUVs routing problem in order to deliver customized sensor packages at scattered positions, while minimizing total energy cost | Uncertainty of the underwater environment is not addressed properly. Does not appear competitive in computational time and cost evaluation. |
| **Approximated Algorithm Based on K-means Method** | | 42 | To address task assignment problem that minimizes the travel time for multiple AUVs considering presence of constant ocean current | The problem considered in a 2D environment that is not sufficient to model 3D environmental challenges. Only static ocean current rather is considered. Uncertainty of the underwater environment is not considered. |
| **SOM-based Neural Network with Velocity Synthesis Approach** | | 43 | For multiple AUVs' dynamic task assignment and motion planning in a 3D environment. | The environment is assumed to be ideal, while there are many other uncertainties and considerations in real world, such as static or moving obstacles. |
| **Adaptive Track-Spacing Algorithm** | | 44 | AUV route planning approach in mine countermeasures mission | In this study no on-board processing capabilities covered in the route planning approach, therefore, the method is not adaptive to dynamic changes. |
| **Mixed Integer Linear Programming (MILP)** | | 49 | Governing multiple AUVs routing and task assignment | The MILP is inaccurate in large sized problems as its computational time increases exponentially with the problem size. |
| **Evolutionary Approaches** | GA- Dijkstra | 46 | AUV optimal route planning and task assignment approach | Offers near-optimal solutions in faster time for large-scale problems. Uncertainty of the underwater environment is not addressed properly. Ocean dynamics also is not considered. |
| | ICA-GA | 47 | | |
| | BBO-PSO | 48 | | |

*CHAPTER 2. LITERATURE REVIEW*

An important issue with all studies outlined in this scope is that they mostly focus on routing problem for different unmanned vehicles in which task assignment is the principle direction of these papers and quality of deployment (motion) and uncertainty of the environment have not been addressed. The vehicle's safe and confident deployment in hazardous underwater environment is a critical factor that should be taken into consideration at all stages of the mission.

### 2.3.1 Open Problems and Research Challenges in the Scope of Autonomous Mission Planning (Task Assignment and Routing: Definition-1)

AUVs capabilities in handling mission objectives are directly influenced by performance of the routing and task assigning system. The route planning system should direct vehicle(s) to a predefined destination while providing efficient manoeuvres and managing the travel time. Effective routing has a great impact on vehicles time management as well as mission performance by appropriate selection and arrangement of the tasks sequence. Some open problems in this regard are pointed as follows:

− The real-time aspect of a route planner is extremely critical to satisfy time constraints of the critical-time missions and to provide online redirection of the AUV. The complexity of the graph topology or in general problem space results extensive computational burden. Developing a fast and efficient mission planning approach for satisfying the real-time requirements of an autonomous operation is still an open area for research.

− A route or a task sequence should be generated during the mission as the vehicle proceeds through the dynamic environment. Majority of previous attempts resulted in developing offline mode of route planning solutions also known as pre-generative strategies. Many typical AUV missions are limited to executing a list of pre-programmed instructions and completing a predefined sequence of tasks. Considering environmental changes and situation updates, re-planning and rescheduling during the mission would be a necessary requirement for autonomous operations. Such circumstances pose major difficulties on offline pre-generative planning strategies when the system is required to dynamically and in real-time cope with the unexpected situations.

− The vehicle routing problem is usually investigated as a shortest route problem in a graph with a specified start and target point, which is a fundamental problem in graph theory and is a field of interest of many researches in various applications of transportation, communications, robot motion planning, network routing, and so on [53-55]. Complicated missions such as logistic applications of an AUV, arise requisition for advance mission scheduling and task organizing strategy to maximize productivity of the mission in a restricted period of time. The system's productivity relies on a number of factors such as vehicles autonomy in decision-making based on raised situation, real-time performance of the applied method to reschedule the mission if required, battery capacity and resource management. So having an intelligent mission planner for fully autonomous operations of AUVs is another area of interest for further research that is not perfectly fulfilled yet.

− The majority of research mentioned is particularly focused on task and target assignment and time scheduling problems without considering requirements for vehicle's safe deployment or quality of its motion in presence of environmental disturbances. As AUV operates in an uncertain environment, there is a huge amount of uncertainty in the travel times that can have a devastating effect on mission plans and mission time. Proper time management is necessary to ensure on-time mission completion and consequently the mission success. To improve mission reliability and performance, and to avoid mission failure, environmental risks and factors should be considered at all stages of the mission for having a safe and confident deployment in a vast and uncertain environment. Therefore, in the next subsection, existing AUV trajectory/path planning approaches (Definition-2) are discussed, which are more concentrated on vehicles deployment encountering dynamicity of the ocean environment.

## 2.4 Autonomous Vehicles Motion Planning (Definition-2)

Motion planning on various frameworks and different environments has been investigated comprehensively over the last two decades. The identical aspect of motion planning systems (in Definition-2) is that they are obligated to react to a dynamic environment that continuously changes its state. Proper and realistic modelling of the environment, having practical assumptions, and selecting efficient searching strategy are main steps of each motion planning approach. The first step toward achieving a satisfactory operation is to find a suitable and efficient method of path planning. In general, the challenges associated with path planning can be considered from two perspectives; firstly, the motion planning methodologies being utilized (subsection 2.4.3), and secondly the competency of the techniques for real-time applications (subsection 2.4.4). Respectively, the prior investigations in this area are considered in the following disciplines.

### 2.4.1 Path Construction Methods

Generally, a potential path is generated by connecting a set of ordered control points that are initialized in advance in the defined operational space (2D or 3D). There are popular techniques for connecting the defined control points namely: **Straight-line method** [56], which is not flexible enough to provide a smooth transition in linkage of piece lines; **Dubins categories** of paths [57] that intercalate circles around the connections so that the path gets smoother in connections of control points. This method caught much interest in mobile robot path planning and it is widely used in past decade [58-60]. The Dubins-based methods are criticized for their high computational cost, which defects the real-time performance of the system especially in path re-planning process [61].

Another popular category of path construction methods is the **spline** and **piecewise polynomial** strategy that use just a few variables of control points' peculiarities to construct a very flexible path curve. Obviously, dealing with small number of variables boosts the computation speed and eases the optimization process. This method is modified to be more efficient and proposed in different sub-groups such as Cubic

Hermite Spline [62]; B-splines [63-65]. A real-time modified version of spline is also proposed by [66].

Between all mentioned methods, the B-Spline is advantageous in controlling the direction of the curve (vehicle) at the starting and ending points by adding fixed points after start point or before ending point. Considering the physical constraints of the vehicle, this characteristic of B-Spline method makes it appropriate for AUV's path planning. Accordingly, the B-Spline method of path construction is utilized by this study, which is detailed in Chapter 4.

### 2.4.2 Path Planning and Optimization

In the robotics prospective, the path planning or in general the motion planning is introduced as a multi-objective constraint optimization problem that aims to find a valid optimal trajectory to autonomously guide the robot towards the target of interest in the corresponding operating field [65,67]. Optimization is a process defined based on important criterion and requirements for any specific problem. Usually, travelling time and distance, minimum energy consumption, or in some cases safety options get used as important criteria for evaluation of the optimal path [68, 69].

Many attempts have been devoted in recent years and different techniques have been suggested for autonomous robots' motion planning. The efficiency of the path planning techniques involves two perspectives: first the techniques that mostly emphasize on optimality and resolution of the solutions; and second group of techniques majorly focus on producing feasible and probabilistically near optimal solutions with minimum computational cost. Resolution based motion planners use graph search algorithms, while the second group involves evolutionary algorithms or many sample-based methods that aim to satisfy the real-time constraints of autonomous operations. Each of these subgroups are appropriate for different purposes according to problems requirements of quality resolution or computation-time restrictions.

### 2.4.3 State-of-the-art in Autonomous Motion Planning: The Methodological Point of View

Solutions proposed for motion planning can be categorized in to two main groups of: A) classical and graph (grid) based technique; and B) artificial intelligence and evolutionary algorithms based techniques [70].

### A) Classical and Graph Search Techniques

The motion planning approaches that are based on graph search techniques use a discrete optimal planning on a graph-like search space [71]. To apply graph search techniques on path planning problem, regarding the discrete nature of these algorithms, the search space should be decomposed to grid-shape or transformed to a graph format [72]. The nature of these algorithms are well suited on vehicle routing problems in which the default form of the search space is a weighted directional or bidirectional graph [72].

Various graph search algorithms like Dijkstra, A* or D* algorithms have been applied to the autonomous vehicles path planning problem in recent years [72-78]. In this group of algorithms, the heuristic function estimates the best cost solution and the algorithm keeps track of the solutions to determine the best possible cost equal to the estimated cost. To apply these algorithms on path planning problem, first the search space should be decomposed to a graph feature to specify the collision boundaries and forbidden sections, and then the path is produced by patching the free sections based on defined cost function.

The Dijkstra algorithm is a popular and efficient method for shortest path computation and graph routing problems. Eichhorn [79] implemented graph-based methods for the AUV ''SLOCUM Glider'' routing in a dynamic environment. The author employed modified Dijkstra Algorithm where the applied modification and conducted time variant cost function simplifies the determination of a time-optimal route in the geometrical graph. The Dijkstra algorithm calculates all applicable paths from the start point to the destination, so it takes long time to get to the appropriate answer when the graph size or the problem search space increases; so a large amount of computations are required to be repeated. This issue makes the Dijkstra computationally expensive and unsuitable for real-time applications [49, 80].

A* is categorized as a best-first heuristic search method and acts more efficiently because of its heuristic searching capability [74]. Carroll et al., [75] presented the operation space with quad trees and applied A* to obtain efficient paths with minimum cost. Fernandez et al., [76] and Pereira et al., [77] utilized A* method to provide a time optimal path with minimum risk for a glider path planning according to an offline data set. Pereira et al., [72] developed an A* based Risk-aware path planning method for AUVs operation in a high boat traffic areas applying current prediction models. The A* method also suffers from expensive computational cost in larger search spaces.

D* algorithm (a variation of A*) is an alternative for AUV's path planning problem. D* operates based on a linear interpolation-based strategy and allows frequently update of heading directions [78]. D* is incapable of posing considerable improvement on the A* computational complexity in high-dimensional problems [80].

The Fast Marching (FM) algorithm is another derivative of the A* approach proposed to solve the AUV path planning problem that uses a first order numerical approximation of the nonlinear Eikonal equation. Petres [81] suggested FM algorithm to capture a time optimum path taking current field and dynamic environment into account. FM is accurate but also computationally more expensive than A* [81]. An upgraded version of FM known as FM* or heuristically guided FM is employed for AUV path planning problem [82]. The FM* keeps the efficiency of the FM along with the accuracy of the A* algorithm, but it is restricted to the use of linear anisotropic cost to attain computational efficiency. Improved version of the FM* also has been developed applying the wave front expansion to determine the time optimum path from vehicles start point to the destination [83-85].

The grid-search-based methods are criticized due to their discrete state transitions, which restrict the vehicle's motion to discrete set of directions. The grid-based strategies are also inefficient in cases that workspace is too large or complex. In such problems the number of cells comprehensively increases, therefore rendering of the data and solutions becomes intractable. On the other hand, graph-based techniques, which are very popular, usually look for the shortest route between two points in a network (graph). The main drawback of these methods is that their time complexity increases exponentially with increasing problem space, which is inappropriate for real-time applications. These methods are time consuming because of redundant computations. Reaching to a reasonable solution takes too long due to several recalculations of the criteria on each iteration and usually in these types of optimizers, many iterations are required, therefore computing an effective path becomes a very time consuming task [86].

For a group of problems in which the deterministic techniques and heuristic-grid search approaches are not capable of satisfying real-time requirements, the artificial intelligence and evolution-based methods are good alternatives with fast computational speed, specifically in dealing with multi-objective optimization problems.

### B) Artificial Potential Field Method

One popular approach suggested by many researches to deal with complexity of path planning problem is the Artificial Potential Field (APF) method that keeps the vehicle from crossing the collision boundaries by defining artificial potential fields for each obstacle in the terrain. This method is applied on the multi-AUV motion-planning problem where static obstacles have been considered to test the collision avoidance applicability of the method; nevertheless, static obstacles are not sufficient to model the real underwater hazardous environment [87]. The APF is also applied by Witt and Dunbabin [88], Warren [89] and Karuger et al., [90] for AUV path planning problem using a combinatorial cost function for collision avoidance involving decision factors of time, distance, collision regions, and energy consumption. Cheng et al., [91] utilized APF and velocity synthesis method for AUV path planning problem considering ocean current variations. The [91] did not consider the dynamic changes of the environment including variations of dynamic and moving obstacles. Although APF is fast, cost effective, and computationally efficient, but it produces locally optimal solutions. It is computationally inexpensive and by nature capable of handling any dynamic environment. APF does not always satisfy the shortest path criterion. It will guarantee to reach to the destination if there is not time restriction.

### C) Evolutionary Algorithms

Evolution-based or in general the meta-heuristic algorithms are another set of approaches utilized successfully in path planning problem. Meta-heuristic optimization algorithms are efficient methods in dealing with path planning problems of an NP-hard nature. For the AUV path-planning problem in a large-scale operation area, satisfying time/energy and collision constraints is more important than providing exact optimal solution; hence, having a quick acceptable path that satisfies all constraints is more appropriate rather than taking a long computational time to find the best path. Meta-heuristic optimization algorithms are capable of being implemented on a parallel machine with multiple processors, which can speed up the computation process; hence, these methods are fast enough to satisfy time restrictions of the real-time applications [92].

Meta-heuristics and evolution-based path planning also catch a lot of interests in recent years. In [68], a Genetic Algorithm (GA) is utilized by Alvarez et al., to determine an energy efficient path for an AUV encountering a strong time varying ocean current field and a grid partitioning method is applied on the search space to avoid convergence to local cost minima. Rubio and Kragelund used GA for UAVs' optimal path re-planning according to prediction of wind velocities, in which a combination of fuel consumption characteristics and desirable relative wind speed been used in cost calculation to reduce the fuel consumption for the entire mission [93]. Zhu and Luo used an improved GA to address the shortest path problem as a combinatorial optimization problem, where the Dijkstra algorithm is applied as a local search operator to generate local shortest path trees that improves the performance of the individuals in the population [94]. Later on, an improved version of GA called Improved Niche Genetic Algorithm (INGA) is employed for carrying out the real-time route planning of a group of UAVs in a 3D environment [95]. The $\delta$-field perturbation operator has been used to improve the local search capabilities of INGA and also adaptive k-means analysis method as a crowding strategy is used to generate coverage paths in the area of interest.

Roberge et al., [92] developed an energy efficient path planner using meta-heuristic search nature of the Particle Swarm Optimization (PSO) algorithm, where time-varying ocean current is encountered in the path planning process. GA and PSO share a similar population-based search mechanism in which a set of probabilistic and deterministic rules are applied to enhance the solutions' quality. PSO differs from GA in the sense of having a stochastic evolution mechanism that does not guarantee the best fitted solution while GA incorporates this problem using mutation and crossover operators. PSO retains particles in each iteration and just adjust their characteristics applying update rules on their velocities and positions to best fit them to the cost function. Further comparison of PSO and GA can be found in [96]. Zamuda et al., [97] also applied a Differential Evolution (DE) based path planner for underwater glider for opportunistic sampling of dynamic mesoscale ocean structures considering dynamic ocean structure.

The main drawback of the evolutionary algorithms is that they may quickly converge to a suboptimal solution. Although suggested evolution based approaches properly satisfy the path planning constraints for autonomous vehicles, AUV-oriented applications still require improvements.

## 2.4.4 State-of-the-art in Autonomous Motion Planning: The Technical Point of View

The underwater environment poses several challenges for AUV deployment such as varying ocean currents, no-fly zones, and moving obstacles, which usually may not be fully known and characterized at the beginning of the mission. Obstacles may appear suddenly or change behaviour as the vehicle moves through the environment. Water current may have positive or disturbing effects on vehicle's deployment, where an undesirable current is a disturbance itself and it also can push floating objects across the AUV's path. However, in most research the hydrodynamic effects of underwater environment and AUV's limited turning ability is ignored or not sufficiently considered [98-100].

Current flow sometimes can be very strong that cannot be controlled by the actuator of the AUV, so that avoidance of such zones would be necessary for safe deployment. Ignoring the water current has detrimental effect on fuel consumption and even may cause serious risks for vehicle's operation. The vehicle should be autonomous enough to cope dynamic changes over the time while it is continuing its mission objectives. Spatio-temporal wave front and sliding wave front expansion algorithm with a contiguous optimization technique has been employed by Thompson et al., [101] and Soulignac et al., [83,84] for AUVs path planning in presence of strong current fields. These methods are capable of providing optimum path for AUV according to some specific requirements and assumptions and use previous information for re-planning process, which is computationally reasonable in generating accurate local trajectory. Nevertheless, this strategy may not be appropriate in dynamic environments where the current field updates continuously during the operation. The first online obstacle avoidance is developed by [102] for path planning of a remotely operated underwater vehicle based on real-world dataset captured by multi-beam forward looking sonar. Gautam and Ramanathan [103] proposed a path planner for SLOCUM Glider operations in near-bottom ocean, in which the path planner is simulated using Q-learning technique and biologically-based GA, ACO, PSO algorithms. Despite of using realistic information, capacity of these approaches in online path planning and simultaneous mapping is not substantiated.

The recent investigations on path planning encountering variability of the environment have assumed that planning is carried out with perfect knowledge of probable future changes of the environment [104, 105], while in reality, environmental events such as current's prediction or obstacles' state variations are usually difficult to predict accurately. Ocean currents also show great variability over such durations in a large operation field. Even though available ocean predictive approaches operate reasonably well in small scales and over short time periods, they are not sufficiently accurate over long time periods in larger scales [72]. Fu et al., [106] employed Q-PSO for UAV's path planning and obtained satisfactory outcome in this regard. However, Fu implemented only off-line path planning in a static and known environment, which is not enough for representing a real world environment. Later on, the Q-PSO algorithm has been employed by Zeng and colleagues [107,108] for AUVs on-line path planning in dynamic marine environment.

To deal with challenges of large-scale operations, Zhang [109] investigated the hierarchical global GA based path planning approach for AUVs, which relies on decomposition of the operational environment and searching for the best path at each level. Sequeira and Ribeiro, [110,111], decreased the operational search space to a graph with limited connected nodes known as obstacle regions, in which sea currents considered in a simplistic fashion as static current with constant velocity and direction. This study [111] assigned energy costs as weight to graph edges and applied Dijkstra algorithm to address the AUV's motion planning with minimum energy cost taking undesired current correlation in to account.

Garau et al., [112] proposed similar consideration but restricted to motion on a 2D environment where the operation area is divided into a large and abrupt grid of connected nodes assigned with energy cost. In [112], the energy cost is calculated with respect to journey time assuming constant motion speed. Afterward, A* search algorithm is used to find the minimum cost route in the graph. Although the satisfactory results are obtained from this simulation, however, a 2D representation of a marine environment does not sufficiently embody all the information of a 3D ocean environment and vehicle motion with six degrees of freedom. Moreover, the applied method in [112] does not consider using desirable currents for vehicles motion, which is important for prolonging mission lifetimes.

An advanced path planning should be able to generate energy efficient trajectories by using desirable current flow and avoiding undesirable current. Desirable current can propel the vehicle on its trajectory and leads saving more energy. This issue considerably diminishes the total AUV mission costs. The importance of the ocean current in saving energy was emphasized by Alvarez et al., in which a GA based path planning was investigated for finding a safe path with minimum energy cost in a variable ocean. An A* based path planer is used by Koay and Chitreaiming to find path with minimum energy consumption considering varying ocean current [113]. A non-linear least squares optimization technique is employed by Gonzalez et al., for AUV path planning to suggest the time-optimal trajectory taking changing current into account [114]. Later on, an offline pre-generative motion planning strategy based on the Non-Dominated Sorting Genetic Algorithm (NSGA-II) is proposed by Ataei et al., for a miniature AUV in 3D underwater environment [115]. They considered the path planning approach as multi objective optimization problem encountering four criteria of length of path, motion planer smoothness, margin of safety and gradient of diving. However, the proposed offline motion planner in [115] gets in to serious difficulties when the environmental situations are updated and re-planning is required.

Although useful meta-heuristics have been suggested for autonomous vehicles path planning, AUV-oriented applications still has several difficulties when operating across a large-scale geographical area. The computational complexity grows exponentially with enlargement of search space dimensions. A huge data load from whole environment should be analysed continuously every time that path re-planning is required, which is computationally inefficient. Table 2.3 and Table 2.4 present survey results for the different methods proposed for AUV path planning and re-planning problems.

**Table 2.3.** Autonomous vehicles' motion/path planning from algorithmic point of view

| Optimization Techniques | | Reference | Advantage | Comments |
|---|---|---|---|---|
| **Graph Search and Heuristic Methods** | **Dijkstra** | [73,79] | Higher accuracy and resolution | Discrete state transitions; Expensive computational cost; |
| | **A*** | [72,74,75,76] [104,111] | | Computationally better that Dijkstra, but still expensive; |
| | **D*** | [78] | | Operates based on a linear interpolation-based strategy; Allows frequently update of heading directions; Computationally better than A*, but still expensive in high-dimensional problems; |
| | **FM-FM*** | [80,81,82,83,84,85] | | FM uses a first order numerical approximation of the nonlinear Eikonal equation; FM is computationally expensive than A*; FM* is qualitatively better than FM and A*, but expensive in large dimension problems; Uses linear cost function to preserve computational efficiency; |
| **APF** | | [87,88,89,90,91] | | Computationally effective, but sensitive to local optima; Not efficient in copping dynamic changes of the ocean; |
| **Evolutionary Approaches** | **GA** | [68,93,94] | Faster computation and Probabilistic resolution | Computationally efficient and cost effective in in high-dimensional problems; May converge to a suboptimal solution within a finite time; |
| | **INGA** | [95] | | |
| | **NSGA-II** | [115] | | |
| | **PSO** | [92] | | |
| | **QPSO** | [106,107,108] | | |
| | **DE** | [97] | | |

**Table 2.4**. Autonomous vehicles' motion/path planning from technical point of view

| Optimization Techniques | Reference | Comments |
|---|---|---|
| PSO | [98] | Ocean dynamics and water current effects on vehicle's motion is not considered; Offline route generation strategy is used that get in to serious difficulties when re-planning is need due to dynamic nature of the underwater environment; |
| BBO | [99] | |
| GA-PSO | [100] | |
| NSGA-II | [115] | |
| Level Set (LS) Method | [61] | Deterministic methods are efficient for small scale spaces, but remarkably less effective in larger dimensions; |
| Gradient-based Method | [90] | |
| QPSO | [106] | only off-line path planning based on known environment is considered; |
| QPSO | [107,108] | Although this method is computationally efficient due to use of evolution mechanism, but computational burden increases when re-planning is required in large dimensions; |
| Sliding Wave Front Expansion Algorithm | [83,84] | This strategy may not appropriate in dynamic environments where the current field updates continuously during the operation; |
| Q-learning Technique and GA, ACO, PSO | [103] | Despite of using realistic information, capacity of these approaches in online path planning and simultaneous mapping is not substantiated; |
| A* algorithm | [112] | Restricted to motion on a 2D spatio that doesn't sufficiently embody all the information of a 3D ocean environment and vehicle's motion with six dof; Did not consider effect of desirable current on vehicle's motion; |
| A* algorithm | [75,76] | Assumed perfect knowledge of the environment is available; |
| A* algorithm | [104] | |
| Centroid-Tracking, Boundary-Tracking, Ocean Plume Tracking Algorithm BuiLt Ocean Model Predictions | [105] | Assumed that planning is carried out with perfect knowledge of probable future changes of the environment, while the in reality environmental events such as currents prediction or obstacles state variations are usually difficult to predict accurately, especially in larger dimensions; Produce insufficiency of accuracy to current prediction over long time periods in larger scales; |
| Hierarchical GA based approach | [109] | Dynamic changes of the ocean and current variations are not considered; Re-planning is not developed for adaption to changes of the terrain; |
| Dijkstra | [110,111] | Sea currents considered in a simplistic fashion as static current with constant velocity and direction; Problematic in larger scales due to use of Dijkstra; |

## 2.4.5 Open Problems of AUVs' Motion Planning Approach

A remarkable effort has been devoted in recent years to improve the level of autonomy for AUVs motion planning and extending their capability for longer operations. The previous section has discussed the current state of the art in the AUV guidance and motion-planning problem. The important issues with current AUV path planners and also critical hints for improving the level of autonomy in motion planning are concluded below:

− The existing planning studies in AUV platform are predominantly concentrated on short-range local path planning in small-scale areas. Widening of the operating range, however, is intertwined with the appearance of new complexities, for instance propagation of uncertainty and accumulation of data that turns the underwater mission to a challenging scenario. Only a limited number of previous studies addressed the problems of long-range autonomous motion planning either in terms of vehicle routing in a graph-like network or in the path-trajectory planning. Previous emphasizes mostly approach this problem by reducing complexity of the network by setting small number of nodes (waypoints) or 2D implementation of the operation field. The problems associated with each assumption is discussed earlier in Section 2.4.3 and 2.4.4.

− Usually information about obstacles' characteristics (position, velocity) is not certain and perfect. An obstacle's position or velocity may change over time or may appear suddenly during vehicle's deployment. Obstacles may drift by current force or may have a self-motivated speed and direction. The moving obstacles also can be categorized as intelligent agents so that in such a case prediction of its behaviour would be impractical. Hence, to carry out collision avoidance in such an uncertain environment the vehicle requires real-time capability of path computation. Moreover, current variation can change the behaviour of moving obstacles over the time. Therefore, proper estimation of the behaviour of a large-scale dynamic uncertain terrain, far-off the sensor coverage is impractical and unreliable, so any pre-planned trajectory upon predicted maps may change to be invalid or inefficient.

- The computational complexity grows exponentially with enlargement of search space dimensions. This becomes even more challenging in longer missions, where a huge load of data about update of whole terrain condition should be computed repeatedly. This huge data load should be analysed continuously every time that path replanting is required. This process is computationally inefficient and unnecessary because awareness of environment in vicinity of the vehicle such that it can react to the changes, is enough.

- With respect to optimization algorithms, obtaining the optimal solutions for NP-hard problems is a computationally challenging issue and it is difficult to solve in practice. Bio-inspired meta-heuristic algorithms are diverse nature-inspired algorithms and are known as new revolution in solving complex and hard problems due to stochastic and/or deterministic nature of these algorithms. However, comparing the previously applied optimization algorithms for the purpose of AUVs' motion planning is difficult as each of them addresses a specific aspect of this problem. Undoubtedly, there is always a significant requirement for more efficient optimization techniques in autonomous vehicle's motion planning problem.

- There are several publications in the area of AUV path planning; however only a limited number of them addressed convincing field trail results of AUVs operation in an uncertain time varying ocean environment. Thus, judgment about efficiency and reliability of the developed technologies in dealing with dynamic ocean environment is difficult and yet is dependent to complex considerations and evaluations.

- The majority of stated motion planners did not accurately address the re-planning procedure as there is a significant difference between off-line operation and online motion planning. Operating in a dynamic environment requires an on-line motion planner to perform prompt reaction to raised changes. Despite some studies that addressed this issue, on-line path re-planning in longer-range operations is computationally complex due to repeated computation of a large data load.

## 2.5 Chapter Summary

The primary step towards increasing endurance and range of the vehicle's operation is increasing the vehicle's ability to save energy. More advanced development to this end aims to increase the vehicle's autonomy in robust decision-making and situation awareness. Efficient motion planning and mission planning (task-assign/routing) are principle requirement toward advance autonomy and facilitate the vehicle to handle long-range operations. A growing attention has been devoted in recent years on increasing the ranges of missions, vehicles endurance, extending vehicles applicability, promoting vehicles autonomy to handle longer missions without supervision, and reducing operation costs [8]. The challenges and difficulties associated with the mission planning (task assignment-VRP) technologies (Section 2.3.1) and motion planning strategies (Section 2.4.5) can be synthesized form two perspective: first, the mechanism of the algorithms being utilized; and second, competency of the techniques for real-time applications. Additional to addressed problems with selected strategies in both motion-planning realm, many technical challenges still remained unaddressed.

In a case, that terrain is cluttered with different waypoints and vehicle is required to carry out a specific sequence of prioritized tasks assigned to distance between waypoints in operation network, path planning is not able to carry out the task assignment and time management considering graph routing restrictions. Therefore, that mission planning and a routing strategy is required to handle graph search constraints and carry out the task assignment. Unlike to path planning strategies, the mission planner is not designed to the purpose of copping environmental dynamic changes. However, it is capable of giving general overview of the area that AUV should fly through (general route), which means cut off the operation area to smaller beneficent zone for vehicle's deployment that leads a remarkable reduction of computational load. In contrast the path-trajectory planning strategies are efficient methods with excellent performance in local operations a vehicle's guidance toward the destination encountering the local unexpected dynamic changes of the surroundings. However, enlargement of the operation field increases the computational burdens and uncertainty of the situations.

*CHAPTER 2. LITERATURE REVIEW*

To the best of the authors' knowledge, although several researchers such as Kingston [36] and Alighanbari [37] investigated similar motion planning for UAVs and UGVs applications, there is currently no research particularly emphasized on both vehicle mission planning, routing-task management problem and robust motion planning in a systematic fashion in the scope of underwater vehicles. Respectively, this research contributes a novel Autonomous Reactive Mission Scheduling and Path-planning (ARMSP) architecture encompassing a Task-Assign/Routing (TAR) approach (discussed in Chapter 3) and local Online Real-time Path Planner (ORPP), given in Chapter 4, to cover shortcomings associated with each of the above mentioned strategies. This architecture contains both high-level decision making and low-level action generator, and it can simultaneously organize the mission and improve the vehicle's manoeuvrability that results increasing autonomy for operating in a cluttered and uncertain environment and carrying out a time-efficient fruitful mission. Parallel execution of the TAR and ORPP speed up the computation process.

The operating field for operation of the ORPP is split to smaller spaces between pairs of waypoints; thus, re-planning a new trajectory requires rendering and re-computing less information. This process leads to problem space reduction and reducing the computational burden, which is another reason for fast operation of the proposed approach, in which total operation time is in the range of seconds that is a remarkable achievement for such a real-time system.

# Chapter 3
# AUV Task-Assign/Routing (TAR)

## 3.1 Introduction

Expansion of today's underwater scenarios and missions necessitates the requisition for robust decision making of the AUV; hence, designing an efficient decision making framework is essential for maximizing the mission productivity in a restricted time. However, today's AUVs' endurances still remain restricted to short ranges due to battery restrictions. In this context, a vehicle's autonomy is strictly tied to accurate mission planning and its ability in prioritizing the available tasks in the bounded time that battery capacity allows, while the vehicle is concurrently guided towards its final destination in a complex terrain that is often modelled in a graph-like way. This problem is thus a combination of the BKP (binary knapsack) and TSP (traveling salesman) problems.

The mission planner aims to select the best order of tasks to effectively use the available time for a series of deployment in a long mission, which tightly depends on the optimality of the selected route between start and destination in a waypoint-cluttered terrain. Therefore, the vehicle starts its mission from a particular point and it should reach to the destination point after meeting adequate number of waypoints.

Making use of the meta-heuristic advantages in solving NP-hard graph problems, Particle Swarm Optimization (PSO), Genetic Algorithm (GA), Ant Colony Optimization (ACO), and Biogeography-Based Optimization (BBO) algorithms are employed here to find a quasi-global route (mission plan) for the underwater mission, which here is dealing with a kind of task assignment as well. The meta-heuristic algorithm seeks for highest priority of tasks between the staring and destination point so that the metrics of optimality for the generated route are satisfied.

To evaluate the robustness of the proposed approach, the performance of the method is examined under a number of Monte Carlo runs. Simulation results clearly confirm that the generated mission plan is feasible and reliable enough, and applicable for an uncertain undersea environment. This process is implemented as an independent component named (Task-Assign/Mission-Planner) TAR that operates concurrently in the top layer of the architecture and synchronous to other components (more information about the architecture and other components are given in the Chapters 4 and 5). Figure 3.1 illustrates the operation diagram for the completed section (TAR block) in this chapter.

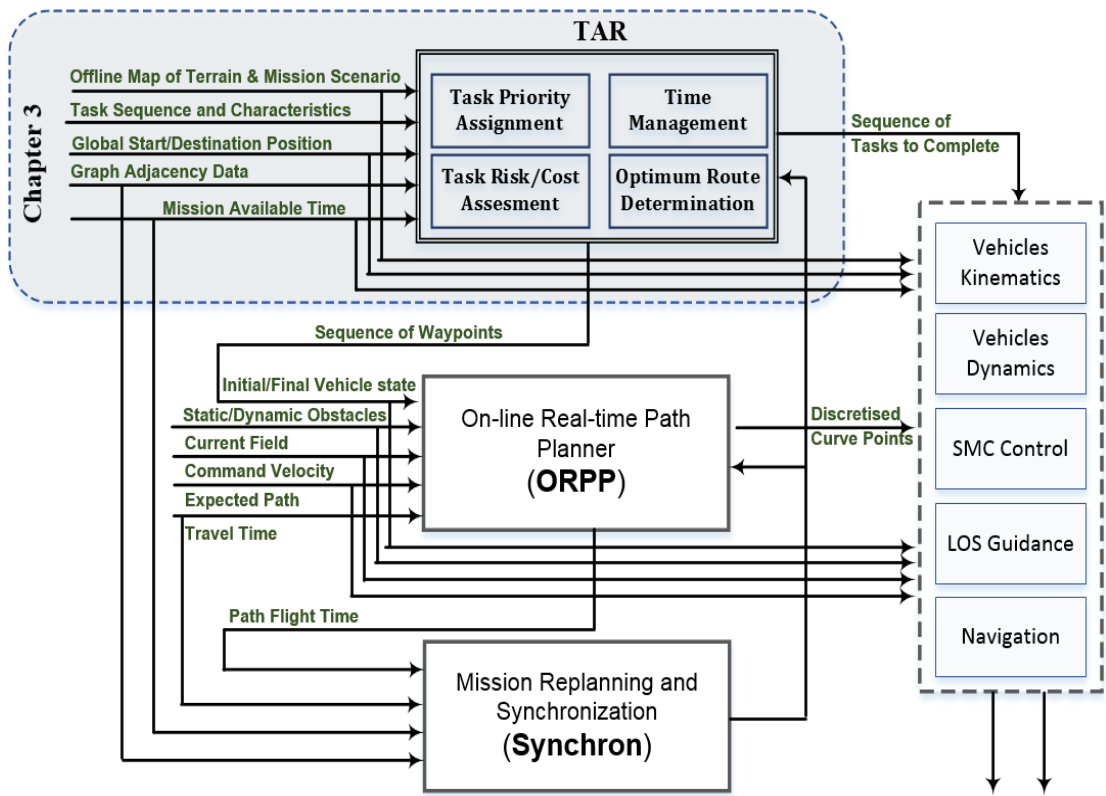

**Figure 3.1.** The operation diagram of the synchronous ARMSP architecture and completion of TAR system by this chapter.

*CHAPTER 3. AUV TASK-ASSIGN/ROUTING (TAR)*

Organization of the Chapter 3 is as follows. task-assign/routing problem is formulated in Section 3.2. The feasibility and optimization criterion is described in subsections 3.2.1 and 3.2.2. An overview of the utilized evolutionary methods and their implementation for this problem are discussed in Sections 3.3 and 3.4. The discussion on simulation results are provided in Section 3.5 with a summary in Section 3.6.

## 3.2 Problem Formulation of the Task-Assign/Routing Approach

For any of scientific, mine or military applications of the AUV, a sequence of tasks such as seabed habitat mapping, sampling the water, mine detections, pipeline inspection, building subsea pipelines, seafloor mapping, any operation of inspection/identification, payload delivery, sensor packages delivery to specific locations, surveillance, etc. Tasks are predefined and characterized in advance and they are fed to vehicle in series of command formats. In total there are 30 different tasks specified in this research, where each task is assigned with a specific priority, risk percentage (risk percentage belong to the task regardless of environmental hazards), and completion time that are initialized once in advance using a normal distribution given by (3-1).

$$\aleph = \{\aleph_1, \ldots, \aleph_{30}\},$$

$$\forall \aleph_i, \quad \exists \rho_i, \xi_i, \delta_i \Rightarrow \begin{matrix} \rho_i \sim U(1,10); \\ \xi_i \sim U(0,100); \\ \delta_i \sim U(20,200); \end{matrix}$$

(3-1)

Where $\aleph$ denotes the set of tasks and $\rho_i$, $\delta_i$, and $\xi_i$ are the $i^{th}$ task's completion time, priority and risk percentage, respectively. For modelling a realistic marine environment, a three-dimensional terrain $\Gamma_{3D}$: $\{10 \times 10 \ km^2 \ (x\text{-}y), \ 100 \ m(z)\}$ is considered based on real map of the Whitsunday island. The map is clustered to the water zone (allowed for deployment) and coastal/uncertain areas, then transformed to a matrix format, in which the water-covered area on the map are filled with value of 1 in the corresponding matrix and the coastal/ uncertain area are assigned by a value between [0,0.3). The waypoints (vertices of the network) are initialized in eligible sections for operation (water-covered area).

Tasks for a specific mission, are distributed in eligible water covered sections of the map, in which placement of the tasks are mapped and presented in a graph format where beginning and ending location of a task are appointed with waypoints. Accordingly, the operating area is mapped with undirected connected weighted network denoted by $G = (P, E)$, where $P$ is the vertices of the graph that corresponds to waypoints and $E$ denotes the edges of the graph in which some of the edges are assigned with a specific task, formulated as follows:

$$G = (P,E) \Rightarrow \begin{vmatrix} |P| = k \\ |E| = m \end{vmatrix} \Rightarrow \begin{matrix} P(G):\{p^1,...,p^k\}; \\ E(G):\{e^1,...,e^m\}; \end{matrix} \Rightarrow e^{ij} = (p^i, p^j)$$

$$\Gamma_{3D} : 10000_x \times 10000_y \times 100_z \tag{3-2}$$

$$\forall p^i \in P \Rightarrow \begin{matrix} p^i_{x,y} \sim U(0,10000) \\ p^i_z \sim U(0,100) \end{matrix} \Rightarrow p^i_{x,y,z} \in \{Map = 1\}$$

where, $k$ and $m$ are the number of nodes ($P$) and edges ($E$) in the graph ($G$); $p^i_{x,y,z}$ is the position of an arbitrary node of the graph; $e^{ij}$ is the edge between two nodes of $p^i_{x,y,z}$ and $p^j_{x,y,z}$. The $\Gamma_{3D}$ presents the 3D volume of the operation terrain. The $\{Map=1\}$ corresponds to the water covered area of map. The water covered eligible sections of the map is identified using an efficient k-means clustering method. Example of such a terrain is given by Figure 3.2.

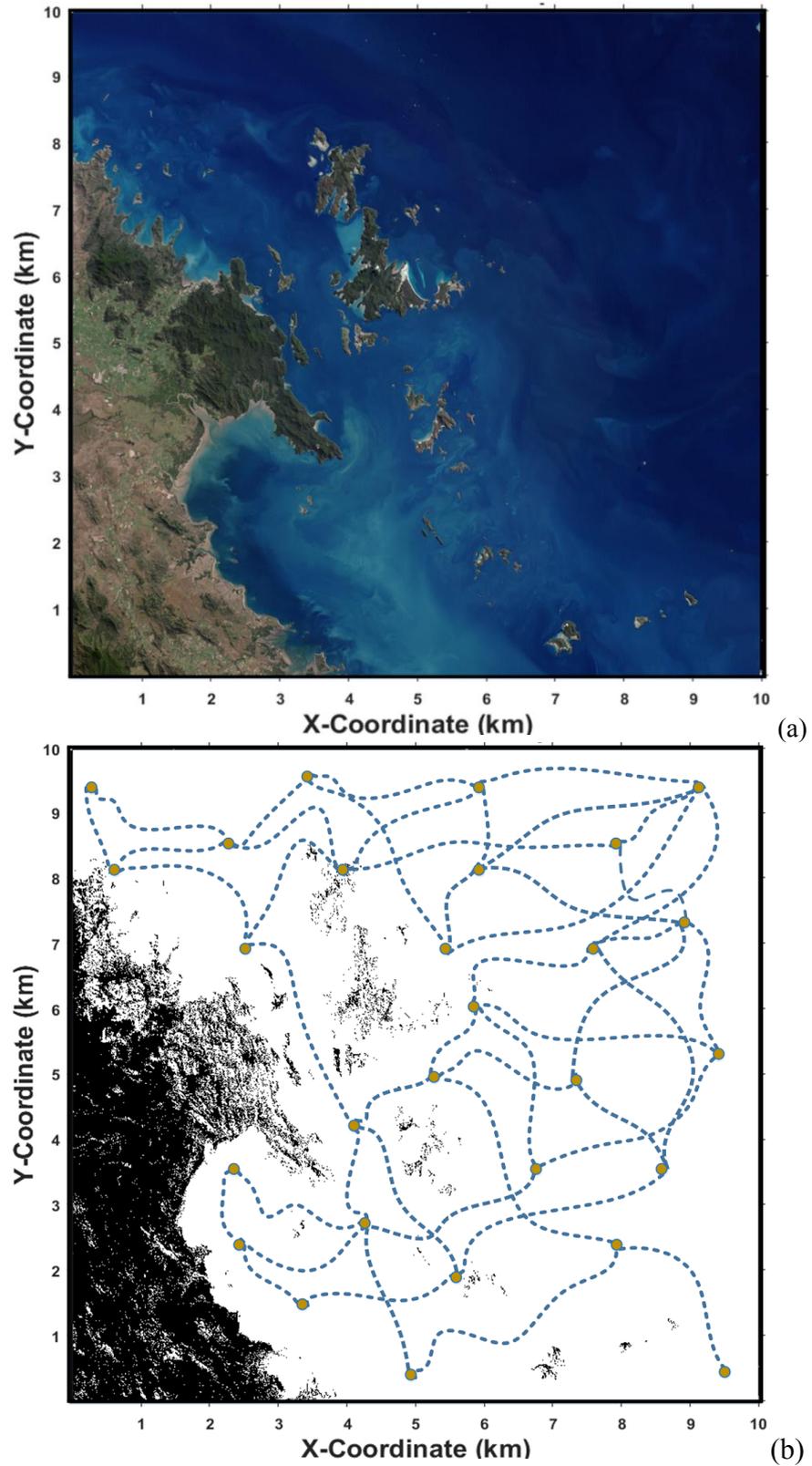

**Figure 3.2.** (a) The original map of the operating area; (b) An example for graph representation of the clustered map, where different tasks ℵ are randomly assigned to edges of the graph.

The weight for edges is calculated based on attributes of the corresponding task (task priority, risk percentage).

$$\forall e^{ij} \quad \exists \quad w_{ij}, d_{ij}, t_{ij}$$

$$w_{ij} = \begin{cases} \dfrac{\rho_{ij}}{\xi_{ij}} & \text{if } e^{ij} \wedge \aleph_l \\ 1 & \text{otherwise} \end{cases}$$

$$\forall e^{ij} \quad \Rightarrow \begin{cases} w_{ij} > 1 & \text{if } \quad \exists \quad \aleph_{ij} \\ w_{ij} = 1 & \text{if } \quad \exists! \quad \aleph_{ij} \end{cases} \tag{3-3}$$

$$d_{ij} = \sqrt{(p_x^j - p_x^i)^2 + (p_y^j - p_y^i)^2 + (p_z^j - p_z^i)^2}$$

$$t_{ij} = \dfrac{d_{ij}}{|\upsilon|} + \delta_{ij}$$

All waypoints ($p_{x,y,z}$) are connected but only the edges ($e^{ij}$) that are assigned with a task are weighted with a value greater than one. If an edge $e^{ij}$ includes a task $\aleph_{ij}$, it is weighted with $w_{ij}$ that is calculated by deviation of the corresponding task's priority $\rho_{ij}$ to its risk percentage $\xi_{ij}$. The $d_{ij}$ and $t_{ij}$ are distance along the $e^{ij}$ and time for travelling the $e^{ij}$. The $|\upsilon|$ is the AUV's water referenced velocity (explained clearly in Chapter 4). The AUV is expected to visit certain points and to perform certain time dependent tasks (such as measuring sea water salt, mapping, taking pictures of underwater structure, or ecosystem), either during its travers towards those points or when it reached to them.

For mission planning, the autonomous vehicle needs to perceive the elements in the variable environment, comprehend the relation between perceived information and meaning of different raised situation in order to compute an optimum global route. It will use prepared information about terrain, waypoints, vehicles limitations, and environmental restrictions to compute the most appropriate route for AUV's mission from start point to the destination. Considering environmental changes and remaining time, the route planner will dynamically re-compute and re-plan a new route repeatedly based on new situation until vehicle reaches to the target point.

### 3.2.1 Shrinking the Search Space to Feasible Task Sequences

The fundamental problem of an AUV system is how to find an optimum sequence of tasks considering tasks priority, hazard percentage, battery sufficiency and time efficiency within a bundled time to battery capacity. From SA point of view, some concepts should be considered regarding mission planning problem, such as "what to search?". The second problem that should be investigated is "How to search?". In this section, it is necessary to define some boundaries for the problem based on its important criteria to limit the search space and solutions to more specific borders. It will help in reducing the initial search space to a more reliable and feasible state that is very helpful in saving time and energy. As discussed previously, the environment is modelled by a graph including several nodes and connections in which some of the connections corresponds to a specific task. Tasks should be selected and ordered in a feasible feature using prior information of tasks and terrain and then are fed to the algorithm as the initial population. The following criteria assesses feasibility of generated solutions:

1) Valid solution is a sequence of nodes (a route) that commenced and ended with index of the predefined start and destination points.
2) Valid solution does not include non-existent edges in the graph.
3) Valid solution does not include a specific node for multiple times, as multiple appearance of the same node implies wasting time repeating a task.
4) Valid solution does not traverse an edge more than once.
5) Valid solution has total completion time smaller than the maximum range of mission available time.

The employed strategy for finding valid routes to given criteria greatly affects performance of the solutions and feasibility of the search space. Hence, a priority-based strategy is conducted by this research to generate feasible solutions (routes), in which a randomly initialized vector is assigned to sequence of nodes in the graph. Adjacency information of the graph and provided random vector is utilized for proper node selection. The random vector is initialised with positive or negative values in the specified range of [-200,100], where each element of the vector corresponds to existing nodes in the graph. Afterward, nodes are added to the sequence one by one according to the values of the vector and adjacency connections. Visited nodes get a large negative value that prevents repeated visit of them. The traversed edges of the graph are eliminated from the adjacency matrix. This process reduces the time and memory consumption, especially when routing in large graphs. Figure 3.3 illustrates a hypothetical example of this process.

| | |
|---|---|
| $Ad$ | Example of adjacency matrix for a graph with 18 nodes |
| $n$ | Node index where $n=1$ is the start and $n=18$ is the destination point |
| $\Re_i^k$ | Partial route including $k$ nodes. |
| $U_i$ | Random non-repeated vector (in range of [-200,100]) |

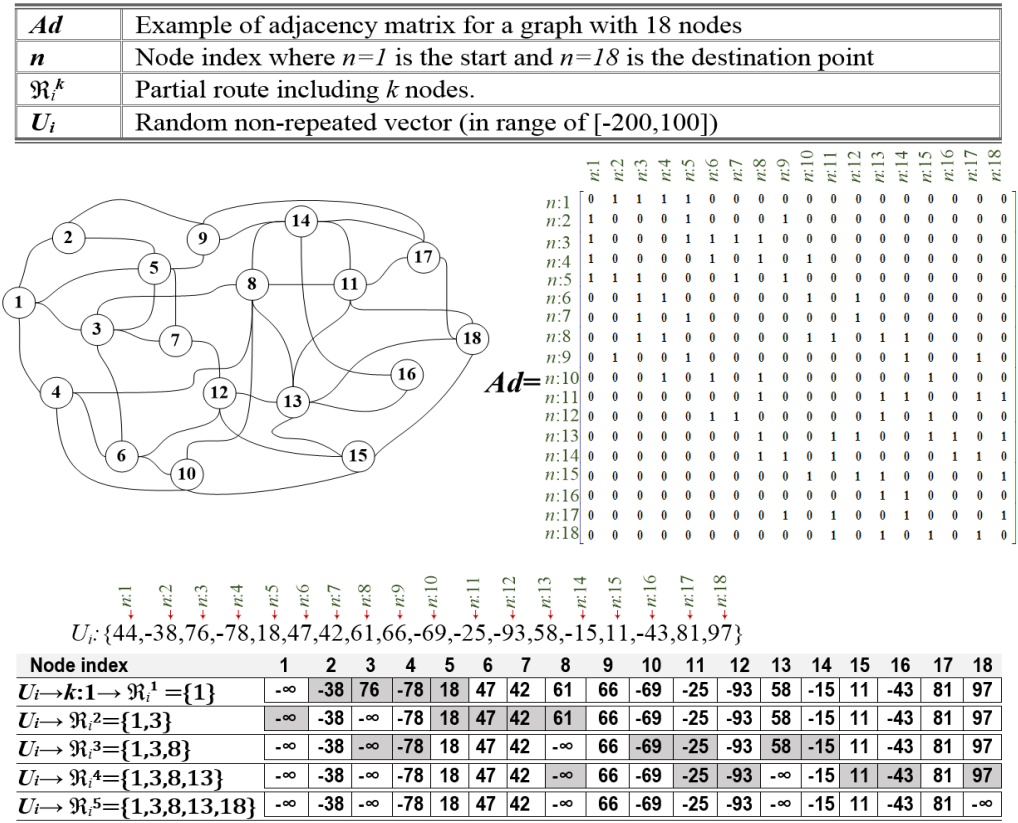

Figure 3.3. Sample of feasible route ($\Re_i$) generation process based on topological information (the value vector $U_i$ and Adjacency matrix $Ad$).

Figure 3.3 presents an example of the route generating process according to a sample Adjacency matrix ($Ad$) of a graph and a random vector ($U_i$). To generate a feasible sequence in a graph with 18 nodes based on topological information, the first node is selected as the start position. Then from the Adjacency matrix, the connected nodes to $n$: 1 are considered. In graph shown in Figure 3.3, this sequence is {*2,3,4,5*}. The node with the highest value (from $U_i$) in this sequence is selected and added to the sequence as the next visited node. This procedure continue until a legitimate route is built (destination is visited). This process greatly affects performance of the solution and feasibility of the search space.

### 3.2.2 Optimization Criterion for Task-Assign/Routing (TAR)

In such a terrain that is covered by several waypoints/tasks, the vehicle is requested to complete maximum possible number of tasks with highest weight in the total available time denoted by $T_\nabla$. Hence, the TAR model tends to find the best fitted route to $T_\nabla$ collecting best sequence of vertexes in the graph in a way that total weight of route is maximized which means the edges with best tasks are selected in a manner to guide the vehicle to destination vertex while on-time termination of the journey is assured. The problem involves multiple objectives that should be satisfied during the optimization process. One approach in solving multi-objective problems is using multi-objective optimization algorithms. Another alternative is to transform a multi-objective optimization problem into a constrained single-objective problem. In this regard, the objective function is defined in a form of hybrid cost function (see (3-5)) comprising weighted functions that are required to be maximized or minimized. So, this problem is described as follows:

$$\Re_k = \left(p_{x,y,z}^S, \ldots, p_{x,y,z}^i, \ldots, p_{x,y,z}^D\right)$$

$$\forall e^{ij} = \left(p_{x,y,z}^i, p_{x,y,z}^j\right) \; \exists \; w_{ij}, d_{ij}, t_{ij} \Rightarrow \begin{cases} w_{ij} > 1 & if \; \exists \; \aleph_{ij} \\ w_{ij} = 1 & f \; \exists! \; \aleph_{ij} \end{cases} \quad (3\text{-}4)$$

$$T_\Re = \sum_{\substack{i=0 \\ j\neq i}}^{n} s_{eij} \times t_{ij} = \sum_{\substack{i=0 \\ j\neq i}}^{n} s_{eij} \times \left(d_{ij}/|v| + \delta_{ij}\right), \quad s_{eij} \in \{0,1\}$$

$$C_{TAR} \propto |T_\Re - T_\nabla| + \left(1 \Big/ \sum_{\substack{i=0 \\ j\neq i}}^{n} s_{eij} \times w_{ij}\right)$$

s.t.

$$\forall \Re_i \Rightarrow \max(T_\Re) < T_\nabla$$

$$Viol_{TAR} = \max(1 - \frac{T_\nabla}{T_\Re}, 0) \quad (3\text{-}5)$$

$$C_{TAR} = \left\{ \Phi_1 \times \left| \sum_{\substack{i=0 \\ j\neq i}}^{n} s_{eij} \times \left(d_{ij}/|v| + \delta_{ij}\right) - T_\nabla \right| + \Phi_2 \times \left(1 \Big/ \sum_{\substack{i=0 \\ j\neq i}}^{n} s_{eij} \times \frac{\Phi_3 \times \rho_{ij}}{\Phi_4 \times \xi_{ij}}\right) \right\} \times$$

$$\times (1 + \Phi_5 \times Viol_{TAR})$$

where, the $\Re$ is an arbitrary route (task sequence generated by explained method in subsection 3.2.1) commenced with the position of the start node $p^s$ and ended at position of the destination node $p^D$. The $s_{eij}$ is a selection variable that equals to 1 for the selected and 0 for unselected edges. The $d_{ij}$ and $t_{ij}$ are distance along the $e^{ij}$ and time for travelling the $e^{ij}$. $T_\Re$ is the route time that should approach the $T_\nabla$ (total available time for mission) but not overstep it. Hence, the problem is a restricted multi-objective optimization problem and $C_{\mathbf{TAR}}$ denotes the cost of the proposed solution by TAR module. The $\Phi_1$, $\Phi_2$, $\Phi_3$, $\Phi_4$ and $\Phi_5$ are positive numbers (defined by user) that determine amount of participation of each factor in the $C_{\mathbf{TAR}}$ cost calculation. The $Viol_{\mathbf{TAR}}$ is the penalty value assigned to the mission route planner to prevent solutions from overstepping the time threshold $T_\nabla$. The search space is initialized with feasible task sequences in format of global route from start to target point. To fulfil the objectives of the proposed model and solving the stated problem (explained by (3-5)), four popular meta-heuristic optimization algorithms of GA, PSO, ACO, and BBO are applied. These methods reported promising performance in solving NP hard problems.

## 3.3 Meta-Heuristics Optimization Algorithms

Meta-heuristics are cost based optimization algorithms that employ a nature based powerful design philosophy to find solutions to hard problems. Former methods to solve VRP or task assignment problems (discussed in Chapter 2) require considerable computational efforts and tend to fail when the problem size grows. Previous approaches to this problem emphasize reducing the complexity of the network by setting small number of nodes (waypoints), assigning small number of edges with limited number of parameters and considerations, and then searching the resultant sparse network for finding the shortest route to the destination using different methods. In contrast, the presented strategy in this research do not require such simplification or discretization of the operation space.

Unlike previous research on vehicle routing problems, which mostly look for the shortest possible route in a graph, this study aims to complete the maximum number of tasks for which time and distance are a function of the individual task. On the other hand, while various exact techniques have been developed over three last decades, meta-heuristic methods still remain as the most appropriate selections in real time applications with larger dimensionality. Generally, evolutionary algorithms and intelligent optimization methods have less sensitivity to size of graph so the search time increases linearly with the of number of points. These algorithms are easy to implement and they are more flexible for being implemented on a parallel machine with multiple processors, which speeds up the computation process [92]. A significant distinction exists between theoretical understanding of meta-heuristics and peculiarity of different applications in contrast. This gap is more highlighted when scale (size), complexity, and nature of the problem is taken to account. Application of different optimization methods may result in diverse performances on the same problem due to specific nature of each problem. Therefore, accurate selection of the algorithm and proper matching of the algorithms functionalities and characteristics in accordance with the nature of a particular problem is a highly critical issue that should be taken into account in advance. These facts usually remain unaddressed and unsolved in most engineering and robotics frameworks in both theory and practice. Often, meta-heuristic algorithms require careful and accurate modifications to appropriately accommodate a specific problem.

With respect to the combinatorial nature of routing and mission planning problem, which is analogous to TSP-BKP problems, this problem is categorized as a multi-objective NP-Hard problem often solved by optimization algorithms [99,116]. Obtaining the optimal solutions for NP-hard problems is a computationally challenging in practice. Obtaining the exact optimum solution is only possible for the certain cases with no uncertainty; which is not the case in this study. Meta-heuristics are appropriate approaches suggested for handling the aforementioned complexities.

Undoubtedly, there is always a significant requirement for more efficient optimization techniques in autonomous vehicle's mission planning problem. To fulfil the objectives of this study towards solving the stated problems, this study presents its solution by applying heuristic search nature of the GA, PSO, ACO, and BBO algorithms.

This chapter aims to investigate efficiency of any applied meta-heuristic algorithm for vehicles prompt route-mission planning as this approach would be foundation for the next two chapters, which is about building a hierarchal control architecture to handle environmental prompt changes along with efficient task-time management. Hence, the main goal is to prove efficiency of meta-heuristics against any other deterministic methods for the purpose of this research rather than criticizing or comparing efficiency of two applied algorithms.

## 3.4 Overview of the Meta-Heuristics Employed

The ACO algorithm was introduced by Dorigo et al., [117] inspired from the ants' foraging process, and is implemented successfully on TSP problem for the first time. This algorithm is has proven very successful in solving vehicle routing problems and is scales well for task assignment due to its discrete nature [118-120]. On the other hand, the PSO is one of the most flexible optimization methods for solving a variety of complex problems and widely used in several studies in the past decades. The PSO is well adapted to multidimensional space and nonlinear functions due to its stochastic optimization nature that does not require any evolutionary operators to transmute the individuals. Due to flexible design of the PSO, it can be easily modified to fit the problem while when using DE or GA or ACO, the problem formulation should be modified to fit the way these methods formulate their evolution. GA is also a stochastic search algorithm that operates based on biological evolution and has been extensively studied and widely used on different realms of graph routing problems and similar applications [49, 50]. The GA also has a discrete nature and is well fitted with the proposed problem. The BBO is another evolutionary technique developed according to equilibrium theory of island biogeography concept [121]. The population of candidate solutions in BBO is defined by geographically isolated islands known as habitat. A special feature of the BBO algorithm is that the original population never is discarded but are modified by migration. This issue improves the exploitation ability of the algorithm.

Several cost factors such as route length, travel time, task priority and task specific metrics are considered to be minimized or maximized simultaneously in order to make maximum use of the mission available time. As an AUV operates in an uncertain environment, there is a huge amount of uncertainty in the travel time that can have a devastating effect on mission plans. Proper time management of the vehicles routing operations is necessary to ensure on-time mission completion and consequently the mission success. Accurate coding of the initial population is the most important step in implementing all evolutionary algorithms. Thus, in proceeding research, the individual population for all ACO, BBO, PSO and GA are initialized with feasible and valid task sequences (routes $\Re$) provided by an accurate proven mechanism (discussed in Section 3.2.1). In order to satisfy all aspects of complex missions, the offered task-assign/routing approach applies a multi objective cost function that is composed of multiple factors given by (3-5), where the generated solutions are subjected to a time threshold of $T_\nabla$. Selecting the optimum route based on these criteria is a very challenging problem, especially when time restrictions forces overall operation performance.

### 3.4.1 Application of ACO on TAR Approach

In ACO, the ants find their path through the probabilistic decision according to the level of pheromone concentration, so that the pheromone trail attract ants and guide them to the grain source. Respectively, more pheromone is released on the overcrowded paths, which is a self-reinforcing process in finding the quickest route to target. The pheromone evaporates over time, which helps ants to eliminate bad solutions. This allows the algorithm to prevent the solutions from falling to the local optima. The pheromone concentration on edges ($\tau$) keeps the ant's collected information within the search process. The probability of targeting a node to travel is calculated by (3-6) and then, the total pheromone trail is updated for all solutions according to (3-7).

$$prob_{ij}^k = \begin{cases} \dfrac{\left(\tau_{ij}^{(k)}\right)^\alpha \left(\eta_{ij}^{(k)}\right)^\beta}{\sum\limits_{l \in N_i^k}\left(\tau_{il}^{(k)}\right)^\alpha \left(\eta_{il}^{(k)}\right)^\beta} & j \in N_i^k \\ 0 & j \notin N_i^k \end{cases} \quad (3\text{-}6)$$

$$\tau_{ij} = (1-r_e)r_e + \sum_{k=1}^{N}\Delta\tau_{ij}^{(k)}$$

$$\tau_{ij}(t+1) = \tau_{ij}(t) + \Delta\tau_{ij} \quad (3\text{-}7)$$

$$\Delta\tau_{ij} = \begin{cases} -\varpi\tau_{ij}(t) + \dfrac{Q}{\Omega} & \text{For nodes of the best ant} \\ -\varpi\tau_{ij}(t) & \text{Otherwise} \end{cases}$$

here, *k* is the ant index, *i* and *j* denote the index of current and the target nodes, $\tau^k_{ij}$ is the pheromone concentration on edge between nodes *i* and *j*, α and β are the pheromone and heuristic factors, respectively. $\eta_i$ denotes the heuristic information in $i^{th}$ node. $N_i^{(k)}$ indicates the set of neighbour of ant *k* when located at node *i*. The $r \in (0,1]$ is evaporation rate. The $\Delta\tau_{ij}^{(k)}$ denotes the pheromone level collected by the best ant *k* based on priority of ants released pheromone (Ω) on the selected nodes. $\tau^k_{ij}(t)$ and $\tau^k_{ij}(t+1)$ are the previous and current level of the pheromone between node *i* and *j*. *Q* is the amount of released pheromone on nodes taken by the best ant(s), and λ is the coefficient of pheromone evaporation. Algorithm 3.1 describes the procedure of the ACO based route planner.

---

**ACO on TAR Approach**

Initialization phase:
- Initialize population of ants in a colony using generated feasible routes
- Define number of layers in network, where each layer contains $p$ nodes i.e. ($\chi_{i1}, \chi_{i2}, ..., \chi_{ip}$)
- Set initial pheromone for each edge by $\tau_{ij}^{(l)}=1$
- Set the maximum number of iterations $t_{max}$

  **For** $t=1$ **to** $t_{max}$
    **For** $k=1$ **to** $i_{max}$
      Compute the probability $prob_{ij}$ of selecting node $j$ from the current node $i$ by Eq.(3.6).
- Assume $\alpha=1$
- Generate $k$ random numbers $r_{ek} \in [0,1]$, one for each ant.
- Calculate range of the cumulative probability associated each routes applying roulette-wheel selection.
- Determine the route captured by ant $k$ as a cumulative probability that keeps $r_{ek}$ random number in it.
- Evaluate the cost value for all solutions $C_{TAR}^k = C_{TAR}(\chi_t^{(k)}); k=1,2,...,n$ and determine best and worst solutions $\chi_t^{best}, \chi_t^{worst}$
- Check the route (solution) feasibility criterion
- Save the best solution $\chi_t^{best}$

Update the pheromone on arc when all the ants complete their tour using Eq.(3.7).
  **For** $j=1$ **to** $k$

$$\tau_{ij}^k = \tau_{ij}^{old} + \sum_{k=1}^{i_{max}} \Delta\tau_{ij}^k$$

$$\tau_{ij}^{old} = (t-\rho)\tau_{ij}^{t-1}$$

    **If** $C_{TAR}(\chi_t^{best})^{new} < C_{TAR}(\chi_t^{best})^{old}$
      $\chi^{G\text{-}best} = (\chi_t^{best})^{new}$
    **else**
      $\chi^{G\text{-}best} = (\chi_t^{best})^{old}$
    **end** (*if*)
  **end** (*For*)
**end** (*For*)
$\alpha_t = 0.99 \times \alpha_{t-1}$
$\beta_t = 0.99 \times \beta_{t-1}$
**end** (*For*)
Output result

---

**Algorithm 3.1.** Pseudo-code of ACO mechanism on TAR approach.

### 3.4.2 Application of GA on TAR Approach

The chromosome population in GA is initialized with feasible routes (task sequences); hence, the first and last gene of the chromosomes always corresponds to index of the start and destination nodes. A new generation is produced from the initial population by applying the GA operators of selection, crossover and mutation. Afterward, new generation is evaluated by defined route feasibility criterion and the route cost function. The best-fitted chromosomes with minimum cost are transferred to the next generation and the rest is eliminated. This process is repeated iteratively until the population is converged to the best-fit solution to the given problem [122]. The average fitness of the population is improved at each iteration by adaptive heuristic search nature of the GA. The operation is terminated when maximum iteration is reached or best solution is found.

1) *Selection*: Selecting the parents for crossover operation is another important step of the GA algorithm that improves the average fitness of the population in the next generation. This research uses the roulette wheel selection, wherein the next generation is selected from the best-fit chromosomes and the wheel divided into a number of slices so that the chromosomes with the best cost take larger slice of the wheel.

2) *Crossover*: This operator shuffles sub parts of two parent chromosomes and generates mixed offsprings. Generally, two main categories of single and multi-point crossover methods are introduced. Discussion over which crossover method is more appropriate is still an open area for research. This study applies uniform crossover, which uses fixed mixing ratio among parent chromosomes that is usually set to 0.5. The uniform crossover is remarkably suitable for large space problems. After that the offsprings are generated, the new generation should be validated.

3) *Mutation*: This operator flips multiple genes of a chromosome to generate new offspring. The current research used three types of mutation techniques: insertion, inversion and swapping, in which all of them preserve the most adjacency information. The mutation is applied on genes between the first and the last genes of the parent chromosome to keep the new generation in feasible space and then new offspring are evaluated. Generally, mutation can happens on two conditions: 1) when a chromosome does not improve for certain number of iterations, then you just mutate it to push it to a different subspace of problem space. This is very difficult to track and rarely used in GA studies. 2) When the population fitness does not change after several iterations and approach to a stall generation. Therefore, a subpopulation is selected randomly for mutation. This operation guarantees the diversity in the population.

The feasibility of the generated route is checked during the process of cost evaluation. The process of the GA algorithm on this approach is displayed by Algorithm 3.2.

| GA on TAR Approach |
| --- |
| Initialize the chromosome population by generated feasible routes |
|     ▪ Choose appropriate parameters for the population size $i_{max}$ |
|     ▪ Set the maximum number of generations (iteration $t_{max}$) |
|     ▪ Assign maximum crossover and mutation coefficients. |
| **For** *t*=1 *to* $t_{max}$ |
|   Evaluate feasibility of each chromosome |
|   Calculate the fitness of each chromosome by defined cost function |
|   **For** *i*=1 *to* $i_{max}$ |
|     Select best two chromosomes as parents using roulette wheel |
|     Apply the crossover operator |
|     Apply the mutation operator |
|     Check feasibility of off springs |
|     Evaluate fitness of offspring and select best between parents and offspring |
|   **end (For)** |
|   Select the best fitted chromosomes to transfer to next generation |
|   Eliminate the worst chromosomes with highest cost value |
| **end (For)** |
| Output the most fitted choromosome as the best route sequence |

**Algorithm 3.2.** Pseudo-code of GA mechanism on TAR approach.

### 3.4.3 Application of BBO on TAR Approach

The population of habitat candidate solutions in BBO is initialized with feasible route sequence and each candidate solution holds a quantitative performance index representing the fitness of the solution called Habitat Suitability Index (HSI). High HSI solutions tend to share their useful information with poor HSI solutions. Habitability is related to some qualitative factors known as Suitability Index Variables (SIVs), which is a random vector of integers. Each candidate solution ($h_i$) holds design variables of SIV, emigration rate (μ), immigration rate (λ) and fitness value of HSI. Generally, a poor solution has higher immigration rate of λ and lower emigration rate of μ. The SIV vector of $h_i$ is probabilistically altered according to its immigration rate of λ. In migration process, one of the solutions ($h_i$) is picked randomly to migrate its SIV to another solution $h_j$ regarding its μ value. Each given solution $h_i$ is modified according to probability of $P_s(t)$ that is the probability of existence of the *S* species at time *t* in habitat $h_i$ while $P_s(t+\Delta t)$ is the change in number of species after time $\Delta t$, given by (3-8).

$$P_S(t+\Delta t) = P_S(t)(1-\lambda_S\Delta t - \mu_S\Delta t) + P_{S-1}\lambda_{S-1}\Delta t + P_{S+1}\mu_{S+1}\Delta t$$

$$\begin{aligned}\lambda_S &= I*\left(1-\frac{S}{S_{\max}}\right); \\ \mu_S &= E*\left(\frac{S}{S_{\max}}\right);\end{aligned} \xrightarrow{if\ E=I} \lambda_S + \mu_S = E \qquad (3\text{-}8)$$

Where *I* is the maximum immigration rate and *E* is the maximum emigration rate. $S_{max}$ is the maximum number of species in a habitat. To have *S* species at time $(t+\Delta t)$ in a specific habitat $h_i$, one of the following conditions must hold:

$$\forall h_i(t) \quad \exists \lambda_S, \mu_S, P_S(t)$$
$$\begin{cases} \forall h_i(t): \exists S & \Rightarrow & \forall h_i(t+\Delta t): \exists S \\ \forall h_i(t): \exists S & \Rightarrow & \forall h_i(t+\Delta t): \exists S-1 \\ \forall h_i(t): \exists S & \Rightarrow & \forall h_i(t+\Delta t): \exists S+1 \end{cases} \qquad (3\text{-}9)$$

As suitability of habitat improves, the number of its species and emigration increases, while the immigration rate decreases. Afterward, the mutation operation is applied, which tends to increase the diversity of the population to propel the individuals toward global minima. Mutation is required for solution with low probability, while solution with high probability is less likely to mutate. Hence, the mutation rate *m(S)* is inversely proportional to probability of the solution $P_s$.

$$m(S) = m_{\max}\left[\frac{1-P_S}{P_{\max}}\right] \qquad (3\text{-}10)$$

Where $m_{max}$ is the maximum mutation rate assigned by operator, $P_{max}$ is probability the habitat with maximum number of species $S_{max}$. The whole procedure is summarized in Algorithm 3.3.

---

**BBO on TAR Approach**

Initialize a set of solutions as initial habitat population by generated feasible routes
- Choose appropriate parameters for the population size $i_{max}$
- Set the maximum number of generations (iteration $t_{max}$)
- Assign maximum immigration and emigration rate ($I_r$, $E_r$)
- Assign maximum mutation rate $m_{r,max}(S_p)$
- Set $Sp_{max}$ and SIV vector

**For** $t=1$ **to** $t_{max}$
  Compute immigration rates $\lambda$ and emigration rate $\mu$ for each solution
  Evaluate the fitness (HSI) of each habitat and identify Elite Habitats based on HIS
  Modify habitats based on $\lambda$ and $\mu$ (Migration):
    **For** $i=1$ **to** $i_{max}$
      Use $\lambda_i$ to probabilistically decide whether immigrate to habitat $h_i$
      **if** $rand(0,1) < \lambda_i$
      **For** $j=1$ **to** $i_{max}$
        Select the emigrating habitat $h_j$ with probability $\propto \mu_j$
        **if** $rand(0,1) < \mu_j$
          Replace a randomly selected SIV variable of $h_i$ with its corresponding value in $h_j$
        **end** (*if*)
      **end** (*For*)
      **end** (*if*)
    **end** (*For*)
  Carry out the mutation based on probability
  Check the route Feasibility Criterion
  Implement the elitism to retain the best solution in the population from one generation to the next
**end** (*For*)
Output the best habitat with optimal fitness value and its corresponding route

**Algorithm 3.3.** Pseudo-code of BBO mechanism on TAR approach.

### 3.4.4 Application of PSO on TAR Approach

The PSO start its search process using a population of $i_{max}$ particles initialized by feasible routes. Particle encoding is very important factor that affects effectiveness of the algorithm. During the search process, the population of routes is directed toward the regions of interest in the search space. Each particle involves a position and velocity that are initialized randomly in a specific range in the search space. The position of each particle (denoted by $\chi_i$; $i \in [1, i_{max}]$) is readjusted during the search process through an updated velocity (denoted by $\upsilon_{i,j}$; $i \in [1, i_{max}]$; $j \in [1, l]$, where $l$ is the dimension of the search space). The particles position and velocity are updated iteratively according to (3-11). The performance of particles is evaluated according to the defined cost functions. Each particle preserves its best position $\chi^{P\text{-}best}$ and swarm global best position $\chi^{G\text{-}best}$. The current state value of the particle is compared to $\chi^{P\text{-}best}$ and $\chi^{G\text{-}best}$ at each iteration. Particle position and velocity are updated as follows.

$$\upsilon_{i,j}(t) = \omega \upsilon_{i,j}(t-1) + c_1 r_{1,j} \left[ \chi_{i,j}^{P-best}(t-1) - \chi_{i,j}(t-1) \right] + $$
$$+ c_2 r_{2,j} \left[ \chi_{i,j}^{G-best}(t-1) - \chi_{i,j}(t-1) \right] \quad (3\text{-}11)$$
$$\chi_{i,j}(t) = \chi_{i,j}(t-1) + \upsilon_{i,j}(t)$$

Where $c_1$ and $c_2$ are acceleration coefficients, $\chi_{i,j}$ and $\upsilon_{i,j}$ are particle position and velocity at iteration $t$. $r_{1,j}$ and $r_{2,j}$ are two independent random numbers in [0,1]. $\omega$ exposes the inertia weight and balances the PSO algorithm between the local and global search. More detail about the algorithm can be found in [123]. Algorithm 3.4 explains the process of PSO.

---

**PSO on TAR Approach**

Initialize swarm population with random velocity and position in following steps:
- Assign particle position $\chi_i$ with priority vector defined for route vector.
- Initialize each particle with random velocity $\upsilon_i$ in range of [-200,100]
- Choose appropriate parameters for the population size $i_{max}$
- Set the maximum number of iterations $t_{max}$
- Set the $\chi_i^{P-best}(1)$ and $\chi^{G-best}(1)$ with particle current position and swarm best at iteration $t=1$

**For** $t=1$ **to** $t_{max}$
    Evaluate each candidate particle according to given cost function
    **For** $i=1$ **to** $i_{max}$
        Updated the particles $\chi_i^{P-best}$ and $\chi^{G-best}$ at iteration $t$
        **if** $C_{TAR}(\chi_i(t)) \leq C_{TAR}(\chi_i^{P-best}(t-1))$
            $\chi_i^{P-best}(t) = \chi_i(t)$
        **else**
            $\chi_i^{P-best}(t) = \chi_i^{P-best}(t-1)$
        **end (if)**

$$\chi^{G-best}(t) = \underset{1 \leq i}{\arg\min} \, C_{TAR}(\chi_i^{P-best}(t))$$

        Update the state of the particle in the swarm using Eq.(3.11).
        Evaluate each candidate particle $\chi_i$ according to given cost function $C_{TAR}(\chi_i(t))$
    **end (For)**
    Transfer best particles to next generation
**end (For)**
Output the corresponding route ($\chi^{G-best}$)

---

**Algorithm 3.4.** Pseudo-code of PSO mechanism on TAR approach.

As discussed earlier, the main goal is to prove efficiency and real-time performance of meta-heuristics methods for the purpose of this mission route planning rather than criticizing or comparing them.

## 3.5 Discussion on the TAR Simulation Results

Comparison of different meta-heuristic methods is common due to their similarity in optimization mechanism. Any of these algorithms show different performance on different problems; however, apparently they are all capable of producing feasible solutions after the iterations has completed. Second benefit of these algorithms is their ability of handling non-linear cost functions. As mentioned earlier, obtaining a near optimal solution in faster computation time is more important for the purpose of this research where the modelled environment corresponds to a spatiotemporal time varying terrain with high uncertainty. The real-time performance of the applied algorithms on TAR approach is not fully known; hence, the computation time would be one of the important performance indicators for evaluating the applied methods. The TAR module in this research uses prepared information about terrain, waypoints location, tasks that are assigned to network's adjacent connections and their characterizations, and time/battery restrictions to compute the most appropriate order of prioritized tasks to guide the AUV towards its destination. There should be a compromise among the mission available time, maximizing number of highest priority tasks, and guaranteeing reaching to the predefined destination before the vehicle runs out of battery/time. All algorithms are configured with $t_{max}$=100 iterations. In the ACO, the population is set to 70 ants. The pheromone value is initialized with $\tau_0$=1.5, the pheromone exponential and heuristic exponential weights are initialized with 1.5 and 0.9, respectively. The pheromone and heuristic factors are updated iteratively according to $\alpha_t$=0.99×$\alpha_{t-1}$ and $\beta_t$=0.99×$\beta_{t-1}$. The evaporation rate is assigned with 1.8. The BBO, in this circumstance, is configured with habitat population size of 70; the emigration rate, immigration rate, and maximum mutation rate are set on $\mu$=0.2, $\lambda$=1-$\mu$, and $m(S)$=0.5, respectively. The GA is configured with population size of 70, crossover rate of 0.99, mutation rate of 0.8 and selection pressure of 8. The PSO is configured by 70 particles, acceleration coefficients of 1.5 and 2. The inertia weight is set to be decreased iteratively according to $\omega$=($t_{max}$ -$t$)/$t_{max}$ (where, $t_{max}$ is the maximum number of iterations and $t$ is the current iteration).

*CHAPTER 3. AUV TASK-ASSIGN/ROUTING (TAR)*

The problem is defined as a constraint routing in a graph-like terrain including waypoints (nodes) and connections in which each connection represents a task. Therefore, increasing number of tasks means increasing quantity of connections and nodes in the network, which means increasing the problem's complexity (as commonly known in graph search problems like TSP). The CPU time and cost variations of the applied methods are investigated on different network topologies, in which network complexity and size increases incrementally from 30 to 150 nodes (waypoints), presented by Figure 3.4. With respect to the defined cost function for the mission planning problem (given by (3-5)), an optimum solution corresponds to a mission that takes maximum use of available time by maximizing number of completed tasks and route weight, while respects the upper bound time threshold denoted by $T_\nabla$.

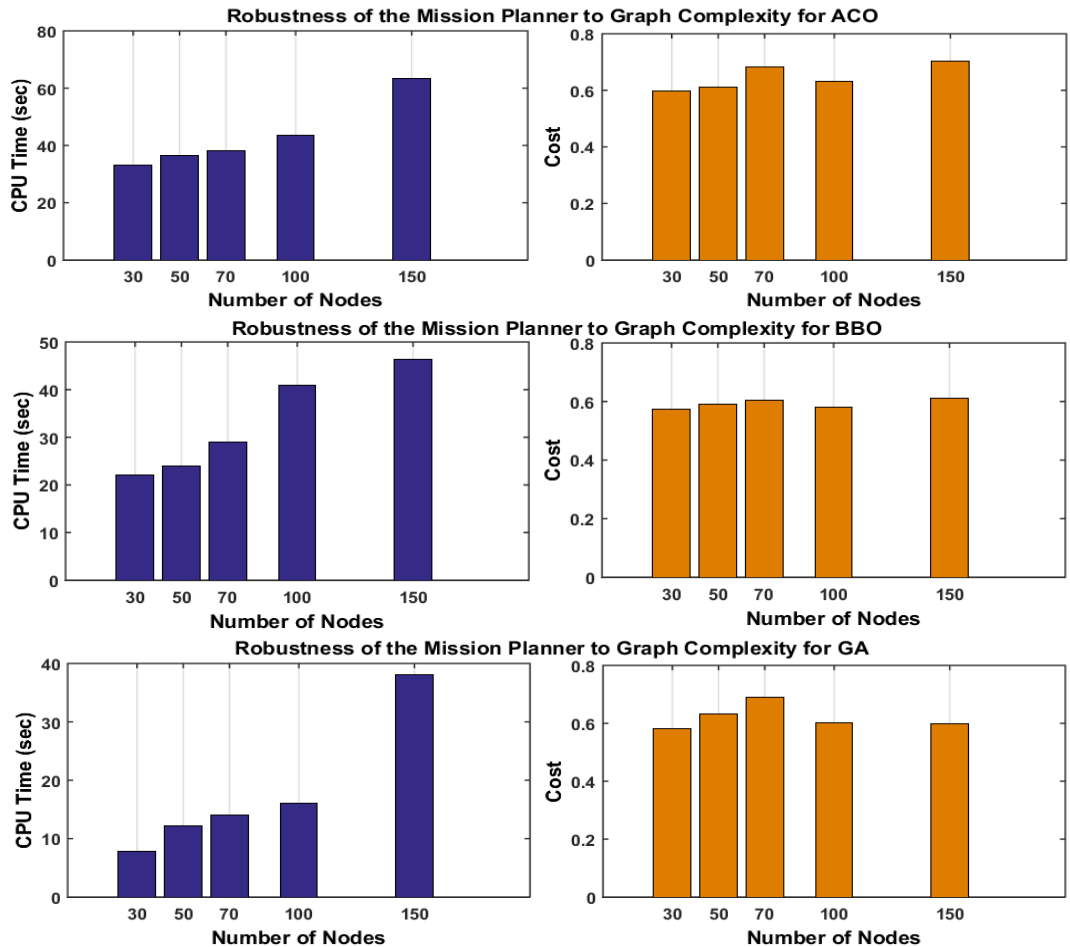

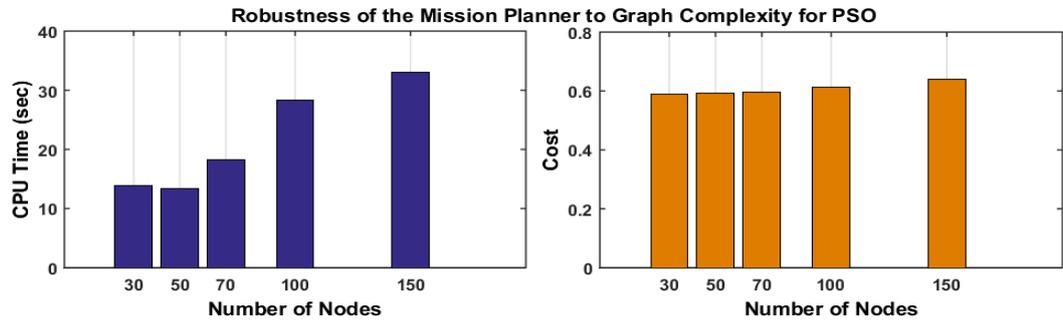

**Figure 3.4.** The mission cost and CPU time variation over 100 iterations for different graph complexities.

As presented in Figure 3.4, the employed algorithms accurately manage the complexity growth as the cost value for all graph topologies lies in the similar range over 100 iterations and the computational time increments linearly with growth of the graph nodes. The CPU time for all samples remained within the bounds of a suitable real-time solution.

The cost and CPU-Time variations declares that the performance of the model is almost independent of both size and complexity of the graph, whereas this is recognized as a problematic issue in many of the outlined approaches.

Accordingly, performance of the ACO, BBO, GA, and PSO algorithms is investigated in a quantitative manner through the 150 Monte Carlo simulation runs (given by Figure 3.5 to 3.7). By analysing the result of the Monte Carlo simulations, it is possible to be confident in the robustness and efficiency of the method in dealing with random transformation of the graph size and topology.

In the performed simulation, number of vertexes of the graph is set to differ between 30 to 50 nodes, connection between nodes and overall graph topology is randomly changed (using a Gaussian distribution) on the problem search space in each execution. The time threshold is set on $3.42 \times 10^4$ (*sec*).

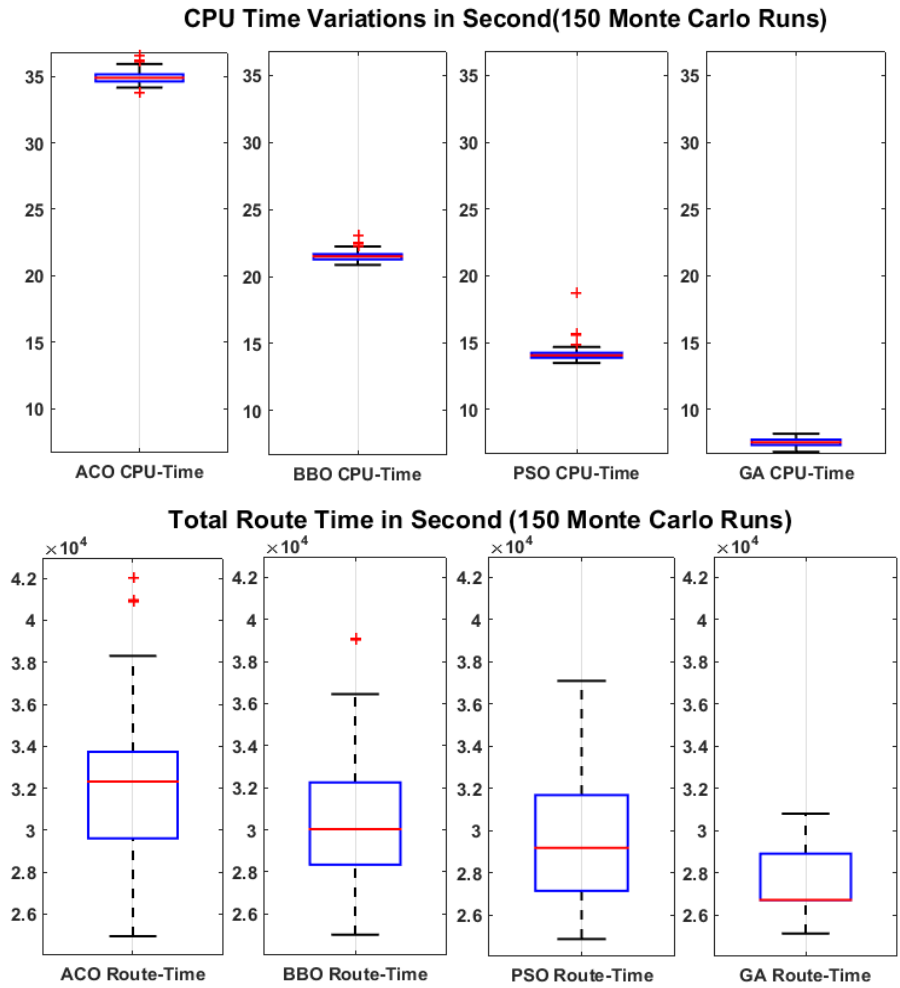

**Figure 3.5.** Average variations of CPU time and the route time ($T_\Re$) over the 150 Monte Carlo simulations, where complexity of the network is changing with a Gaussian distribution on the problem search space in each execution.

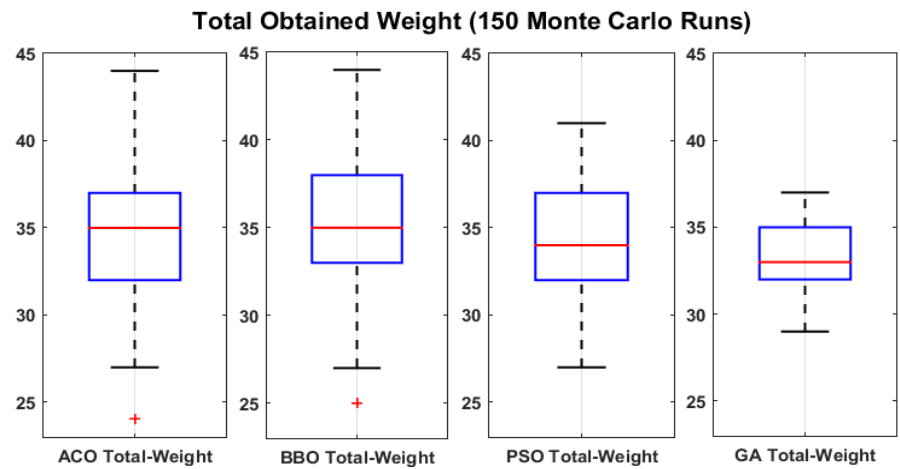

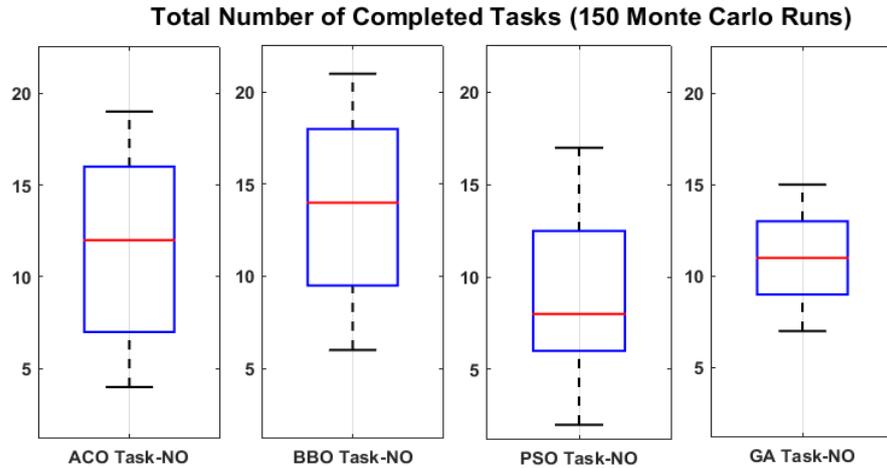

**Figure 3.6.** Average variations of the total obtained weight and number of completed tasks over the 150 Monte Carlo simulations, where complexity of the network is changing with a Gaussian distribution on the problem search space in each execution.

Figures 3.5, 3.6 and 3.7 compare the functionality of the applied algorithms in dealing with problem's space deformation. It is apparent from CPU-Time variations in Figure 3.5 that all algorithms are capable of satisfying real-time requirement of the proposed problem as the variation for all algorithms is drawn in a very narrow boundary in range of seconds; however, the fastest operation belongs to GA and then PSO, BBO, and ACO, in order.

The appropriate case for the TAR approach is producing routes with maximum travel time limited to threshold. It is denoted from route time variations in Figure 3.5, all algorithms significantly manage the route time to approach proposed time threshold, but it seems that GA acts more cautious in constraining the route time as the variation range for GA is strictly kept far below the defined threshold. However, variations of PSO, BBO, and ACO almost placed in a same range, but no outlier is seen in PSO route time variations, which is good point for this algorithm.

Figure 3.6 illustrates the stability of the TAR in terms of completed tasks and total obtained weight for each algorithm. It is apparent in Figure 3.6 that variation range for total obtained weight and completed tasks by GA appeared in a smaller range in comparison with outcome of others respecting the narrow range of GA route time. Three other algorithms of BBO, ACO and PSO propose almost similar performance in quantitative measurement of two significant mission metrics of route obtained weight and number of covered tasks; however, being more specific, BBO acts more efficiently as the produced results by BBO dominates two others.

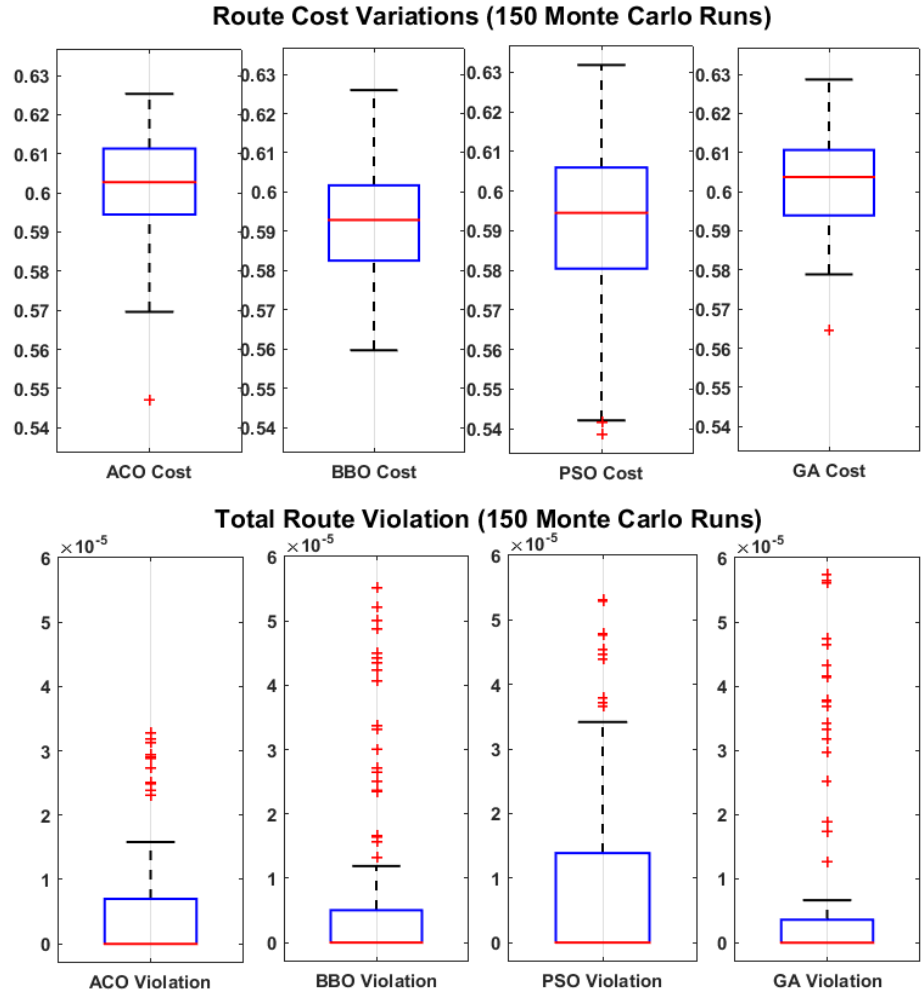

**Figure 3.7.** Average variations of route cost ($C_{TAR}$) and violation ($Viol_{TAR}$) over the 150 Monte Carlo simulations, where complexity of the network is changing with a Gaussian distribution on the problem search space in each execution.

A solution gets a penalty value of $Viol_{TAR}$, when the route time oversteps the total mission available time $T_\nabla$. Figure 3.5 presents average variations of route cost and violation over the 150 Monte Carlo simulations. It is noteworthy to mention, the route cost is varying in a similar range for all four algorithms, almost between 0.57 to 0.62, and it is noted that average variation of route violation for Monte Carlo executions approaches zero. This confirms efficiency of the applied algorithms in satisfying defined constraints for TAR and they are persistent against problem space deformation, as this is a challenging problem for other deterministic algorithms addressed earlier. Being more specific, the provided results also shows that the GA reveals better performance in eliminating the route violation. To compare applied meta-heuristic methods, a benchmark table is employed as presented in Table 3.1, which gives a summary of the characteristics of four aforementioned algorithms that quantitatively analysed by Figures 3.5 to 3.7.

**Table 3.1.** Comparison of the GA, BBO, PSO, and ACO based mission-planning performance.

|  | *CPU Time* | *Route Time* | *Route Weight* | *Covered Tasks* | *Route Cost* | *Route Violation* |
|---|---|---|---|---|---|---|
| ***Best to Worst*** | GA | BBO | BBO | BBO | PSO | GA |
|  | PSO | PSO | ACO | ACO | BBO | BBO |
|  | BBO | ACO | PSO | PSO | ACO | ACO |
|  | ACO | GA | GA | GA | GA | PSO |

The ACO and PSO has a similar feature in which both of them use three principals of the transient state, local and global updates. Local updating in ACO preserves the quality of the population in each iteration so it requires less iterations to converge the population toward the optimum solution. However, both PSO and ACO face the stagnation problem when the swarm quickly is stuck in a local optimum; hence, the algorithm is not able to promote solution quality as time progresses. The particle or ant swarm will eventually find an optimal or near optimal solution despite no estimate can be given on convergence time. The ACO also shares the similar shortcoming with the GA algorithm as there is no guarantee in random method for fast optimal solutions even though GA select a proper population at each iteration. Moreover, when complexity of graph increases the length of chromosomes or solution vectors is increased accordingly, so that mission-planning process becomes slower; nevertheless, all aforementioned algorithms operate in a competitive CPU time unlike the deterministic and heuristic methods. Hence, the evolutionary algorithms are suitable to produce optimal solutions quickly for real-time applications. Compared performance of four applied algorithms in routing and mission planning process declares the superior performance of BBO comparing others; however, **there is only a slight difference between the produced results.**

## 3.6 Chapter Summary

The vehicle is requested to complete the best subset and ordering of the provided tasks while biasing to maximum priority in a restricted mission time. Realistically, it is impossible for a single vehicle to cover all tasks in a single mission in a large-scale operation area. Reaching to the destination is another critical factor for the mission planner that should be taken into consideration. All tasks are distributed in the operating field and mapped in the feature of a connected network, in which some edges of the network are assigned by tasks. The tasks should be prioritized in a way that selected edges (tasks) of the network can govern the AUV to the destination. Hence, the TAR tends to find the best collection of tasks and order them in a way that AUV gets guided to the destination while it is respecting the upper time threshold (mission available time) at all times during the mission. The problem involves multiple objectives that should be satisfied during the optimization process.

In this chapter, a mission-planning model is introduced for AUV task assign-routing employing different evolutionary optimization algorithms, which provides a comprehensive control on mission timing, multitasking in a waypoint cluttered underwater environment to maximize productivity of a single vehicle in a single mission within a limited time interval. This approach operates at the higher level to compromise among the mission available time, maximizing number of highest priority tasks, and guaranteeing reaching to the predefined destination, which is combination of a discrete and a contiguous optimization problem at the same time (analogous to both TSP and BKP NP-hard problems). Relying on competitive performance of the meta-heuristic algorithms in handling computational complexity of NP-hard problems, different meta-heuristic methods including BBO, ACO, GA, and PSO are carefully chosen and configured according to expectation from the corresponding problem; and then used to evaluate the TAR model in finding correct and near optimal solutions in competitive time (real and CPU). The performance and stability of the model in satisfying the metrics of "obtained weight", "number of completed tasks", "mission cost" and "total violation of the model" is investigated and analysed in a quantitative manner through 150 Monte Carlo simulation runs with the initial condition analogous to real underwater mission scenarios.

The simulation results are carefully analysed and discussed in Section 3.5. The study aims to prove the robustness of proposed model employing any sort of algorithms with real-time performance, rather than criticizing or comparing efficiency of the applied algorithms. Relying on analysis of the simulation results, all algorithms reveal very competitive performance in satisfying addressed performance metrics. Variation of the computational time for operation of all algorithms is settled in a narrow bound in a range of seconds that confirm accuracy and real-time performance of the system. Independent of the type of meta-heuristic algorithm applied in this chapter, the proposed framework is able to achieve the mission objectives toward task-time management; and carrying out beneficial deployment for a single vehicle operation in a restricted time. The concept of this chapter is work out and published through two conference papers (Appendices A and B).

This framework is a foundation for improving vehicle's autonomy in higher levels of decision making in prioritizing tasks, time management, and mission management-scheduling, which are core elements of autonomous underwater missions. However, underwater environment poses a considerable variability and uncertainty. The proposed mission planning strategy, however, is not designed efficient for the purpose of handling environment sudden changes, but it gives a general overview of the area that an AUV should fly through (general route), which means reducing the operation area to smaller operating zone for vehicle's deployment. Thus, another benefit of such a planner is reducing the run time by splitting the operation terrain to smaller zones for the vehicles operation. The environmental risks and factors such as time-varying ocean currents, diverse static and dynamic obstacles, and uncertainty of the operation terrain should be considered when planning a transit path in smaller scale. The importance of this issue is investigated through the next chapter.

# Chapter 4

# AUV Online Real-Time Path Planning (ORPP)

## 4.1 Introduction

Robust path planning is an important characteristic of autonomy. It represents the possibilities for vehicles deployment during a mission in presence of different disturbances, which is a complicated procedure due to existence of the inaccurate, unknown, and varying information. Recent advancements in sensor technology and embedded computer systems has opened new possibilities in underwater path planning and made AUVs more capable for handling complicated long range underwater missions. However, there still exist major challenges for this class of the vehicles, where the surrounding environment has a complex spatiotemporal variability and uncertainty. Ocean current variability affect vehicle's motion positively or negatively. This variability can also have a profound impact on vehicle's battery usage and its mission duration. Current variations over time can affect moving obstacles (or waypoints in some cases) and drift them across a vehicle's trajectory; therefore, the planned trajectory may change and become invalid or inefficient. Proper estimation of the events in such a dynamic uncertain environment in long-range operations, outside the vehicle's sensor coverage, is impractical and unreliable. The robustness of a vehicle's path planning to this strong environment variability is a key element to its safety and mission performance. AUV-oriented applications still have several difficulties when operating across a large-scale geographical area in which a large data load of variation of whole terrain condition should be computed repeatedly any time that path replanting is required. This process is computationally inefficient and unnecessary as only awareness of environment changes in vicinity of the vehicle is enough. Figure 4.1 illustrates the operation diagram of this chapter.

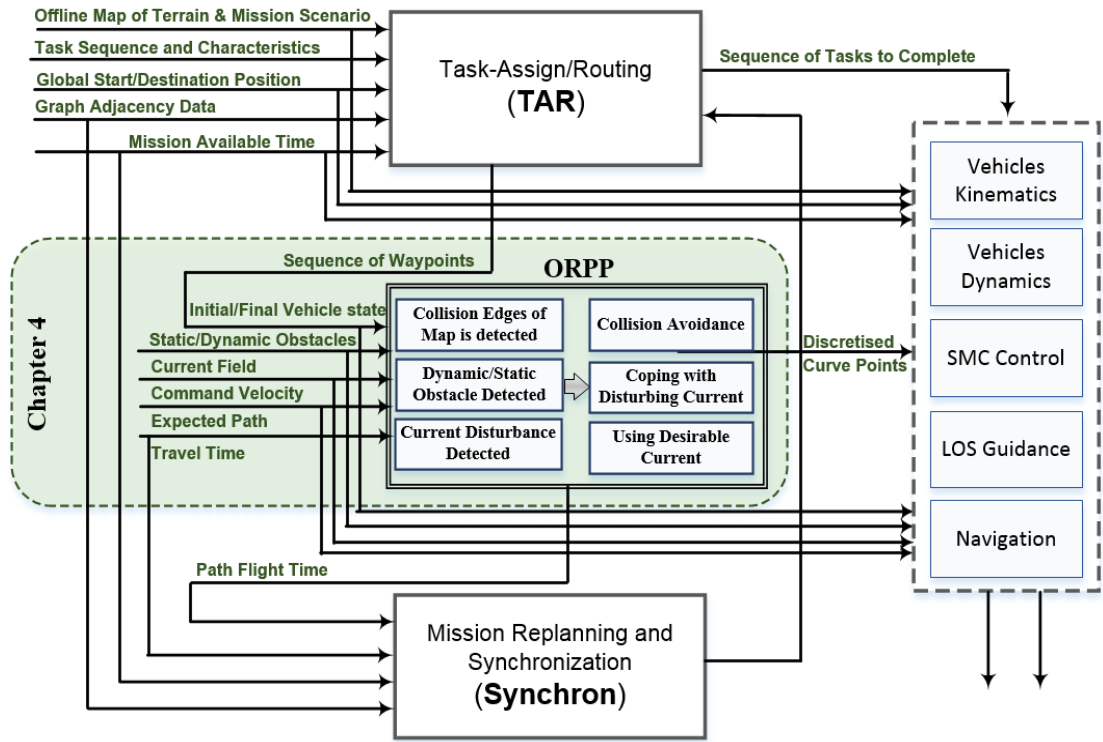

**Figure 4.1.** The operation diagram of the synchronous ARMSP architecture and completion of ORPP system by this chapter.

Path planning is a complicated multi-objective NP-hard problem that has a great impact on vehicle's overall autonomy. Currently there is no polynomial time algorithm to handle a NP-hard problem of even moderate size. Furthermore, finding a pure optimum solution is only possible when the environment is fully known and no uncertainty exists. The modelled underwater environment in this chapter corresponds to a highly dynamic uncertain environment. To satisfy the addressed challenges of the path planning (discussed in Chapter 2) in a time-varying underwater environment, this chapter proposes an online real-time path planning strategy that improves a vehicle's ability to cope with local environment changes. Although the solutions proposed by any meta-heuristic algorithm do not necessarily correspond to the optimal solution, it is more important to control the time, and thus relying on the previously mentioned ability of meta-heuristic algorithms, four evolutionary methods namely PSO, BBO, Differential Evolution (DE), and Firefly Algorithm (FA) are employed in the core of the proposed local path planner to find correct and near optimal solutions in competitive time (real and CPU). In this chapter, the environment is modelled to be more realistic comprising uncertainty of moving, floating/suspended objects and dynamic multiple-layered time varying ocean current. Accordingly, the path planner is facilitated with dynamic re-planning capability.

The objective is to synthesize and analyses of the performance and capability of the mentioned methods for guiding an AUV from an initial loitering point towards the destination point through a comprehensive simulation study. The proposed planner module entails a heuristic for refining the path considering situational awareness of environment. The planner thus needs to accommodate the unforeseen changes in the operating field such as emergence of unpredicted obstacles or variability of current field and turbulent regions. Thereafter, using a penalty function, an augmented cost function embodied all constraints is made by defining the loitering and target points as boundary conditions; static and moving obstacles as path constraints; bounds on the vehicle states associated with the limits of vehicle actuators.

The chapter is organized as follows. Modelling of underlying operating environment including the obstacles and current vector field is described in Section 4.2. The path planning problem formulation, mathematical model of the vehicle together with the mechanism of the online path planner are explained in Section 4.3. An overview of the utilized evolutionary methods and their implementation for the path-planning problem is discussed in Section 4.4. The discussion on simulation results are provided in Section 4.5. The Section 4.6 presents the summary of this chapter.

## 4.2 Modelling Operational Ocean Environment

Realistic modelling of the operational field including offline map, dynamic/static current map, moving and static uncertain obstacles provides a thorough testbed for evaluating the path planner in situations that closely match with the real underwater environment. The ocean environment for the local path planner is modelled as a three dimensional environment $\Gamma_{3D}$:{3.5-by-3.5 $km^2$ (x-y), 1000 $m$(z)} embodied uncertain static and moving obstacles in presence of ocean variable current that makes the simulation process more challenging and realistic. The designed path planner is capable of extracting authorized areas of operation through the real map to prevent colliding with coastal areas, islands, and static/dynamic obstacles. In addition, it is capable of compensating for adverse current flows or taking advantage of favourable ones. In the following, the proposed operating field is described in detail.

### 4.2.1  Offline Map

Existence of prior information about the terrain, location of coasts and static obstacles as the forbidden zones for deployment, position of the start, target, and waypoints in operating area promotes AUV's capability in robust path planning. To model a realistic marine environment, a sample of a real map is utilized as shown in Figure 4.2. To cluster the coastal areas and authorized water zones (allowed for deployment) into separate regions, k-means clustering method [125] is employed. K-means is usually applied for partitioning a data set into k groups. The method is initialized as *k* cluster centres and the clusters are then iteratively refined. It converges when a saturation phase emerges where there is no further chance for changes in assignment of the clusters. The method aims to minimize a squared error function as an objective function defined in (4-1):

$$\arg\min_{C} \sum_{i=1}^{k} \sum_{\partial \in C_i} \| \partial - \mu_i \| \qquad (4-1)$$

where $\partial$ belongs to a set of observations ($\partial_1, \partial_2, \ldots, \partial_n$) in which each observation is a d-dimensional real vector, and $\mu$ is the cluster centres belongs to $C=\{C_1, C_2, \ldots, C_k\}$.

The utilized method, first, takes the original map, presented by an image of size 350-by-350 pixels where each pixel corresponds to $10 \times 10 \ m^2$, in which the allowed zones for the vehicle's operation are illustrated by the blue sections and the coastal areas are depicted with the brown colour. Then, the clustering is applied and the operating and no-fly areas are transformed into the white and black regions in the clustered map, respectively. As a result, the planner works within the white (feasible) regions.

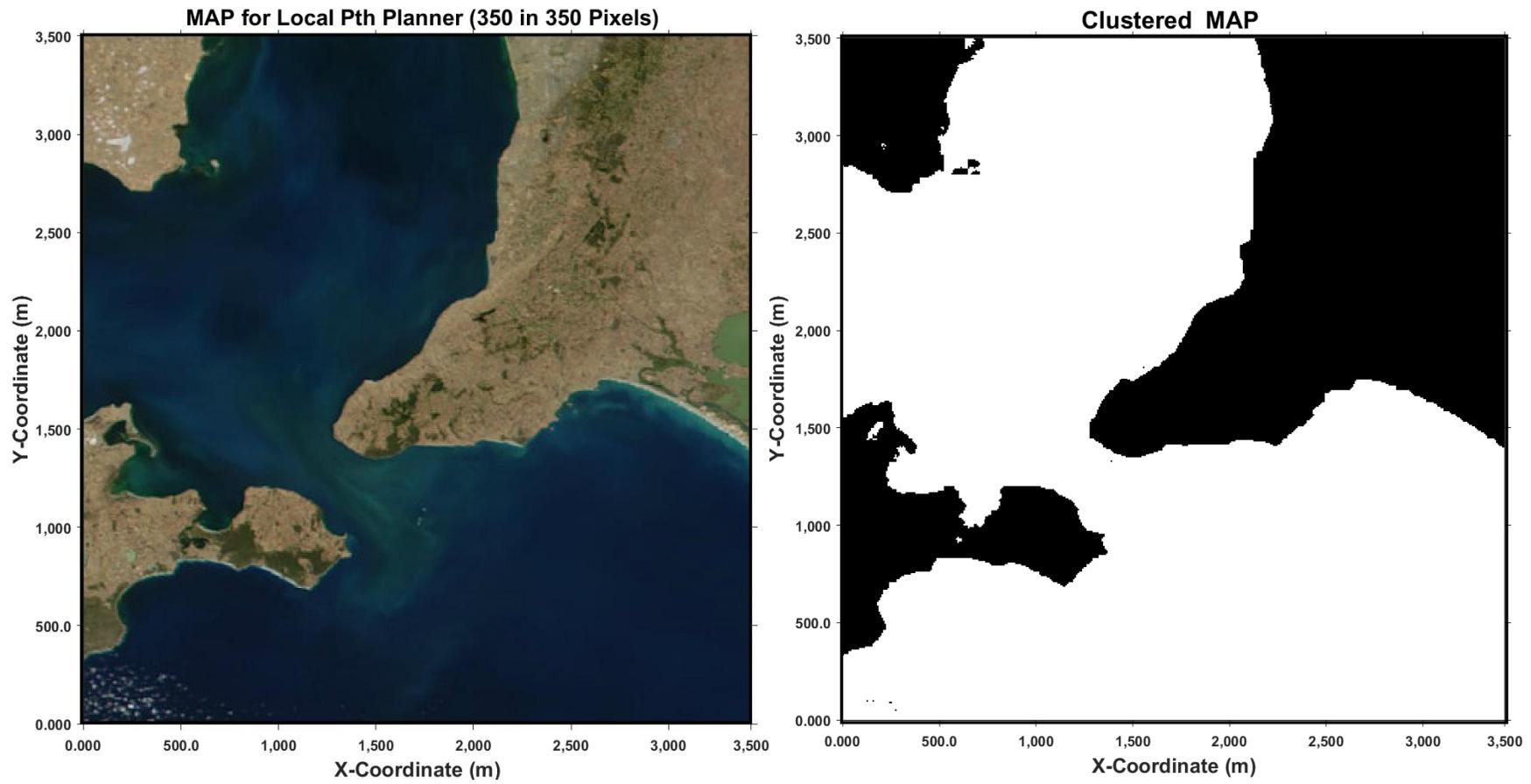

**Figure 4.2.** The original and clustered map selected for simulating the operating field.

The clustered map is converted to a matrix format, in which the matrix size is same to map's pixel density. The corresponding matrix is filled with value of 0 for coastal sections (forbidden area in black colour), value of (0,0.35] for uncertain sections (risky area in grey colour), and value of 1 for water covered area as valid zone for vehicles motion that presented by white colour on the clustered map. The utilized clustering method is capable of clustering any alternative map very efficiently in a way that the blue sections sensed as water covered zones and other sections depending on their colour range categorized as forbidden and uncertain zones. The map analysis in this study is assumed in a 2-dimensional scale, in which depth of water is not considered in the map clustering process. However, the third dimension is taken into account in AUV's deployment and the probable challenges in the 3D volume is simulated through the modelling of the 3D ocean current and different types of obstacles.

### 4.2.2 Mathematical Model with Uncertainty of Static/Dynamic Obstacles

Beside the offline map, where the coasts are assumed as static known obstacles, various uncertain objects are also considered in this study to encompass different possibilities of the real world situations. The AUV is equipped with sonar sensors for measuring the velocity and coordinates of the objects with a level of uncertainty depicted with a normal distribution and formulated as follows:

i. **Quasi-static Uncertain Objects**: Object's position should be placed between the location of AUV's starting and destination spot in the given map. This group of objects are introduced with a fixed-centre generated at commencement with uncertain radius that varying over time according to (4-2).

$$\forall \Theta_{x,y,z}, \ \exists (\Theta_p, \Theta_r, \Theta_{Ur})$$
$$\Theta_p^i \in [(x_s, y_s, z_s), (x_d, y_d, z_d)] - \Theta_{r_{x,y,z}}^i$$
$$\Theta_p \sim N(0, \sigma_1) \qquad (4\text{-}2)$$
$$\Theta_{Ur} \sim U(0, \sigma_2)$$
$$\Theta_r \sim N(\Theta_p, \Theta_{Ur})$$

Where, $\Theta_p$, $\Theta_r$, $\Theta_{Ur}$ represent obstacles' position, radius and uncertainty, respectively. The $(x_s, y_s, z_s)$ and $(x_d, y_d, z_d)$ are the location of start and destination.

ii. **Moving Uncertain Objects:** This group of objects move spontaneously to random positions in the operating environment. The uncertainty encountered on objects position changes over the time according to (4-3).

$$\Theta_p(t) = \Theta_p(t-1) \pm \mathbf{U}(\Theta_{p_0}, \Theta_{Ur}) \tag{4-3}$$

iii. **Dynamic Uncertain Obstacles:** This group of objects moving self-propelled while their motion gets affected by current fields and calculated as follows:

$$\begin{aligned}
U_R^C &= |V_C| \sim \mathbf{N}(0, 0.3) \\
X_{(t-1)} &\sim \mathbf{N}(0, \sigma_3) \\
\Theta_p(t) &= \Theta_p(t-1) \pm \mathbf{N}(\Theta_{p_0}, \Theta_{Ur}) \\
\Theta_r(t) &= B_1 \Theta_r(t-1) + B_2 X_{(t-1)} + B_3 \Theta_{Ur}
\end{aligned} \tag{4-4}$$

$$B_1 = \begin{bmatrix} 1 & U_R^C(t) & 0 \\ 0 & 1 & 0 \\ 0 & 0 & 1 \end{bmatrix}, B_2 = \begin{bmatrix} 0 \\ 1 \\ 1 \end{bmatrix}, B_3 = \begin{bmatrix} 0 \\ 0 \\ U_R^C(t) \end{bmatrix}$$

where the $U_R^C$ is the impact of current velocity ($V_C$) on objects motion. $\Theta_{Ur} \sim \sigma$ is the rate of change in objects position, and $X_{(t-1)} \sim \mathbf{N}(\Theta_p, \sigma_0)$ is the Gaussian normal distribution that assigned to each obstacle and gets updated at each iteration $t$.

### 4.2.3 Mathematical Model of Static/Dynamic Current Field

This study uses a numerical estimator based on multiple Lamb vortices and Navier-Stokes equations [126] for modelling ocean current behaviour in 3D space. Two types of static and dynamic current flow are mentioned in the following.

i. The AUV's deployment usually is assumed on horizontal plane, because vertical motions in ocean structure are generally negligible due to large horizontal scales comparing vertical [126]. Hence, the physical model employed by the AUV to identify the current velocity field $V_C = (u_c, v_c)$ can be mathematically described by (4-5) and (4-6):

$$\begin{aligned}
u_c(\vec{S}) &= -\Im \frac{y - y_0}{2\pi(\vec{S} - \vec{S}^O)^2} \left[ 1 - e^{\frac{-(\vec{S} - \vec{S}^O)^2}{\ell^2}} \right] \\
v_c(\vec{S}) &= \Im \frac{x - x_0}{2\pi(\vec{S} - \vec{S}^O)^2} \left[ 1 - e^{\frac{-(\vec{S} - \vec{S}^O)^2}{\ell^2}} \right]
\end{aligned} \tag{4-5}$$

$$\frac{\partial \omega}{\partial t} + (\vec{V}_C \nabla)\omega = v\Delta\omega$$

$$\omega(\vec{S}) = \frac{\Im}{\pi \ell^2} e^{\frac{-(\vec{S}-\vec{S}^o)^2}{\ell^2}}$$

(4-6)

where the $v$ represents the viscosity of the fluid, $\omega$ gives the vorticity, $\Delta$ and $\nabla$ are the Laplacian and gradient operators, respectively; $S$ is a 2D spatial space, $S^o$ is the centre of the vortex, $\ell$ and $\Im$ are the radius and strength of the vortex.

ii. A 3D turbulent dynamic current field is estimated by a multiple layer structure based on the generated 2D current. The currents' circulations patterns gradually change with depth. Therefore, to estimate continuous circulation patterns of each subsequent layer, a recursive application of Gaussian noise is applied to the parameters of 2D case. A probability density function of the multivariate normal distribution is employed to calculate the vertical profile $w_c$ of the 3D current $V_C=(u_c,v_c,w_c)$, which is given by (4-7):

$$w_c(\vec{S}) = \gamma \Im \frac{1}{\sqrt{\det(2\pi \lambda_w)}} \times e^{\frac{-(\vec{S}-\vec{S}^o)^T}{2\lambda_w(\vec{S}-\vec{S}^o)}}, \lambda_w = \begin{bmatrix} \ell & 0 \\ 0 & \ell \end{bmatrix}$$

(4-7)

where, $\lambda_w$ is a covariance matrix of the vortex radius $\ell$. A parameter $\gamma$ is used to scale the $w_c$ from the current horizontal profile $u_c$ and $v_c$ due to weak vertical motions of ocean environment. For generating dynamic time varying ocean current, Gaussian noise can be applied on $S^o, \Im,$ and $\ell$ parameters recursively as defined in (4-8):

$$V_c = f(\vec{S}^O, \Im, \varsigma)$$
$$S_i^o = A_1 S_{i-1}^o + A_2 X_{(i-1)}^{S_x} + A_3 X_{(i-1)}^{S_y}$$
$$\ell_{(i)} = A_1 \ell_{(i-1)} + A_2 X_{(i-1)}^{\ell}$$
$$\Im_{(i)} = A_1 \Im_{(i-1)} + A_2 X_{(i-1)}^{\Im}$$
$$A_1 = \begin{bmatrix} 1 & 0 \\ 0 & 1 \end{bmatrix}, A_2 = \begin{bmatrix} U_R^C(t) \\ 0 \end{bmatrix}, A_3 = \begin{bmatrix} 0 \\ U_R^C(t) \end{bmatrix}$$

(4-8)

where, $U_R^C(t)$ is the update rate of current field in time $t$, and the rest of the unknown parameters in (4.8) are defined based on normal distribution as follows: $X^{Sx}_{(i-1)}$~$N(0,\sigma_{Sx})$, $X^{Sy}_{(i-1)}$~$N(0,\sigma_{Sy})$, $X^{\ell}_{(i-1)}$~$N(0,\sigma_{\ell})$, $X^{\Im}_{(i-1)}$~$N(0,\sigma_{\Im})$. The current field continuously updated every 4s during the AUV path planning process.

## 4.3 Formulation of the Local Path-Planning Problem

In the general definition of the AUV path-planning problem, information of the current state of the AUV(s) and environments are conducted to compute a desirable trajectory from one point to another, while specific objectives are satisfied respecting existing constraints and vehicle(s) properties. An efficient path takes the minimum travel time/distance and is safe enough to satisfy collision constraints. The path planner in this research is capable of extracting feasible areas of a real map to determine the allowed spaces for deployment, where coastal area, islands, static/dynamic objects and ocean current are taken into account. Water current may have positive or disturbing effect on vehicles deployment, where an undesirable current is a disturbance itself and it also can push floating objects across the AUV's path. On the other hand, the desirable current can motivate AUVs motion towards its target; hence, adapting to water current variations is an important factor affecting optimality of the generated trajectory and considerably diminish the total AUV mission costs. The AUV is assumed to have a freely deploying rigid body in 3D space with six degrees of freedom. AUV's state variables of body frame $\{b\}$ and NED (North-East-Depth) frame$\{n\}$ that provides six degrees of freedom for its motion is presented using a set of ordinary differential equations given by (4-9) and (4-10), which present dynamics and kinematics of the vehicle over the time [127].

$$\begin{cases} \{n\} \rightarrow \eta : (X,Y,Z,\varphi,\theta,\psi) \\ \{b\} \rightarrow \upsilon : (u,v,w,p,q,r) \end{cases} \quad (4\text{-}9)$$

$$\begin{bmatrix} \dot{X} \\ \dot{Y} \\ \dot{Z} \end{bmatrix} = \begin{bmatrix} {}^n_b R \end{bmatrix} \begin{bmatrix} u \\ v \\ w \end{bmatrix}; \quad \begin{bmatrix} {}^n_b R \end{bmatrix} = \begin{bmatrix} \cos\psi\cos\theta & -\sin\psi & \cos\psi\sin\theta \\ \sin\psi\cos\theta & \cos\psi & \sin\psi\sin\theta \\ -\sin\theta & 0 & \cos\theta \end{bmatrix} \quad (4\text{-}10)$$

Here, the $\eta$ is state of the vehicle in the$\{n\}$ frame; $X,Y,Z$ gives the vehicle's position along the generated path, and $\varphi,\theta,\psi$ are the Euler angles of roll, pitch, and yaw, respectively. The $\upsilon$ is AUV's velocity in the $\{b\}$ frame; $u,v,w$ are AUV's directional velocities of surge, sway and heave; and $p,q,r$ are the AUV's rotational velocities in the $x$-$y$-$z$ axis. The $[{}^n_b R]$ is a rotation matrix that transforms the body frame $\{b\}$ into the $\{n\}$ frame. Figure 4.3 presents the AUV's and current coordinates in $\{n\}$ and $\{b\}$ frames.

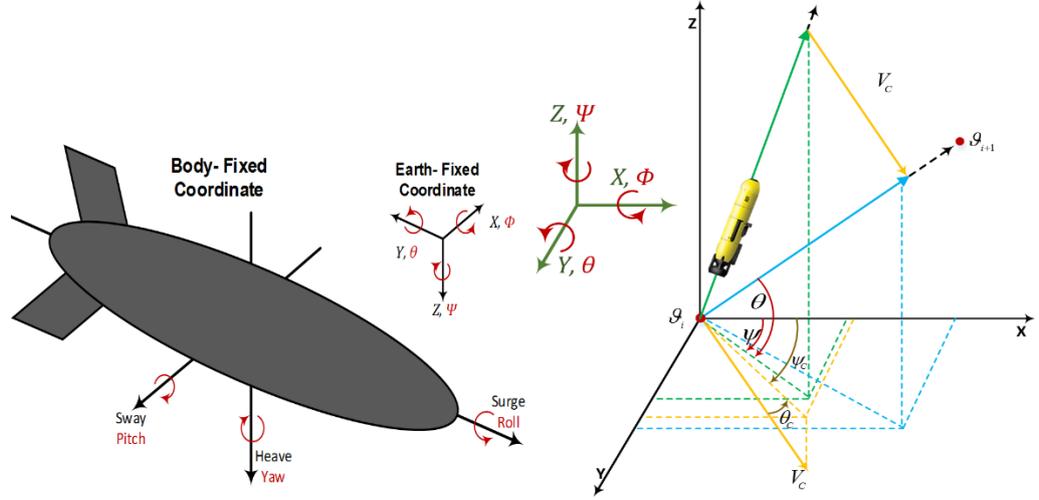

**Figure 4.3.** Vehicle's and ocean current coordinates in NED and Body frame, respectively.

The potential trajectory $\wp_i$ in this research is generated based on B-Spline curves captured from a set of control points $\vartheta:\{\vartheta^1_{x,y,z},...,\vartheta^i_{x,y,z},...,\vartheta^n_{x,y,z}\}$ as presented by (4-11) and (4-12). The path planning is a constraint optimization problem in which the main goal is to provide a safest and quickest trajectory between two points. Placement of these control points play a substantial role in determining the optimal path. All $\vartheta$ should be located in respective search space limited to predefined Upper-Lower bounds of $\vartheta \in [U_\vartheta, L_\vartheta]$ in Cartesian coordinates. Accordingly, path curve is generated as follows:

$$L_\vartheta \equiv P^a_{x,y,z} \quad \& \quad U_\vartheta \equiv P^b_{x,y,z}$$

$$\begin{cases} \vartheta^i_x = L^i_{\vartheta(x)} + Rand^x_i [U^i_{\vartheta(x)} - L^i_{\vartheta(x)}] \\ \vartheta^i_y = L^i_{\vartheta(y)} + Rand^y_i [U^i_{\vartheta(x)} - L^i_{\vartheta(x)}] \\ \vartheta^i_z = L^i_{\vartheta(z)} + Rand^z_i [U^i_{\vartheta(x)} - L^i_{\vartheta(x)}] \end{cases} \quad (4\text{-}11)$$

$$\left. \begin{array}{l} X = \sum_{i=1}^{n} \vartheta^i_x \times B_{i,K} \\ Y = \sum_{i=1}^{n} \vartheta^i_y \times B_{i,K} \\ Z = \sum_{i=1}^{n} \vartheta^i_z \times B_{i,K} \end{array} \right\} \mapsto \wp_{x,y,z} = \sum_{x_s,y_s,z_s}^{|\wp|} \sqrt{\Delta X^2 + \Delta Y^2 + \Delta Z^2} \quad (4\text{-}12)$$

where the $L_\vartheta$ is the lower bound that corresponds to location of the start point ($L_\vartheta \equiv p^a_{x,y,z}$) and $U_\vartheta$ is the upper bound that corresponds to location of the target point ($U_\vartheta \equiv p^b_{x,y,z}$) in Cartesian coordinates. The $B_{i,K}$ is a blending function used to slice the curve, and $K$ is the order of the curve representing smoothness of the curve. For further information, refer to [128]. The vehicle should be oriented along the path segments generated according to given (4-13) to (4-15). Water currents continually affect the vehicle's motion; so that the vehicle's angular velocity components along the path curve $\wp$ is calculated by:

$$\psi = \tan^{-1}\left(\frac{|\vartheta_y^{i+1} - \vartheta_y^i|}{|\vartheta_x^{i+1} - \vartheta_x^i|}\right) = \tan^{-1}\left(\frac{\Delta Y}{\Delta X}\right)$$

$$\theta = \tan^{-1}\left(\frac{-|\vartheta_z^{i+1} - \vartheta_z^i|}{\sqrt{(\vartheta_y^{i+1} - \vartheta_y^i)^2 + (\vartheta_x^{i+1} - \vartheta_x^i)^2}}\right) = \tan^{-1}\left(\frac{-\Delta Z}{\sqrt{\Delta X^2 + \Delta Y^2}}\right) \quad (4\text{-}13)$$

$$\begin{cases} u_c = |V_C|\cos\theta_c \cos\psi_c \\ v_c = |V_C|\cos\theta_c \sin\psi_c \end{cases} \Rightarrow \begin{array}{l} u = |\upsilon|\cos\theta\cos\psi + |V_C|\cos\theta_c\cos\psi_c \\ v = |\upsilon|\cos\theta\sin\psi + |V_C|\cos\theta_c\sin\psi_c \\ w = |\upsilon|\sin\theta \end{array} \quad (4\text{-}14)$$

$$\langle X(t), Y(t), Z(t) \rangle \approx \left\langle \sum_{i=1}^{n}(\vartheta_x^i(t), \vartheta_y^i(t), \vartheta_z^i(t)) \right\rangle$$

$$\wp_{x,y,z}^i = \sum_{i=1}^{|\wp|} \sqrt{(\vartheta_x^{i+1} - \vartheta_x^i)^2 + (\vartheta_y^{i+1} - \vartheta_y^i)^2 + (\vartheta_z^{i+1} - \vartheta_z^i)^2} \quad (4\text{-}15)$$

$$\wp(t) = [X(t), Y(t), Z(t), \psi(t), \theta(t), u(t), v(t), w(t)]$$

where, the $|V_C|$ is magnitude of the current speed, $\psi_c$ and $\theta_c$ give directions of current vector in horizontal and vertical planes, respectively. Figure 4.4 illustrates the schematic of the AUV path planning process.

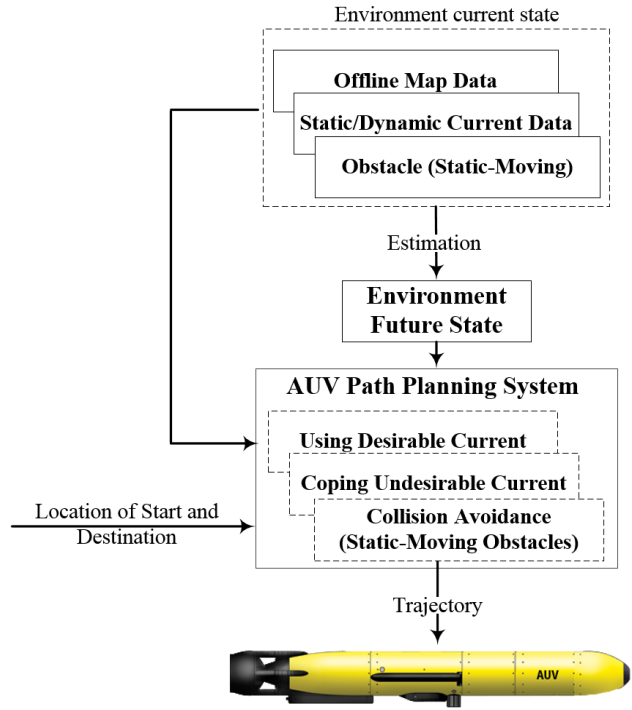

**Figure 4.4.** The operation flow of the AUV path planning.

### 4.3.1 Path Optimization Criterion

The path planning is a NP-hard optimization problem that aims to guide the vehicle towards a specific location encountering dynamics of the ocean and kinematics of the vehicle. An efficient path takes the minimum travel time/distance and is safe enough to satisfy collision constraints. Path planning aims to satisfy following aspects of performance:

– ***To minimize path length/time:*** the path planning in this study aims to find the shortest and safest path between two waypoints taking the advantages of desirable current while avoiding collision and adverse current fields. The AUV is considered to have constant thrust power; therefore, the battery usage for a path is a constant function of the time and distance travelled. Performance of the generated path is evaluated based on overall time consumption, which is proportional to travelled distance. Assuming the constant water referenced velocity $|v|$ for the vehicle, the travel time is estimated by

$$\forall \wp^i_{x,y,z} \rightarrow L_\wp = \sum_{i=p^a_{x,y,z}}^{|\wp|} \sqrt{(X_{i+1}(t) - X_i(t))^2 + (Y_{i+1}(t) - Y_i(t))^2 + (Z_{i+1}(t) - Z_i(t))^2}$$

$$T_\wp = \sum_1^{|\wp|} \frac{L_\wp}{|v|}$$

(4-16)

where, the $L_\wp$ and $T_\wp$ are path length and path flight time, respectively.

– ***To satisfy path constraints and safe deployment:*** A vehicle's safe and confident deployment is a critical issue that should be taken into consideration at all stages of the mission in a vast and uncertain environment. The resultant path should be safe and feasible. The environmental constraints in this chapter are associated with the forbidden zones of map, collision borders of obstacles, and current disturbance. The current disturbance may cause drift to vehicle's desired motion. AUV's directional velocity components of $u,v$ and its $\psi,\theta$ orientation should be constrained to $u_{max}$, $[v_{min}, v_{max}]$, $\theta_{max}$, and $[\psi_{min}, \psi_{max}]$, while current magnitude $|V_c|$ and direction $\psi_c$, $\theta_c$ are encountered in all states along the path. The path also shouldn't cross the forbidden collision areas $(\sum M, \Theta)$ associated with map or obstacles. The cost function is defined as a combination of path time and corresponding penalty function described in (4-17).

$$\wp(t) = [X(t), Y(t), Z(t), \psi(t), \theta(t), u(t), v(t), w(t)]$$

$$\nabla_{\Sigma_{M,\Theta}} = \begin{cases} 1 & \wp_{x.y.z}(t) = Coast : Map(x,y) = 1 \\ 1 & \wp_{x.y.z}(t) \cap \bigcup_{N\Theta} \Theta(\Theta_p, \Theta_r, \Theta_{Ur}) \\ 0 & Otherwise \end{cases}$$

$$\nabla_{\wp} = \begin{cases} \varepsilon_{z\min} \times \min(0; Z(t) - Z_{\min}) \\ \varepsilon_{z\max} \times \max(0; Z(t) - Z_{\max}) \\ \varepsilon_u \times \max(0; u(t) - u_{\max}) \\ \varepsilon_v \times \max(0; |v(t)| - v_{\max}) \\ \varepsilon_\theta \times \max(0; \theta(t) - \theta_{\max}) \\ \varepsilon_\psi \times \max(0; |\dot{\psi}(t)| - \psi_{\max}) \\ \varepsilon_{\Sigma_{M,\Theta}} \times \nabla_{\Sigma_{M,\Theta}} \end{cases} \quad (4\text{-}17)$$

$$C_\wp = L_\wp + \sum_{i=1}^{n} Q_i f(\nabla_\wp)$$

In Equation (4-17), the $Q_i f(\nabla_\wp)$ is a weighted violation function that respects the AUV Kino-dynamic and collision constraints including depth violation ($Z$), to prevent the path from straying outside the vertical operating borders, surge ($u$), sway ($v$), yaw ($\psi$), pitch($\theta$) violations, and the collision violation ($\nabla_{\Sigma M,\Theta}$) specified to prevent the path from collision danger. The $\varepsilon_{zmin}$, $\varepsilon_{zmax}$, $\varepsilon_u$, $\varepsilon_v$, $\varepsilon_\theta$, $\varepsilon_\psi$, and $\varepsilon_{\Sigma M,\Theta}$ respectively denote the impact of each constraint violation in calculation of total path cost $C_\wp$.

### 4.3.2 On-line Path Re-planning based on Previous Solution

The underwater variable environment poses several challenges that makes the AUV mission problematic and cumbersome. One solution to overcome the mentioned difficulties is to establish an on-line path planning approach that incorporates dynamic and reactive behaviours. Through this approach, the vehicle exercises situational awareness of the environment and refines the path until vehicle reaches the specified target waypoints. By doing so, the vehicle is able to make use of desirable current flows at any time that current map is updated for speeding up its motion and saving the energy usage.

*CHAPTER 4. AUV ONLINE REAL-TIME PATH PLANNING*

In this research, an on-line path planning mechanism is utilized, so that when the path re-planning flag is triggered, the new optimal path is generated from current position to the destination by refining the previous solution. In other words, the solution of previously optimal path is used as an initial solution for computing the new path and current states of the vehicle are replaced as the new initial boundary conditions. By following this approach, first, the vehicle is able to cope with the uncertainties of the operating field and in fact, this strategy provides a near closed-loop guidance configuration; that can be supported with a minimal computational burden, as there is no need to compute the path from the scratch, as opposed to reactive planning strategy [108]. The online path planning mechanism is summarized in Figure 4.5 and Algorithm 4.1.

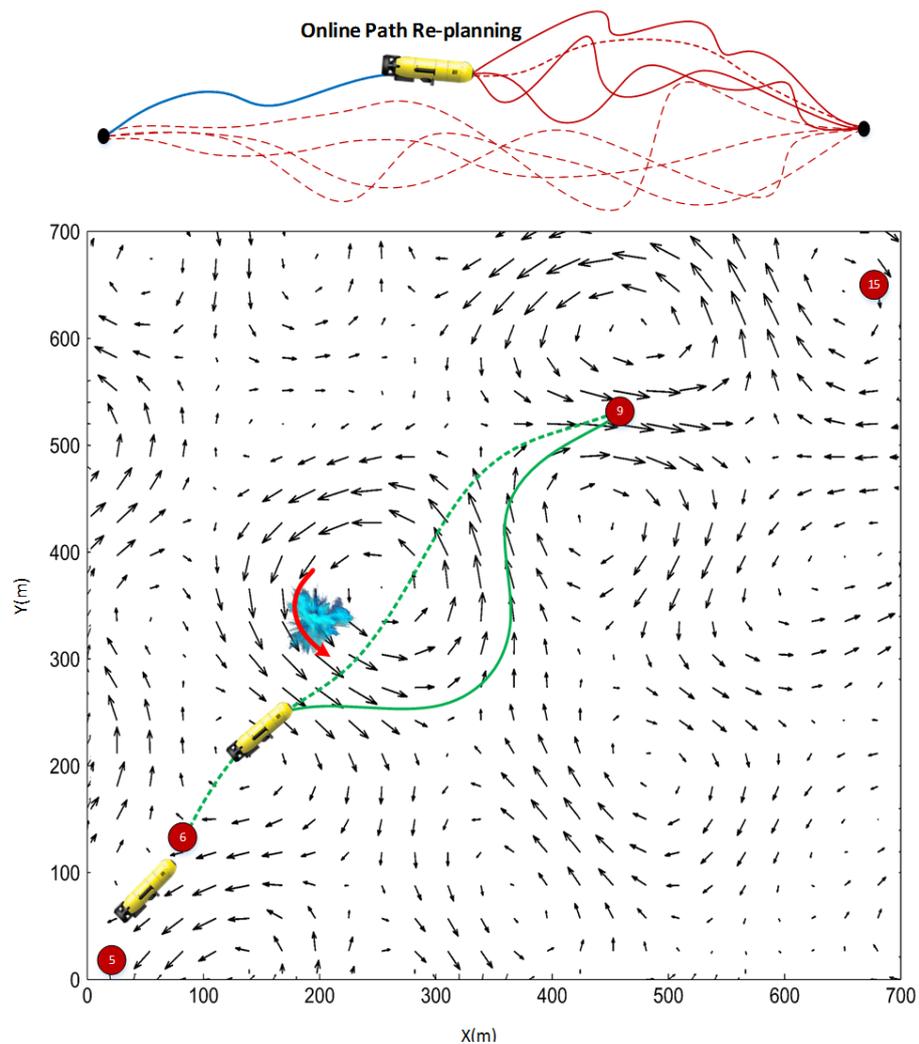

**Figure 4.5.** Online path re-planning and regular path planning (online re-planning reuses the subset of the past solutions).

```
Online Path Re-Planning Mechanism
─ Get the position of start and target points
─ Generate the initial optimum path using Eq.(4-12) to (4-15)
─ Evaluate the generated path by defined C_℘ in Eq.(4-17)
─ Check Path-Replanning Flag continuously
BEGIN
  If Flag ==1
    ─   Apply the current states of the vehicle as the initial condition
    ─   Set the current location as the start point for local path planning process
    ─   Use the solution at hand (from previous path) as an initial solution or
        initial guess for the proposed evolutionary method
    ─   Generate a new path
    ─   Check the problem optimality conditions Eq.(4-17)
    If Eq.(4-17) is satisfied
      Follow the new path
      Terminate the planning process
      else
      Follow the pervious path
      Terminate the planning process
      Back to BEGIN
    End
  End
END
```

**Algorithm 4.1.** Online path planning pseudo-code.

## 4.4 Overview of the Applied Evolutionary Algorithms

The path planning is an optimization problem aims to guide the vehicle between two specific points with minimum time due to battery limitation while ensuring safe deployment coping various real world disturbances. Before going through the brief review of the employed evolutionary algorithms, it is noteworthy to mention how they are applied over the optimization problem. As mentioned before, B-Spline curves are exploited to parameterize the path. The generated path by B-Spline curves captured from a set of control points like $\vartheta=\{\vartheta_1, \vartheta_2,...,\vartheta_i..., \vartheta_n\}$ in the problem space with coordinates of $\vartheta_1:(x_1,y_1,z_1),...,\vartheta_n:(x_n,y_n,z_n)$, where $n$ is the number of corresponding control points. Therefore, appropriate location of these control points plays a substantial role in determining the optimal path. The mathematical description of the B-Spline coordinates is given by (4-12). Application of different meta-heuristic methods may result in very diverse outcome on the same problem due to specific nature each problem. Hence, accurate selection of the algorithm and proper matching of the algorithms' functionalities according to the nature of a particular problem is an important issue to be considered. To this end, the following popular meta-heuristics are investigated as they have performed promisingly in solving NP hard problems and particularly on motion planning problems.

### 4.4.1 Application of FA on ORPP Approach

Firefly Algorithm (FA) is a kind of meta-heuristic algorithm [129] captured by sampling the flashing patterns of fireflies. In the FA process, the fireflies are attracted to each other according to their brightness, which fades when fireflies' distance is increased. The brighter fireflies attract the others in the population. The firefly's attraction depends on their distance $L$ and received brightness from the adjacent fireflies; thus, the firefly $\chi_i$ approaches the $\chi_j$ according to their attraction parameter $\beta$, calculated by (4-23):

$$L_{ij} = \|\chi_j - \chi_i\| = \sqrt{\sum_{q=1}^{d}(x_{i,q} - x_{j,q})^2}$$
$$\beta = \beta_0 e^{-\gamma L^2} \qquad (4\text{-}23)$$
$$\chi_i^{t+1} = \chi_i^t + \beta_0 e^{-\gamma L_{ij}^2}(\chi_j^t - \chi_i^t) + [rand - \frac{1}{2}]$$
$$\alpha_t = \alpha_0 \delta^t, \delta \in (0,1)$$

In Equation (4-23), the coordinate firefly $\chi_i$ in d dimension space contains $q$ components of $x_{i,q}$. The $\beta_0$ is attraction value of firefly $\chi_i$ when $L$=0. The $\alpha_t$ and $\alpha_0$ are the randomization factor and initial randomness scaling value, respectively. The $\gamma$ and $\delta$ are light absorption coefficient and damping factor, respectively. It is noteworthy to mention that there should be a suitable balance between mentioned parameters because in a case that $\gamma \rightarrow 0$ the algorithm mechanism turns to particle swarm optimization, while $\beta_0 \rightarrow 0$ turns the movement of fireflies to a simple random walk [129]. In path planning procedure, the fireflies correspond to B-Spline control points ($\{\chi_1=\vartheta_1^{x,y,z}, ..., \chi_i=\vartheta_i^{x,y,z}, ..., \chi_n=\vartheta_n^{x,y,z}\}$) tends to be optimized toward the best solution in the search space. The FA is privileged to apply an automatic subdivision approach that improves its capabilities in dealing with highly nonlinear and multi-objective problems [130]. The attraction has inverse relation with distance; therefore, the firefly population is subdivided and subgroups swarm locally in parallel toward the global optima, which this mechanism enhances the convergence rate of the algorithm. Pseudo-code of FA mechanism on path planning approach is given in Algorithm 4.1. Figure 4.6, also illustrates the encoding scheme of the fireflies with the corresponding path control points.

```
Procedure of FA on ORPP
Initialization phase:
  ▪ Initialize population of fireflies χ_i with the control points ϑ_i
  ▪ Define light absorption coefficients ε
  ▪ Initialize the attraction coefficient β_0
  ▪ Set the damping factor of κ
  ▪ Initialize the randomness scaling factor of α_0
  ▪ Set the parameter of randomization α_t
  ▪ Set the maximum iteration t_max
  ▪ Set the number of population i_max
  For t =1 to t_max
    For i =1 to i_max
      Reconstruct a path according to χ_i
      Evaluate the path C_℘(χ_i(t))
      Update light intensity of χ_i
      For j =1 to i
        Reconstruct a path according to χ_j
        Evaluate the path C_℘(χ_j(t))
        Update light intensity of χ_j
          If (βj > βi),
            Move firefly i towards j
          end (if)
      end (For)
    end (For)
    Rank the fireflies and find the current best
  end (For)
Output result
```

**Algorithm 4.2.** Pseudo-code of FA based path planning.

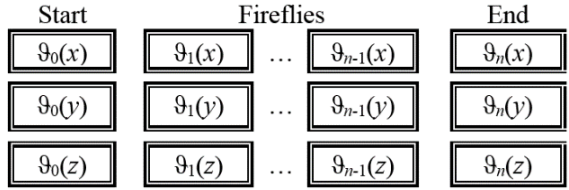

**Figure 4.6** Fireflies encoding based on the path control points.

## 4.4.2 Application of DE on ORPP Approach

The DE algorithm is a population-based optimization algorithm that was introduced as a modified version of genetic algorithm and applies similar evolution operators (crossover, mutation, selection) [131]. DE uses floating-point numbers and real coding for presenting problem parameters that enhances solution quality and provides faster optimization. The algorithm uses a non-uniform crossover and differential mutation;

then applies selection operator to propel the solutions toward the optimum area in the search space. Algorithm 4.3 presents the mechanism of DE algorithm.

---
**Procedure of DE on ORPP**

---
Initialization phase:
- Initialize population of solution vectors randomly $\chi_i^{x,y,z}$ with the control points $(\vartheta^i_x, \vartheta^i_y, \vartheta^i_z)$
- Set the maximum number of iteration $t_{max}$
- Choose appropriate parameters for the population size $i_{max}$
- Set the mutation coefficient
- Set the crossover coefficient

  **For** $t = 1$ **to** $t_{max}$
    **For** $i = 1$ **to** $i_{max}$
      Reconstruct a path according to $\chi_{i,t}$
      Evaluate the path
      Determine the *donor*
      Apply mutation using
      $\dot{\chi}_{i,t} = \chi_{r3,t} + F(\chi_{r1,t} - \chi_{r2,t})$
      $r1, r2, r3 \in \{1, ..., i_{max}\}, r1 \neq r2 \neq r2 \neq i, \quad F \in [0, 1+]$
      **For** $j = 1$ **to** $i$
        Apply crossover using $\chi_{i,t}$ mutant solution $\dot{\chi}_{i,t}$ to get the $\ddot{\chi}_{i,t}$
        Reconstruct a path according to $\chi_{i,t}, \dot{\chi}_{i,t}$, and $\ddot{\chi}_{i,t}$
        Evaluate the corresponding paths to $\chi_{i,t}, \dot{\chi}_{i,t}$, and $\ddot{\chi}_{i,t}$
        **if** $C_\wp(\chi_{i,t}) \leq C_\wp(\dot{\chi}_{i,t})$
          $\dot{\chi}_{i,t+1} = \chi_{i,t}$
          **if** $C_\wp(\chi_{i,t}) \leq C_\wp(\ddot{\chi}_{i,t})$
            $\dot{\chi}_{i,t+1} = \chi_{i,t}$
            **else**
            $\dot{\chi}_{i,t+1} = \ddot{\chi}_{i,t}$
          **end (if)**
        **else**
          $\dot{\chi}_{i,t+1} = \dot{\chi}_{i,t}$
        **end (if)**
      **end (For)**
    **end (For)**
    Select the best solutions to transfer to next iteration
  **end (For)**
Output best solution and the corresponding paths

---

**Algorithm 4.3.** Pseudo-code of DE based path planning.

The mechanism of the DE algorithm is given below:

1) ***Initialization:*** In this step the population is initialized with solution vectors $\chi_i$, ($i = 1, ..., i_{max}$), in which any vector $\chi_i^{x,y,z}$ is assigned with an arbitrary path $\wp^i_{x,y,z}$ while the control points $\vartheta$ correspond to elements of $\chi_i^{x,y,z}$ vector as given by (4-24):

$$\forall i \in \{1, ..., i_{max}\}$$
$$\forall t \in \{1, ..., t_{max}\} \quad\quad\quad\quad\quad\quad\quad\quad\quad\quad (4\text{-}24)$$
$$\chi_{i,t}(\chi^x_{i,t}, \chi^y_{i,t}, \chi^z_{i,t}) \equiv (\vartheta^i_x(t), \vartheta^i_y(t), \vartheta^i_z(t))$$

where $i_{max}$ is the population size, and $t_{max}$ is the maximum number of iterations.

2) *Mutation:* The main reason for superior performance of DE comparing to GA is using an efficacious mutation scheme. Three solution vector like $\chi_{r1,t}$, $\chi_{r2,t}$, and $\chi_{r3,t}$ are picked randomly in iteration *t* and the difference vector between them is utilized in the mutation procedure as given by (4-25). This kind of mutation uses a *donor* parameter, which the *donor* is one of the randomly selected triplets and accelerates rate of convergence, given by (4-26):

$$\dot{\chi}_{i,t} = \chi_{r3,t} + S_f(\chi_{r1,t} - \chi_{r2,t})$$
$$r1, r2, r3 \in \{1,...,i_{max}\}$$
$$r1 \neq r2 \neq r3 \neq i, F \in [0,1+] \tag{4-25}$$

$$donor = \sum_{i=1}^{3} \left( \lambda_i \Bigg/ \sum_{j=1}^{3} \lambda_j \right) \chi_{ri,G} \tag{4-26}$$

where $\chi`_{i,t}$ and $\chi_{i,t}$ are mutant and parent solution vector, respectively. $S_f$ is a scaling factor that balances the difference vector ($\chi_{r1,t}$-$\chi_{r2,t}$) and higher $S_f$ enhances algorithm's exploration capability. $\lambda_j \in [0,1]$ is a uniformly distributed parameter.

3) *Crossover:* This operator shuffles the mutant individuals and the existing population members using (4-27):

$$\begin{cases} \chi_{i,t} = (x_{1,i,t},...,x_{n,i,t}) \\ \dot{\chi}_{i,t} = (\dot{x}_{1,i,t},...,\dot{x}_{n,i,t}) \\ \ddot{\chi}_{i,t} = (\ddot{x}_{1,i,t},...,\ddot{x}_{n,i,t}) \end{cases}$$
$$\ddot{\chi}_{j,i,t} = \begin{cases} \dot{\chi}_{j,i,t} & rand_j \leq r_C \vee j = k \\ \chi_{j,i,t} & rand_j \leq r_C \wedge j \neq k \end{cases}; \begin{cases} j = 1,...,n \\ n \in [1, i_{max}] \end{cases} \tag{4-27}$$

where $\ddot{\chi}_{i,t}$ is the produced offspring and $r_C \in [0,1]$ is the crossover rate.

4) **Validation and Selection:** The solutions produced by the mutation and crossover get evaluated and the best solutions are shifted to the next generation (*t+1*). All solutions are evaluated by predefined cost function and the worst solutions are discarded from the population. The selection procedure is given by (4-28). Algorithm 4.3 expresses the process of DE.

$$\dot{\chi}_{i,t+1} = \begin{cases} \dot{\chi}_{i,t} & C_\wp(\dot{\chi}_{i,t}) \leq C_\wp(\chi_{i,t}) \\ \chi_{i,t} & C_\wp(\dot{\chi}_{i,t}) > C_\wp(\chi_{i,t}) \end{cases};$$
$$\ddot{\chi}_{i,t+1} = \begin{cases} \ddot{\chi}_{i,t} & C_\wp(\ddot{\chi}_{i,t}) \leq C_\wp(\chi_{i,t}) \\ \chi_{i,t} & C_\wp(\ddot{\chi}_{i,t}) > C_\wp(\chi_{i,t}) \end{cases} \tag{4-28}$$

### 4.4.3 Application of PSO on ORPP Approach

The PSO is an optimization method that is well suited on path planning problem due to its efficiency in handling complex and multi-objective problems (More explanation and the general definitions of PSO process can be found in Chapter 3: Section 3.4.4). In the path-planning problem, the particles are coded by coordinates of a potential path, generated by B-Spline curves. The position and velocity parameters of the particles correspond to the coordinates of the B-Spline control points ($\vartheta$) utilized in path generation. As the PSO iterates, each particle is attracted towards its respective local attractor based on the outcome of the particle's individual and swarm search. The most attractive benefit of the PSO is that it is easy to implement and it requires less computation to converge to the optimum solution. Algorithm 4.4 provides the mechanism of the PSO-based path planning.

---

**PSO based Path Planning**

Initialize each particle by random velocity and position in following steps:
- Assign B-Spline control points $\vartheta_i$ as particle position $\chi_i$
- Initialize each particle with random velocity $v_i$ in range of predefined bounds $\beta^i{}_\vartheta = [U^i{}_\vartheta, L^i{}_\vartheta]$.
- Choose appropriate parameters for the population size ($i_{max}$)
- Set the number of control-points $m$ used to generate the B-Spline path
- Set the maximum number of iterations ($t_{max}$)
- Initialize $\chi_i^{P\text{-}best}(1)$ with current position of each particle at first iteration $t=1$.
- Set the $\chi^{G\text{-}best}(1)$ with the best particle in initial population at $t=1$.

**For** $t=1$ **to** $t_{max}$
  Evaluate each candidate particle according to given cost function
  **For** $i=1$ **to** $i_{max}$
    Updated the particles $\chi_i^{P\text{-}best}$ and $\chi^{G\text{-}best}$ at iteration $t$
    **if** $C_\wp(\chi_i(t)) \leq C_\wp(\chi_i^{P\text{-}best}(t-1))$
      $\chi_i^{P\text{-}best}(t) = \chi_i(t)$
    **else**
      $\chi_i^{P\text{-}best}(t) = \chi_i^{P\text{-}best}(t-1)$
    **end (if)**

$$\chi^{G\text{-}best}(t) = \underset{1 \leq i}{\arg\min}\, C_\wp(\chi_i^{P\text{-}best}(t))$$

    Update the state of the particle in the swarm
$$v_i(t) = \omega v_i(t-1) + c_1 r_1 \left[\chi_i^{P\text{-}best}(t-1) - \chi_i(t-1)\right] + c_2 r_2 \left[\chi^{G\text{-}best}(t-1) - \chi_i(t-1)\right]$$
$$\chi_i(t) = \chi_i(t-1) + v_i(t)$$

    Evaluate each candidate particle $\chi_i$ according to given cost function $C_\wp(\chi_i(t))$
  **end (For)**
  Transfer best particles to next generation
**end (For)**
Output $\chi^{G\text{-}best}$ and its correlated path as the optimal solution

**Algorithm 4.4.** Pseudo-code of PSO based path planning.

### 4.4.4 Application of BBO on ORPP Approach

The BBO is an evolution-based technique that mimics the equilibrium pattern of inhabitancy in biogeographical island [121]. This algorithm has strong exploitation ability as the habitats never are eliminated from the population but are improved by migration. More explanation and the general definitions of BBO process can be found in Chapter 3: Section 3.4.3. In the proposed BBO based path planner, each habitat $h_i$ corresponds to the coordinates of the B-Spline control point's $\vartheta_i$ utilized in path generation, where $h_i$ defined as a parameter to be optimized ($P_s$:($h_1,h_2,…,h_{n-1}$)). Habitats are improved iteratively. Algorithm 4.5 explains the pseudo-code of the BBO on path planning process.

**Procedure of BBO on ORPP**
Initialize a set of solutions as initial habitat population
- Assign B-spline control points $\vartheta_i$ as habitat $h_i$
- Choose appropriate parameters for the population size $i_{max}$
- Set the number of control-points ($n$) that used to generate the B-Spline path
- Set the maximum number of iteration $t_{max}$
- Assign maximum immigration and emigration rate ($I_r, E_r$)
- Assign maximum mutation rate $m_{r,max}(S_p)$
- Set $Sp_{max}$ and SIV vector

**For** $t=1$ **to** $t_{max}$
  Compute immigration rates $\lambda$ and emigration rate $\mu$ for each solution
$$\lambda_{Sp}(t) = I_r * \left(1 - \frac{S_p}{S_{pmax}}\right); \quad \mu_{Sp}(t) = E_r * \left(\frac{S_p}{S_{pmax}}\right)$$
  Evaluate the fitness (HSI) of each habitat and identify Elite Habitats based on HIS
  Modify habitats based on $\lambda$ and $\mu$ (Migration):
$$P_{Sp}(t) = P_{Sp}(t-1)(1 - \lambda_{Sp}(t) - \mu_{Sp}(t)) + P_{Sp-1}\lambda_{Sp-1}(t) + P_{Sp+1}\mu_{Sp+1}(t)$$
  **For** $i=1$ **to** $i_{max}$
    Use $\lambda_i$ to probabilistically decide whether immigrate to habitat $h_i$
    **if** $rand(0,1) < \lambda_i$
    **For** $j=1$ **to** $i_{max}$
      Select the emigrating habitat $h_j$ with probability $\propto \mu_j$
      **if** $rand(0,1) < \mu_j$
        Replace a randomly selected SIV variable of $h_i$ with its corresponding value in $h_j$
      end (*if*)
    end (*For*)
    end (*if*)
  end (*For*)
  Carry out the mutation based on probability by $m_r(S_p) = m_{r,max}\left[\dfrac{1-P_{Sp}}{P_{Sp,max}}\right]$
  Transfer the best solution in the population from one generation to the next
end (*For*)
Output the best habitat and its correlated path as the optimal solution

**Algorithm 4.5.** Pseudo-code of BBO based path planning.

## 4.5 Simulation Results of the Local ORPP Approach

There is a significant distinction between theoretical understanding of evolutionary algorithms and their properness for being applied to different problems due to the size, complexity, and nature of different problems. This may lead to different results using different algorithms for the same problem. Therefore, careful selection of evolutionary methods is of significant importance. An efficient path planner is able to make use of desirable currents and cope with undesirable current flows to increase battery lifetime. Moreover, the uncertainty of the operating field, which more or less is taken into account in the literature, has a devastating effect on a vehicle's safe deployment.

According to the defined path cost function $C_\wp$, the optimum solution corresponds to quickest and safest path in an uncertain terrain underwater environment. Performance of the path planner in coping with current variations and collision avoidance, therefore, need to be addressed thoroughly in accordance with complexity of the modelled terrain. The path planning in current study mainly focuses on finding the optimum trajectory adjustments using the advantages of desirable current field whilst not colliding with obstacles.

In this section, the simulation results obtained for the online path-planning problem through different scenarios are shown and analysed. The objective is to show the performance of the utilized evolutionary methods in the online AUV path-planning problem resulting in an adaptive manoeuvrability of the vehicle in a fully cluttered dynamic operating field. In each scenario, important aspects of uncertain and cluttered operating fields are considered and the outcome of the planner module is demonstrated. In the last scenario, all possible challenges of a realistic operating field are implemented and full details of the planner outputs including the generated path, corresponding states' history and final path time are presented.

**Simulation Setup and Research Assumptions**

The REMUS vehicle selected for this study has the following specifications:

- It is a modular AUV with dimensions of 0.2*m* in diameter, 1.6*m* in length, and less than 45*kg* in weight;

- It can operate with maximum forward velocity of approximately 5 *knots* for maximum mission distances of approximately 55 *km* and can operate up to the depth of 100 *m*;

- It is equipped with a long range Acoustic Doppler Current Profiler (ADCP) operating in 75 *kHz* and is able to measure currents profiles up to 1km ahead of the vehicle;

- It is equipped with upward looking downward looking, and sidescan sonar sensors operating in range of 75-540 *kHz* to percept and measure obstacles' features such as obstacles' coordinate and velocity;

- It is additionally equipped with underwater modem communication, GPS navigation, a Seabird CTD, a Wetlabs ECO sensor, and Wi-Fi.

The update of operating field includes changes in the position of the obstacles or current behaviour, continuously measured from the on-board sonar and horizontal acoustic Doppler current profiler (HADCP) sensors. The current field used in study is computed from a random distribution of 50Lamb vortices in a 350×350 grid within a 2D spatial domain according to the pixel size of the clustered map. In order to change the current parameters, Gaussian noise, in a range of 0.1~0.8, is randomly applied to update current parameters of $S_i^o$, $\ell$, and $\Im$ given in (4-5) to (4-8).

Configuration of the utilized method is given below. Four different scenarios are introduced in this chapter to evaluate the performance of the applied evolutionary methods on online path planning process.

- **PSO:** The PSO is initialized by 100 particles (candidate paths). The acceleration coefficients expansion-contraction coefficients also set on *2.0* to *2.5*. The inertia weight varies from 1.4 to 0.5.

- **BBO:** The habitats population ($i_{max}$) is set on 100. The emigration rate is generated by a vector of μ including $i_{max}$ elements in (0,1), and the immigration rate is defined as λ=1- μ. The maximum mutation rate is set on 0.1.

- **FA:** Fireflies population set on 100, the attraction coefficient base value $β_0$ is set on 2, light absorption coefficient γ is assigned with 1. The damping factor of δ is assigned with 0.95 to 0.97. The scaling variations is defined based on initial randomness scaling factor of $\alpha_0$. The parameter of randomization is set on 0.4.

- **DE:** Population size is set on 100, lower and upper bound of scaling factor is set on 0.2 and 0.8, respectively. The crossover probability is fixed on 20 percent.

Maximum number of iterations is set on $t_{max}$=100 for all algorithms. Finally, number of the path control points ($\vartheta$) for each B-Spline path is set on 7. For the purposes of this study, the optimization problem was performed on a desktop PC with an Intel i7 3.40 GHz quad-core processor in MATLAB® R2016a.

**Scenario 1: Path Planning in Spatiotemporal Ocean Flows**

To investigate the performance of the proposed online path planning strategy, the ocean environment is modelled as a three-dimensional volume $\Gamma_{3D}$ covered by uncertain time varying ocean currents. The update of operating field includes changes in the current behaviour, continuously measured from the on-board sonar and horizontal acoustic Doppler current profiler (HADCP) sensors.

Figure 4.7 represents the performance of the path planners in adapting ocean currents that varies within 4 time steps (various updating rate) once the AUV starts to travel from a waypoint on a loitering pattern (shown with a red circle) toward the target position (indicated by a yellow square). In this scenario, the variability of the current field is illustrated through four subplots in Figure 4.7 labelled by Time Step 1-4. The time step (time interval) for current update is determined by division of the number of iterations to current update rate ($U_R^C$=4) that is initialized by the operator. In order to change the current parameters, Gaussian noise, in a range of 0.1~0.8, is randomly applied to update current parameters of $S_i^o$, $\ell$, and $\Im$ given in (4-5) to (4-8).

From Figure 4.7, it is apparent that the generated path between the starting point (red circle) and the final position (yellow square) adapts the behaviour of current in all time steps. By changing the current map, the new information of spatiotemporal current flow, collected by HADCP, is fed to the online path planner module. As can be seen in each mentioned subplot, the proposed path planner autonomously adapts current variations and generates a refined path that can avoid severely adverse current flows and can take advantage of favourable ones to speed up the vehicle's motion. Being more specific in comparing the generated paths, obviously the FA method shows better performance in making use of favourable current flows.

In the second rank, the path generated by BBO shows significant flexibility in coping with current change specifically when the current magnitude gets sharper (more clear in turbulent given in Time Step: 3). Figure 4.8 demonstrates the variation of fitness value and the rate of solution convergence for the all evolutionary methods through this scenario. As can be seen, the convergence rate of the PSO is quite moderate and the rest of three methods are roughly identical in finding the optimal solution.

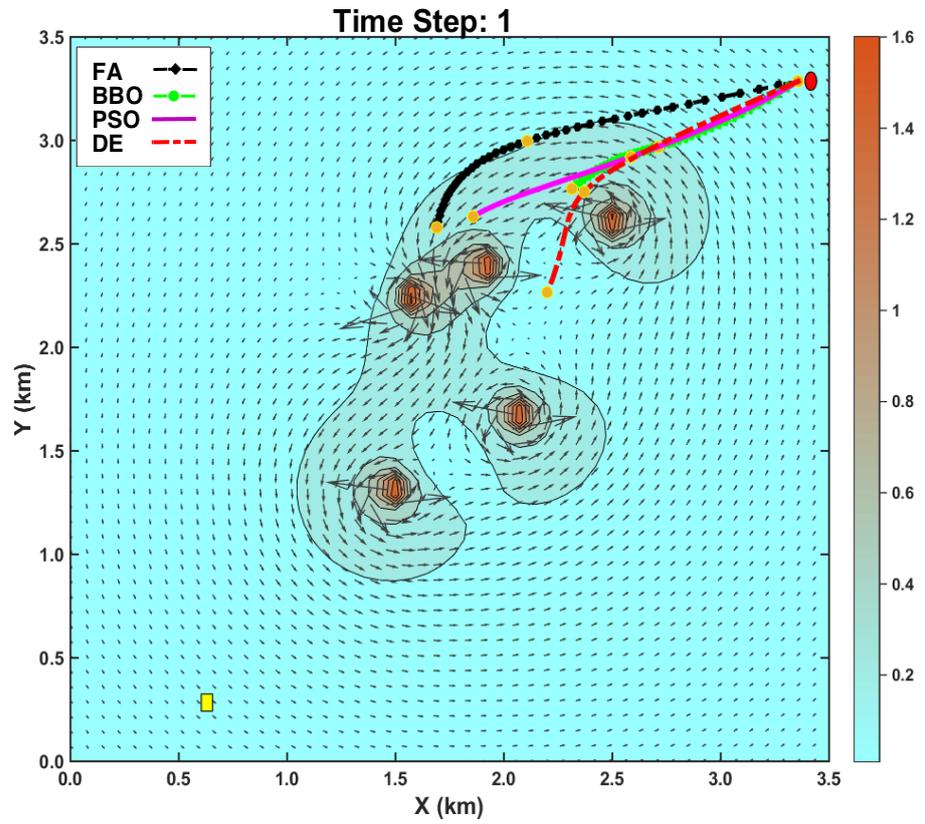

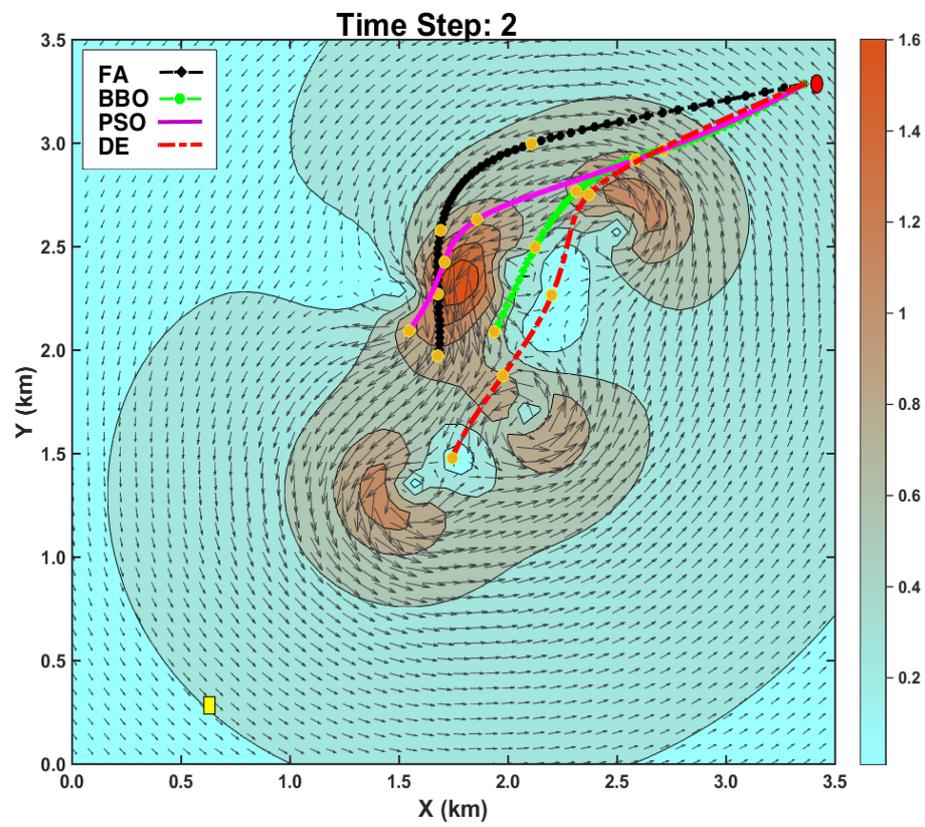

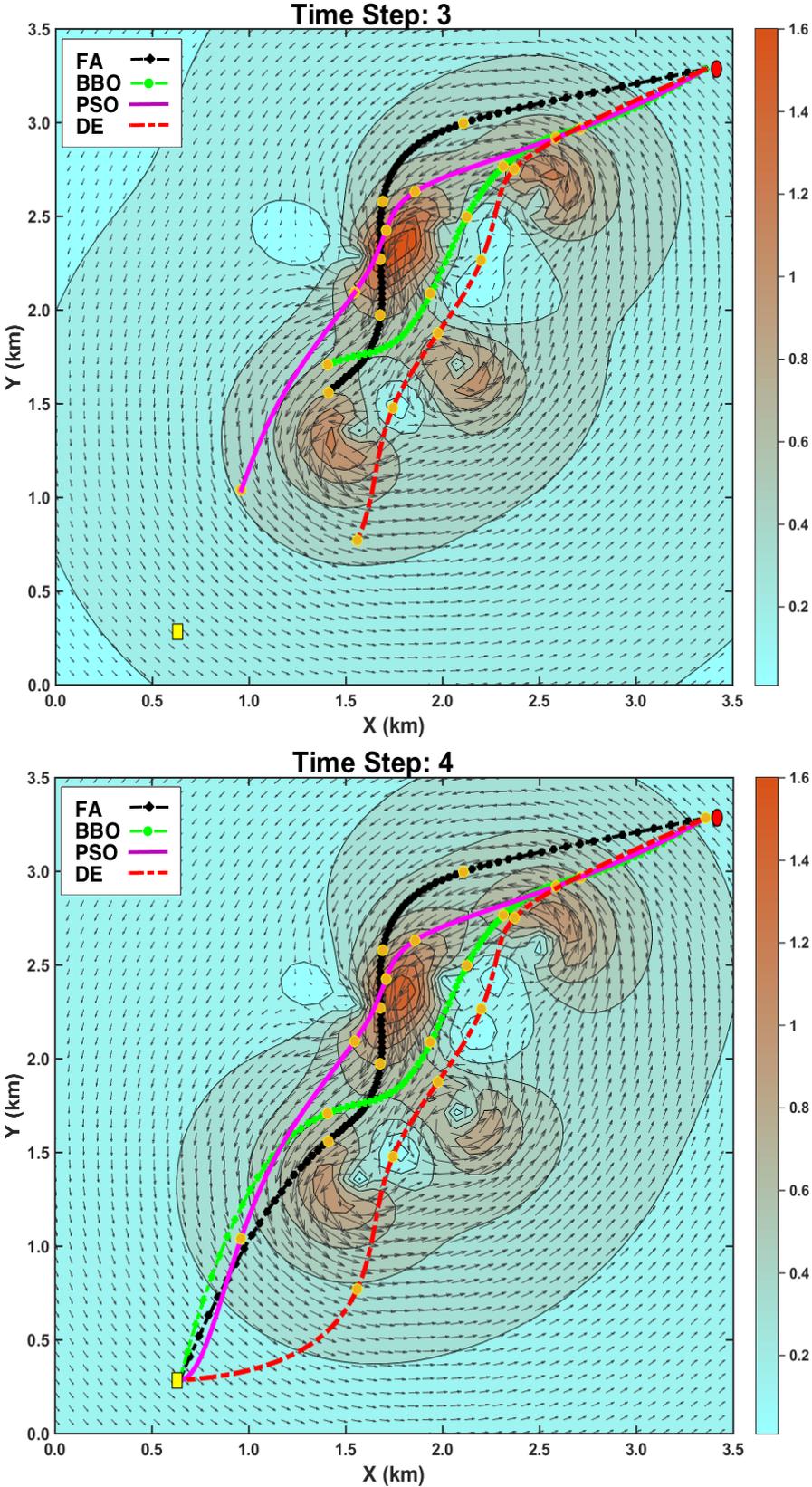

**Figure 4.7.** Adaptive path behaviour to current updates in 4 time steps.

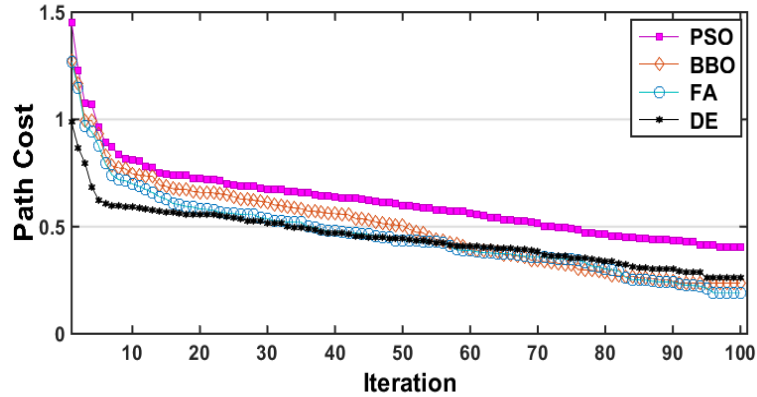

**Figure 4.8.** Variation of path cost and violation in the first Scenario.

It is noteworthy to mention that setting lower rate for Gaussian noise parameter (i.e., in range of 0.1 to 0.8) applied to the current vector field parameters, leads towards better fitness values for the generated paths by all four algorithms. Moreover, by setting higher current update rate (where Gaussian noise parameters get greater value than 0.8), the path fitness value reduces and practically other conditions remaining constant in path execution process. Thus, in a dynamic path planning, the probability of path failure increases when the current field updating rate get significantly greater value than the path updating rate.

**Scenario 2: Path Planning in a Cluttered Field with Variable Current Flow and Static Obstacles**

In this scenario, the operating field is cluttered with the static obstacles, randomly distributed in the problem space. The variability of the current vector field is the same as the Scenario 1. In Figure 4.9(a), the planner module generates the candidate path considering the underlying current map and obstacles' situations toward the final target position. The current map is updated in the location where the position of the vehicle is shown by the yellow triangles; these changes are obvious in Figure 4.9(b). The thin curves in Figure 4.9(b) indicate the offline initial paths generated by the proposed methods (before updating the current map), and the refined paths (online paths), are clearly shown by the thick curves in Figure 4.9(c). The utilized FA, BBO, PSO and DE algorithms are capable of generating a collision free path against the distribution of obstacles. Here again, the re-planning procedure is helpful to refine the offline path that may have not been tailored for the vehicle operation in the variable ocean flows.

Figure 4.10 indicates the performance of the proposed methods in finding the optimal solution considering the augmented objective function in (4-17).

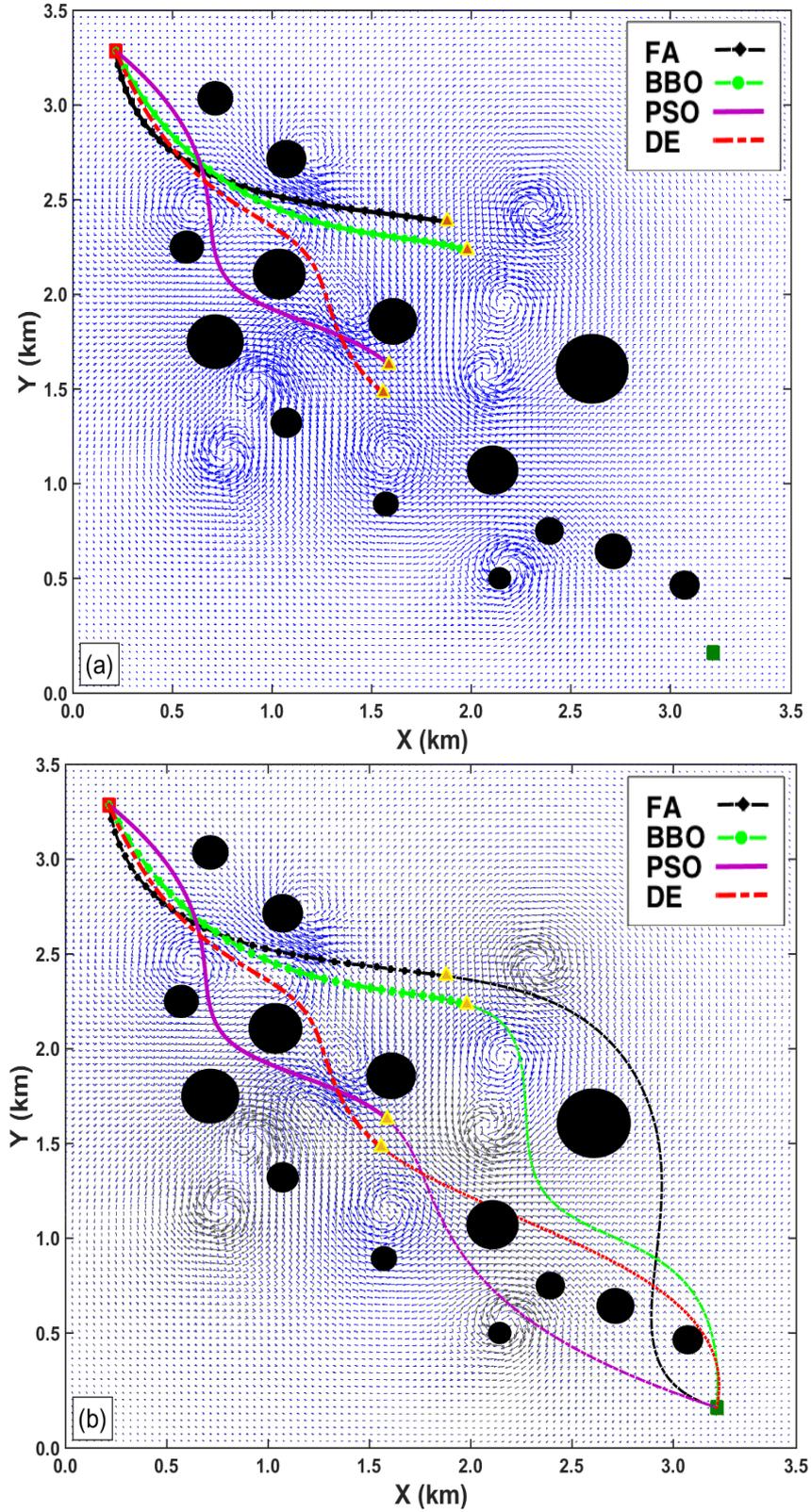

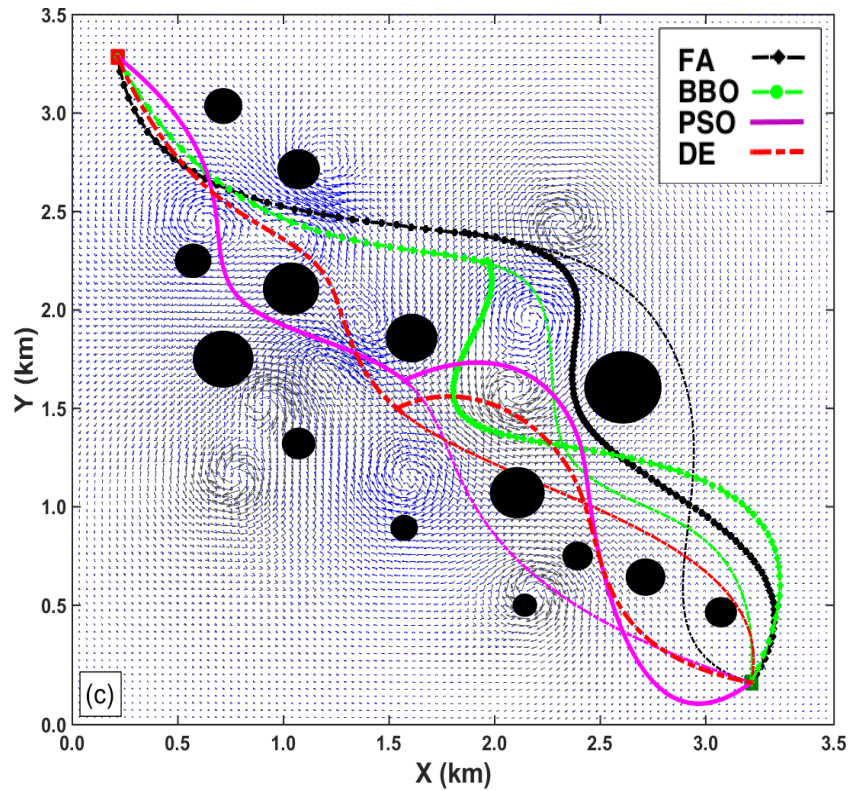

**Figure 4.9.** Offline path behaviour and re-planning process in accordance with current and static obstacles.

Referring to Figure 4.9, it is apparent that the initial produced path by all algorithms adapts the behaviour of current and accurately handles obstacle avoidance. Being more specific, the FA shows better performance in adapting adverse or concordant current arrows and take shortest path comparing to others. As presented in Figure 4.9(b-c), the proposed online path planner is capable of re-generating the alternative trajectory according to latest update of current map, in which the regenerated path takes a detour taking the advantages of previously generated optimum path data and the desirable current flow to speed up and save more energy in its motion toward the destination. Further, it is noted from Figure 4.9(b-c) the path generated by FA shows its superior flexibility in coping with current change specifically when the current magnitude gets sharper. The path re-planning procedure reuses the history of previous paths (online re-planning) to achieve more optimized path. As can be seen in Figure 4.10, the initial solutions generated by the evolutionary methods are different, as a result of different natures and mechanisms. However, they all are capable of searching the optimization space and converge to the best possible minima. Referring to collision violation plot,

Figure 4.10, it can be inferred that applying the re-planning strategy around 20$^{th}$ iteration, the violation value decreases to a great extent and at the end there is no violation over the obtained solution.

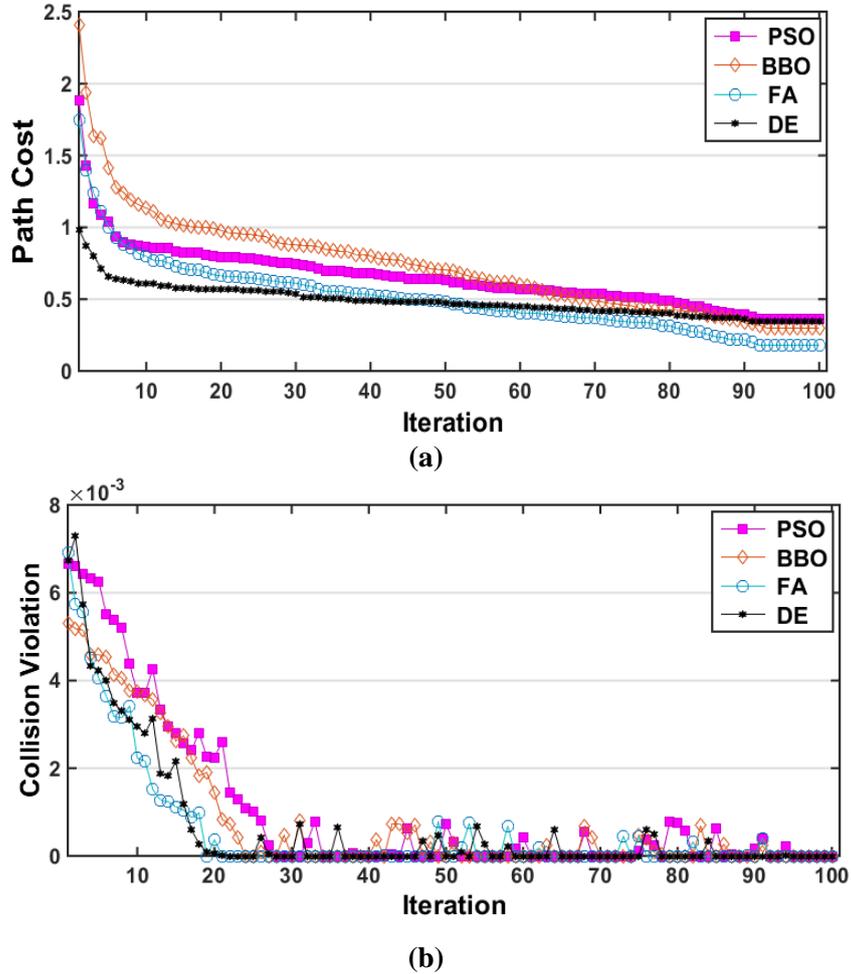

**Figure 4.10. (a)** Iterative cost variations for FA, DE, BBO, and PSO algorithms in the 2$^{nd}$ Scenario; **(b)** Iterative variations of collision violation for FA, DE, BBO, and PSO algorithms in the 2$^{nd}$ Scenario.

**Scenario 3: Path planning in a Cluttered Field with Variable Current Flow, Uncertain Static/Moving Obstacles**

This scenario becomes more challenging as we add the uncertain static and moving obstacles to the variable spatiotemporal current field. In Figure 3.11, the collision boundaries represented as green spheres around the obstacles computed with radius $\Theta_r \sim 2\sigma_o$ indicating a confidence of 98% that the obstacle is located within this area, referring to the equations mentioned in Section 4.2.2. The obstacles employed in this scenario are generated with random position by the mean and variance of their uncertainty distribution and velocity proportional to the current velocity. The uncertainty over the both static and moving obstacles are linearly propagated relative to the updating time and leads to the radius growth of the static obstacles and simultaneous position and radius changes in the moving obstacles. These variations are measured using on-board sonar sensors.

Figure 4.11 illustrates 3D-path generated by the evolutionary methods for the scenario at hand. In this figure, for having a clear visualization, just the final refined path is indicated. As can be seen in this figure, due to the linear propagation of the uncertainty with time, there exist a collision boundary encircling the objects. The safe trajectory is achieved if the vehicle manoeuvre does not have any intersection with the proposed obstacle boundary. The utilized evolutionary path planners are able to generate a collision free path that can satisfy the conditions of the proposed problem. This fact is indicated in Figure 4.12, showing the expedient convergence rate of the methods in finding optimal solution and sharp convergence rate of collision violation, as iteration goes on.

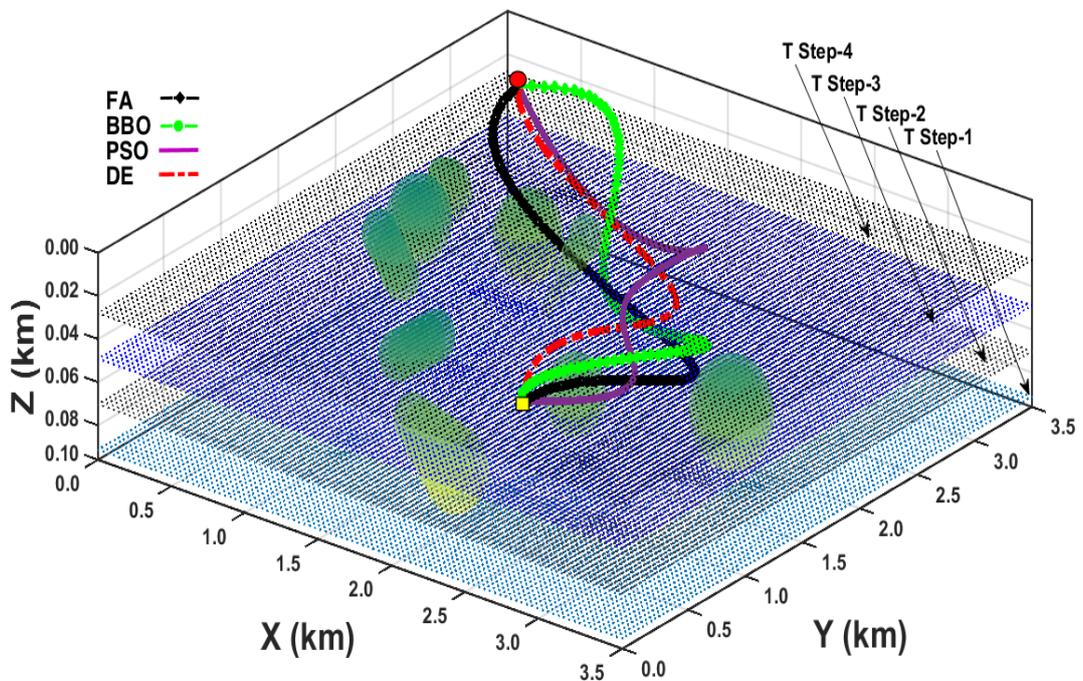

**Figure 4.11.** Evolution curves of FA, DE, BBO, and PSO algorithms with four time step change in water current flow in three-dimension space.

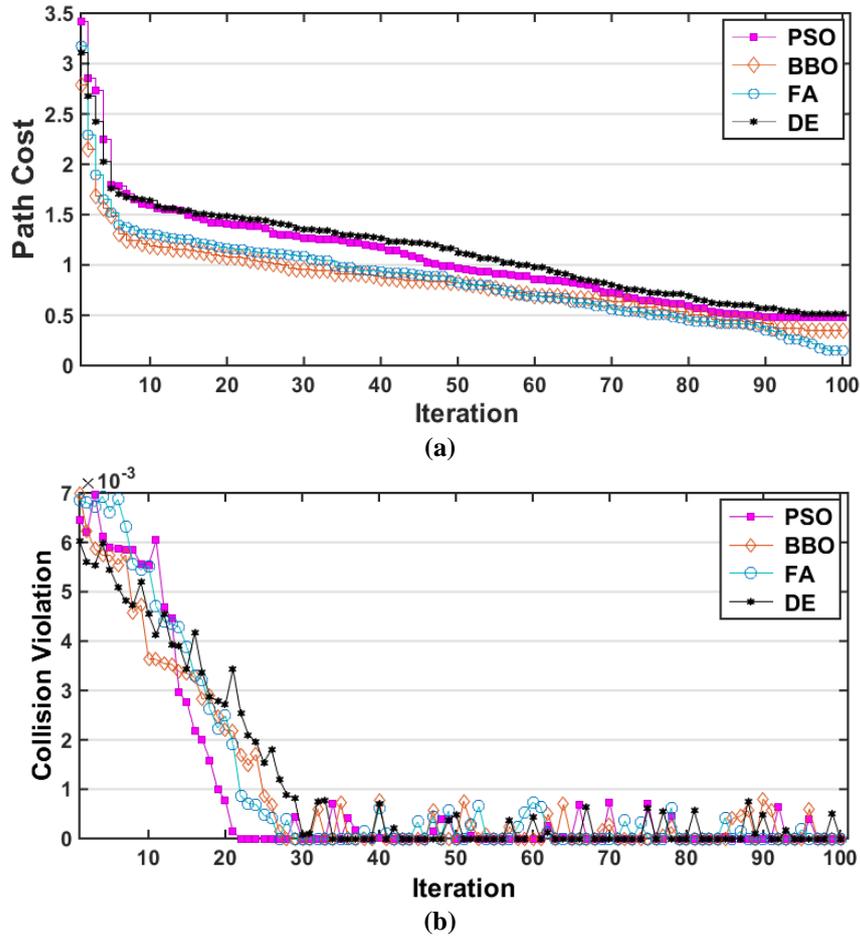

**Figure 4.12.(a)** Iterative cost variations for FA, DE, BBO, and PSO algorithms in the 3$^{rd}$ Scenario; **(b)** Iterative variations of collision violation for FA, DE, BBO, and PSO algorithms in the 3$^{rd}$ Scenario.

**Scenario 4: Path planning in a highly uncertain realistic operating field**

In the last scenario, the goal is to thoroughly simulate a highly uncertain cluttered realistic operating field including all possible barriers to assess the performance of the online path planner. In this scenario, the clustered map is used as an underlying environment embodied with variable ocean flows and uncertain static and moving obstacles. The coastal areas on the map are treated as no-fly zones and the planner should be aware of them. These zones in a context of offline obstacles are fed to the planner module. Although the problems complexity increases by encountering different environmental uncertain factors, it is derived from Figure 4.13 all aforementioned algorithms are able to produce efficient path to avoid collision and to adapt current variations.

*CHAPTER 4. AUV ONLINE REAL-TIME PATH PLANNING*

The AUV starts to transmit from the red circle indicated on Figure 4.13 and its transition is indicated by yellow triangles. The expected time for passing this distance is assumed $T_\varepsilon$=1800 (sec). The AUV water-reference velocity is set on υ=2.5 (m/s). For the current flow, the vortexes radius ($\ell$) and strength parameters ($\Im$) are set on 2.8 m and 12 m/s, respectively [108]; the uncertain static and moving obstacles follows the same concept used in the previous scenario. No collision will occur if the trajectory does not cross inside the collision boundaries of obstacles or coastal areas on the map. The collision boundaries represented as black circles around the obstacles indicating a confidence of 98% that the obstacle is located within this area. The gradual increment of collision boundary of each obstacle is presented in Figure 4.13 in which the uncertainty propagation is assumed linear with time.

Figure 4.13 (Time Step: 1) shows the first path generated by the evolutionary methods. The yellow triangles indicate the place where the current vector field is updated and the planner uses the new situational awareness of the environment to refine the previous path. It is noteworthy to mention that the path generated by DE, indicated by red dash line, does not collide with the neighbour coast at the start of the mission. The uncertainty of the obstacles can be clearly seen in the subsequent Time Step: 2-4. It is obvious form Figure 4.13, by increasing the complexity of the problem, all the evolutionary path planners still generate safe collision free paths satisfying the conditions of the proposed path planner.

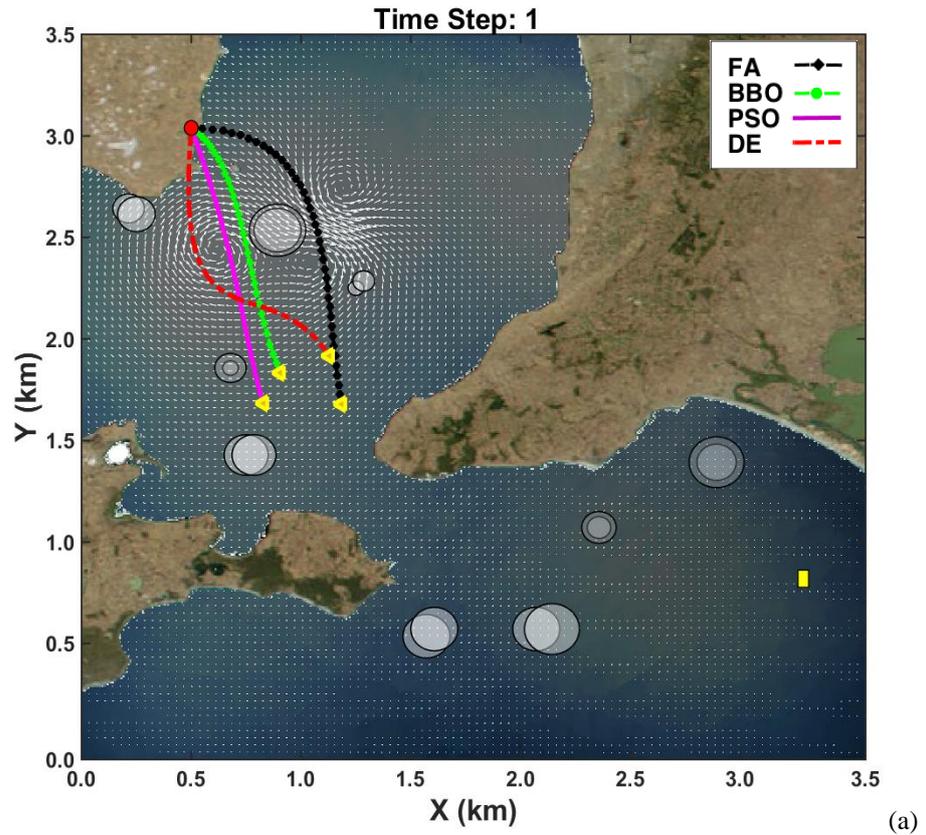

(a)

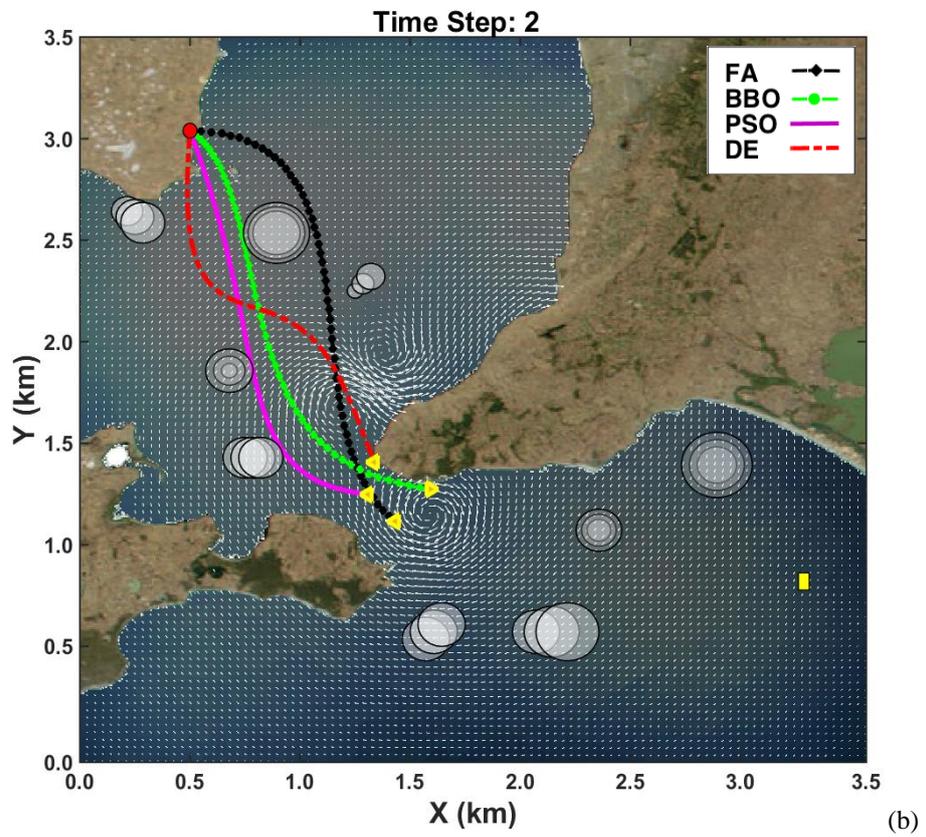

(b)

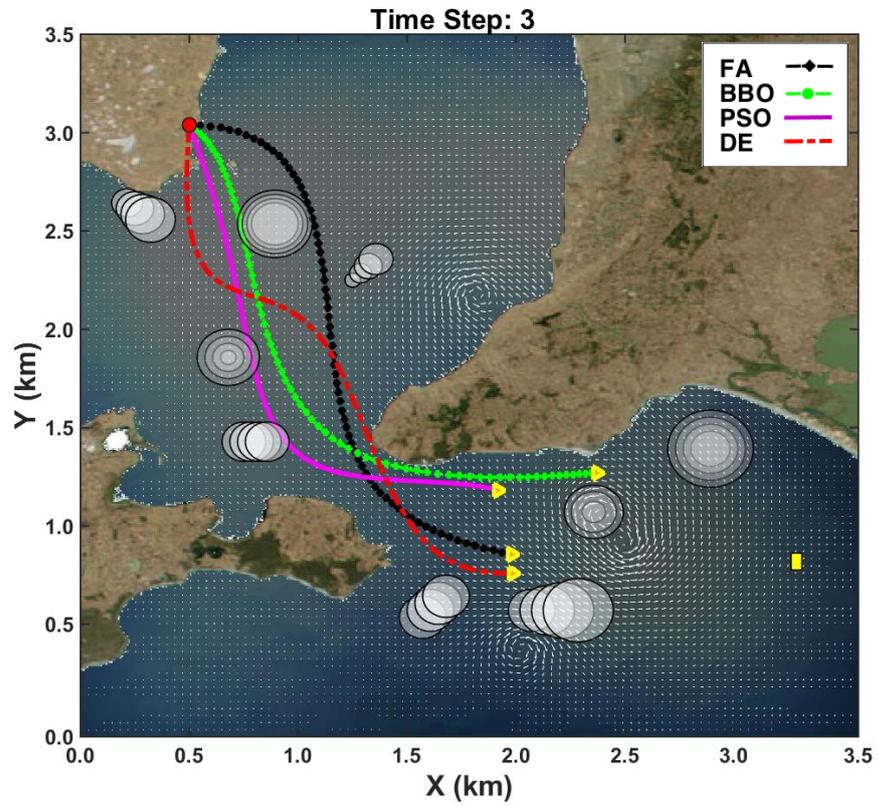

(c)

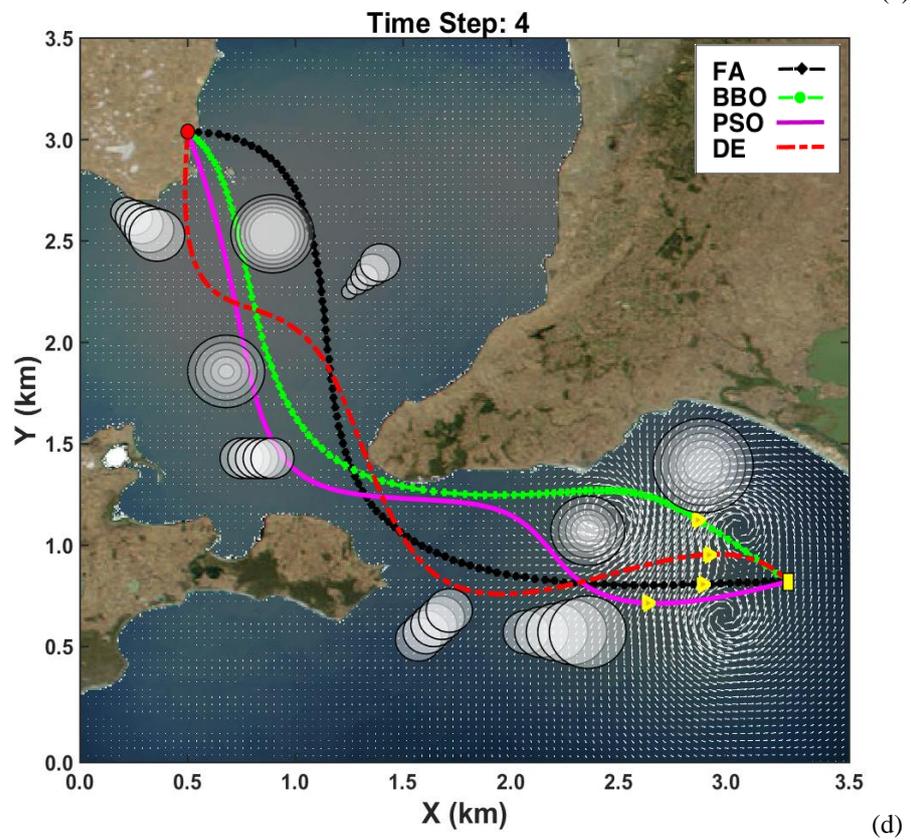

(d)

**Figure 4.13.** The path behaviour and re-planning process in a more complex scenario encountering different uncertain dynamic-static obstacles and current update in 4 time steps.

Figure 4.14 demonstrates the final path time achieved by the evolutionary path planners. As can be seen, the final time is different stemming from the inherent differences in structure and mechanism of the proposed methods; however, all of them properly satisfy the assigned upper time threshold of $T\varepsilon$. The associated variations of the performance index and collision violation per iteration are depicted in Figure 4.15. Again, the convergence rate of collision violation to zero is sharp due to the nature of online planning strategy. It is derived from simulation results in Figure 4.13, there is only slight difference comparing the results produced by all four algorithms and all of them accurately avoid crossing into the coastal sections of map even when the disturbing current pushes the vehicle to the undesired directions in the given operation window.

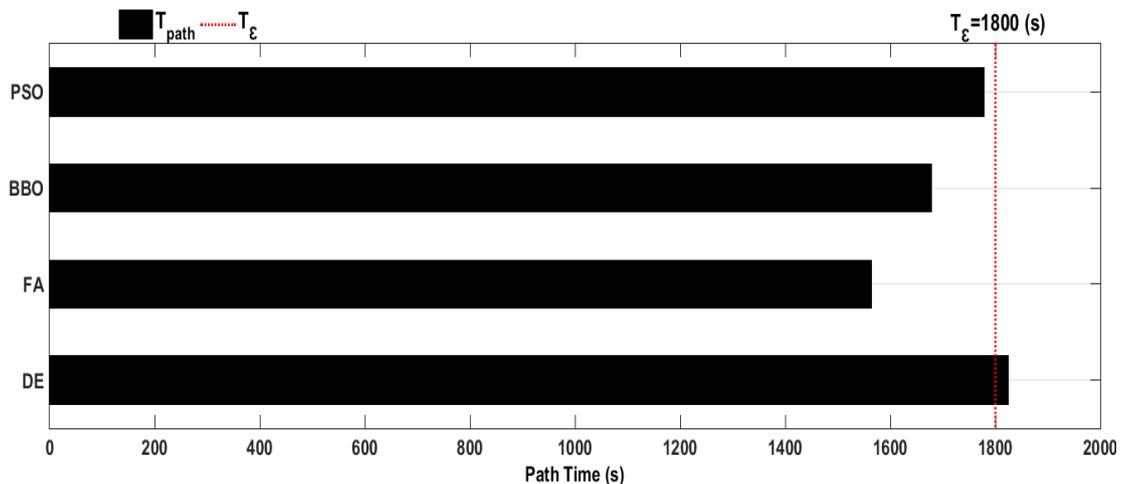

**Figure 4.14.** Performance of the algorithms for time optimality criterion.

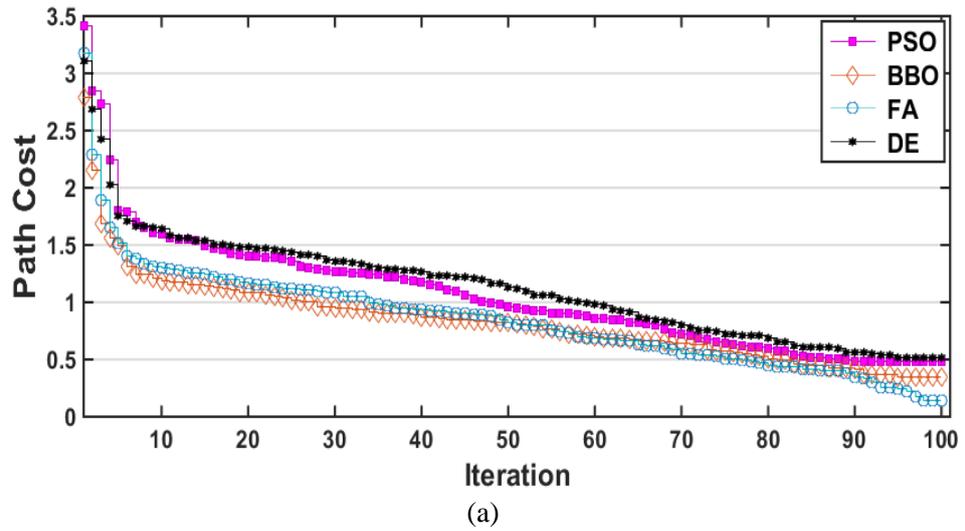

(a)

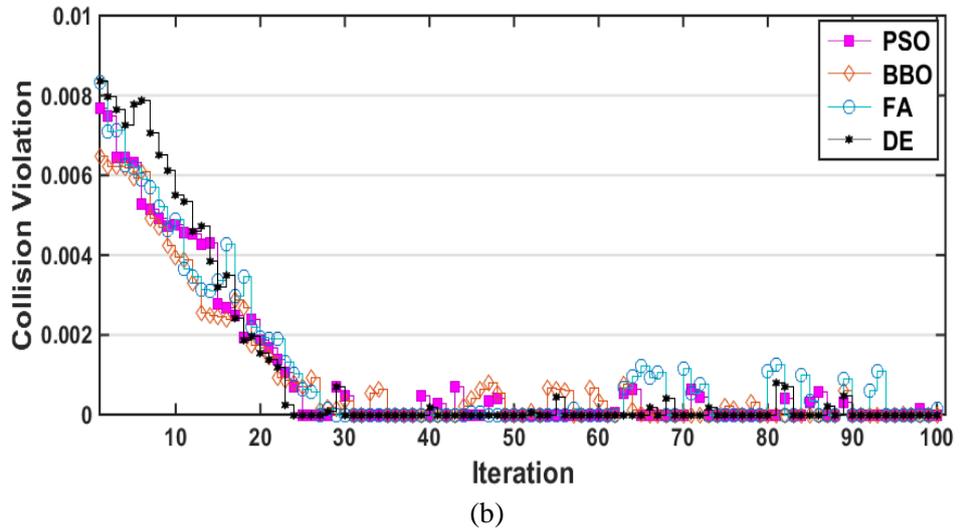

(b)

**Figure 4.15.** (a) Iterative cost variations for FA, DE, BBO, and PSO algorithms in the 4$^{th}$ Scenario; (b) Iterative variations of collision violation for FA, DE, BBO, and PSO algorithms in the 4$^{th}$ Scenario.

The cost variations of all four algorithms (given by Figure 4.15(a)) declares that the FA shows superior performance in reducing the cost within 100 iterations, while the DE and PSO do this process almost in a same accuracy. After all, BBO shows the lowest efficiency, despite its starting point even stands in a lower value $C_\wp \approx 1.2$ comparing others.

It is also noted from variation of the collision violation in Figure 4.15 (b), the violation of all four algorithms diminishes iteratively, but again the FA shows the best performance in managing solutions toward eliminating the violation.

On the other hand, the B-spline paths' curvature is obtainable by the vehicle's radial acceleration and angular velocity constraints. Hence, the behaviour of the produced path curves by FA, DE, BBO, and PSO algorithms in satisfying the vehicular constraints is presented through the Figure 4.16 and 4.17.

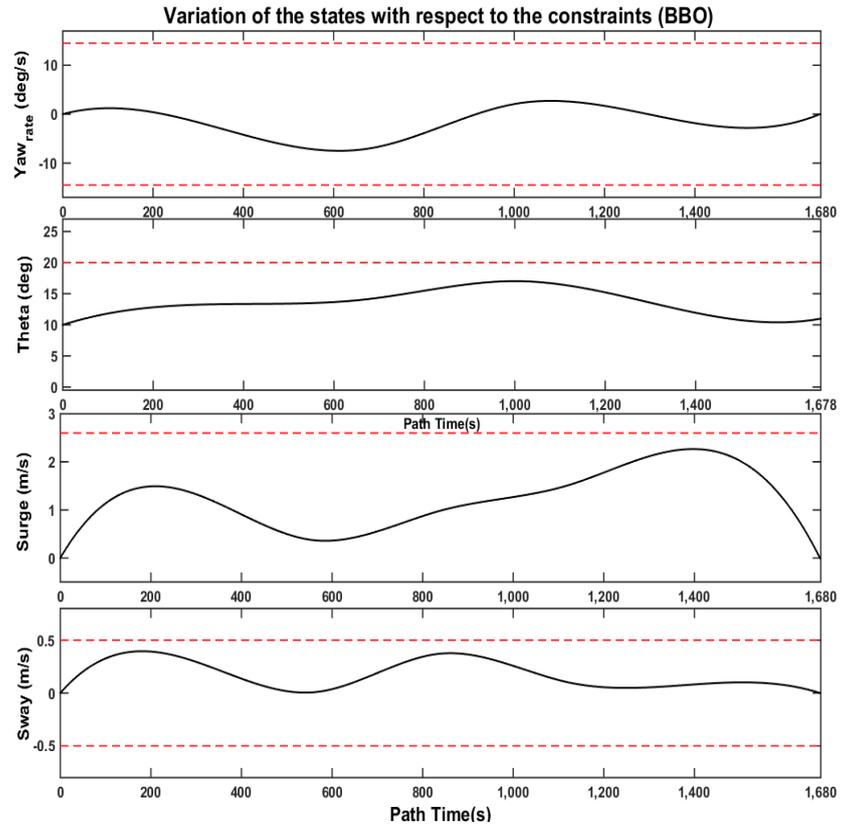

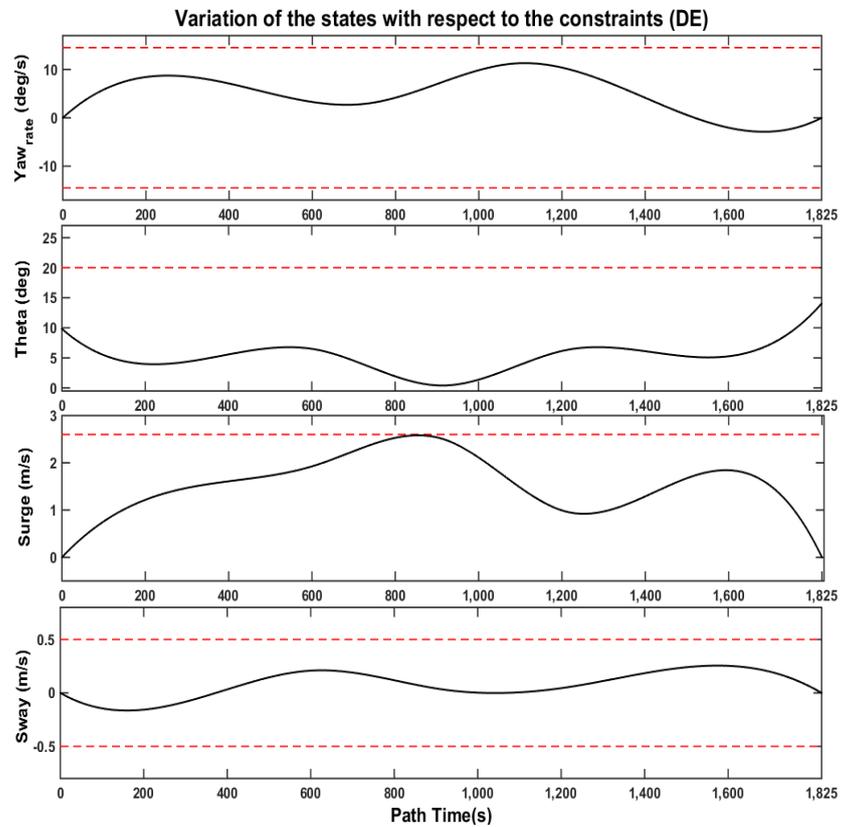

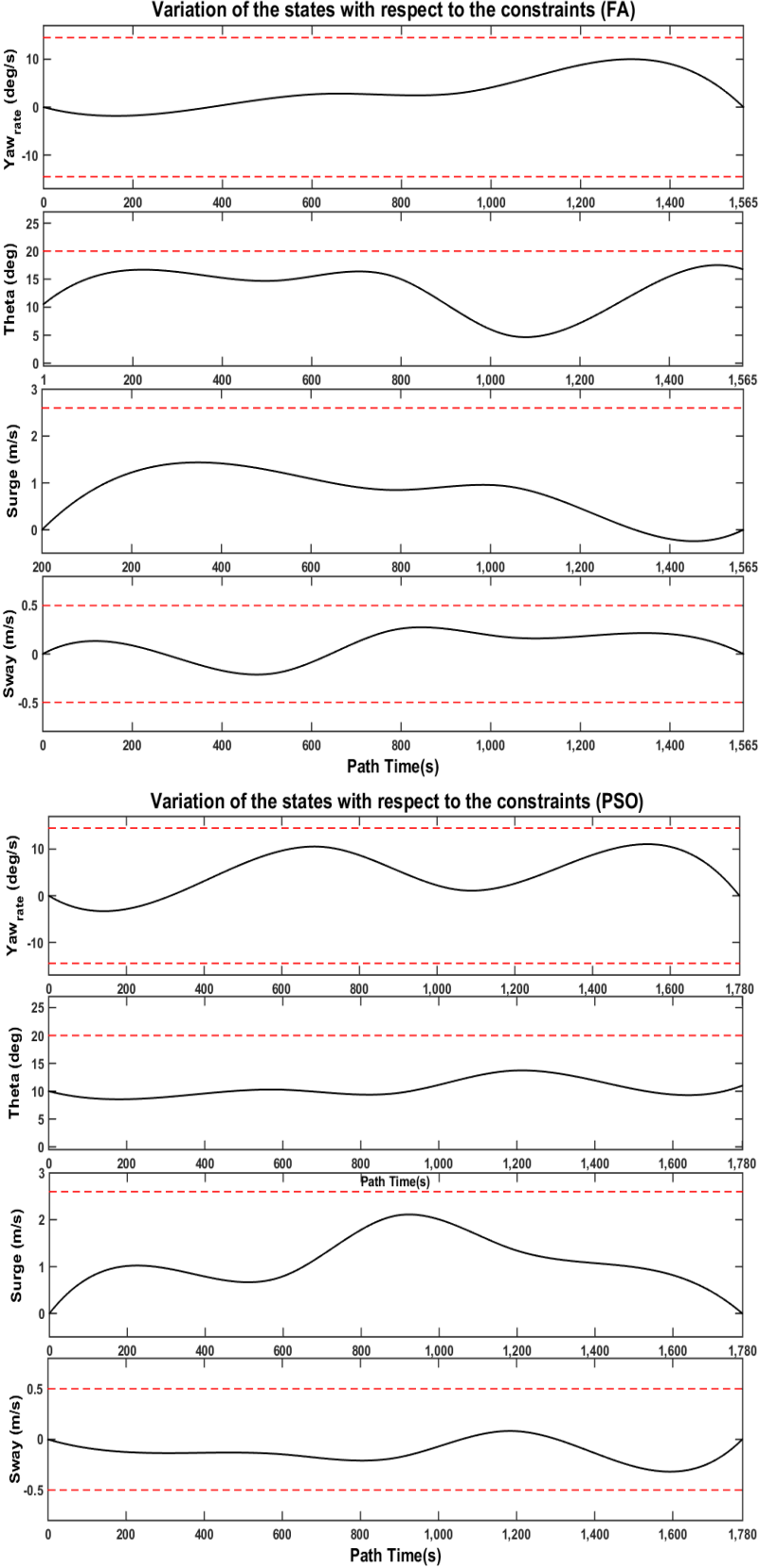

**Figure 4.16.** Variations of the states across the generated path with respect to the vehicular constraints (red dash line).

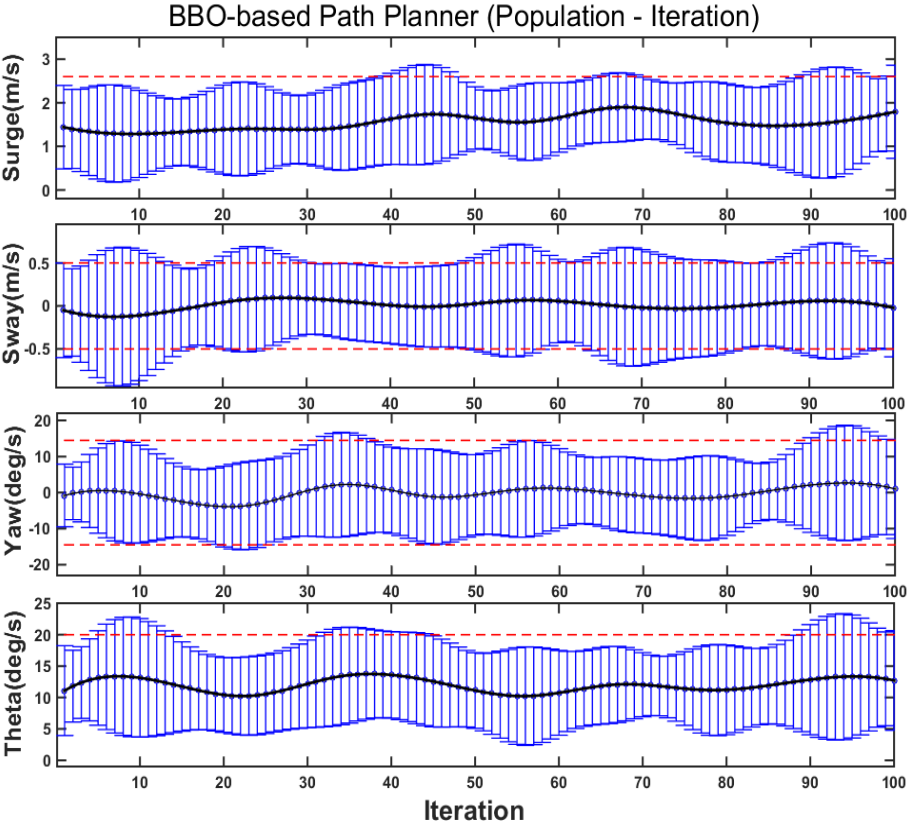

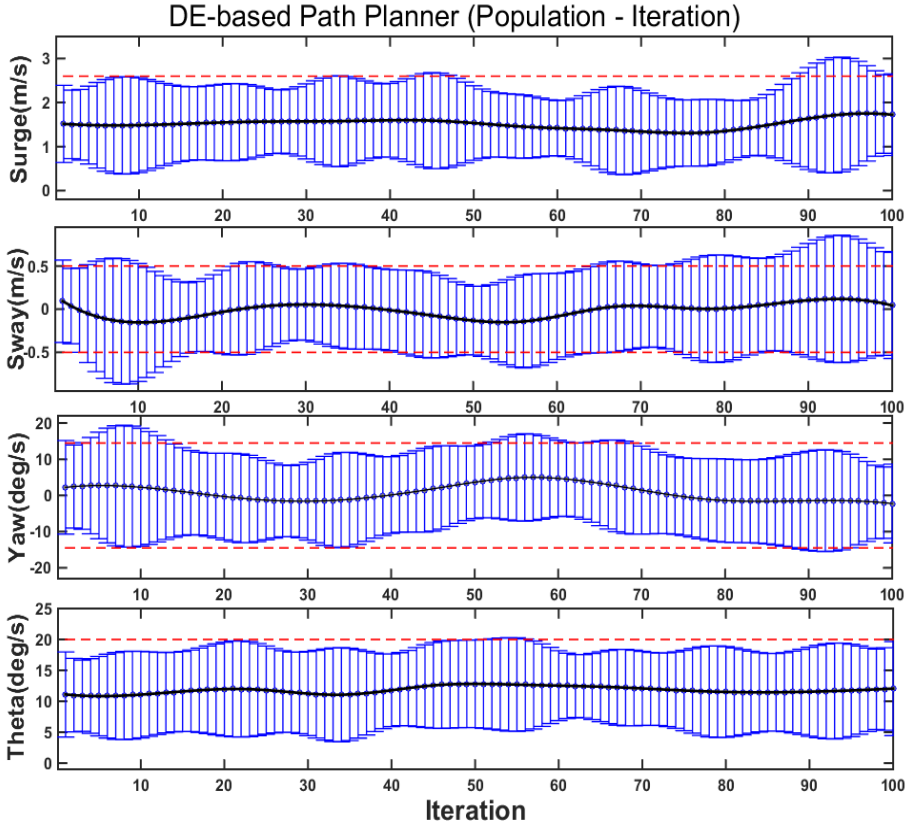

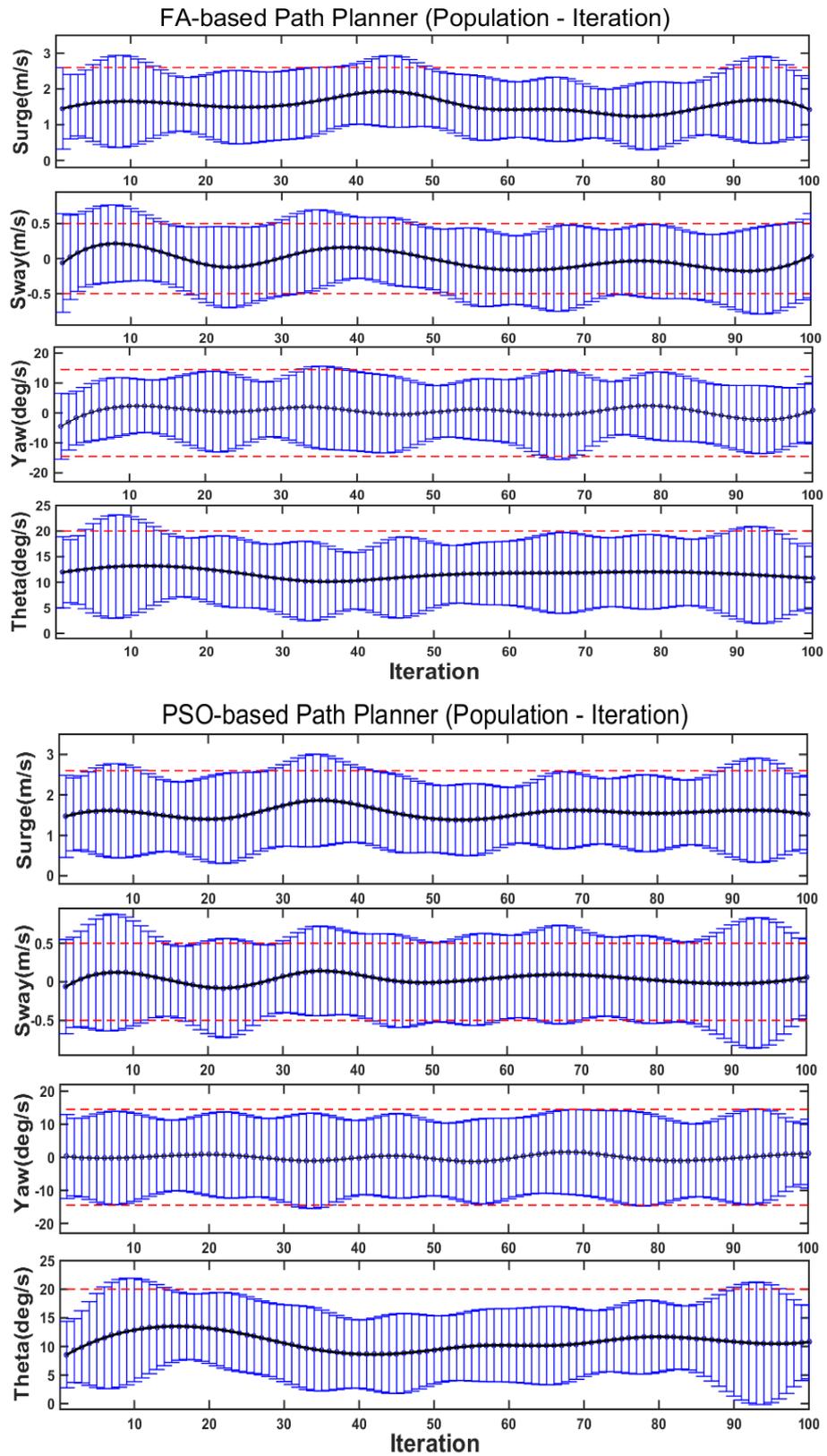

**Figure 4.17.** Iterative variations of the vehicular constraint of Surge, Sway, Theta, and Yaw rates for B-spline curves generated by FA, DE, BBO, and PSO.

Figure 4.16 illustrates the history of the states with respect to the vehicular constraints. The generated paths by the all methods respect the physical constraints over the states and generates smooth trajectory that is applicable for the vehicle low-level auto pilot module. Considering the iterative variations of surge, sway, yaw and theta rates corresponding to path splines produced by FA, BBO, DE, and PSO from the simulation given by Figure 4.17, it is evident that all generated paths accurately satisfy the defined constraints as the produced error-bars are bounded to specified violation boundaries presented with the red dashed line. Tracking the mean variation of the vehicular parameters in Figure 4.17, presented by the black line in the middle of the error-bar graphs, declares that algorithm enforces the solutions to respect the constraints over the 100 iterations.

## 4.6 Chapter Summary

Efficient motion planning is identified as a critical factor to enhance vehicle's autonomy and its resistance against various disturbances. This chapter aims to develop and evaluate an Online Real-time Path Planner (ORPP) as the lower level autonomy provider. The path planning is an NP-hard optimization problem in which the main goal is to minimize the path length, avoid crossing collision borders, and coping current variations over the time. On the one hand, existence of moving obstacles may lead to change the path conditions, as time goes on, and on the other hand, variability of ocean current components has considerable impact in drifting the vehicle from target of interest. In such a situation, adaptability of the path planner to environmental variabilities is a key element to carry out the mission safely and successfully. An efficient trajectory produced by the path planner enables an AUV to cope with adverse currents as well as exploit desirable currents to enhance the operation speed that results in considerable energy saving. To have an expedient adaptation with the variations of the underlying environment, an accurate online re-planning mechanism is developed and implemented. We mathematically formulate the realistic scenarios that AUV usually is dealing with in a large-scale undersea environment. This provides with use of a real map containing current and obstacles information; furthermore, the uncertainty of environment and navigational sensor suites are carefully taken into account by simulating imperfectness of sensory information in dealing with different types of moving objects. The path planner designed in the proceeding research simultaneously tracks the measurements of the terrain status and tends to generate the shortest collision-free trajectory by extracting eligible areas of map for vehicles deployment, carrying out obstacle avoidance and coping water current disturbance. The proposed planner is capable of refining the original path considering the update of current flows, uncertain static and moving obstacles. This refinement is not computationally expensive as there is no need to compute the path from scratch and the obtained solutions of the original path is utilized as the initial solutions for the employed methods. This leads to reduction of optimization space and accelerate the process of searching, therefore the optimal solution can be achieved in a fast manner. Having significant flexibility for approximating complex trajectories, B-Spline curves are utilized to parameterize the desired path. Applying the evolutionary methods namely PSO, BBO, DE, and FA indicate that they are capable of satisfying the path planning problem's conditions. The combination of meta-heuristic B-Spline technique provides an effective tool to solve the AUV's path planning problem. Considering the results of simulation, performance comparison between all applied methods is undertaken.

From the simulation results given by Section 4.5 it can be concluded that all FA, DE, BBO and PSO based path planner perform with almost similar cost and great fitness encountering collision constraints and all other constraints for the corresponding numbers of iteration. In addition, they remarkably satisfy the time optimality condition that is expressed by a time threshold in this study, and constraints over the corresponding states due to the physical limitations of the vehicle. From comparison point of view, the utilized evolutionary methods generate slightly different solutions due to the inherent differences in their nature and mechanism. More specifically, in this study, the FA planner shows better performance in terms of making use of favourable current flow for AUV manoeuvrability and collision avoidance. In terms of satisfying the time optimality condition, the performance of the PSO and DE path planner is better. In summary, the simulation results confirm that the proposed online planning approach using all four meta-heuristic algorithms are efficient and fast enough in generating optimal and collision-free path encountering dynamicity of the uncertain operating field and results in leveraging the autonomy of the vehicle for having a successful mission. The concept of this chapter is work out and published trough one conference and one journal papers (Appendices C and D).

In the next chapter, a consistent architecture will be designed to increase the autonomy in both high-level decision-making and low-level action generator by simultaneous mission scheduling (ordering the tasks), optimizing the vehicle's manoeuvrability to provide a safe and efficient deployment.

# Chapter 5
# Autonomous Reactive Mission Scheduling and Path Planning (ARMSP) Architecture

## 5.1 Introduction

Providing a higher level of decision autonomy is a true challenge in development of today AUVs and promotes a single vehicle to accomplish multiple tasks in a single mission as well as accompanying prompt changes of a turbulent and highly uncertain environment, which has not been completely attained yet. Today's AUVs operation still remains restricted to very particular tasks with restricted autonomy due to battery restrictions that necessitate them to manage endurance times. Improving the levels of autonomy facilitates the vehicle in performing long-range operations with minimum supervision in severe unknown environment. The proceeding approach builds on recent two chapters (Chapters 3 and 4) towards constructing a comprehensive structure for AUV mission planning, routing, task-time managing, and synchronic online motion planning adaptive to sudden changes of the time-variant environment. This chapter introduces an "Autonomous Reactive Mission Scheduling and Path Planning" (ARMSP) architecture and employs a set of evolutionary algorithms in layers of the proposed control architecture to investigate the efficiency of the structure towards handling the addressed objectives and prove stability of its performance in real-time mission task-time-threat management regardless of the applied meta-heuristic algorithm. Numerical simulations for analysis of different situations of the real-world environment is accomplished separately for each layer and also for the entire ARMSP model at the end. It is proven, by analysing the simulation results that the proposed model provides excellent mission timing and task managing while guaranteeing a secure deployment during the mission. The results of simulations support that the proposed architecture is reliable and robust, particularly in dealing with uncertainties. Figure 5.1 illustrates the operation diagram of the completed section in this chapter.

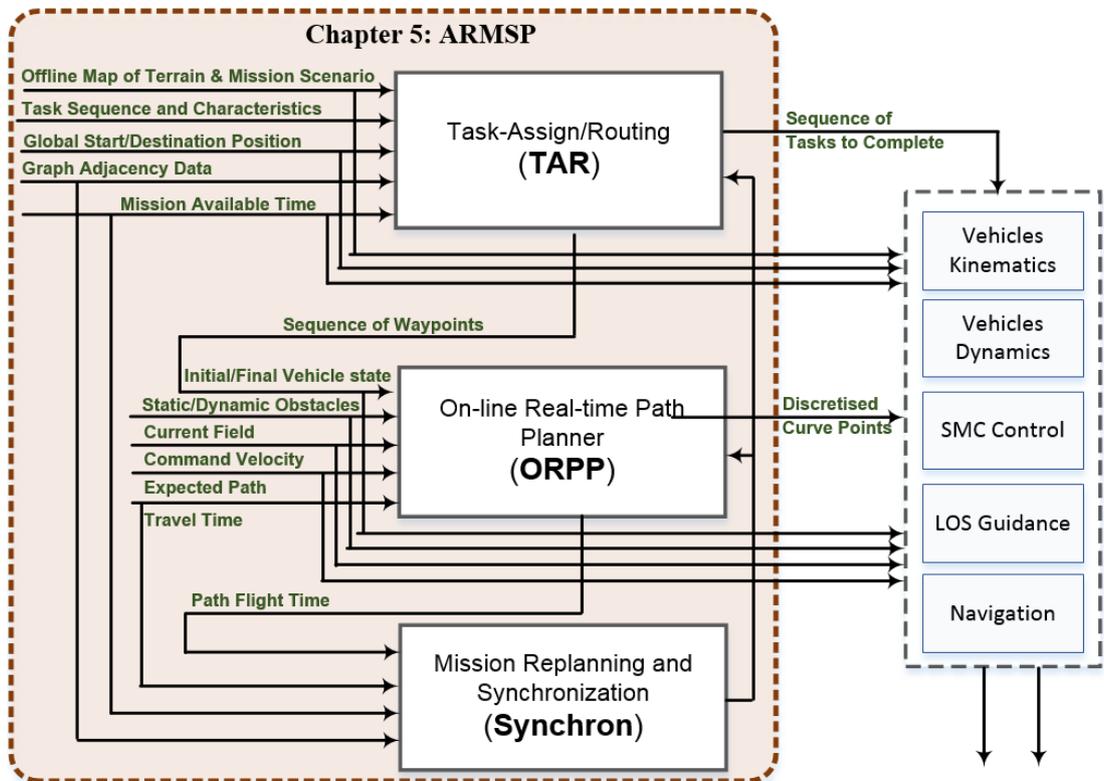

**Figure 5.1.** The operation diagram of the synchronous ARMSP that is completed by this chapter.

The rest of the chapter is organized in following sections: shortcomings associated with the mission-motion planning strategies are explained in Section 5.2. The main goal and motivation of this chapter in expressed by Section 5.3. The mechanism of the proposed modular architecture is provided by Section 5.4. The research assumptions and evaluation of the ARMSP in satisfying different aspects of an autonomous mission planning objectives and analysis of the simulation results is provided in Section 5.5. The conclusion and summary of the chapter is given by Section 5.6.

## 5.2 Shortcomings Associated with the Mission-Motion Planning

The challenges and difficulties associated with the motion-mission planning can be synthesized form two perspectives. First, the mechanism of the algorithms being utilized; and second the competency of the techniques for real-time applications. Additional to the addressed problems with selected strategies in both task assignment/mission planning and motion planning realm in the Chapters 3 and 4, many technical challenges are still remained unaddressed.

− Current variations over time can affect moving obstacles and drift them across a vehicle's trajectory; therefore, the planned trajectory may change and become invalid or inefficient. Proper estimation of the events in such a dynamic uncertain terrain in long-range operations, outside the vehicle's sensor coverage, is impractical and unreliable. Even though available ocean predictive approaches operate reasonably well in small scales, they produce insufficient accuracy to current prediction over long periods in larger scales. This becomes even more challenging in re-planning, when a large data load of variation of whole terrain condition should be computed repeatedly any time that path replanting is required, which is computationally inefficient and unnecessary as only awareness of environment changes in vicinity of the vehicle is enough.

− Assuming that the tasks for a specific mission are distributed in different areas of a waypoint cluttered graph-like terrain, there should be a compromise among prioritizing the tasks according to available battery/time in a way that vehicle is guided toward the destination waypoint. As mentioned earlier, the path planning strategies principally deal with the quality of a vehicle's motion between two points and are not provided for handling vehicle's task assignment in a waypoint-cluttered terrain; so that a routing mission planning strategy is required to handle graph search constraints and carrying out the task assignment. However, vehicles routing and mission planning strategies are not accurate enough in coping with the dynamic variations of the terrain, but these approaches divide the operation area to smaller beneficent sections and give an overview of the pathway for vehicle to manoeuvre that helps reducing the computational load. Accordingly, each approach is able to cover only a part of this problem.

## 5.3 Chapter Motivation

With respect to above discussion, existing approaches mainly are able to handle only a specific level of autonomy either task assignment together with time management or path planning with safety considerations (as discussed in Chapter 2). In this chapter, an autonomous reactive ARMSP control architecture encompassing a Task-Assign/Routing (TAR) approach and local Online Real-time Path Planner (ORPP) is developed to produce a reliable mission plan for a large-scale time-varying underwater environments, and to cover shortcomings associated with each of the above-mentioned strategies. The TAR module at the top level is responsible for task-assign process in a limited time interval and guiding the vehicle towards a target of interest while considering on-time termination of the mission. At the lower level, the ORPP module is in charge of generating optimal trajectories while operating in an uncertain undersea environment, where spatiotemporal variability of the operating field is taken into account. The TAR is able to reactively re-arrange the tasks based on mission/terrain updates while the low level ORPP is capable of coping unexpected changes of the terrain by correcting the old path and re-generating a new trajectory. As a result, the vehicle is able to undertake the maximum number of tasks with certain degree of manoeuvrability having situational awareness of the operating field. A constant interaction exists between high-low level modules across small and large scale. The ARMSP decides whether to carry out the path re-planning or mission re-planning procedure according to the raised situation. Mission re-planning is performed to manage the lost time in cases that the path planner process takes longer than expectation. The proposed re-planning procedure in both of the mission and path planners improve the robustness and reactive-ability of the AUV to the environmental changes and enhance its performance in accurate mission timing.

The proposed architecture offers a degree of flexibility for employing diverse sorts of algorithms subjected to real-time performance of them. Both of the planners operate individually and concurrently while sharing their information. The modular structure of this architecture provides a reusable and versatile framework that, at first, is easily upgradable and secondly is applicable to a broad group of autonomous vehicles. Parallel execution of the planners speeds up the computation process. The operating field for operation of the ORPP is split into smaller spaces between pairs of waypoints; thus, re-planning a new trajectory requires rendering and re-computing less information. This leads to problem space reduction and reducing the computational burden.

Fast and concurrent operation of the subsystems (TAR/ORPP) is the most critical factor in preserving the stability and consistency of the proposed ARMSP architecture that prevents each of TAR and ORPP components from dropping behind each other. Any delay in operation of TAR or ORPP disrupts the concurrency of the entire system. Also adding NP computational time into the equation would itself render a solution suboptimal. Relying on the previously mentioned ability of meta-heuristics in solving complex NP-hard problems, a combination of the employed algorithms by TAR and ORPP (examined in Chapters 3 and 4) is utilized to evaluate the performance of the proposed ARMSP architecture in this chapter.

## 5.4 Mechanism of the Proposed Modular ARMSP Architecture

The proposed architecture is designed in separate modules running concurrently. The TAR allows the vehicle to select the best sequence of tasks in a restricted time so that the vehicle is guided towards the final destination. ORPP deals with vehicle's guidance from one point to another taking the ocean uncertainty and variability into account. The modules interact simultaneously by back feeding the situational awareness of the surrounding operating field; accordingly, the system decides to re-plan the path or mission or continue the current mission. Hence, the "Synchron" module is designed to manage the lost time within the TAR/ORPP process and reactively adapt the system to the latest updates of the environment and the required decision parameters (e.g. remaining time). This process is frequently repeated until the AUV reaches the end point. The design of the proposed control architecture is illustrated by Figure 5.2.

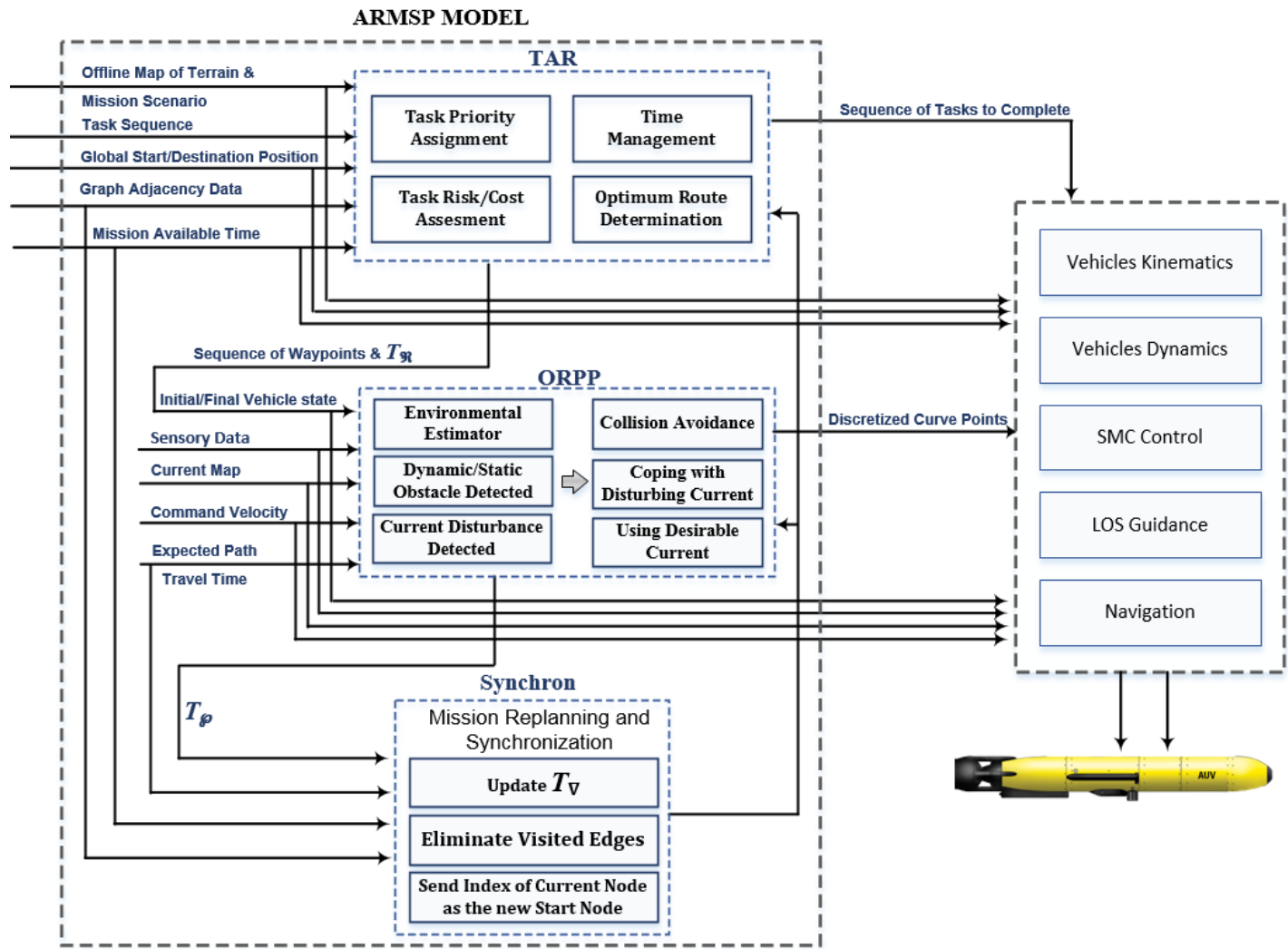

**Figure 5.2.** The operation diagram of the designed ARMSP model for autonomous operation of an AUV.

Operation of the TAR and ORPP module is explained in Chapters 3 and 4, respectively. Their real-time performance and accuracy also have been examined using a bunch of evolutionary algorithm. Another module will be added to the architecture to provide concurrency between operation of two TAR and ORPP modules and to recognize the requisition for either path re-planning or mission re-planning in different circumstances.

### 5.4.1 Modelling of the 'Synchron' Module

After the TAR generated a mission plan in a sequence of tasks/waypoints, the ORPP module gets this sequence along with environmental information and starts generating safe and efficient trajectory between the waypoints while the AUV concurrently executes the corresponding tasks. In this way, the AUV efficiently carries out its mission and safely is guided through the waypoints. The provided trajectory is shifted to the guidance controller to produce the guidance commands. During deployment between two waypoints, the ORPP can incorporate any dynamic changes of the environment. The ORPP repeatedly calculates the trajectory between vehicles current position and its specified target location (any time that new update of information received from the sensors). After each waypoint is visited the total path time of $T_\wp^{ij}$ is compared to expected time $T_\varepsilon^{ij}$ for traversing the corresponding distance, in which the $T_\varepsilon^{ij} \approx t_{ij}$ is extracted from the mission time $T_\Re$ given by (3-5). In a case that the $T_\wp^{ij}$ is found to be smaller than $T_\varepsilon^{ij}$, the current order of tasks would be valid and the vehicle can continue its mission. However, in the opposite case, when the $T_\wp^{ij}$ exceeds the $T_\varepsilon^{ij}$, that is, a certain amount of the available time $T_\nabla$ is spent for handling any probable challenge during the path planning process, the previously defined mission scenario cannot be appropriate any more. Under such conditions mission re-planning according to mission updates is necessary. This can cause a computational burden due to the mission re-planning process. Multiple traversing a specific edge (distance) is time dissipation, which means repeating a task for multiple times. To prevent this issue, the following steps should be carried out any time that mission re-planning is required:

- Update $T_\nabla$
- Eliminate passed edges from the operation network (so the search space shrinks);
- The location of the present waypoint is considered as the new starting position for both ORPP and TAR.
- TAR is to reorder the tasks and to prepare a new mission scenario based on new information and the updated network topology. A computation cost is to be encountered any time that mission re-planning is required.
- This process is to be continued until vehicle arrives to the required destination waypoint.

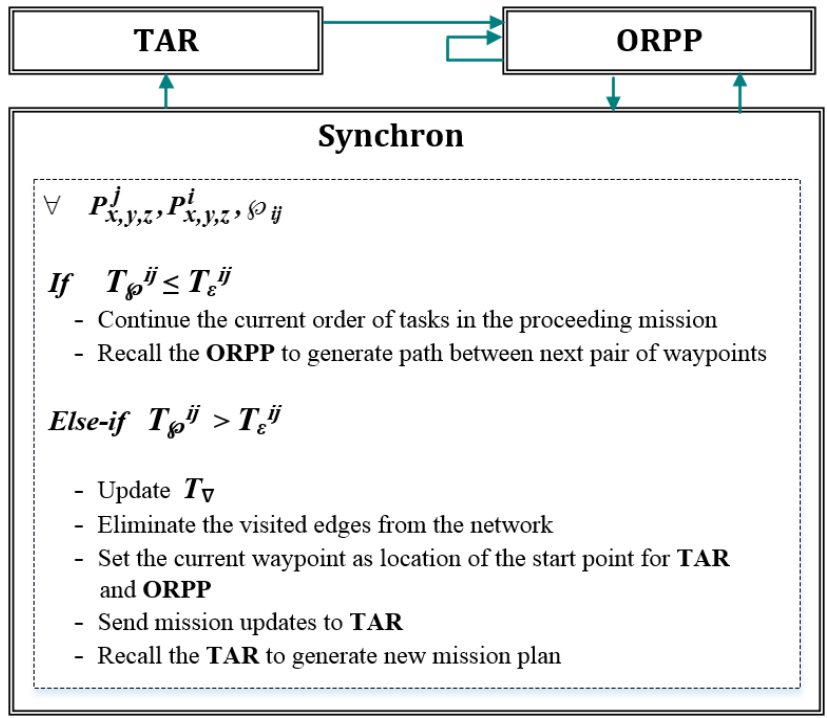

**Figure 5.3.** The synchronization process.

The trade-off between total available time and mission objectives is a critical issue that should be adaptively carried out by TAR. Hence, the main architecture should be fast enough to track environmental changes, cope with the dynamic changes, and carry out prompt path or mission re-planning based on the raised situation. The proposed synchronous architecture is evaluated according to criterion given in the next section.

### 5.4.2 Architecture Evaluation Criterion

The ORPP operates concurrently in background of the TAR scheme. Performance of the generated local path by ORPP is evaluated by a cost function $C_\wp$ based on travel time ($T_\wp$), which is proportional to energy consumption and the distance travelled ($L_\wp$). The mission cost $C_{TAR}$ has direct relation to the passing distance between each pair of selected waypoints. Hence, the path cost $C_\wp$ for any appropriate local path is used in the context of the TAR. The main goal of the ARMSP module is to maximize the weight of the selected edges, which means selecting the best sequence of highest priority tasks in a limited time while guaranteeing on-time termination of the mission and safe deployment during the mission. To this end, the mission cost $C_{TAR}$ is defined as a combination of multiple weighted functions that should maximized or minimized. The entire ARMSP model is searching for a beneficent solution in the sense of the best combination of task, path, and route cost as given by (5-1) and (5-2):

$$\forall \wp^j_{x,y,z} = [X(t), Y(t), Z(t), \psi(t), \theta(t), u(t), v(t), w(t)]$$

$$C_\wp \approx T_\wp = \sum_{i=P^a_{x,y,z}}^{|\wp_{x,y,z}|} \frac{\sqrt{(X_{i+1}-X_i)^2 + (Y_{i+1}-Y_i)^2 + (Z_{i+1}-Z_i)^2}}{|\upsilon|}$$

s.t. (5-1)

$$[X(t), Y(t), Z(t)] \cap \Sigma_{M,\Theta} = 0$$

$$u(t) \leq u_{max} \quad \& \quad v_{min} \leq v(t) \leq v_{max}$$

$$\theta(t) \leq \theta_{max} \quad \& \quad \dot{\psi}_{min} \leq \dot{\psi}(t) \leq \dot{\psi}_{max}$$

$$d_{ij} \propto \wp \quad \& \quad t_{ij} \propto T_\wp$$

$$T_\Re = \sum_{\substack{i=0 \\ j \neq i}}^{n} s_{eij} \times t_{ij}$$

$$C_{TAR} \propto |T_\Re - T_\nabla| = \left| \sum_{\substack{i=0 \\ j \neq i}}^{n} s_{eij} e_{ij} \times (C_\wp^{ij} + \wp_{CPU}) - T_\nabla \right| + \sum_{\substack{i=0 \\ j \neq i}}^{n} s_{eij} \times \frac{e_{ij}}{w_{ij}}; \quad s_{eij} \in \{0,1\}$$

s.t. (5-2)

$$\max(T_\Re) < T_\nabla$$

The path cost is calculated based on travelling time between two waypoints and it is subjected to vehicles Kino-dynamic restrictions and avoiding collisions (given by (5-1). On the other hand, the mission cost depends on number of completed tasks and total obtained weight restricted to total available time (battery capacity). As mentioned above, the path cost $C_\wp$ is a function of the mission cost $C_{TAR}$ ($C_{TAR}$ is affected by $C_\wp$ as presented by (5-2)). During the path planning process, traversing the corresponding distance may consume more time than what is expected. Restitution of the wasted time necessitates a mission re-planning process. The mission re-planning criteria (given by Figure 5.3) is investigated after the completion of the ORPP process (when a waypoint in the route sequence is visited). Hence, the computation time is also encountered in calculation of the total mission cost. Thus, the total cost $Cost_{Total}$ for the model is defined as:

$$Cost_{Total} = C_{TAR} \times f(C_\wp) + \sum \wp_{CPU} + \sum_{1}^{rep} T_{compute} \times (TAR_{CPU}) \qquad (5\text{-}3)$$

where $T_{compute}$ is the mission re-planning computation time, $rep$ is the number of time that the mission re-planning procedure is repeated. The AUV should use the available mission time as productively as possible. The TAR module is evaluated through the Monte Carlo simulation for BBO, PSO, ACO and GA algorithms in Chapter 3.

The ORPP module is also implemented using four evolutionary algorithms of BBO, PSO, DE, and FA, and evaluated and discussed in Chapter 4. These meta-heuristic algorithms have two inner loops through the population size of $i_{max}$ and number of iterations $t_{max}$ (refer to pseudo-codes); thus, in the worst case, their computational complexity is $O(i_{max}^2 \times t_{max})$, which is linear with time. Hence, they are computationally cost effective.

Various combination of these algorithms is tested for evaluating the entire ARMSP model, and their performance is statistically analysed through the series of Monte Carlo simulations (discussed in Section 5.5).

## 5.5 Discussion and Analysis of Simulation Results

The model aims to use the total available time for completing the best possible set of tasks in the most efficient way and guarantee on-time arrival to the destination while concurrently checking for safe deployment through an efficient path planning strategy.

To this end, an accurate and stringent concurrency between different components of the model is required so that each component can carry out its responsibility in a synchronous manner. Assumptions play an important role in performance of the model (given by subsection 5.1.1).

## 5.5.1 Simulation Setup and Research Assumptions

A model of the undersea environment is provided to evaluate the performance of the proposed framework. To model a realistic marine environment, a three-dimensional terrain in scale of $\{10 \times 10$ km ($x$-$y$), 1000 m($z$)$\}$ is considered based on a realistic example map presented by Figures 5.4 and 5.5, in which the operating field is covered by uncertain static-moving objects, several fixed waypoints and variable ocean current. A k-means clustering method is employed to classify the coast, water and uncertain area of the map. To this purpose, a large map with the size of $1000 \times 1000$ pixels is conducted that presents $10 \times 10$ km square area for the TAR and a smaller part of the map in size of $350 \times 350$ pixels that corresponds to $3.5 \times 3.5$ km square is selected to implement and test the ORPP performance. In this map each pixel corresponds to $10 \times 10$ m square space.

The clustered map is converted to a matrix format, in which the matrix size is same to map's pixel density. The corresponding matrix is filled with value of zero for coastal sections (forbidden area in black colour), value of (0,0.35] for uncertain sections (risky area in grey colour), and value of one for water covered area as valid zone for vehicles motion that is presented by white colour on the clustered map. The utilized clustering method is capable of clustering any alternative map very efficiently in a way that the blue sections sensed as water covered zones and other sections depending on their colour range categorized as forbidden and uncertain zones.

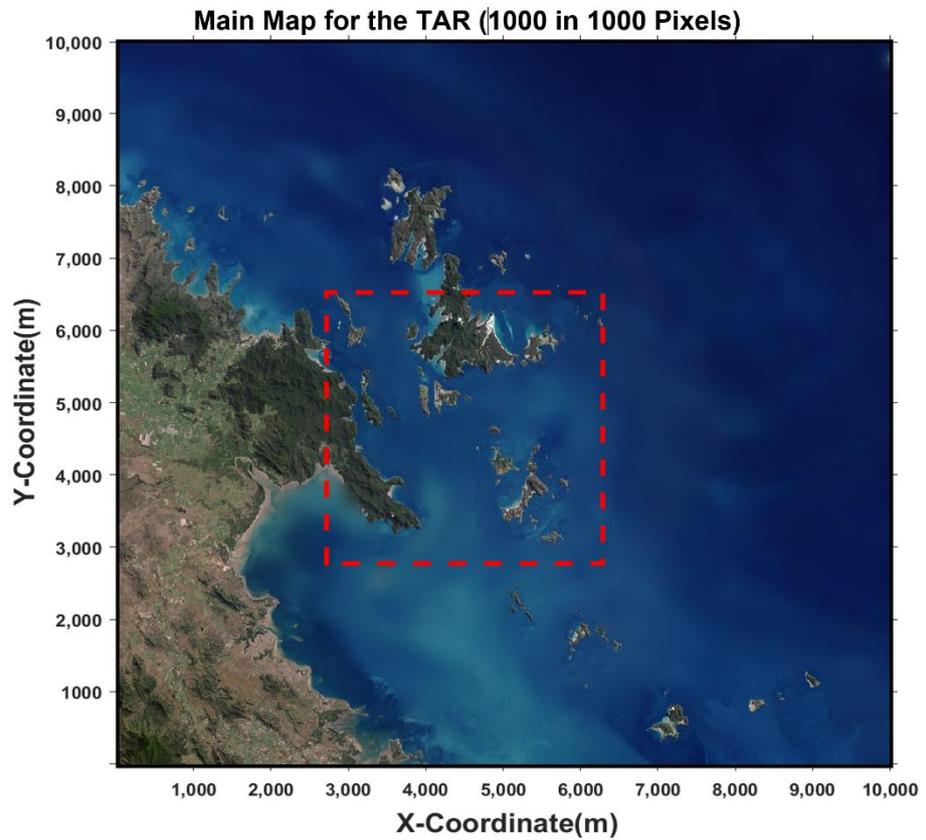

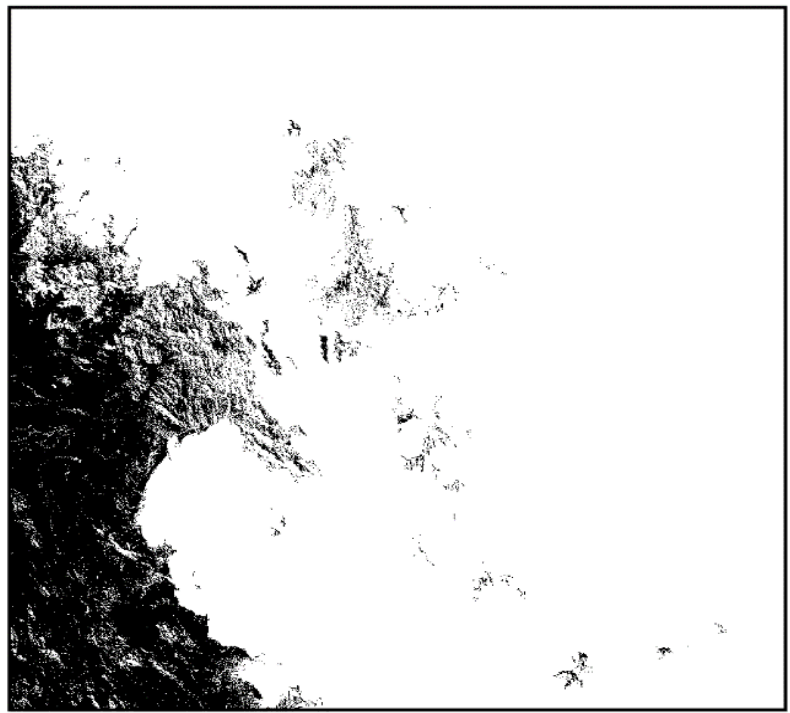

**Figure 5.4.** The original and clustered map used by TAR. The sub-map that is illustrated by dashed red line is conducted by ORPP module.

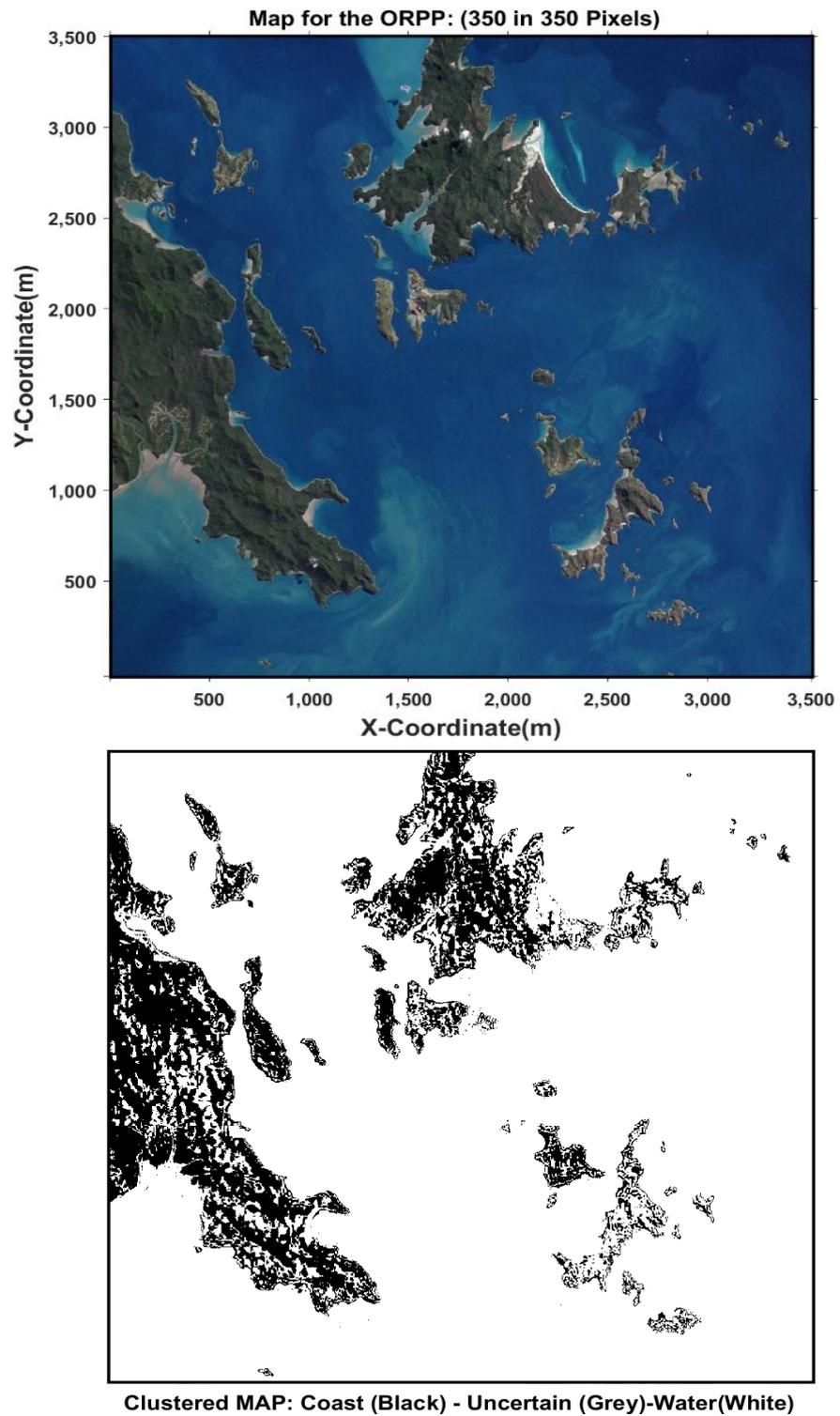

**Figure 5.5.** The original and clustered map used by ORPP.

The operation environment is covered by waypoints, so it is modelled as a geometrical network, where nodes corresponds to waypoints. Various tasks are assigned to passible distances between connected nodes in advance. Hence, every edge has weight and cost that is combination of tasks priority, tasks completion time, length of the edges and the time required for traversing the edges. Number of waypoint is set with a random number between 30 to 50 nodes that is calculated with a uniform distribution. The network topology also transforms randomly with a Gaussian distribution on the problem search space. A fixed set of 15 tasks, which are characterized with risk percentage, priority value, and completion time, are specified and are randomly assigned to some of the edges of the graph. The edges that are not assigned with a task are weighted with 1. The location of nodes are randomized according to $\{P^i_{x,y} \sim \mathbf{U}(0,10000)$ and $P^i_z \sim \mathbf{U}(0,100)\}$ in the water-covered area.

In addition to offline map, different types of static-moving obstacles are considered in this study in order to cover different possibilities of the real world situations (modelled in Chapter 4). Obstacle's velocity vector and its coordinates can be measured by the sonar sensors with a specific uncertainty that is modelled with a Gaussian distribution; hence, each obstacle is presentable by its position ($\Theta_p$), dimension (diameter $\Theta_r$) and uncertainty ratio. Position of the obstacles are initialized using normal distribution of $\mathbf{N} \sim (0,\sigma^2)$ bounded to position of two targeted waypoints by the path planner (e.g. $P^a_{x,y,z} < \Theta_p < P^b_{x,y,z}$), where $\sigma^2 \approx \Theta_r$. Further details on modelling of different obstacles is given by Chapter 4.

Beside the uncertainty of operating field, which is taken into account in the modelling of the operation terrain, water current needs to be addressed thoroughly in accordance with the type and the range of the mission. For the purpose of this study, a 3D turbulent time-varying current is considered that is generated using a multiple layered 2D current maps, in which the current circulation patterns gradually change with depth. It is assumed the AUV is traveling with constant velocity $|v|$ between two waypoints. The vehicular constraints and boundary conditions on AUV actuators and state also are taken into account for realistic modelling of the AUV operation. As discussed earlier the violation function for the path planners is defined as a combination of the vehicle's depth, surge, sway, theta, yaw and collision violations. According to (5-1), the boundaries of the constraints are defined as follows: the $z_{min}=0(m)$; $z_{max}=100(m)$;

$u_{max}$=2.7(*m/s*); $v_{min}$=-0.5(*m/s*); $v_{max}$=0.5(*m/s*); $\theta_{max}$=20 (*deg*); $\psi_{min}$=-17 (*deg*) and $\psi_{max}$=17 (*deg*). The proposed autonomous/reactive system is implemented in MATLAB®2016 on a desktop PC with an Intel i7 3.40 GHz quad-core processor and its performance is statistically analysed.

### 5.5.2 ARMSP Performance on Scheduling and Time Management for Reliable and Efficient Operation

The proposed architecture offers a degree of flexibility for employing a diverse set of algorithms and can perfectly synchronize them to have a unique performance. The performance of the model, thus, is independent of the meta-heuristics employed by its different components. The reason for this advantage is to join two disparate perspective of vehicle's autonomy in high-level task organizing and low-level self/environment awareness in motion planning in a cooperative and concordant manner in which adopting diverse algorithms by these modules are not a detriment to the real-time performance of the system. To cover shortcomings of the path planning when operating in a large scale, the operating field for operation of the ORPP is split to smaller spaces between pairs of waypoints (by higher layer TAR); thus, re-planning a new trajectory requires rendering and re-computing less information. This leads to problem space reduction and reducing the computational burden, which is another reason for fast operation of the proposed approach. Two of the tested algorithms in Chapters 3 and 4 (DE and FA) are selected arbitrarily and are employed by TAR and ORPP modules to test the performance of the ARMSP model. In the proposed ARMSP architecture, the TAR is expected to be capable of generating a time efficient route with best sequence of tasks to ensure the AUV has a beneficial operation and reaches to the destination on time, as it is obligatory requirement for vehicles safety.

A beneficial operation is a mission that covers maximum possible number of tasks in a manner that total obtained weight by a route and its travel time is maximized but not exceed the time threshold. Along with an efficient mission planning, the ORPP in a smaller scale is expected be fast enough to rapidly react to prompt changes of the environment and generate an alternative trajectory that safely guides the vehicle through the specified waypoints optimally. Hence, the performance and stability of the model in satisfying the metrics of "obtained weight", "number of completed tasks", "cost of the TAR/ORPP" and "total violation of the model" is investigated. These factors are statistically analysed through 50 Monte Carlo simulation runs (presented by Figures 5.6 and 5.7).

Based on the results presented in Figure 5.7, the cost variation for both TAR and ORPP shows the stability of the model in producing optimal solutions as its range stands in a specific interval for all experiments (mission). The ORPP gets violation if the generated path crosses the collision edges, its depth drawn outside of the vertical operating borders or when the vehicles surge, sway, yaw and pitch parameters egress the defined boundaries; and the TAR gets violation if the route time $T_\Re$ exceeds the total available time $T_\nabla$. TAR and ORPP are recalled for several times in a particular mission. Figure 5.6 shows that the model accurately satisfies all defined constraints as the violation values considerably approach zero in all 50 executions (which is negligible).

Proper coordination of the higher and lower level components of TAR and ORPP in the system is another important factor. To provide a stringent and accurate concordance between two planners and real-time implementation of the model, some other performance indexes should be highlighted additional to those which addressed above. So one critical factor for both planners is having a short computational time to provide a concurrent synchronization. Fast operation of each planner, keeps any of them from dropping behind the process of the other one, since appearance of such a delay damages concurrency of the entire system.

Another significant performance metric that influences the synchronism of whole system is compatibility of the value of local path time ($T_\wp$) and the expected time ($T_\varepsilon$) in multiple operations of the ORPP. So existence of a reasonable correlation between $T_\wp$ and $T_\varepsilon$ values in each ORPP operation is critical to total performance of the ARMSP model, which means a big difference between these two parameters cause interruption in cohesion of the whole system, as it is critical for recognizing the requisition for mission re-planning. The models behaviour in satisfying these two performance metrics is also investigated in a quantitative fashion as illustrated by Figure 5.8.

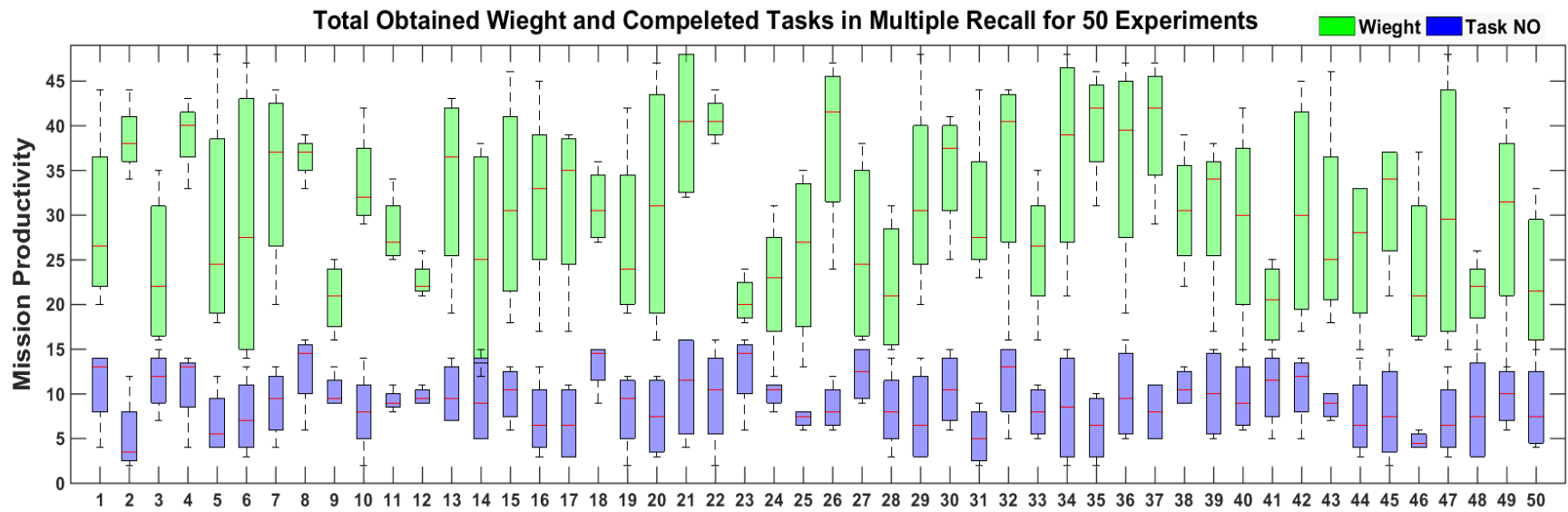

**Figure 5.6.** Statistical analysis of the model in terms of mission productivity according to total obtained weight and average number of completed tasks in a mission in 50 Monte Carlo simulations.

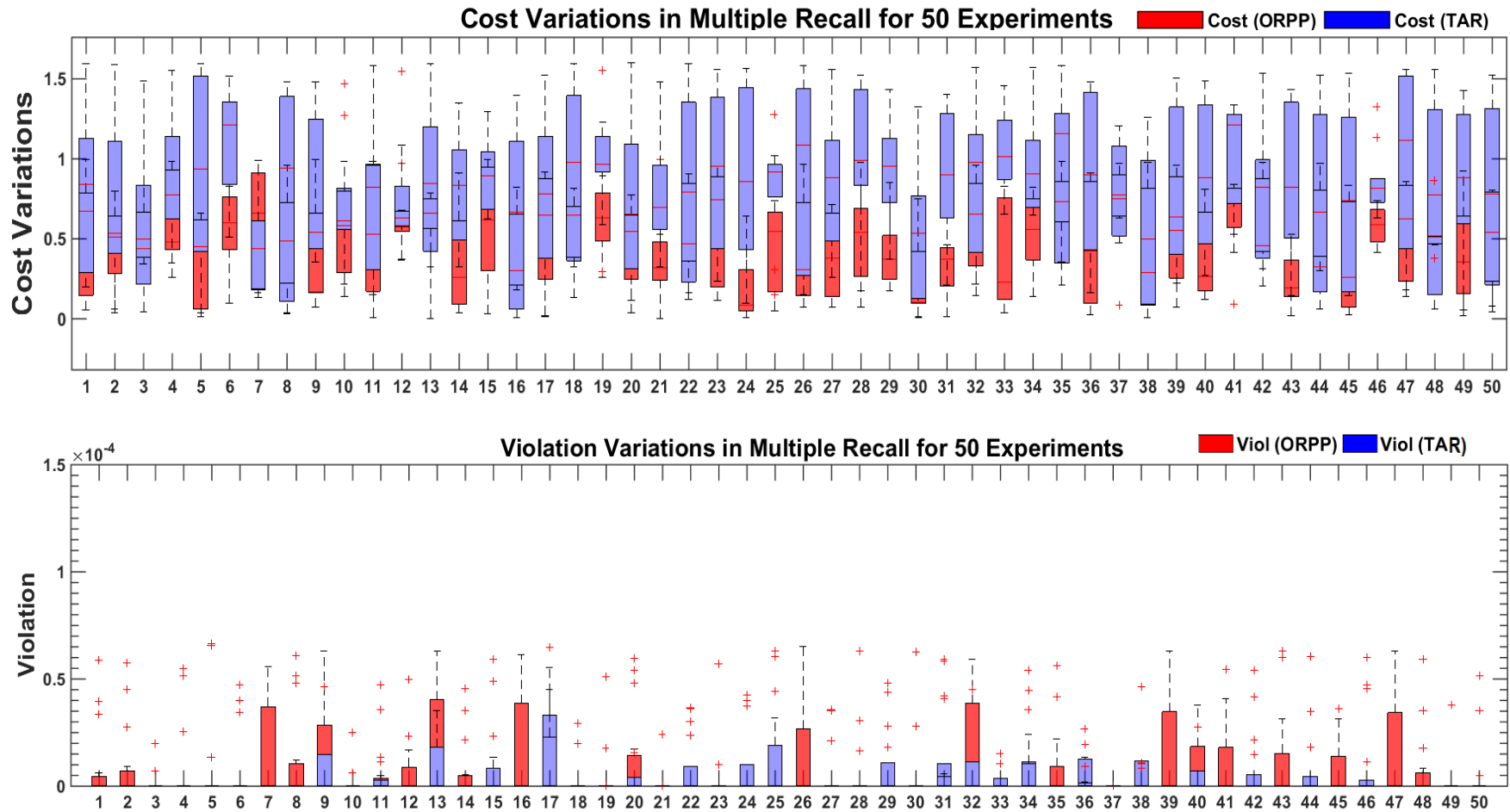

**Figure 5.7.** The cost and violation variation for both TAR and ORPP operations in 50 Monte Carlo simulations.

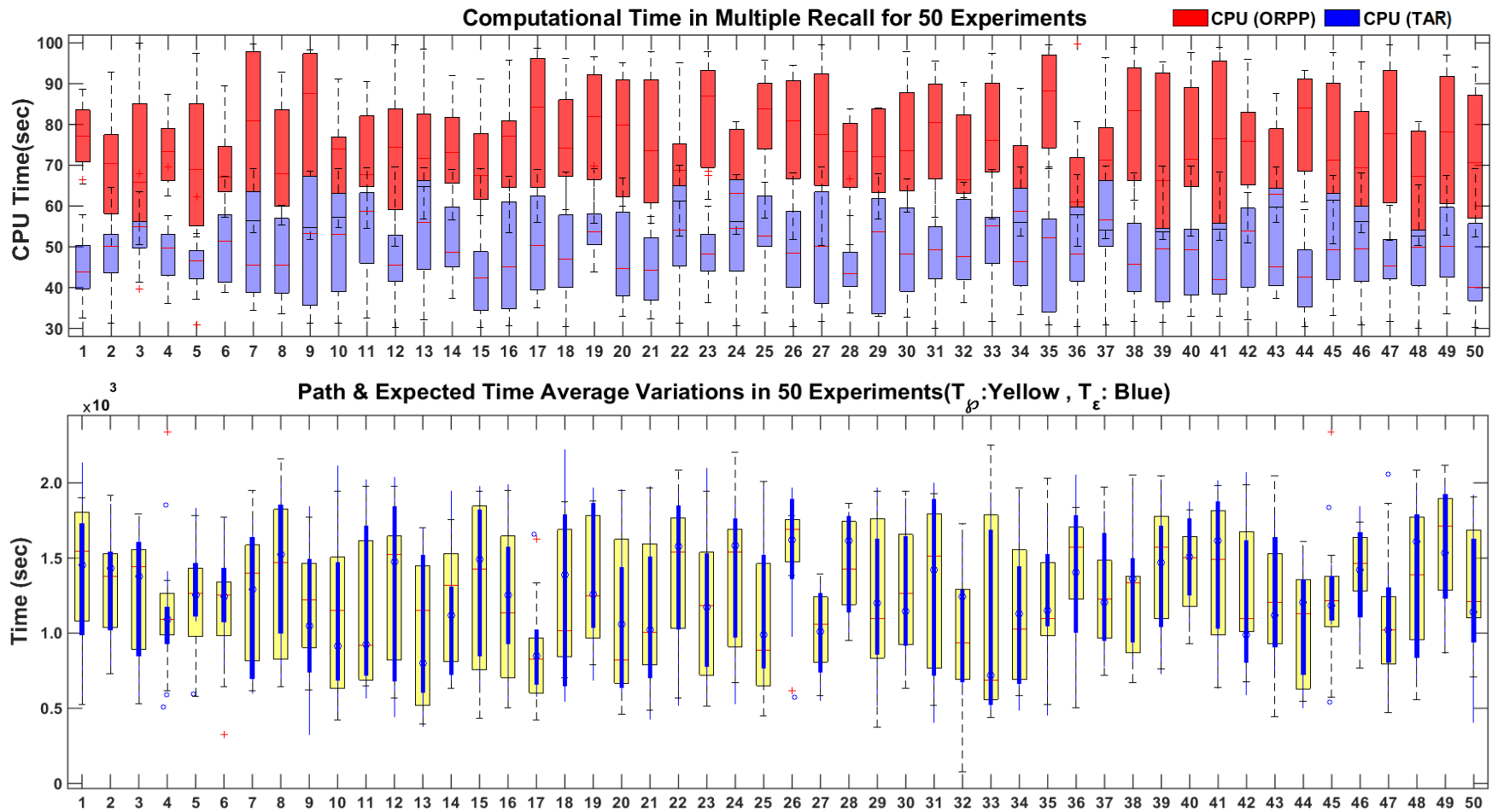

**Figure 5.8.** Real-time performance of the model, and compatibility of the value of $T_\wp$ and $T_\varepsilon$ in multiple operation of the ORPP through the 50 Monte Carlo simulation runs.

It is apparent from Figure 5.8 that variations of both $T_\wp$ and $T_\varepsilon$ are in a similar range (very close to each other). In this figure, the $T_\wp$ and $T_\varepsilon$ are presented with yellow transparent boxplot and blue compact boxplot, respectively. Figure 5.8 also presents the relative proportions of CPU time used for multiple runs of the ORPP compared to CPU time for TAR operations. It is noteworthy that, based on the results in Figure 5.8, both modules take a very short CPU time for all experiments that makes it highly appropriate for real-time application. It is evident that the range of variations of CPU time for both planners is almost in similar range in all experiments that proves the inherent stability of the model's real-time performance.

The system aims to take the maximum use of the total available time ($T_\nabla$), increase the mission productivity by collecting best sequence of tasks (fitted to $T_\nabla$), and to guarantee on-time termination of the mission. The stability of the ARMSP model in time management is the most critical factor representing robustness of the architecture. To this end, the robustness of the ARMSP model in terms of mission time management is evaluated by testing 50 missions' through individual experiments with the same initial condition that closely matches actual underwater mission scenarios, presented in Figure 5.9.

The most appropriate outcome for a mission is completion of the mission with the minimum residual time ($T_{Residual}$), which means the vehicle took maximum use of available time and terminated its mission before running out of battery; thus, the most effective performance indicator for this model is its accuracy in mission time management. The mission time ($T_\Re$) is the summation of time taken for passing through the waypoints (produced by the ORPP) and completion time for covered tasks along paths. It is clear in Figure 5.9 the proposed reactive ARMSP system is capable of making full use of the available time as the $T_\Re$ in all experiments approaches the $T_\nabla$ and met the above constraints denoted by the upper bound of $T_\nabla = 7.2 \times 10^3 (sec)$ (presented by the green line). Conversely, the $T_{Residual}$ has a linear relation to $T_\Re$ that should be minimized but it should not be equal to zero and is accurately satisfied considering variations of $T_{Residual}$ over 50 missions. Considering the fact that reaching to the destination is a big concern for vehicles safety, a big penalty value is assigned to strictly prevent missions from taking longer than $T_\nabla$.

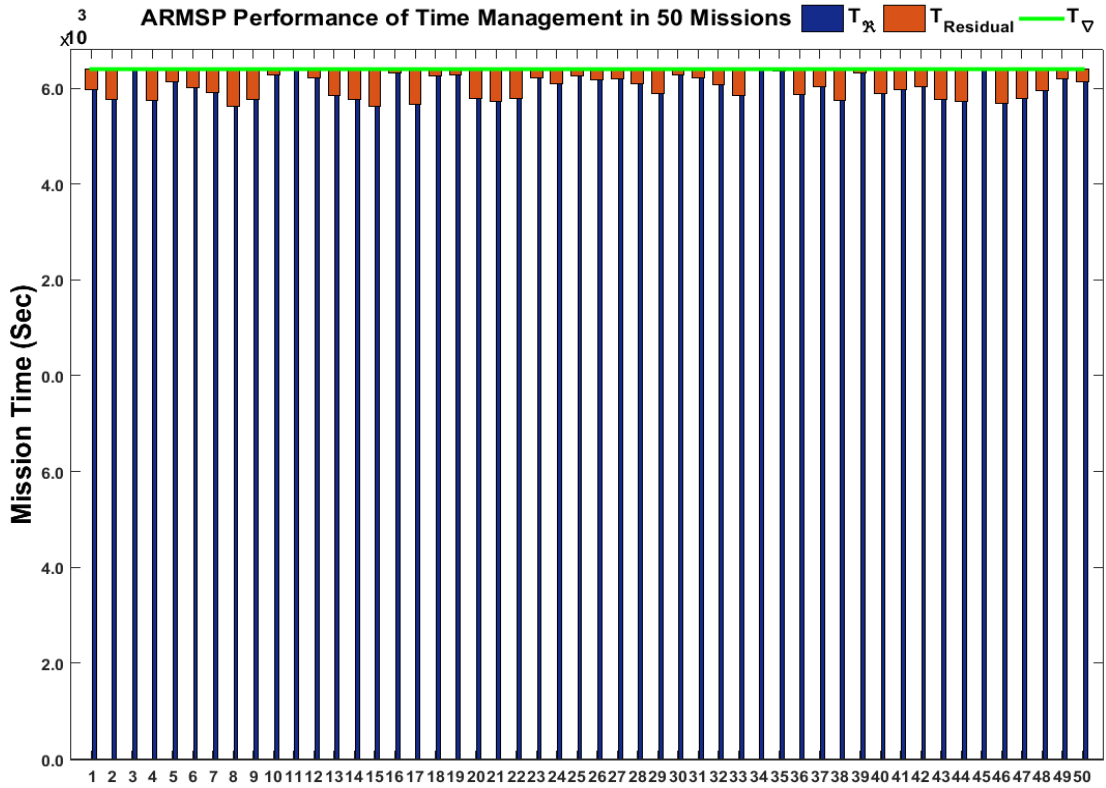

**Figure 5.9.** Statistical analysis of the ARMSP performance in terms of mission time management and satisfying time constraint in 50 missions.

As declared by Figure 5.9, the $T_{Residual}$ got a very small positive value in all missions that means no failure has occurred in this simulation. This is a remarkable achievement for having a confident and reliable mission as the failure is not acceptable for AUVs due to expensive maintenance in severe underwater environment. Obviously, the $T_\Re$ is maximized by minimizing the $T_{Residual}$, which represents that how much of total available $T_\nabla$ is used for completing different tasks in each mission. Analysing the variation of the $T_\Re$ and the $T_{Residual}$ confirms supreme performance of the proposed novel ARMSP model in mission reliability and excellent time management.

For proper visualization of the dynamic mission planning process, one of the experiments is presented in Figure 5.10 in which rearrangement of order of tasks and waypoints according to updated $T_{Residual}$ is presented through three re-planning procedure.

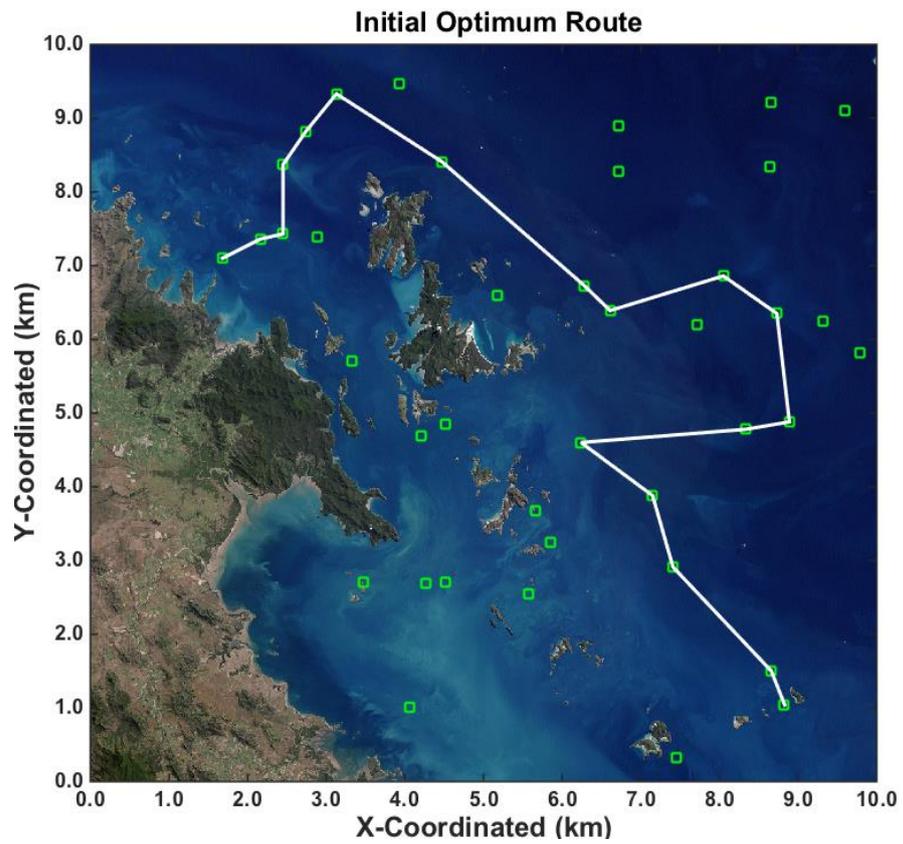

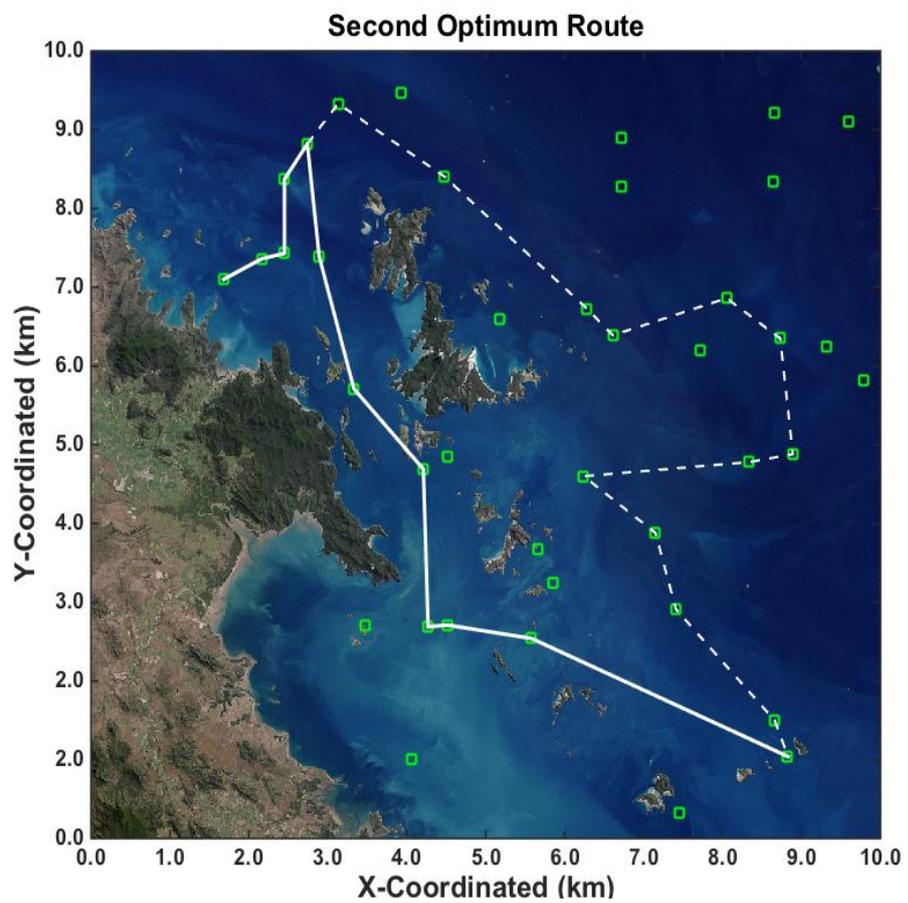

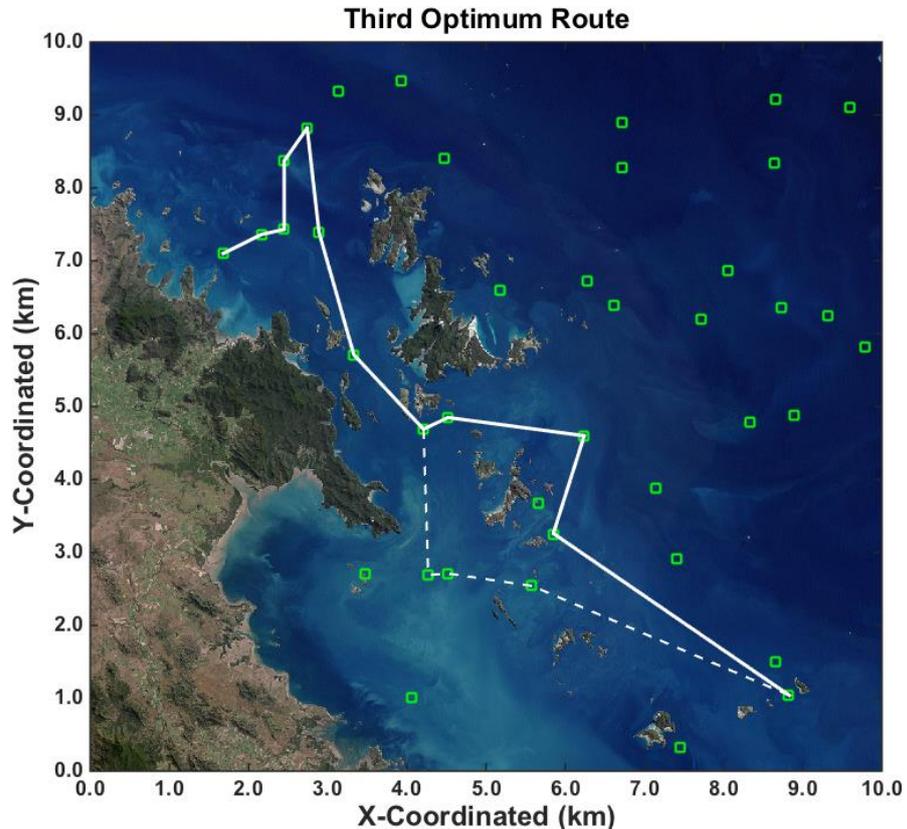

**Figure 5.10.** The mission planning/re-planning and tasks rearrangement based on updated terrain and available time.

In the beginning of the mission, the TAR is called to generate the initial appropriate task sequence fitted to available time. Given a candidate initial route in a sequence of waypoints (edges assigned by tasks) along with environment information, the ORPP provides a trajectory to safely guide the vehicle through the waypoints. In the whole process, the $T_{Residual}$ that is initialized with the value equal to $T_{\nabla}$, is counted inversely. Indeed, the $T_{Residual}$ is the total available time that is reduced as the mission is proceeds. The ORPP incorporates any dynamic changes of the terrain during the vehicle's deployment between two waypoints. In some cases, process of ORPP may take longer. This means the path took longer than expected time $T_\varepsilon$ for passing the corresponding distance (occurred in passing the fifth edge in Figure 5.10). In such a case, the remaining time ($T_{Residual}$) gets updated by reducing the wasted time and the TAR is recalled to rearrange the sequences of tasks according to updated $T_{Residual}$, presented by the white thick line in Figure 5.10. Referring to Figure 5.10, number of 17 paths generated during 3 mission re-planning processes that encapsulates number of 17 tasks (edges) with total weight of 47. In this figure, the discarded routes presented by dashed white lines. This synchronous process is frequently repeated and the interplay between the ORPP and TAR continues until the AUV reach the destination that means mission success.

### 5.5.3 Evaluation of the ARMSP through the Examination of Multiple Meta-heuristics: Single Run and 100 Monte Carlo Runs

To measure the performance optimality of the model, several performance metrics are utilized such as, total model cost and violation that involves the resultant violation of both ORPP and TAR approaches; mission profit which belongs to the number and total value of the completed tasks in a single mission; real-time performance of the system; and mission timing. The aforementioned factors are analysed in a single run for a different combination of the applied algorithms for both ORPP and TAR. The results are illustrated in Figures 5.11 to 5.13. Then, 100 runs of Monte Carlo simulation is carried out to prove the stability of the introduced model employing any meta-heuristic method, which is a remarkable advantage for a real-time autonomous system. From Figure 5.11, it is clear that computational time for any component of the system and also the whole operation of ARMSP is drawn in a reasonable interval in range of seconds. There is no remarkable difference between outcome of different combination of the proposed algorithms, which confirms ARMSP's capability in real time handling of task assigning/reassigning and coping with the dynamic changes of the terrain.

Figure 5.12 illustrates the comparison of elapsed time for carrying out a mission using different combination of algorithms. As mentioned earlier, it is highly critical to end the mission before vehicle runs out of battery and time. In addition, it is important for the model to take maximum use of total time and finish the mission with minimum positive residual time. Considering the variations of residual time (yellow boxes), route time (mission blue boxes), and total available time (red line over the boxes) in Figure 5.12, all missions terminated with a very small residual time and did not cross the predefined time boundary. This confirms the remarkable capability of the proposed model in mission timing which is critical for fully autonomous operations; however, it seems that the combination of ACO-DE, BBO-DE, BBO-BBO, and BBO-FA show the best timing performance between others.

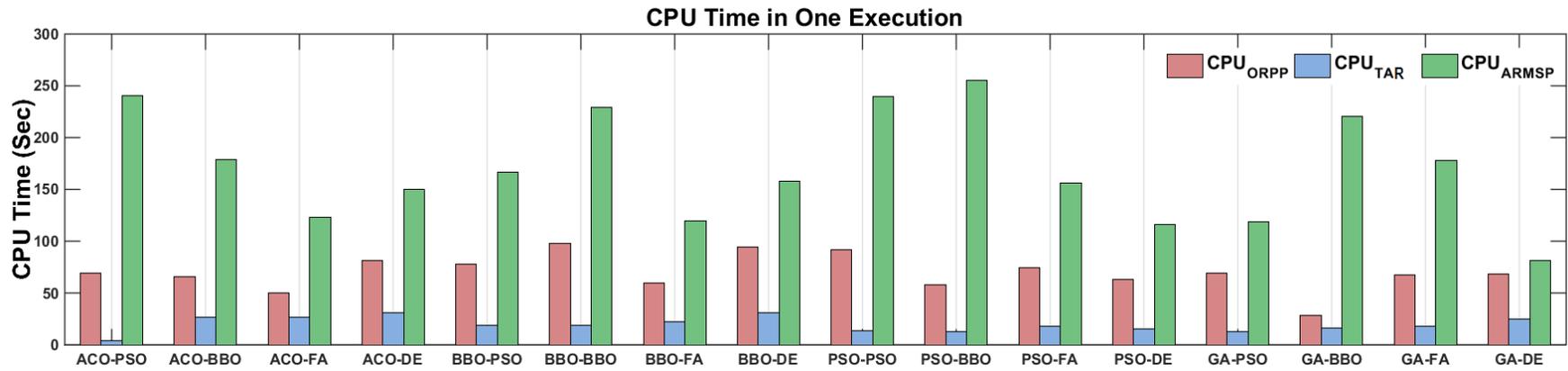

**Figure 5.11.** The computational time of the ORPP, TAR modules and entire ARMSP operation in one execution run for different combination of aforementioned algorithms; (each execution of ARMSP includes multiple executions of TAR and ORPP due to re-planning requisitions).

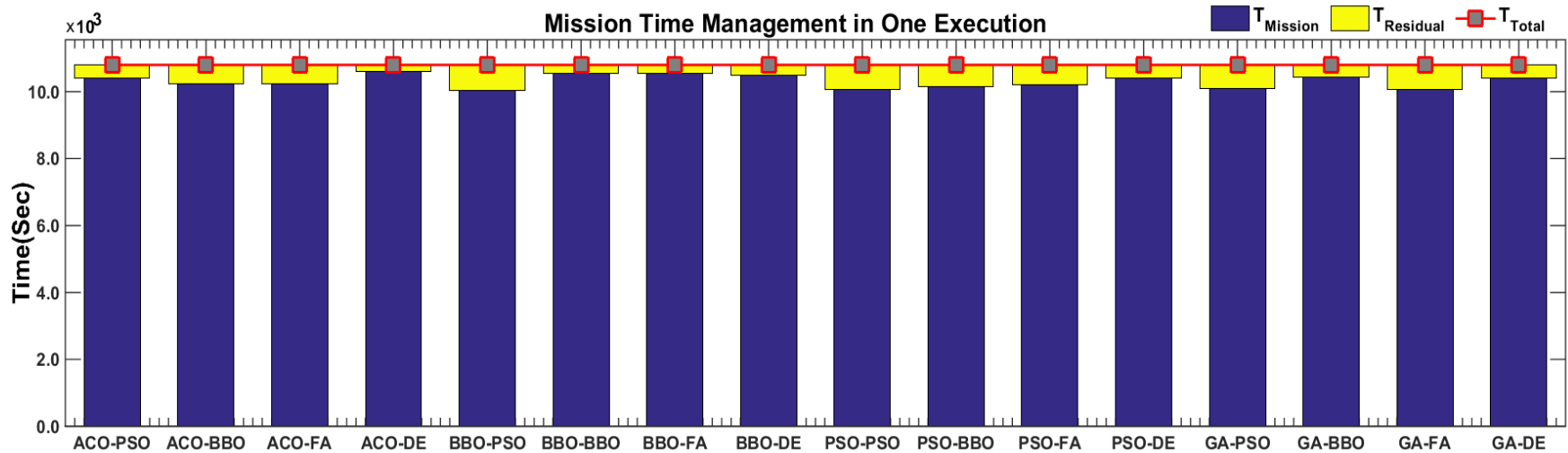

**Figure 5.12.** Mission timing performance of the ARMSP model.

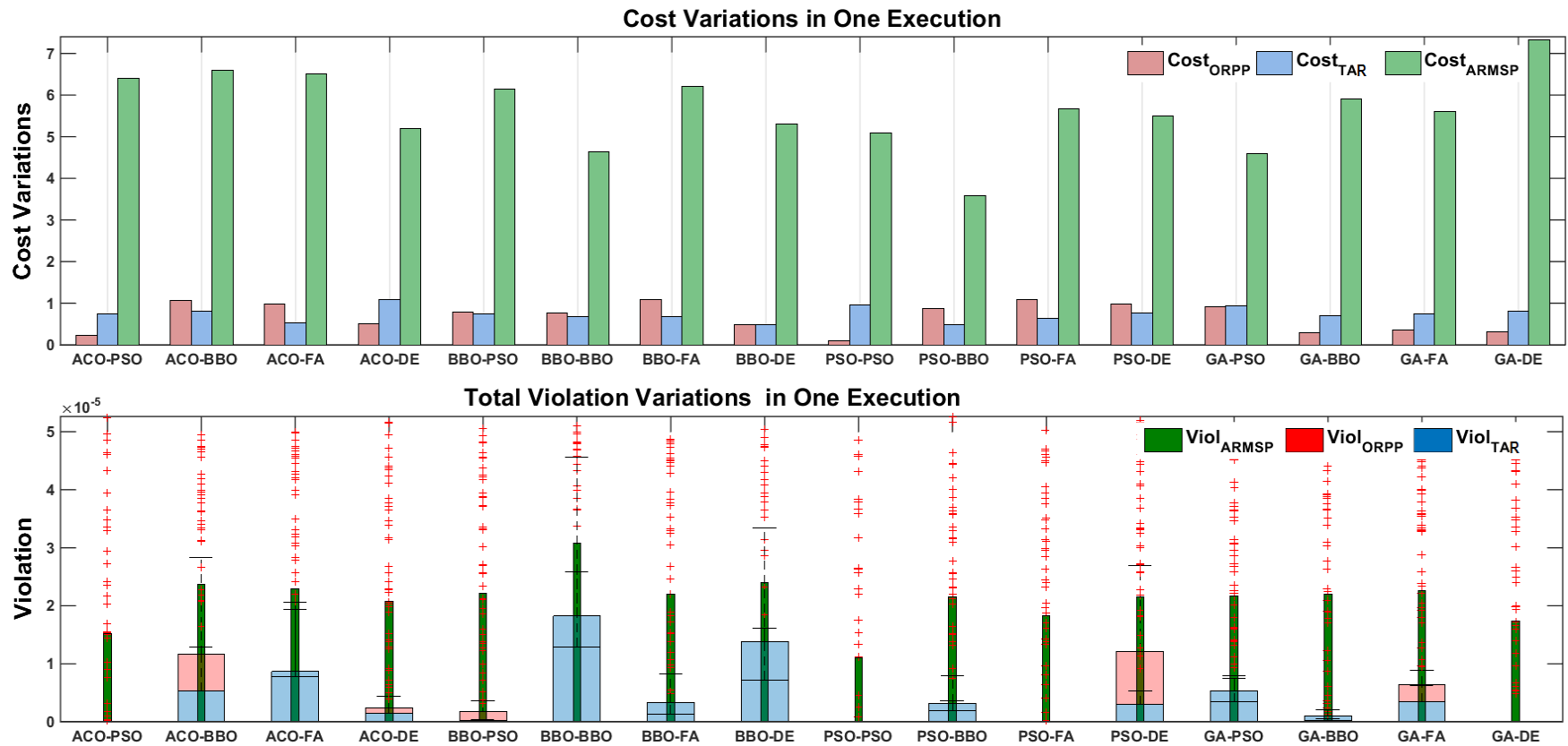

**Figure 5.13.** The cost and violation variation of the ORPP, TAR modules and entire ARMSP operation in one execution run for different combination of aforementioned algorithms.

Along with mentioned factors, the stability of the model in terms of managing total systems violation is also important issue that should be taken into consideration. According to results presented in Figure 5.13, the cost variation for the model and its components shows the stability of the model in producing optimal solutions as the cost variation range stands in a specific interval for all cases. Both TAR and ORPP are recalled for several times in a mission. Violation variations in Figure 5.13 shows that the model accurately satisfies all defined constraints and efficiently manages the engaged components toward eliminating the total violation regardless of the employed algorithms.

Carrying out the Monte Carlo simulations gives more confidence about total model's performance, stability and accuracy in dealing with random transformation of the environment or operational graph size and topology. The 100 Monte Carlo simulation runs are carried out for all applied algorithms to prove independency of the proposed strategy from the applied meta-heuristic methods, given by Figure 5.14 to 5.17.

The results captured from Monte Carlo simulation in Figure 5.14 also confirms that the proposed model is remarkably appropriate for real-time application as all the CPU time variations are drawn in a very narrow boundary in range of seconds for all combination of employed algorithms by TAR and ORPP. It is also noteworthy to mention from the analysis of the results in Figure 5.15 that the variation ranges of performance metrics of total obtained weight and completed tasks is almost in a same range for all applied algorithms. This shows the stability and robustness of the model in maintaining usefulness of the mission in a reasonable range when dealing with environmental changes and random deformation of the topology of operation network. Referring to Figure 5.16, the total model cost is also varying in a similar range for all applied algorithms (almost between 4.5 to 7). The results obtained from analysis of the cost variations in Figure 5.16 also confirms the inherent robustness and stability of the proposed model independent of the applied algorithm.

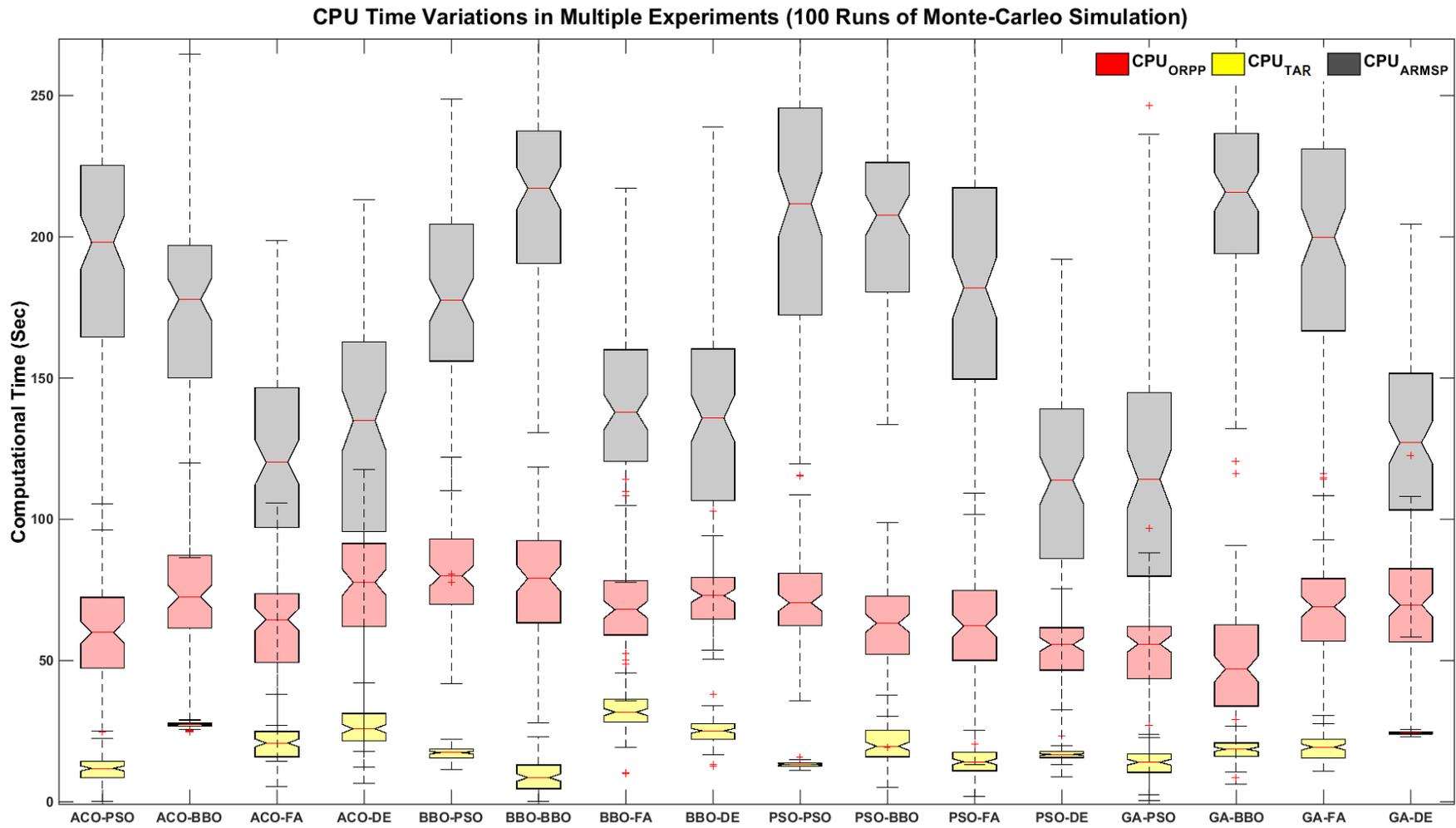

**Figure 5.14.** Statistical analysis computational performance of the ARMSP model for multiple embedded meta-heuristic algorithms.

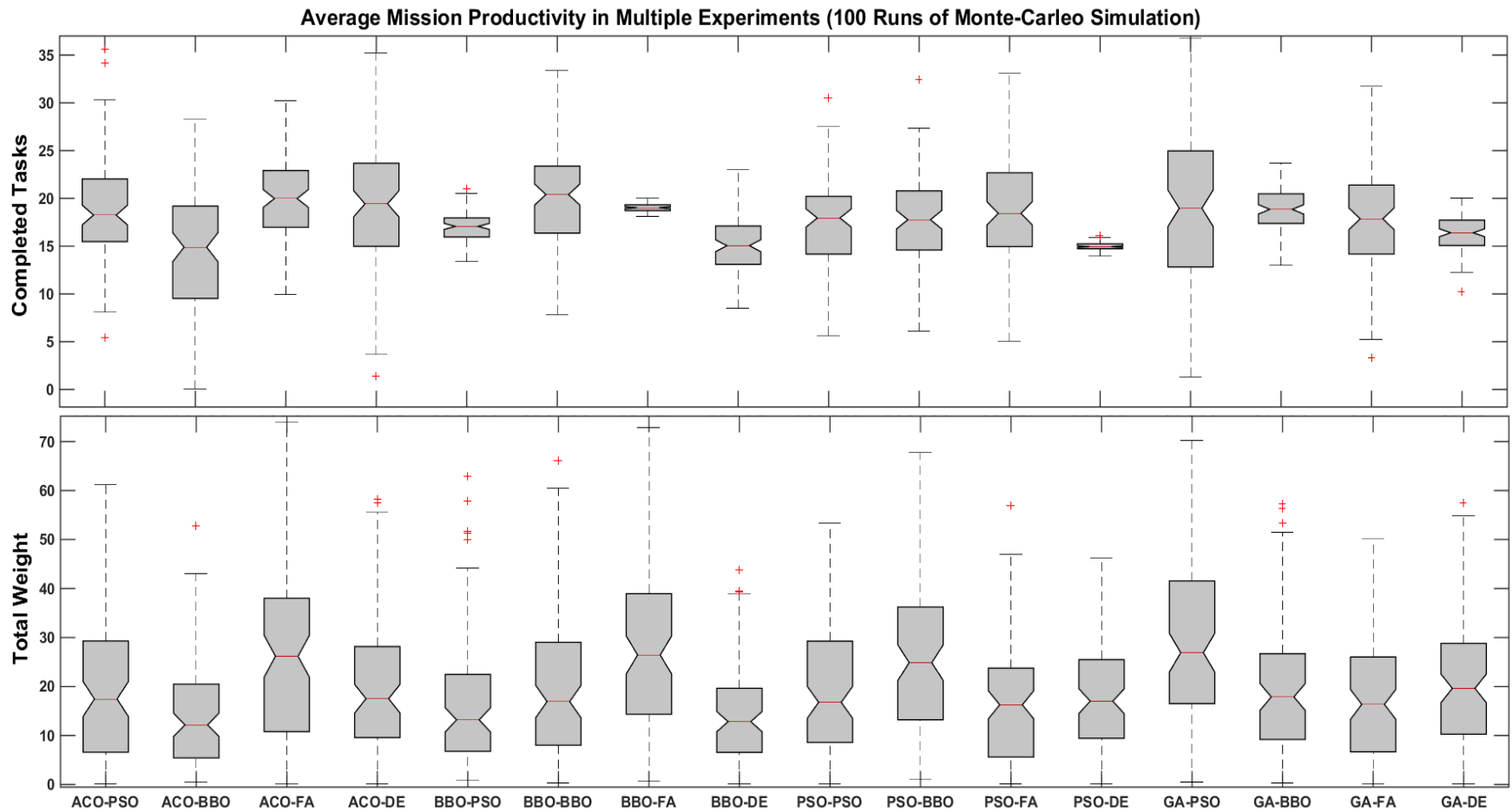

**Figure 5.15.** Average range of total obtained weight and number of completed tasks by different combination of embedded meta-heuristics in multiple experiments.

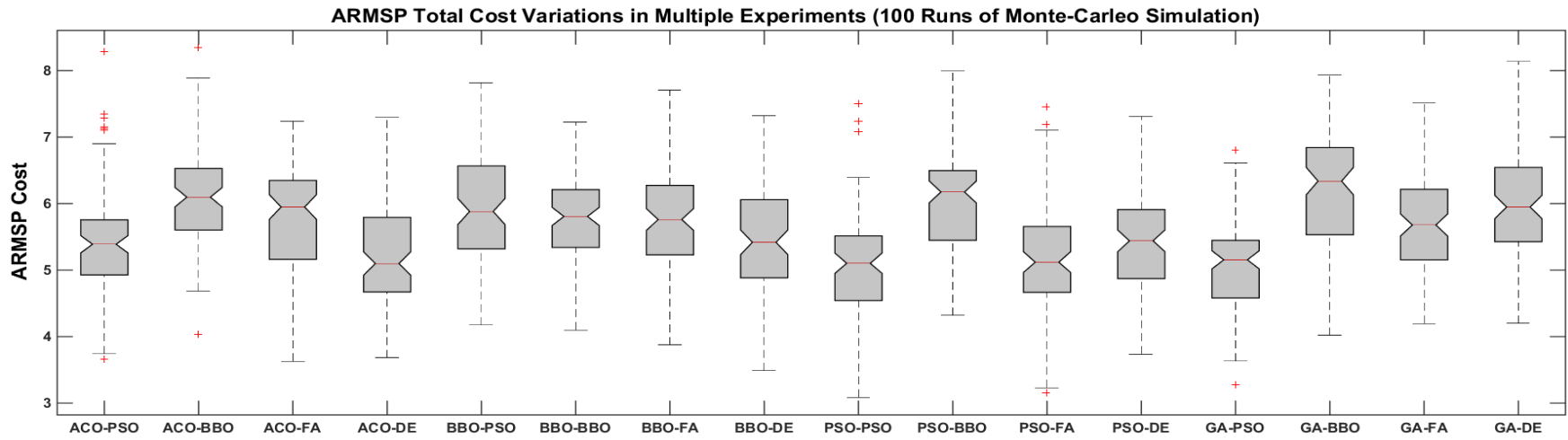

**Figure 5.16.** Statistical analysis of the total model cost for different combination of embedded meta-heuristics in multiple experiments.

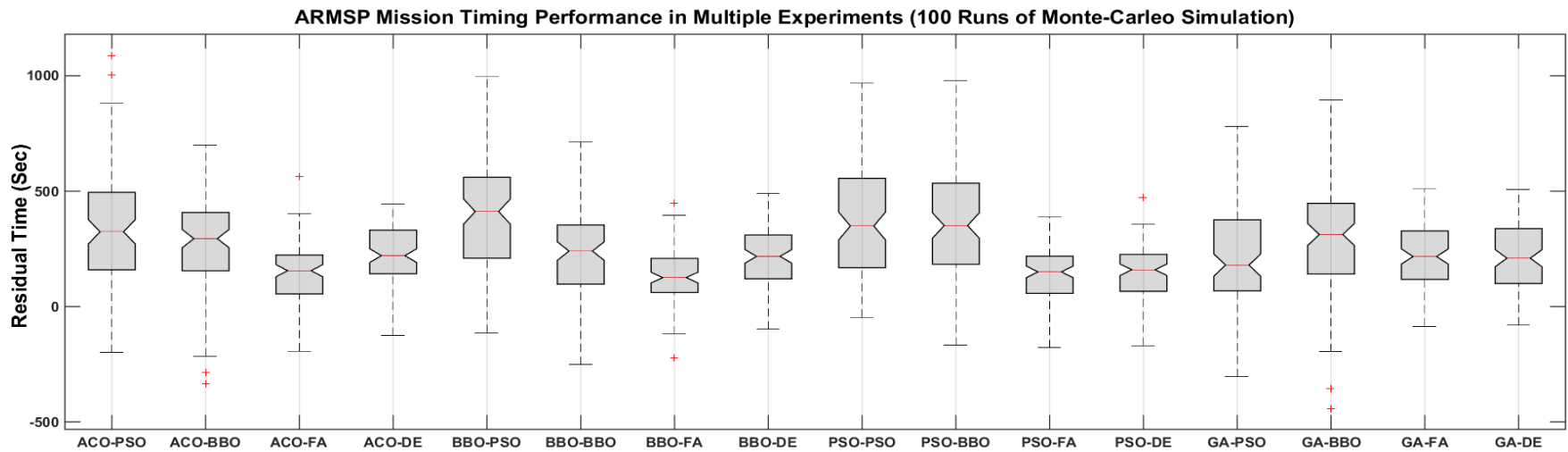

**Figure 5.17.** Average range of residual time variations in multiple experiments for different combination of embedded meta-heuristics.

To further prove the timing accuracy of the proposed ARMSP model in different situations and against problem space deformations, variation range of the residual time in 100 Monte Carlo executions is presented by Figure 5.17 for all applied algorithms. It is noted the residual time variations for all cases is lied in a positive scale (except some outliers denoted by red cross sign under the boxplots) and comparing to the assigned time threshold of $T_\nabla$:$T_{Total}$=10800 (sec) considerably minimized and approached to zero which means maximum part of the available time is used. This achievement confirms the performance of the ARMSP model in mission timing.

## 5.6 Chapter Summary

In this chapter, an autonomous reactive architecture of ARMSP has been developed that contains both high-level decision-making and low-level action generator. This modular architecture is developed based on two mission planning (TAR) and path planning (ORPP) modules that were implemented and evaluated in two previous chapters. The ARMSP can simultaneously organize the tasks based on their priority, order, risk percentage and completion time; and ensure mission completion. In the lower level, it is also capable of enhancing vehicle's manoeuvrability and increasing its autonomy for operating in a cluttered and uncertain environment and carrying out a fruitful mission.

The proposed architecture offers a degree of flexibility for employing a diverse set of algorithms and can effectively synchronize them to achieve stable performance, in so that adopting different algorithms in the modules is not detrimental to real-time performance of the system. A quantitative analyse is provided in order to examine the capability and efficiency of the proposed framework performed in MATLAB®2016b. This implementation includes the robustness assessment in terms of statistical analysis of the performance of the framework in diverse experiments to ensure its reliability and stability regardless of the applied algorithms. Synchronization between higher and lower level modules is efficiently configured to manage the mission time and to guarantee on-time termination of the mission. Real map data and different uncertain dynamic/static objects are encountered to model a realistic marine environment.

Moreover, the impact of the time variant water current on the AUV's motion is incorporated.

The most critical factor for preserving the stability and consistency of the proposed real-time ARMSP scheme is maintaining comparably fast operation for each component of the system (TAR and ORPP) to prevent any of them from dropping behind the others. The analysis of the captured results from different simulated missions indicates the model's consistency and robustness against the problem space deformation. The results obtained from Monte Carlo analysis (presented by Figure 5.9) demonstrate the inherent robustness and efficiency of the proposed hierarchical reactive system in terms of time management and enhancing mission productivity by maximizing $T_\Re$ while each mission is terminated before the vehicle runs out of time/energy, as in all experiment $T_{Residual}$ gets nonzero positive value. The results, also confirm the effectiveness of the model in providing safe and beneficial operation. The captured CPU time from the Monte Carlo simulations indicate applicability of the ARMSP for real-time implementation, which is a significant factor for an autonomous system. The content of this chapter has been investigated and published through the seven journal papers (given in Appendices E to K).

# Chapter 6

# Conclusions and Future Work

This chapter summarizes the context of the accomplished research and points out the key findings of this Thesis. Thus, recommendations for possible future work are proposed.

## 6.1  Summary

An AUV needs to acquire a certain degree of autonomy for any particular underwater mission to fulfil the mission objectives successfully and ensure its safety in all stages of the mission in a large scale operating filed. As regards, existing AUVs are restricted in terms of battery capacity and endurance, they should be capable of managing mission time. Obviously, a single AUV is not able to meet all specified tasks in a single mission with confined time and energy, so the vehicle has to manage its resources effectively to perform effective persistent deployment in longer missions. In this respect, time management is a fundamental requirement toward mission success that tightly depends on the optimality of the selected tasks between start and destination point. Assuming that the tasks for a specific mission are distributed in different areas of a waypoint cluttered graph-like terrain, proposing an accurate task-assign/routing scheme and task organizer facilitates the vehicle to accurately order the tasks and to take the best use of a restricted time so that the order of tasks guide the vehicle toward the destination. Thus, design of an efficient mission-planning framework considering vehicle's properties and resources is essential requirement for maximizing mission productivity. On the other hand, having an efficient motion planning system facilitates the vehicle to cope with environmental sudden changes as a principle safety requirement.

In this Thesis a synchronic modular architecture of ARMSP is introduced to address higher level of decision autonomy for AUV employing a Task-Assign/Routing (TAR) system and an adaptive Online Real-time Path Planning (ORPP) approach, which provides a comprehensive control on mission timing, multi-tasking, safe and efficient deployment in a waypoint-based, cluttered, uncertain, and varying underwater environment. The ORPP handles unforeseen changes and generates an alternative trajectory that safely guides the vehicle through the specified waypoints with minimum time/energy cost. A constant interaction exists between high and low level planners across small and large scale.

Chapter 2 provided a basic definition of autonomy and importance of resource management as principle requirements for autonomous operations. Afterward, dependencies of autonomy to accurate decision-making and efficient situation awareness are pointed out. Thus, the task-assign/routing is introduced as a higher-level decision making system for ordering the tasks execution and guiding the AUV toward the targeted location; and respectively the relation of the path (motion) planning with awareness of the vehicle from different situations is explained by Sections 2.3 and 2.4. A principle review of the existing methods for autonomous vehicles' routing-task assignment problem and state of the art in autonomous vehicles motion (path-route-trajectory) planning is provided and discussed in this chapter.

Chapter 3 described the mission planning and formulated the problem. The reactive TAR mission planner is developed. Given a set of constraints (e.g., time) and a set of task priority values, the TAR tends to find the optimal sequence of tasks for underwater mission that maximizes the sum of the priorities and minimizes the total risk percentage while meeting the given constraints. To fulfil the objectives of this chapter toward solving the stated problems, the solution was presented in several steps to enables the vehicle to autonomously find an optimal route through the operation network, carry out the maximum number of highest priority tasks, and reach to destination on time. Novel modification was applied to route generating flow and route encoding distribution to prevent generating infeasible task sequences by reducing the possibility of loop-formation and speed up the entire process. Finally the TAR system has been simulated and evaluated on different graphs of varying complexity applying ACO, BBO, PSO and GA algorithm.

Furthermore, a path-planning module of ORPP is designed in Chapter 4 to operate in a smaller scale and concurrently plan trajectory between waypoints included in the task sequence. The ORPP operates in the context of the ARMSP system in a manner to be fast enough to handle unforeseen changes and regenerates an alternative trajectory that safely guides the vehicle with minimum cost. All aspects of the ORPP implementation including the model of different static-dynamic obstacles, time-varying ocean currents, and map clustering were delivered by Section 4.2. The Kino-dynamic model of the vehicle, B-Spline path formation, optimization criterion for the path planning problem, and online re-planning process were described in detail by Section 4.3. This chapter also provided an overview and mechanism of the applied evolutionary algorithms on ORPP process, presented in Section 4.4. In particular, path planning based on PSO, BBO, DE, and FA algorithms are developed and their performance are examined and compared through several scenarios in Section 4.5.

After having discussed on different aspects of mission planning and path planning, and development of TAR and ORPP modules, subsequently, Chapter 5 addressed problems and shortcomings associated with each planner by developing a novel autonomous architecture with reactive re-planning capability. The mechanism of the proposed modular ARMSP control architecture is presented by Section 5.4. The system incorporates two different execution layers, deliberative and reactive, to satisfy the AUV's high and low level autonomy requirements in motion planning and mission management, in which two previously developed modules of TAR and ORPP are integrated to operate in a synchronous and concurrent manner in context of the ARMSP. In addition, a 'Synchron module' is also designed to manage the synchronizations among higher and lower level modules and to recognize whether to carry out the mission rescheduling or path re-planning.

Finally, performance and stability of the ARMSP architecture in maximizing mission productivity, real-time operation, mission timing, and handling dynamicity of the terrain was evaluated using different combination of aforementioned meta-heuristic algorithms. Knowing the fact that existing approaches are mainly only able to cover a part of problem of ether motion/path planning or task assignment based on offline map, the provided framework is a remarkable contribution to improve an AUV's autonomy in both higher and lower levels due to its comprehensive mission-motion planning capability that covers shortcomings associated with previous strategies used in this scope.

## 6.2 Conclusions

The conclusions can be derived from the proposed TAR scheme in the deliberative layer of ARMSP; the ORPP scheme in the reactive layer, and entire structure of the proposed ARMSP control architecture in providing a higher level of autonomy and promoting vehicle's capabilities in decision-making and situational awareness. The main conclusions for each section are summarised below.

### 6.2.1. AUV Task-Assign/Routing (TAR): (Chapter 3)

The capability of AUVs to fulfil specific mission objectives is directly influenced by routing and task assignment system performance. Mission planning and task priority assignment have great impact on mission time management and mission success. Assuming the tasks for a specific mission are distributed in eligible water covered sections of the map (where beginning and ending location of a task appointed with waypoints), a mission planning system of TAR is designed in the top level that simultaneously rearranges the order of tasks taking passage of time into account. Addition to important fact of prioritizing tasks and fitting them to the available time, order of tasks, so that the vehicle is guided from a start point to the destination point, is another important factor that should be taken into account. Respecting these definitions, task-assign/routing problem has a combinatorial nature analogous to both BKP and TSP NP-Hard problems. Obtaining the optimal solutions for NP-hard problems is computationally challenging issue and difficult to solve in practice.

Moreover, obtaining the exact optimal solution is only possible for the certain cases that the environment is completely known and no uncertainty exist; however, the modelled environment by this research corresponds to a spatiotemporal time varying environment with high uncertainty. Meta-heuristics are appropriate approaches suggested for handling above-mentioned complexities. Making use of the heuristic nature of swarm based and evolutionary algorithms in solving NP-hard graph problems, PSO, BBO, ACO, and GA are employed to evaluate the performance of TAR module toward fulfilment of the proposed objectives at this stage. For performance evaluation, the AUV operation was simulated in a three-dimension large terrain (almost $10 \times 10\ km^2 \times 100\ m$), where several prioritized tasks are distributed in different areas of the terrain. The tasks are predefined and initialized in advance with specific characteristics. Then, considering the limitations over the mission time, vehicle's battery, uncertainty and variability of the underlying operating field, appropriate mission timing and energy management is undertaken.

A series of Monte Carlo simulation trials are undertaken. The results of simulations demonstrate that the proposed method is reliable and robust particularly in dealing with uncertainties and problem space deformations; as a result, the proposed TAR model can significantly enhance the level of vehicle's decision autonomy in prioritizing tasks and mission time management by providing the best sequence of tasks (waypoints) with respect to the mission objectives and relying on the capability of providing fast and feasible solutions. By doing this, the optimal deployment zone is selected and the search area is minimized.

Operating in the turbulent and uncertain underwater environments gives many concerns for vehicle's deployment. Thus, robustness of the vehicle to current variability and terrain uncertainties is essential to mission success and AUV safe deployment. At this stage, the TAR module, however, is designed for the purpose of mission planning and task-time management and it is not able to cover dynamic variation of the underwater environment. Thus, an online path planner of ORPP is designed to assist the higher lever TAR in coping environmental dynamic changes.

### 6.2.2. AUV Online Real-Time Path Planning (ORPP): (Chapter 4)

AUV is obligated to react to changes of an unfamiliar and dynamic underwater environment, where usually a-priory information is not available. The path-planning scenario is defined through a highly cluttered and variable operating field. Providing a certain level of autonomy makes the vehicle and particularly path planner capable of using the favourable current flow for energy management. To have an expedient adaptation with the variations of the underlying environment, a novel online path planning module of ORPP was developed and implemented that is capable of refining the original path considering the update of current flows, uncertain static and moving obstacles. This refinement is not computationally expensive as there is no need to compute the path from scratch and the obtained solutions of the original path is utilized as the initial solutions for the employed methods. This leads to reduction of optimization space that accelerates the process of searching.

The proposed ORPP module entails a heuristic for refining the path considering situational awareness of environment. The planner thus needs to accommodate the unforeseen changes in the operating field such as emergence of unpredicted obstacles or variability of current field and turbulent regions. A penalty function method is utilized to combine the boundary conditions, vehicular and environmental constraints with the performance index that is expected path time ($T\varepsilon$). Applying the evolutionary methods namely PSO, BBO, DE, and FA indicate that they are capable of satisfying the path planning problem's conditions. The objective of this chapter was to synthesize and analyse the performance and capability of the mentioned methods for guiding an AUV from an initial loitering point toward the target waypoint through a comprehensive simulation study.

The simulation results demonstrate that this new approach along with the proposed algorithms could generate a time efficient path using the advantages of desirable current field and avoiding obstacles and no-fly zones. It is inferred from results of ORPP simulation, the generated trajectory by all aforementioned algorithms efficiently respect environmental and vehicular constraints that means safety of vehicle's deployment is efficiently satisfied; while comparing the optimality of the solutions differs in different situations, but in conclusion FA acts more efficiently in satisfying vehicle path planning objectives in all examined cases.

Although ORPP operates efficiently in coping with dynamicity of the environment, there are still many challenges when operating across a large-scale area as predicting the current variations and subsequently behaviour of the dynamic obstacles in large scale (far off the sensor coverage) is impractical and computationally inefficient. This becomes even more problematic in re-planning process, when a large data load of variation of whole terrain condition should be computed repeatedly. However, in this research, the higher layer TAR assist the ORPP to trim down the operation area to smaller beneficent operating zones that reduces computation burdens.

### 6.2.3. Autonomous Reactive Mission Scheduling and Path Planning (ARMSP) Architecture: (Chapter 5)

The aim of this research was to introduce a novel synchronous architecture including higher level task-organizer/mission-planner and lower level On-line path planner to provide a comprehensive control on mission time management and performing beneficent mission with the best sequence of tasks fitted to available time, while safe deployment is guaranteed at all stages of the mission. In this respect, the top level module of the architecture (TAR) is responsible to assign the prioritized tasks in a way that the selected edges in a graph-like terrain lead the vehicle toward its final destination in predefined restricted time. Respectively, in the lower level the ORPP module handles the vehicle safe deployment along the selected edges in presence of severe environmental disturbances, in which the generated path is corrected and refined reactively to cope with the sudden changes of the terrain. Precise and concurrent synchronization of the higher and lower level modules is the critical requirement to preserve the accretion and cohesion of the system that promotes stability of the architecture toward furnishing the main objectives addressed above. Developing such a structured modular architecture is advantageous as it can be upgraded or modified in an easier way that promotes reusability of the system to be implemented on other robotic platforms such as ground or aerial vehicles. In this way, the architecture also can be developed by adding new modules to handle vehicles inner situations, fault tolerant or other consideration toward increasing vehicle's general autonomy.

The problem formulated for this research was a combination of three NP-hard problems and currently there is no polynomial time algorithm that solves an NP-hard problem. Relying on competitive performance of the meta-heuristic algorithms in handling computational complexity of NP-hard problems, different meta-heuristic methods carefully chosen and configured according to expectation from the corresponding problem; and then used to evaluate the TAR and ORPP components of the proposed ARMSP model. The performance of TAR was investigated applying ACO, BBO, PSO and GA algorithms and the ORPP is validated using BBO, PSO, FA, and DE algorithms; after all, different combination of these algorithms applied for evaluating the performance stability of the ARMSP in maximizing mission productivity by taking maximum use of total available time, system real-time performance, accuracy of mission timing, and handling dynamicity of the terrain. The introduced ORPP used useful information of previous solutions to re-plan the subsequent trajectory in accordance with terrain changes and this method considerably reduces the computational burden of the path planning process. Moreover, only local variations of the environment in vicinity of the vehicle is considered in path re-planning, so a small computational load is devoted for re-planning procedure since the upper layer of TAR gives a general overview of the operation area that vehicle should fly through. An efficient synchronization flows between higher and lower level modules.

A systematic comparison of meta-heuristic methods was carried out for validation of the ARMSP overall performance. The simulation results carefully analyzed and discussed in Chapter 5. Accuracy and performance of all different combination of the employed algorithms has been studied globally in different situations that closely matches the real-world conditions. Reflecting on the analysis of the simulation results, all algorithms reveal very competitive performance in satisfying addressed performance metrics. Variation of computational time for operation of both TAR and ORPP modules settles in a narrow bound in the range of seconds for all experiments that confirm accuracy and real-time performance of the system, which makes it remarkably appropriate for real-time application. More importantly, referring to results in Chapter 5, the violation value is almost zero for operation of both modules claiming that the system is robust to the variations of the environment while operating in irregularly shaped uncertain spatiotemporal environment in presence of the time-varying water current and other disturbances. The subsequent Monte-Carlo test results show the significant robustness of the model against uncertainties of the operating field and variations of mission conditions. Independent of the type of meta-heuristic algorithm applied in this study, there was also a significant robustness with the model in terms of the terrain's variability and configuration changes of the allocated tasks.

It is noteworthy to mention from analyses of the results of either TAR, ORPP, and ARMSP architecture, the proposed strategy enhances the vehicle's autonomy in time management and carrying out safe, reliable, and efficient mission for a single vehicle operation in a specific time interval. This framework is a foundation for promoting vehicle's autonomy in different levels of self-awareness and mission management-scheduling, which are core elements of autonomous underwater missions.

## 6.3 Future Research Directions

Building on the contributions of this research, future work will concentrate on the design of additional modules to cover vehicles self-awareness and fault tolerance. It is also targeted to develop vehicles ability of risk assessment and risk management as its failure is not acceptable due to expensive maintenance.

For the future work, it is also possible to integrate the architecture with an additional module of predictor-corrector encompassing the ability of at least one-step ahead prediction of situational awareness of the environment that results in more flexibility for successful mission in real-world experiments. The architecture also has the potential of setting up for experimental operations using real-world sensory data. The proposed model can be expanded in the following directions:

**Direction ①: Internal Situation Awareness (SA) and Fault Tolerance**

The vehicle should be capable of recognizing any internal failure and accommodating the raised situation. To this purpose the proposed architecture of ARMSP can be upgraded by an extra module to provide internal awareness of the vehicle and diagnose its fault tolerance accordingly. A fault tolerant module helps the system to recognize the minor and major failures of vehicles subsystems and take an appropriate action accordingly. For example, the vehicle should be able to abort the mission safely in major failures or continue the mission (considering the new restrictions) for any tolerable failure. Following diagram represents the internal components that should be checked continuously during the mission. The feedback from this block is considered and are returned to decision making system as a part of situation awareness puzzle.

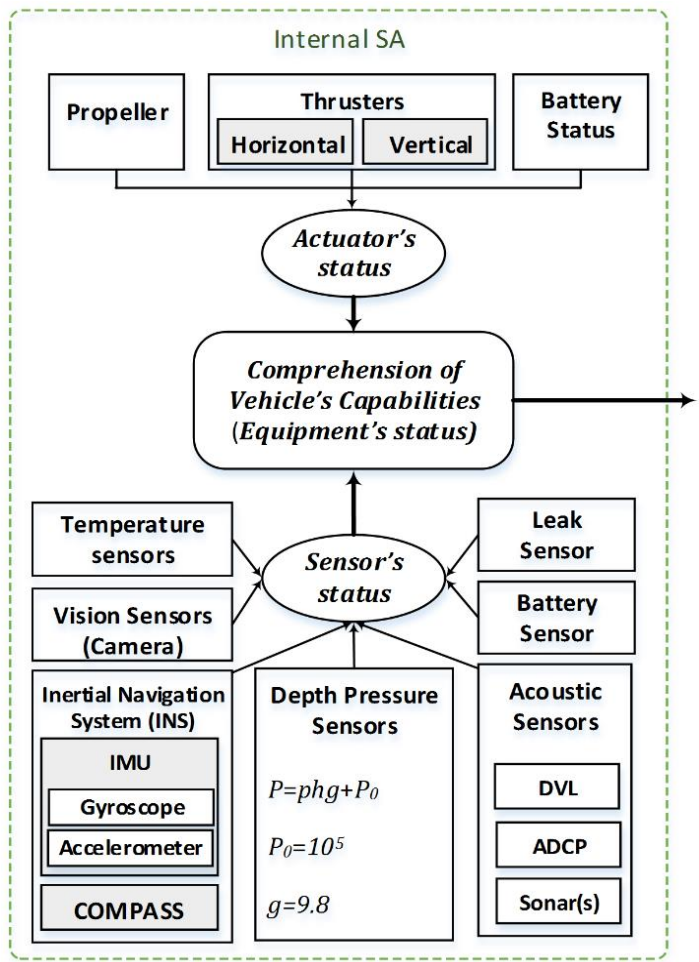

**Figure 6.1.** Vehicle internal hardware failure detection and recognize whether it is tolerable or not (Internal situation awareness).

The internal fault tolerant system includes three levels of fault detection (monitors the vehicle's overall hardware and software availabilities), fault recognition (determines the severity of the fault and whether it is tolerable or not), and fault accommodation (deal with the raised situation by safely aborting the mission for not-tolerable faults or reconfiguring the vehicle control architecture to successfully carry out the mission if the fault is tolerable). If a vehicle is damaged or some of its components fail, mission adaptation will be required to cope with the new restricted capabilities. In fully autonomous missions, the performance of the on-board sensors and other subsystems are checked continuously during the mission. In an autonomous system if internal errors push the system outside of its normal equilibrium state (mission process) the system tends to throwback the process to the normal state (original equilibrium state). Following examples helps understanding of this process:

> An AUV is capable of manoeuvring with six degree of freedom (DOF): sway, surge, heave, roll, pitch, and yaw. All six DOF motions are possible with existence of eight or just six thrusters.
>
> ***Thrusters:*** Four vertical thrusters provide motion of pitch, roll, and heave. Vertical motion is possible using four vertical thrusters or just two them. Four horizontal thrusters provide the surge(x), sway(y), and yaw motions. Vertical and horizontal thrusters are paired. Loos of two horizontal thrusters is tolerable for the vehicle and remaining thrusters can cover its motion in x-y plane with a careful path-planning. Thruster failure can be easily detected by monitoring thruster motor outputs. Decision on which force or moment direction can compensate loss of two thrusters, is necessary.
>
> ***Sensors:*** The bottom acoustic sonar and pressure sensors can measure the vehicle's depth and its altitude from the bottom. Loos of one of these sensors is also tolerable by replacing the other one.

**Direction ②: Data Fusion and External (Environmental) Situation Awareness**

Two sources of information can be conducted in order to provide external and internal situation awareness; extracted knowledge from the sensory data and the domain knowledge provided by expert. In both cases, it will be necessary for the information to be diagnosed, classified, and accessed efficiently by the SA and decision making unit while performing a mission.

*CHAPTER 6. CONCLUSIONS AND FUTURE WORKS*

Different types of environmental events and area of stationary or moving obstacles can be scanned and obtained through the various sensors. The relation among obtained information get used for event extraction and classification, such as recognizing threatening obstacle, coping unexpected and uncertain disturbances, etc. To provide awareness of surrounding environment, not only the individual obstacles should be identified, but also higher order relations among the different objects must be derived in order to recognition the threat degree and uncertainty of an obstacle. Obstacle group classification and recognition considering their dynamic deployment and behaviour based on their relative distance, relative angle, and relative velocity to vehicle can explain the threat degree. For threat recognition, some question can be answered using appropriate data fusion and accurate SA, such as:

- What type of object is it? (static/dynamic, quantity, dimension, etc.);
- Multiple sensors redundancy detection (whether different measurements of sensory signals are obtained from the same object or multiple objects).
- What is the purpose of the moving object? (Moving toward/just deploying);
- What is its behaviour in one step forward (its velocity and direction)?
- What is the threat (danger) degree?
- What is the appropriate reaction facing this specific object? (Avoid/ evade/ ignore);

Current data fusion models lack the capability of fully supporting the cognitive processes. Situations are recognized as states of a dynamic system observed at particular time; Complexity of the situations may range from a single attribute value of an object, to complex collections of objects interlinked by various relations. To this purpose, system integrates all information from the sensors and assess the availability of sufficient time for detection through classification, decision and reaction. To this purpose, a model is required to imitate human cognition as it relates to internal or external SA to produce appropriate decision-making process in a broader view. Designing such a module enhance the system capability of threat/disturbance recognition and protection by classifying the events and taking an appropriate reaction to handle raised event.

**Direction ③ Mission Risk Probability Assessment**

It is important for an autonomous vehicle to operate successfully in dealing with continuously changing situations. AUV's failure in the ocean is unacceptable due to expensive maintenance. From another point of view, the system needs to have a general background about close future states so that to have a fast and on-time interaction with the environment. The system should be able to autonomously carry out a proper decision facing completely non-deterministic events, which is known as autonomous decision-making. This will upgrade systems ability in fast response to the different unknown events and enhance the overall autonomy of the vehicle operating in hazardous uncertain environments. A reliable and appropriate probabilistic risk assessment strategy facilitates the vehicle to understand its surrounding hazardous environment, interact with the new unexperienced events and effectively handle the unforeseen and unknown phenomena. The information derived from SA get used to provide the decision-maker with the capability to understand how the model is arrived at a decision. Mission risk probability assessment is another direction that can be expanded and taken into consideration.

**Direction ④: Multiple Vehicle Cooperative Operation Using ARMSP Model**

Multiple vehicles cooperative operations would be a useful idea to cover battery restrictions and to cover multiple missions while the vehicles are communicating and managing the endurance time. The proposed control architecture has the potential of upgrading and implementing on multiple vehicles for multiple missions in a collaborative manner with no conflict. In this case, if any vehicle gets discarded from the team (abort its mission for any reason), the closest vehicle(s) with the most similar configurations undertake the incomplete mission or the mission gets divided between several vehicles and system re-plans a new mission scenario for all vehicles. The benefits of using multiple vehicles cooperative operations include covering long missions in larger areas, increased fault-tolerance, improved system resilience and robustness, and better risk management. Using Underwater Wireless Sensor Network (UWNS) can facilitate vehicles to communicate and share useful local information. In such a case, the operation network transforms to dynamic or semi dynamic network with time-varying characteristics.

**Further Benefits and Applications:**

The model I have developed was focused on AUVs, but was specifically designed to be generalized and to be compatible with other classes of autonomous applications, and in particular autonomous unmanned aerial, ground or surface vehicles.

Such a modular structure specifically increases the reusability and versatility the system in which upgrading or updating functionalities of any module does not require change in whole structure of the architecture. On the other hand, parallel execution of modules speeds up the computation process and also increases accuracy of the system in autonomous operations and concurrently handling all internal and external situations.

# Appendix A

# Optimal Route Planning with Prioritized Task Scheduling for AUV Missions

*Abstract*—This paper presents a solution to Autonomous Underwater Vehicles (AUVs) large scale route planning and task assignment joint problem. Given a set of constraints (e.g., time) and a set of task priority values, the goal is to find the optimal route for underwater mission that maximizes the sum of the priorities and minimizes the total risk percentage while meeting the given constraints. Making use of the heuristic nature of genetic and swarm intelligence algorithms in solving NP-hard graph problems, Particle Swarm Optimization (PSO) and Genetic Algorithm (GA) are employed to find the optimum solution, where each individual in the population is a candidate solution (route). To evaluate the robustness of the proposed methods, the performance of the all PS and GA algorithms are examined and compared for a number of Monte Carlo runs. Simulation results suggest that the routes generated by both algorithms are feasible and reliable enough, and applicable for underwater motion planning. However, the GA-based route planner produces superior results comparing to the results obtained from the PSO based route planner.

Link:   http://ieeexplore.ieee.org/document/7451578/authors

# Appendix B

# A Novel Efficient Task-Assign Route Planning Method for AUV Guidance in a Dynamic Cluttered Environment

*Abstract*— Promoting the levels of autonomy facilitates the vehicle in performing long-range operations with minimum supervision. The capability of Autonomous Underwater Vehicles (AUVs) to fulfil the mission objectives is directly influenced by route planning and task assignment system performance. This paper proposes an efficient task-assign route planning model in a semi-dynamic operation network, where the location of some waypoints are changed by time in a bounded area. Two popular meta-heuristic algorithms named biogeography-based optimization (BBO) and particle swarm optimization (PSO) are adopted to provide real-time optimal solutions for task sequence selection and mission time management. To examine the performance of the method in a context of mission productivity, mission time management and vehicle safety, a series of Monte Carlo simulation trials are undertaken. The results of simulations declare that the proposed method is reliable and robust particularly in dealing with uncertainties and changes of the operation network topology; as a result, it can significantly enhance the level of vehicle's autonomy by relying on its reactive nature and capability of providing fast feasible solutions.

Link:   http://ieeexplore.ieee.org/document/7743858/authors

# Appendix C

# Differential Evolution for Efficient AUV Path Planning in Time Variant Uncertain Underwater Environment

*Abstract*—The AUV three-dimension path planning in complex turbulent underwater environment is investigated in this research, in which static current map data and uncertain static-moving time variant obstacles are taken into account. Robustness of AUVs path planning to this strong variability is known as a complex NP-hard problem and is considered a critical issue to ensure vehicles safe deployment. Efficient evolutionary techniques have substantial potential of handling NP hard complexity of path planning problem as more powerful and fast algorithms among other approaches for mentioned problem. For the purpose of this research Differential Evolution (DE) technique is conducted to solve the AUV path planning problem in a realistic underwater environment. The path planners designed in this paper are capable of extracting feasible areas of a real map to determine the allowed spaces for deployment, where coastal area, islands, static/dynamic obstacles and ocean current is taken into account and provides the efficient path with a small computation time. The results obtained from analyze of experimental demonstrate the inherent robustness and drastic efficiency of the proposed scheme in enhancement of the vehicles path planning capability in coping undesired current, using useful current flow, and avoid colliding collision boundaries in a real-time manner. The proposed approach is also flexible and strictly respects to vehicle's kinematic constraints resisting current instabilities.

Link:  https://arxiv.org/ftp/arxiv/papers/1604/1604.02523.pdf

# Appendix D

# AUV Rendezvous Online Path Planning in a Highly Cluttered Undersea Environment Using Evolutionary Algorithms

*Abstract*—In this study, a single autonomous underwater vehicle (AUV) aims to rendezvous with a submerged leader recovery vehicle through a cluttered and variable operating field. The rendezvous problem is transformed into a Nonlinear Optimal Control Problem (NOCP) and then numerical solutions are provided. A penalty function method is utilized to combine the boundary conditions, vehicular and environmental constraints with the performance index that is final rendezvous time. Four evolutionary based path planning methods namely Particle Swarm Optimization (PSO), Biogeography-Based Optimization (BBO), Differential Evolution (DE), and Firefly Algorithm (FA) are employed to establish a reactive planner module and provide a numerical solution for the proposed NOCP. The objective is to synthesize and analyze the performance and capability of the mentioned methods for guiding an AUV from an initial loitering point toward the rendezvous through a comprehensive simulation study. The proposed planner module entails a heuristic for refining the path considering situational awareness of environment, encompassing static and dynamic obstacles within a spatiotemporal current fields. The planner thus needs to accommodate the unforeseen changes in the operating field such as emergence of unpredicted obstacles or variability of current field and turbulent regions. The simulation results demonstrate the inherent robustness and efficiency of the proposed planner for enhancing a vehicle's autonomy so as to enable it to reach the desired rendezvous. The advantages and shortcoming of all utilized methods are also presented based on the obtained results.

Link:   https://arxiv.org/ftp/arxiv/papers/1604/1604.07002.pdf

# Appendix E

# Toward Efficient Task Assignment and Motion Planning for Large Scale Underwater Mission

*Abstract*—An Autonomous Underwater Vehicle (AUV) needs to possess a certain degree of autonomy for any particular underwater mission to fulfil the mission objectives successfully and ensure its safety in all stages of the mission in a large scale operating field. In this paper, a novel combinatorial conflict-free-task assignment strategy consisting of an interactive engagement of a local path planner and an adaptive global route planner, is introduced. The method takes advantage of the heuristic search potency of the Particle Swarm Optimization (PSO) algorithm to address the discrete nature of routing-task assignment approach and the complexity of NP-hard path planning problem. The proposed hybrid method, is highly efficient as a consequence of its reactive guidance framework that guarantees successful completion of missions particularly in cluttered environments. To examine the performance of the method in a context of mission productivity, mission time management and vehicle safety, a series of simulation studies are undertaken. The results of simulations declare that the proposed method is reliable and robust, particularly in dealing with uncertainties, and it can significantly enhance the level of a vehicle's autonomy by relying on its reactive nature and capability of providing fast feasible solutions.

Link: http://journals.sagepub.com/doi/abs/10.1177/1729881416657974

# Appendix F

# Biogeography-Based Combinatorial Strategy for Efficient AUV Motion Planning and Task-Time Management

*Abstract*—Autonomous Underwater Vehicles (AUVs) are capable of spending long periods of time for carrying out various underwater missions and marine tasks. In this paper, a novel conflict-free motion planning framework is introduced to enhance underwater vehicle's mission performance by completing maximum number of highest priority tasks in a limited time through a large scale waypoint cluttered operating field, and ensuring safe deployment during the mission. The proposed combinatorial route-path planner model takes the advantages of the biogeography-based optimization (BBO) algorithm toward satisfying objectives of both higher-lower level motion planners and guarantees maximization of the mission productivity for a single vehicle operation. The performance of the model is investigated under different scenarios including the particular cost constraints in time-varying operating fields. To show the reliability of the proposed model, performance of each motion planner assessed separately and then statistical analysis is undertaken to evaluate the total performance of the entire model. The simulation results indicate the stability of the contributed model and its feasible application for real experiments.

Link:   https://link.springer.com/article/10.1007/s11804-016-1382-6

# Appendix G

# A Novel Versatile Architecture for Autonomous Underwater Vehicle's Motion Planning and Task Assignment

*Abstract*—Expansion of today's underwater scenarios and missions necessitates the requestion for robust decision making of the Autonomous Underwater Vehicle (AUV); hence, design an efficient decision making framework is essential for maximizing the mission productivity in a restricted time. This paper focuses on developing a deliberative conflict-free-task assignment architecture encompassing a Global Route Planner (GRP) and a Local Path Planner (LPP) to provide consistent motion planning encountering both environmental dynamic changes and *a priori* knowledge of the terrain, so that the AUV is reactively guided to the target of interest in the context of an uncertain underwater environment. The architecture involves three main modules: The GRP module at the top level deals with the task priority assignment, mission time management, and determination of a feasible route between start and destination point in a large scale environment. The LPP module at the lower level deals with safety considerations and generates collision-free optimal trajectory between each specific pair of waypoints listed in obtained global route. Re-planning module tends to promote robustness and reactive ability of the AUV with respect to the environmental changes. The experimental results for different simulated missions, demonstrate the inherent robustness and drastic efficiency of the proposed scheme in enhancement of the vehicles autonomy in terms of mission productivity, mission time management, and vehicle safety.

Link:   https://link.springer.com/article/10.1007/s00500-016-2433-2

# Appendix H

# An Efficient Hybrid Route-Path Planning Model for Dynamic Task Allocation and Safe Maneuvering of an Underwater Vehicle in a Realistic Environment

*Abstract*—This paper presents a hybrid route-path planning model for an Autonomous Underwater Vehicle's task assignment and management while the AUV is operating through the variable littoral waters. Several prioritized tasks distributed in a large scale terrain is defined first; then, considering the limitations over the mission time, vehicle's battery, uncertainty and variability of the underlying operating field, appropriate mission timing and energy management is undertaken. The proposed objective is fulfilled by incorporating a route-planner that is in charge of prioritizing the list of available tasks according to available battery and a path-planer that acts in a smaller scale to provide vehicle's safe deployment against environmental sudden changes. The synchronous process of the task assign-route and path planning is simulated using a specific composition of Differential Evolution and Firefly Optimization (DEFO) Algorithms. The simulation results indicate that the proposed hybrid model offers efficient performance in terms of completion of maximum number of assigned tasks while perfectly expending the minimum energy, provided by using the favorable current flow, and controlling the associated mission time. The Monte-Carlo test is also performed for further analysis. The corresponding results show the significant robustness of the model against uncertainties of the operating field and variations of mission conditions.

Link:    https://arxiv.org/ftp/arxiv/papers/1604/1604.07545.pdf

# Appendix I

# A Hierarchal Planning Framework for AUV Mission Management in a Spatio-Temporal Varying Ocean

*Abstract*—The purpose of this paper is to provide a hierarchical dynamic mission planning framework for a single autonomous underwater vehicle (AUV) to accomplish task-assign process in a limited time interval while operating in an uncertain undersea environment, where spatio-temporal variability of the operating field is taken into account. To this end, a high level reactive mission planner and a low level motion planning system are constructed. The high level system is responsible for task priority assignment and guiding the vehicle toward a target of interest considering on-time termination of the mission. The lower layer is in charge of generating optimal trajectories based on sequence of tasks and dynamicity of operating terrain. The mission planner is able to reactively re-arrange the tasks based on mission/terrain updates while the low level planner is capable of coping unexpected changes of the terrain by correcting the old path and re-generating a new trajectory. As a result, the vehicle is able to undertake the maximum number of tasks with certain degree of maneuverability having situational awareness of the operating field. The computational engine of the mentioned framework is based on the biogeography based optimization (BBO) algorithm that is capable of providing efficient solutions. To evaluate the performance of the proposed framework, firstly, a realistic model of undersea environment is provided based on realistic map data, and then several scenarios, treated as real experiments, are designed through the simulation study. Additionally, to show the robustness and reliability of the framework, Monte-Carlo simulation is carried out and statistical analysis is performed. The results of simulations indicate the significant potential of the two-level hierarchical mission planning system in mission success and its applicability for real-time implementation.

Link:           https://arxiv.org/ftp/arxiv/papers/1604/1604.07898.pdf

# Appendix J

# An Autonomous Dynamic Motion-Planning Architecture for Efficient AUV Mission Time Management in Realistic Sever Ocean Environment

*Abstract*—Today AUVs operation still remains restricted to very particular tasks with low real autonomy due to battery restrictions. Efficient path planning and mission scheduling are principle requirement toward advance autonomy and facilitate the vehicle to handle long-range operations. A single vehicle cannot carry out all tasks in a large scale terrain; hence, it needs a certain degree of autonomy in performing robust decision making and awareness of the mission/environment to trade-off between tasks to be completed, managing the available time, and ensuring safe deployment at all stages of the mission. In this respect, this research introduces a modular control architecture including higher/lower level planners, in which the higher level module is responsible for increasing mission productivity by assigning prioritized tasks while guiding the vehicle toward its final destination in a terrain covered by several waypoints; and the lower level is responsible for vehicle's safe deployment in a smaller scale encountering time varying ocean current and different uncertain static/moving obstacles similar to actual ocean environment. Synchronization between higher and lower level modules is efficiently configured to manage the mission time and to guarantee on-time termination of the mission. The performance and accuracy of two higher and lower level modules are tested and validated using ant colony and firefly optimization algorithm, respectively. After all, the overall performance of the architecture is investigated in 10 different mission scenarios. The analysis of the captured results from different simulated missions confirms the efficiency and inherent robustness of the introduced architecture in efficient time management, safe deployment, and providing beneficial operation by proper prioritizing the tasks in accordance with mission time.

Link: https://arxiv.org/ftp/arxiv/papers/1604/1604.08336.pdf

# Appendix K

# Persistent AUV Operations Using a Robust Reactive Mission and Path Planning (RRMPP) Architecture

*Abstract*— Providing a higher level of decision autonomy is a true challenge in development of today AUVs and promotes a single vehicle to accomplish multiple tasks in a single mission as well as accompanying prompt changes of a turbulent and highly uncertain environment. The proceeding approach builds on recent researches toward constructing a comprehensive structure for AUV mission planning, routing, task-time managing and synchronic online motion planning adaptive to sudden changes of the time variant environment. Respectively, an "Autonomous Reactive Mission Scheduling and Path Planning" (ARMSP) architecture is constructed in this paper and a bunch of evolutionary algorithms are employed by different layers of the proposed control architecture to investigate the efficiency of the structure toward handling addressed objectives and prove stability of its performance in real-time mission task-time-threat management regardless of the applied metaheuristic algorithm. Static current map data, uncertain dynamic-static obstacles, vehicles Kino-dynamic constraints are taken into account and online path re-planning strategy is adopted to consider local variations of the environment so that a small computational load is devoted for re-planning procedure since the upper layer, which is responsible for mission scheduling, renders an overview of the operation area that AUV should fly thru. Numerical simulations for analysis of different situations of the real-world environment is accomplished separately for each layer and also for the entire ARMSP model at the end. Performance and stability of the model is investigated thorough employing metaheuristic algorithms toward furnishing the stated mission goals. It is proven by analyzing the simulation results that the proposed model provides a perfect mission timing and task managing while guarantying a secure deployment during the mission.

Link:    https://arxiv.org/ftp/arxiv/papers/1605/1605.01824.pdf